\documentclass[pdflatex,iicol]{sn-jnl}
\usepackage{amsmath,amsfonts}
\usepackage{algorithmic}
\usepackage{algorithm}
\usepackage{array}
\usepackage[caption=false,font=normalsize,labelfont=sf,textfont=sf]{subfig}
\usepackage{textcomp}
\usepackage{stfloats}
\usepackage{url}
\usepackage{verbatim}
\usepackage{graphicx}
\usepackage{cite}
\usepackage{framed,multirow}  %
\usepackage[x11names]{xcolor} %
\usepackage{booktabs} %
\usepackage{arydshln} %
\usepackage{bm} %
\usepackage{placeins} %
\usepackage{bbm} %
\usepackage{hyperref} %
\usepackage{amsmath} %
\usepackage[authoryear]{natbib} %
\usepackage{verbatim} %
\usepackage{tikz} %
\usetikzlibrary{calc}
\usepackage{geometry} %
\usetikzlibrary{spy}

\usepackage[capitalize]{cleveref}
\crefname{section}{Sec.}{Secs.}
\Crefname{section}{Section}{Sections}
\Crefname{table}{Table}{Tables}
\crefname{table}{Tab.}{Tabs.}

\theoremstyle{thmstyleone}%

\theoremstyle{thmstyletwo}%

\theoremstyle{thmstylethree}%

\begin{document}

\title{Blur2seq: Blind Deblurring and Camera Trajectory Estimation from a Single Camera Motion-blurred Image}

\author*[1]{\fnm{Guillermo} \sur{Carbajal}}\email{carbajal@fing.edu.uy}

\author[2]{\fnm{Andrés} \sur{Almansa}}\email{andres.almansa@parisdescartes.fr}

\author[1,3]{\fnm{Pablo} \sur{Musé}}\email{pmuse@fing.edu.uy}

\affil[1]{\orgdiv{IIE, Facultad de Ingeniería}, \orgname{Universidad de la República}, \orgaddress{\street{J. Herrera y Reissig 565}, \city{Montevideo}, \postcode{11300}, %
\country{Uruguay}}}

\affil[2]{\orgdiv{MAP5 - CNRS}, \orgname{Université Paris Cité}, \orgaddress{\street{45 rue des Saints-Pères}, \city{Paris}, \postcode{F-75006}, %
\country{France}}}

\affil[3]{\orgdiv{Centre Borelli - CNRS}, \orgname{ENS Paris-Saclay, Université Paris Saclay}, \orgaddress{\street{4 avenue des Sciences}, \city{Gif-sur-Yvette}, \postcode{91190}, %
\country{France}}}

\abstract{Motion blur caused by camera shake, particularly under large or rotational movements, remains a major challenge in image restoration. We propose a deep learning framework that jointly estimates the latent sharp image and the underlying camera motion trajectory from a single blurry image. Our method leverages the Projective Motion Blur Model (PMBM), implemented efficiently using a differentiable blur creation module compatible with modern networks. 
A neural network predicts a full 3D rotation trajectory, which guides a model-based restoration network trained end-to-end. This modular architecture provides interpretability by revealing the camera motion that produced the blur.
Moreover, this trajectory enables the reconstruction of the sequence of sharp images that generated the observed blurry image. To further refine results, we optimize the trajectory post-inference via a reblur loss, improving consistency between the blurry input and the restored output. Extensive experiments show that our method achieves state-of-the-art performance on both synthetic and real datasets, particularly in cases with severe or spatially variant blur, where end-to-end deblurring networks struggle.
Code and trained models are available at \texttt{\url{https://github.com/GuillermoCarbajal/Blur2Seq/}}

}

\maketitle


Before the advent of the deep learning era, most classic deblurring algorithms relied on explicit degradation models. Among the most common were the uniform blur model, the locally uniform motion blur model \citep{levin2006blindmotion, joshi2008psf, hirsch2011fastremoval}, and the Projective Motion Blur Model (PMBM) \citep{whyte2010nonuniform,joshi2010image,tai2011richardson-lucy, gupta_mdf_deblurring, vasu2017local}, which describes the spatially varying blur induced by camera motion during image acquisition. The PMBM represents the blurry image as the temporal average of images resulting from a sequence of homographies warping a latent image, each corresponding to a camera pose at a discrete time step during exposure. 

\citet{joshi2010image} and \citet{tai2011richardson-lucy} employ the PMBM model in their deblurring algorithms, by assuming a planar or distant scene, and a known motion path composed of six degrees of freedom: three rotations and three translations. Later works \citep{whyte2010nonuniform, gupta_mdf_deblurring} introduce methods to estimate the motion path from a single blurry image by recovering motion kernel densities, i.e., how long the camera remained at different poses. To facilitate this estimation, the camera pose space is discretized and its degrees of freedom are reduced from six to three. \citet{whyte2010nonuniform} recover motion densities parameterized by the three rotational components -- roll, pitch, and yaw -- while \cite{gupta_mdf_deblurring} recover motion in terms of roll and \textit{x}-\textit{y} plane translation. These parameterizations are similar for sufficiently long focal lengths because of the rotation-translation ambiguity at such focal lengths. The camera motion parameters and the restored images are recovered by solving an alternate minimization problem. However, these image-specific optimization approaches are computationally expensive and rely on strong priors that are difficult to enforce in practice.

\begin{figure*}[ht]
    \centering
    \includegraphics[width=\textwidth]{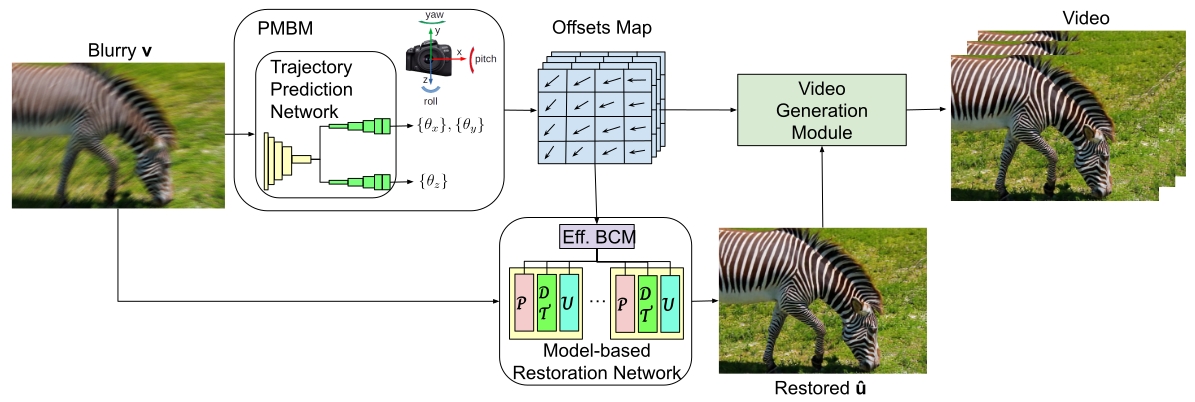}
    \caption{The Trajectory Prediction Network estimates the camera motion during image acquisition. According to the Projective Motion Blur Model, this trajectory defines the image degradation, which we characterize as a set of pixel-specific time-dependent offsets. A model-based restoration network then takes the predicted offsets together with the blurry image to produce the restored image. Finally, the Video Generation Module uses the restoration and the pixel displacements to synthesize a video of the acquisition process. \label{fig:overview_of_the_method}}
\end{figure*}

\citet{li2023self} avoid the well-known dataset-specific bias by leveraging the classical \textit{space-variant overlap-add} (SVOLA) degradation model \citep{harmeling2010space, hirsch2010efficient} and parameterizing the latent image with a neural network. The network parameters are then optimized via Monte Carlo Expectation Maximization (MCEM), without requiring any training data.

Despite the remarkable progress of deep learning methods for image restoration, motion blur due to camera shake remains a challenge, especially under large or rotational (e.g., roll) motion. Most modern deblurring networks operate as black boxes, lacking explicit constraints on the degradation process, which often leads to artifacts and poor generalization to real-world motion blur \citep{tran2021explore,rim_2022_ECCV}.

Some recent works \citep{mustaniemi2019gyroscope,ji2021robust, luan2024gyroscope} attempt to incorporate external motion cues using Inertial Measurement Units (IMUs) data. While promising, these approaches depend on the availability and precise synchronization of IMU data, which is often impractical. Moreover, how to effectively integrate such measurements into state-of-the-art deep-deblurring architectures remains unclear.

Since the blurry image contains information on the path followed by the camera, we build on the idea of simultaneously restoring the image and estimating the trajectory of the camera \citep{whyte2010nonuniform,gupta_mdf_deblurring}, adapting it to the context of deep neural networks. Training a restoration network that also estimates the degradation process serves as a form of regularization \citep{li2022learning,carbajal2023blind}, ensuring that restorations are consistent with a physically grounded blur model. This approach helps to reduce artifacts and hallucinations in the restored image. 

In this work, we revisit the idea of estimating the motion trajectory jointly with the restored image, but unlike previous methods, we do so using deep neural networks. In addition, this joint estimation provides interpretability, as the estimated motion trajectory offers insight into the cause of the blur, something often lacking in purely black-box models. Importantly, we show that this approach is especially effective under large camera motions, where most existing deblurring networks fail.

\subsection*{Overview of the proposed method}

We propose a blind deblurring algorithm for camera shake that first estimates the camera trajectory during image acquisition and then feeds this trajectory to a non-blind deblurring module specifically adapted to camera motion blur.

The method relies on the Projective Motion Blur Model and learns its parameters by training a neural network that predicts the camera trajectory from a single blurry image. Then, we propose a model-based non-blind restoration network that generates the restoration from the blurry image and the estimated degradation $\mathbf{B}$ defined by offset maps derived from the predicted trajectory. After pretraining the trajectory and the restoration networks separately, we perform a joint training phase that enables the restoration network to cope with inaccuracies in the trajectory estimates. The method is summarized in \cref{fig:overview_of_the_method}.

Since the method jointly predicts the trajectory and the restored image, it enables:
i) refinement of both through a fine-tuning step, ensuring consistency with the motion blur model for a given blurry input, and
ii) the generation of a video sequence by warping the restored image with the homographies defined by the trajectory.

\subsection*{Contributions}

This work first proposes an efficient approach to incorporate model-based camera motion information into learning-based, non-blind deblurring algorithms. The method includes the proposal of:
\begin{enumerate}
    \item A trajectory prediction network that estimates camera motion directly from a single blurry image. To the best of our knowledge, the proposed motion trajectory network is the first to recover a motion trajectory from a single blurry image. 
    \item A model-based non-blind restoration network specially tailored for blur due to camera shake. 
    \item A procedure for joint-training the networks that leads to a novel blind, interpretable end-to-end motion deblurring method that exhibits excellent performance in challenging blur conditions.  
    \item A fine-tuning step that allows for refinement of both the trajectory and the restored network.  
    \item A blur to sequence extension that blindly reconstructs the sequence of sharp images that gave rise to the single observed blurry image.

\end{enumerate}

\section{Background}

In this section, we review the Projective Motion Blur Model (PMBM)\citep{tai2011richardson-lucy}, an established model for camera-induced blur, and the Exposure Trajectory Recovery (ETR) framework \citep{zhang2021exposure}, a recent neural approach to blur synthesis. By leveraging the ETR framework, in \cref{sec:PMBM_through_ETR} we propose an efficient implementation of PMBM using a differentiable module that accelerates the trajectory and the restoration network training.

\subsection{Projective Motion Blur Model \label{sec:PMBM}} 

The Projective Motion Blur Model (PMBM) represents the observed blurry image as the average of a sequence of homography transformations of the latent image. 

Throughout this article, we distinguish between the continuous blur operator $B$, defined as an integral operator over a continuous sequence (trajectory) of homographies $\{\mathcal{H}_t\}$, and its discrete counterpart $\mathbf{B}$, which acts on vectorized images and discrete timesteps. Specifically, $B$ operates on image functions defined on a continuous domain, i.e., $u: \Omega \subset \mathbb{R}^2 \rightarrow \mathbb{R}$. It is defined as:
\begin{equation}\label{eq:continuous-blur}
v(x):=B(\{\mathbf{\mathcal{H}}_t\})(u)(x) = \int_0^{\tau} u(\mathbf{\mathcal{H}}^{-1}_t(x)) dt,
\end{equation}
where $\tau$ is the exposure time. The discrete form of the blur operator $\mathbf{B}$ is a sparse matrix constructed as a weighted sum of discrete warpings:

\begin{equation}
    \mathbf{B}=\sum_j w_j \mathbf{T}_{\mathcal{H}_j},
    \label{eq:homographic_blur_operator}
\end{equation}
where $\mathbf{T}_{\mathcal{H}_j}$ is a $N \times N$ sparse square matrix that resamples and warps the $N$-pixels latent image $\mathbf{u} \in \mathbb{R}^N$ following homography $\mathcal{H}_j$. Each row of $\mathbf{T}_{\mathcal{H}_j}$  contains the interpolation coefficients used to compute the pixel values in the warped image by applying the inverse homography. The weights $w_j$ represent the proportion of time spent by the camera at pose $j$. The blurry image $\mathbf{v}$ is written as:
\begin{equation}
    \mathbf{v} = \sum_j w_j (\mathbf{T}_{\mathcal{H}_j} \mathbf{u}) + \bm{\varepsilon},
\end{equation}
where $\bm{\varepsilon}$ is the estimation or approximation error. The sequence \{$\mathbf{T}_{\mathbf{\mathcal{H}}_j}$\} forms a basis set whose elements can be linearly combined to get the corresponding blur matrix for any camera trajectory.

For a particular 6D pose indexed by $j$, the homography $\mathbf{\mathcal{H}}_j$ that warps a fronto-parallel scene at depth $d$ is defined as

\begin{equation}
    \mathbf{\mathcal{H}}_j = \mathbf{K} \left( \mathbf{R}_j +\frac{1}{d}\mathbf{t}_j\mathbf{n}^T \right) \mathbf{K}^{-1}, \label{eq:homographies_from_poses}
\end{equation}
where $\mathbf{R}_j$ and $\mathbf{t}_j$ are the rotation matrix and translation vector at position $j$, $\mathbf{K}$ is the camera intrinsics matrix, and $\mathbf{n}$ the unit vector normal to the camera's focal plane \citep{hartley2003multiple}.
By assuming a pure rotational camera motion ($\mathbf{t}_j=0$) parameterized by three angles (pitch, yaw, and roll) \citet{whyte2010nonuniform} relax the parallel scene hypothesis and simplifies the homography calculation:
\begin{equation}
\mathbf{\mathcal{H}}_j = \mathbf{K}  \mathbf{R}_j  \mathbf{K}^{-1}. \label{eq:homographies_from_poses2}
\end{equation}
In this article, we adopt the same parameterization as in \citep{whyte2010nonuniform}.

\subsection{Exposure Trajectory Recovery Model}

The Exposure Trajectory Recovery model, introduced by \cite{zhang2021exposure},  represents blur formation as a continuous spatial shift over time.
This model represents the blurry image as a sequence of transformations applied to the latent image; in particular, these transformations can be homographies, like in the PMBM. Specifically, the blurry image $\mathbf{v}$ is modeled as
\begin{equation}
    \mathbf{v} = \int_{0}^{\tau} \mathbf{u}^{(t)} dt,
\end{equation}
where $\mathbf{u}^{(t)}$ denotes the latent frame at time $t$ and $\tau$ the exposure time.     

The authors approximate $\mathbf{u}^{(t)}$ as a warping of the latent image, where each pixel undergoes a time-dependent shift:
\begin{equation}
    \mathbf{u}^{(t)}(\mathbf{i}) = \mathbf{u}(\mathbf{i} + \Delta^{(t)}(\mathbf{i}) ).
\end{equation}
Here $\mathbf{i}$ represents the pixel with 2D coordinates $(i,j)$, and $\Delta^{(t)}(\mathbf{i})$ denotes the displacement of pixel $\mathbf{i}$ at time $t$. Assuming the scene brightness remains constant during exposure, and discretizing the exposure time $\tau$ into $T$ uniformly spaced time steps $t_0$, $t_1$ \ldots $t_{T-1}$,  the blurry image is synthesized as
\begin{equation}
    \mathbf{v}(\mathbf{i})=\frac{1}{T} \sum_{t_n=0}^{T-1} \mathbf{u} (\mathbf{i} + \Delta^{(t_n)}(\mathbf{i})).
    \label{eq:forward_offsets_ch7}
\end{equation}
 In this formulation, the value of each blurry pixel $\mathbf{i}$ is the result of averaging the contributions of multiple pixels from the latent image, each defined by an offset $\Delta^{(t_n)}(\mathbf{i})$ from $\mathbf{i}$. The blur degradation is thus characterized by a set of motion offsets $\{\Delta^{(t_n)}(\mathbf{i})\}_{t_n=0}^{T-1}$. 

\begin{figure}[ht]
    \centering
\includegraphics[width=1.0\linewidth]{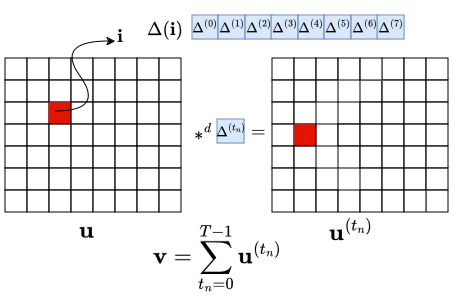}
    \caption{Blur Creation Module proposed by \cite{zhang2021exposure}. The module receives the offsets $\Delta(\mathbf{i})=(\Delta^{(0)},\Delta^{(1)}, \ldots, \Delta^{(T-1)})$ for each pixel $\mathbf{i}$. With the offsets associated to time step $t_n$ ($\Delta^{(t_n)}$), the module generates an image $\mathbf{u}^{(t_n)}$ using deformable convolutions ($*^d$). The blurry image is the average of the $T$ images $\mathbf{u}^{(t_n)}$.   } \label{fig:exp_traj_blur_creation_module}
\end{figure}

\citet{zhang2021exposure} propose a Blur Creation Module (c.f. \Cref{fig:exp_traj_blur_creation_module}) to synthesize a blurry image from a sharp image $\mathbf{u}$ and a set of pixel-specific motion offsets $\{\Delta^{(t_n)}(\mathbf{i})\}_{t_n=0}^{T-1}$, by averaging warped frames defined by the offsets. Each offset is a 2D vector that represents the displacement of the pixel $\mathbf{i}$ in the sharp image at the time interval $[t_n,t_{n+1})$. Since the offsets take continuous values, bilinear interpolation is used to compute pixel values in the warped frames. 

The Blur Creation Module receives the offsets from a motion offset estimation network, trained to minimize a \textit{reblur loss}. Estimating plausible offsets is extremely challenging, as there exist infinitely many sets of offsets that can produce the same blurry image through different warping combinations of a sharp image. To address the ill-posed nature of this problem and ensure smooth motion paths, \citet{zhang2021exposure} constrain the solution space by predicting only two offsets per pixel $\Delta^{t_0}(i)$ and $\Delta^{t_{T-1}}(i)$, and assuming that the origin is the midpoint of the trajectory. Then, a 15-step trajectory is generated through linear, bilinear, or quadratic interpolation. The interpolation formula is given by:
\begin{equation}
    \mathbf{u}(\underbrace{\mathbf{i} + \Delta^{(t_n)}(\mathbf{i})}_{=:\,\mathbf{i}^{(t_n)}})=\sum_{\mathbf{k} \in \mathrm{nn}(\mathbf{i}^{(t_n)})}G(\mathbf{\mathbf{k}}, \mathbf{i}^{(t_n)})\mathbf{u}(\mathbf{k}), 
\end{equation}
where $\mathrm{nn}(\mathbf{i}^{(t_n)})$ are the neighboring pixels of $\mathbf{i}^{(t_n)}$, and $G(\cdot)$ represents the bilinear interpolation kernel.     

Although originally formulated as a neural component for motion offset learning, the Blur Creation Module used in ETR can also be repurposed to implement the PMBM efficiently. In \cref{sec:PMBM_through_ETR} we describe how to represent homography-based blur as pixel-wise motion offsets compatible with the ETR framework.

The following section describes the Trajectory Prediction Network (TPN) we propose to estimate the rotational trajectory of the camera during the acquisition trajectory. According to the PMBM, this trajectory completely determines the blurry image as the average of homographic transformations of the latent sharp image. 

\section{The trajectory prediction network}

\begin{figure*}[h]
    \centering
    \includegraphics[width=\linewidth]{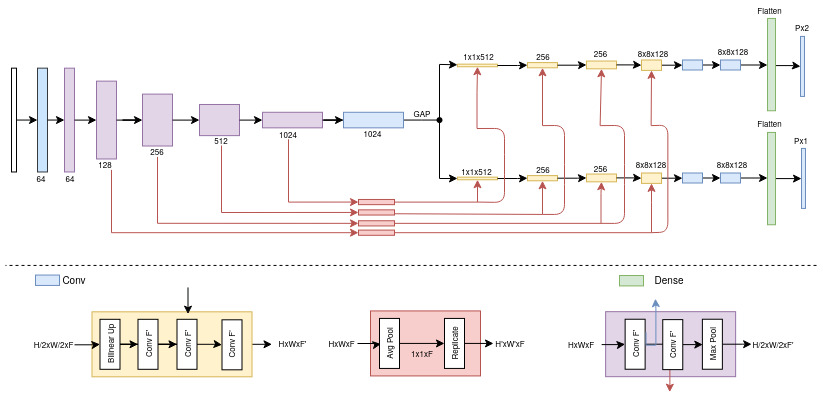}
    \caption{Motion Trajectory Prediction Network. The U-Net type network has two branches. One of the branches predicts the \textit{pitch} and the \textit{yaw}, and the other estimates the \textit{roll} of the camera.}
\label{fig:motion_trajectory_prediction_network}
\end{figure*}

Given a single blurry image with focal length $f$, we propose to predict the camera motion trajectory using the Trajectory Prediction Network $TPN_{\phi}(\cdot)$ shown in \Cref{fig:motion_trajectory_prediction_network}. Similarly to \cite{whyte2010nonuniform, whyte2014deblurring}, we parameterize the camera pose at each time step $t_s$ as a 3D rotation vector $\bm{\theta}^{(t_s)}=[\theta_x^{(t_s)},\theta_y^{(t_s)}, \theta_z^{(t_s)}]^T$. Since in-plane and out-of-plane rotations produce distinctive blur maps,  the network is designed with two branches, one to predict {pitch} and {yaw} angles ($\theta_x$ and $\theta_y$), and the other to predict {roll} angles ($\theta_z$). The rationale for this separation into two branches is the following: roll motion leads to spatially variant blur kernels characterized by minimal blur near the image center and increasing blur toward the image boundaries. In contrast, out-of-plane rotations typically result in more smoothly varying kernel maps across the image.

For the {roll}, the network predicts a vector $\mathbf{r} \in \mathbb{R}^{T}$  containing the rotation angles $\theta_z^{(t_s)}=r_s$ for all time steps $s=0, \ldots, T-1$. For the {pitch} and {yaw} the network predicts a matrix $[\mathbf{p},\mathbf{y}] \in \mathbb{R}^{T\times2}$, from which we derive the \textit{pitch} angles from the first column $\mathbf{p}$ as $\theta_x^{(t_s)}=\arctan(p_s/f)$ and the \textit{yaw} angles from the second column $\mathbf{y}$ as $\theta_y^{(t_s)}=\arctan(y_s/f)$. 

\section{Representing the PMBM degradation as a map of offsets \label{sec:PMBM_through_ETR}}

The blur operator determined by the PMBM (\cref{eq:homographic_blur_operator}) can be implemented as the average of a sequence of homographies applied to the sharp image. Although this approach is accurate, it is computationally inefficient if we want to synthesize a batch of blurry images using different blur maps because different homographies are applied to different images. 

Since the warping determined by a homography can be equivalently represented by pixel-wise offsets, the PMBM can be implemented using the Blur Creation Module \citep{zhang2021exposure}. %
For a blurred image $\mathbf{v}$ that follows the PMBM,
\begin{align}
    \mathbf{v}=\mathbf{Bu} &= \frac{1}{T} \sum_{t_n=0}^{T-1}  \mathbf{T}_{\mathcal{H}_{t_n}} \mathbf{u}.
\end{align}
The value of $\mathbf{v}$ at pixel $\mathbf{i}$ is given by 
\begin{align}
      \mathbf{v}(\mathbf{i}) &= \frac{1}{T} \sum_{t_n=0}^{T-1}  \mathbf{u}(\mathcal{H}^{-1}_{t_n}(\mathbf{i})) \\
    &=  
    \frac{1}{T} \sum_{t_n=0}^{T-1} \mathbf{u}(\mathbf{i} + \Delta^{(t_n)}(\mathbf{i})), \label{eq:offsets_from_homographies} 
\end{align}
where we define the offsets $\Delta^{(t_n)}(\mathbf{i}) = {\mathcal{H}_{t_n}^{-1}}(\mathbf{i}) - \mathbf{i}$. 

Representing the blur operator in terms of per-pixel offsets makes the PMBM practical for training and inference in modern deep learning pipelines. In \Cref{sec:eff_BCM} we propose a GPU-friendly implementation of PMBM using offset maps. In the following, we present the model-based restoration network used in the proposed blind deblurring method.

\section{The Model-based Restoration Network}

The Restoration Network $RN_{\mathbf{\psi}}(\cdot, \cdot)$ receives the blurry image and the degradation $\mathbf{B}$ characterized by a set of offsets $\{\bm{\Delta}^{(t_s)}\}$, and produces the restored image
\begin{align}
\widehat{\mathbf{u}} &  = RN_{\psi}(\mathbf{B}(\{\bm{\Delta}^{(t_s)}(\mathbf{i})\}), \mathbf{v}).
\end{align}

For the restoration network, we adapt the architecture proposed in~\cite{laroche2023deep} to the case of blur due to camera shake. The network unrolls eight iterations of a linearized ADMM optimization.  On each iteration, it alternates between a prior-enforcing step $\mathcal{P}$,
a data-fitting step $\mathcal{DT}$ and an update block $\mathcal{U}$:
\begin{align}
    \mathcal{P}:  \mathbf{u}_{k+1} &=\mathcal{D}_{b_k} \left( \mathbf{u}_k - c_k \mathbf{B}^T\left( \mathbf{B}\mathbf{x}_k - \mathbf{z}_k + \bm{\beta}_k \right) \right), \\
    \mathcal{DT}:  \mathbf{z}_{k+1} &= \frac{\mathbf{v} + a_k \left( \mathbf{B}\mathbf{u}_{k+1} + \bm{\beta}_k \right)}{a_k + 1}, \\
 \mathcal{U}: \bm{\beta}_{k+1} &=\bm{\beta}_k + (\mathbf{B}\mathbf{u}_{k+1}-\mathbf{z}_{k+1}).
\end{align}
Besides the latent image $\mathbf{u}$, the blurry image $\mathbf{v}$, and the degradation $\mathbf{B}$, the update equations involve the auxiliary variables $\mathbf{x}$ and $\mathbf{z}$ introduced during the ADMM optimization, the dual variable $\bm{\beta}_k$, and the coefficients $a_k$,$b_k$, and $c_k$, which are predicted by a fully-connected network. For the prior $\mathcal{P}$ step, the authors assume a denoising prior and retrain a ResUnet architecture \citep{diakogiannis2020resunet}.

The iterative procedure also requires efficient computation of the forward and adjoint blur operators, $\mathbf{B}\tilde{\mathbf{x}}$ and $\mathbf{B}^T\tilde{\mathbf{x}}$, for any image vector $\tilde{\mathbf{x}} \in \mathbb{R}^N$. Previous works such as \citep{laroche2023deep} and \citep{carbajal2023blind} address this by leveraging the convolution theorem and approximating the blur operator as a linear combination of learned kernel bases  \citep{nagy1998restoring}. In contrast, here we propose to employ the Projective Motion Blur Model (PMBM).

In Section~\ref{sec:PMBM_adjoint}, we revisit the computation of $\mathbf{B}^T$ under the PMBM, while in Section~\ref{sec:adjoint_operator_ETR} we present its implementation within the ETR framework. Finally, in Section~\ref{sec:eff_BCM}, we introduce the Efficient Blur Creation Module, which leverages deformable convolutions to compute both the forward and adjoint operators.

\subsection{On the computation of the PMBM adjoint operator \label{sec:PMBM_adjoint}}

To incorporate the PMBM into iterative non-blind deblurring methods, we must compute the adjoint operator of the blurring degradation. In the continuous formulation, this requires characterizing the adjoint $B^*(\{\mathcal{H}_t\})(u)$ such that for any image functions $w$ and $u$ defined on the continuous domain, the following condition holds:
\begin{equation}
    <w,B(\mathbf{\{ \mathcal{H}}_t \} )(u)>\; = \; <B^*(\mathbf{\{ \mathcal{H}}_t\} )(w),u>.
\label{eq:adjoint_operator_condition}
\end{equation}
Using the definitions of the scalar product between functions and of $B(\mathbf{ \{ \mathcal{H}}_t \} )(u)$, the left-hand side of \cref{eq:adjoint_operator_condition} becomes: 
\begin{align}
   <w,& B(\{ \mathcal{H}_t \})(u)> \notag \\ & =  \int_x w(x) \int_t u(\mathcal{H}_t^{-1}(x)) \, dt \, dx \\
   & = \int_t \int_x w(x) u(\mathcal{H}_t^{-1}(x)) \, dx \, dt. 
\end{align}
Each homography $\mathcal{H}^{-1}_t$ maps $x$ to a point $y_t$. By changing the variable $ x = \mathcal{H}_t(y_t) $ with the corresponding Jacobian $dx = |J_{H_t}(y_t)| dy_t$\, we can rewrite this expression as:
{%
\begin{align}
   <w, & B(\{ \mathcal{H}_t \})(u)> \notag \\ & =  \int_t \int_{y_t} w(\mathcal{H}_t(y_t)) u(y_t) \vert J_{\mathcal{H}_t}(y_t)\vert \, dt \, dy_t \\
    & =  \int_t \int_{y} w(\mathcal{H}_t(y)) u(y) \vert J_{\mathcal{H}_t}(y)\vert \, dt \, dy.
   \label{eq:Tai0}
\end{align}
}
The last equality results from noting that the integration in $y_t$ is performed over the entire image, so we replace $y_t$ by $y$ to emphasize its independence from $t$. Then, exchanging the order of integration and using that $u(y)$ does not depend on $t$, yields
{%
\begin{align}
   <w,& B(\{ \mathcal{H}_t \})(u)> \notag \\
   & = \int_{y} u(y) \int_t w(\mathcal{H}_t(y)) \vert J_{\mathcal{H}_t}(y)\vert \, dt \, dy 
   \label{eq:Tai1} \\
   & = <B^*(\{\mathcal{H}^{-1}_t\})(w), u>. \notag
\end{align}
}
Hence, the adjoint operator writes
$$ B^*(\{\mathcal{H}^{-1}_t\})(w) := \int_t w(\mathcal{H}_t(y)) \vert J_{\mathcal{H}_t}(y)\vert \, dt. $$
Note that the right-hand side of this equation coincides with $B(\{\mathcal{H}^{-1}_t\})$ (see equation~\eqref{eq:continuous-blur}) except for the Jacobian weights $\vert J_{\mathcal{H}_t}(y)\vert$ in the integral.

\citet{tai2011richardson-lucy} compute the adjoint operator as $B^*(\{\mathcal{H}_t\})=B(\{\mathbf{\mathcal{H}}^{-1}_t\})$, but, following \cref{eq:Tai1}, this holds if and only if $\vert J_{\mathcal{H}_t}(y)\vert=1$.

This assumption, valid for homographies that do not compress or dilate the image, simplifies the computation of the adjoint operator. In the supplementary material, we empirically examine the conditions under which this approximation holds. Specifically, we parameterize the pose space using rotations as in \citep{whyte2014deblurring}, and analyze how the Jacobian determinant varies with the pose parameters. These experiments reveal that, for normal sequences of homographies found in practice, the approximation error introduced when considering $\vert J_{\mathcal{H}_t}(y)\vert=1$ has no noticeable impact on the estimation of the adjoint blur operator. In what follows, we assume that this condition is met and detail the implementation of the blur operator and its adjoint for the Projective Motion Blur Model.

\subsection{Implementing the Adjoint Operator  with the ETR Blur
Creation Module} 
\label{sec:adjoint_operator_ETR}

Following the same strategy used in \cref{sec:PMBM_through_ETR} to compute the blur operator $\mathbf{B}$, the adjoint operator can also be computed using offsets: 
\begin{align}
    \mathbf{B}^T\mathbf{v} &= \frac{1}{T} \sum_{t_n} \mathbf{T}_{\mathbf{\mathcal{H}}^{-1}_{t_n}} \mathbf{v} \\
    (\mathbf{B}^T\mathbf{v})(\mathbf{j}) &= \frac{1}{T} \sum_{t_n} \mathbf{v}(\mathcal{H}_{t_n}(\mathbf{j})) \\
    &=  
    \frac{1}{T} \sum_{t_n} \mathbf{v}(\mathbf{j} + \tilde{\Delta}^{(t_n)}(\mathbf{j})),
\end{align}
where the offsets in this case are defined as $\tilde{\Delta}^{(t_n)}(\mathbf{j}) = \mathcal{H}_{t_n}(\mathbf{j}) - \mathbf{j}$. This expression mirrors the implementation of the forward operator $\mathbf{B}$, with the key difference being that the inverse homographies are used. In our implementation, instead of explicitly computing $\tilde{\Delta}^{(t_n)}(\mathbf{j})$ from the homographies, we reduce the memory requirements by approximating them from the forward offsets $\Delta^{(t_n)}(\mathbf{j})$, effectively halving the storage cost. 
The details of this approximation are provided in the Supplementary Material.

\subsection{An efficient Blur Creation Module \label{sec:eff_BCM}}

As described in \cref{sec:PMBM_through_ETR} and \cref{sec:adjoint_operator_ETR}, both the blur operator $\mathbf{B}$ and its adjoint $\mathbf{B}^T$ are computed using the Blur Creation Module by \cite{zhang2021exposure}. Here, we propose a modification for this module that results in an efficient implementation. This approach allows for increasing the number of time steps considered in the blurring process while reducing the time required to blur the image.

\begin{figure}[ht]
    \centering    \includegraphics[width=0.8\linewidth]{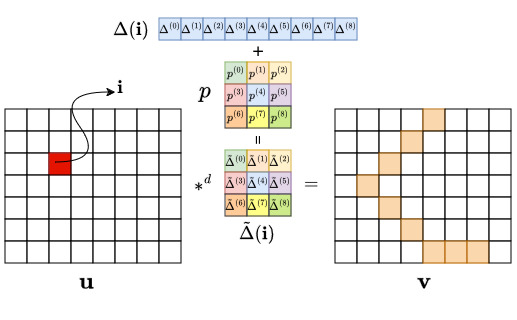}
    \caption[Efficient Blur Creation Module.]{Efficient Blur Creation Module. The module receives the offsets $\Delta(\mathbf{i})=(\Delta^{0},\Delta^{1}, \ldots, \Delta^{T-1})$ for each pixel $\mathbf{i}$. The offsets are first rearranged into a $\sqrt{T}$ square matrix, and then the original offsets values $\Delta(\mathbf{i})$ modified to compensate the displacement from the center in the matrix ($\tilde{\Delta}(\mathbf{i})=\Delta(\mathbf{i})+\mathbf{p}$). The blurry image is then generated as a deformable convolution ($\mathbf{*}^d$) with a size $\sqrt{T}$ filter with constant values $(1/T)$ and sampling locations determined by the the modified offsets $\tilde{\Delta}(\mathbf{i})=(\tilde{\Delta}^{0},\tilde{\Delta}^{1}, \ldots, \tilde{\Delta}^{T-1})$. \label{fig:efficient_blur_creation_module}}    
\end{figure}

The original Blur Creation Module synthesizes the blurry image by averaging $T$ warped frames $\mathbf{u}^{(t_n)}$,  each obtained as a deformable convolution with a $1\times1$ filter. To compute and average all warped frames in a single pass, we introduce a new formulation using a filter $\bm{\omega}$, initially defined as $\bm{\omega}(\mathbf{j})=\mathbbm{1}_{\{\mathbf{j}=(0,0)\}}$. Using this filter, the blurred image $\mathbf{v}$ writes:
\begin{align}
\mathbf{v}(\mathbf{i}) & = (\mathbf{B}\mathbf{u})(\mathbf{i}) \\
&= \frac{1}{T} \sum_{t_n=0}^{T-1} \mathbf{u}(\mathbf{i} + \Delta^{(t_n)}(\mathbf{i})) \\
&= \frac{1}{T} \sum_{t_n=0}^{T-1} \sum_{\mathbf{j}} \bm{\omega}({\mathbf{j}}) \,\mathbf{u}(\mathbf{i} + \mathbf{j} +  \Delta^{(t_n)}(\mathbf{i})).
\label{eq:blur_def_conv}
\end{align}
Now, unlike the original implementation, which produces the blurred image using $T$ deformable convolutions with a $1 \times 1$ filter, here we propose to achieve the same result but using a single deformable convolution with a $\sqrt{T} \times \sqrt{T}$ filter. To this end, we redefine the previous filter $\bm{\omega}$, indexing it by a set of pixel positions $\mathbf{p}^{(t_n)}$, associated with the corresponding time steps $t_n = 0, 1, \dots, T-1$. For example, when $T=25$, we define a $5\times5$ filter with $\mathbf{p}^{(t_n)} \in \{-2, -1, 0, 1, 2 \}^2$. The filter values are constant for all $t_n$, with $\omega_{t_n} := \bm{\omega}(\mathbf{p}^{(t_n)}) = 1/T$. Since the center of the filter is the only position aligned with the pixel positions $\mathbf{i}$, we compensate for the displacement of each position $\mathbf{p}^{(t_n)}$ by modifying the offsets. Specifically, we define adjusted offsets $\tilde{\Delta}^{(t_n)}(\mathbf{i})$ such that:
\begin{equation}
\Delta^{(t_n)}(\mathbf{i}) = \mathbf{p}^{(t_n)} + \tilde{\Delta}^{(t_n)}(\mathbf{i}).
\end{equation}
Using this formulation, Eq.~\eqref{eq:blur_def_conv} becomes:
\begin{align}
(\mathbf{B}\mathbf{u})(\mathbf{i}) &= \frac{1}{T} \sum_{t_n=0}^{T-1}  \mathbf{u}(\mathbf{i} + \Delta^{(t_n)}(\mathbf{i}))  \\
   &= \sum_{t_n=0}^{T-1} \bm{\omega}(\mathbf{p}^{(t_n)}) \mathbf{u}(\mathbf{i} + \mathbf{p}^{(t_n)} + \tilde{\Delta}^{(t_n)}(\mathbf{i})) \\
    &= \sum_{t_n=0}^{T-1} w_{t_n} \mathbf{u}(\mathbf{i} + \mathbf{p}^{(t_n)} + \tilde{\Delta}^{(t_n)}(\mathbf{i})).  \label{eq:forward_offsets_efficient}
\end{align}
The procedure is illustrated in \Cref{fig:efficient_blur_creation_module}.

\FloatBarrier

\section{Training Procedure \label{sec:training_procedure}}

Training the Trajectory Prediction Network $TPN_{\phi}(\cdot)$ and the Restoration Network $RN_{\psi}(\cdot,\cdot)$ requires a dataset of blurry/sharp image pairs with corresponding camera motion trajectories.

The K\"{o}hler dataset is the only available collection of real images with this information. However, its limited size and variability make it unsuitable for training a neural network. Therefore, we generate the dataset synthetically using the projective motion blur model.

\subsection{Synthetic Dataset Generation \label{sec:dataset_generation_ch7}}

The synthetic dataset is constructed using images from the COCO dataset (\cite{lin2014microsoft}), though any other dataset of sharp images could be used. Camera trajectories are generated with the camera shake trajectories generator (\cite{gavant2011physiological,delbracio2015removing}) based on physiological hand tremor data.
Blurry images are generated on the fly by randomly choosing an image $\mathbf{u}$ from the COCO dataset and a trajectory $\{\bm{\theta}^{(t_s)}\}$ from a pre-computed set of trajectories. The blurry image $\mathbf{v}$ is generated as:
\begin{equation}
    \mathbf{v}=R(\mathbf{B}\mathbf{u}), 
\end{equation}
where $R(\cdot)$ is the smooth approximation of the sensor response function used by \cite{whyte2014deblurring} and \cite{carbajal2023blind}:
\begin{equation}
    R(x)=x-\frac{1}{a}\log \left( 1 + e^{a(x-1)}\right).
    \label{eq:saturation_function}
\end{equation}
The parameter $a$ controls the smoothness of the approximation and is set to $a=50$.

The degradation operator $\mathbf{B}$ is applied efficiently to the image $\mathbf{u}$ using \cref{eq:forward_offsets_efficient}, where the offsets are computed from the set of camera poses in two steps: first, computing the homographies using  \cref{eq:homographies_from_poses2}; and second, deriving the offsets from these homographies using \cref{eq:offsets_from_homographies}. 

This synthetic dataset enables supervised training of the $TPN_\phi(\cdot)$ using both image and trajectory supervision, as described next.

\subsection{Training Losses and Joint Optimization}

To train the motion Trajectory Prediction Network we combine \textit{two losses}:

\paragraph{Reblur Loss} Given a blurry image $\mathbf{v}^{GT}$ generated from a latent image $\mathbf{u}^{GT}$, we aim to find the camera's path followed during the exposure by minimizing:
\begin{equation}
    \mathcal{L}_{reblur}=\sum_i \left( \widehat{v}_i - v_i^{GT} \right)^2, \label{eq:reblur_loss_ch7}
\end{equation}
where $\widehat{v}_i$ is the $i$-th element of $\widehat{\mathbf{v}}=R(\mathbf{B}\mathbf{u}^{GT})$, which is computed using the degradation operator $\mathbf{B}$ defined by the trajectory $\{\bm{\widehat{\theta}}^{(t_s)}\}$ predicted by the network. 

\paragraph{Trajectory Loss} 
Ground truth trajectories $\{\bm{\theta}^{(t_s)}\}^{GT}$ and predicted trajectories $\{\bm{\widehat{\theta}}^{(t_s)}\}$ are sequences of 3D camera poses that define motion blur kernel maps. Since altering the order of the elements in these sequences does not affect the resulting kernel field, this invariance should be considered in the metric used to compare the trajectories. 

To address this, we match each camera pose in the ground-truth trajectory to a pose in the predicted trajectory by minimizing the Earth Movers Distance (EMD) between both sequences. To construct the $T\times T$ cost matrix $\mathbf{M}$, we compute distances in the image plane rather than in the 3D rotation space. This choice is motivated by the observation that Euclidean distances in 3D rotation space may not accurately reflect differences in the induced image projections. 

To define the cost matrix, we create a grid of $N_g$ evenly spaced points $\mathbf{g}$ over the image and define an operator $p(\mathcal{H}_{\bm{\theta}},\mathbf{g})$ that transforms the grid $\mathbf{g}$ into a grid $\mathbf{g_{\bm{\theta}}}$, using the homography $\mathcal{H}_{\bm{\theta}}$. The cost matrix element $m_{ij}$ is computed as
\begin{equation}
    m_{ij}=\frac{1}{N_g}  \Vert  \mathbf{g}_{\widehat{\bm{\theta}}^{(i)}}  - \mathbf{g}_{{\bm{\theta}}^{(j)}}\Vert_2^2.
\end{equation}
Using $\mathbf{M}$ as the EMD cost matrix, we estimate the binary optimal transport matrix $\widehat{\mathbf{O}}$, and then the \textit{trajectory loss} as:

\begin{equation}
    \mathcal{L}_{trajectory}=\sum_{l=0}^{T-1} \sum_{m=0}^{T-1}\left({\mathbf{\widehat{O}}} \circ \mathbf{M}\right)_{lm},
\end{equation}
where $T$ are the time steps and $\circ$ is the point-wise product.

\paragraph{Pre-training loss} The Trajectory Prediction Network ($TPN_{\phi}(\cdot)$) is first pre-trained using the loss 
\begin{equation}
    \mathcal{L}_{MTP} = \mathcal{L}_{reblur} + \lambda_{emd} \,\mathcal{L}_{trajectory}, \label{eq:loss_MTP}
\end{equation}
with $\lambda_{emd}=1$. 

\paragraph{Restoration Loss} 

The Restoration Network ($RN_{\psi}(\cdot,\cdot)$) was trained separately using synthetic blurry/sharp image pairs generated as described in \Cref{sec:dataset_generation_ch7}. This pre-training step minimizes the following reconstruction loss:
\begin{equation}
\mathcal{L}_{restoration}=\sum_i \left( \widehat{u}_i - u_i^{GT} \right)^2, \label{eq:loss_restoration_ch7}
\end{equation}
where $\widehat{u}_i$ and $u_i^{GT}$ denote the $i$-th pixels of the restored image $\widehat{\mathbf{u}}$ and the ground truth image $\mathbf{u}^{GT}$, respectively.

\paragraph{Joint-training} After independently pre-training $TPN_{\bm{\phi}}(\cdot)$ and $RN_{\bm{\psi}}(\cdot,\cdot)$, we jointly fine-tuned them by minimizing a combined objective function that includes both the trajectory and image supervision terms:
\begin{equation}
\mathcal{L}_{joint} = \mathcal{L}_{MTP} + \mathcal{L}_{restoration}.
\end{equation}
This joint training promotes consistency between the predicted motion trajectory and the restored image, allowing the two networks to reinforce each other: the trajectory estimation guides the deblurring process, while feedback from the restoration network refines the motion prediction.

\paragraph{Trajectory Optimization via Reblur Loss \label{sec:traj_optim}}

After joint training, we obtain a set of weights $\bm{\phi}$ for $TPN_{\phi}(\cdot)$ and a fixed restoration network $RN_{\bm{\psi}}(\cdot,\cdot)$. Given a blurry input image $\mathbf{v}^{GT}$, $TPN_{\phi}(\cdot)$ predicts a motion trajectory, which induces a blur operator $\mathbf{B}(\bm{\phi}, \mathbf{v}^{GT})$. This operator is then used to guide the deblurring process.

The restored latent image is obtained from the frozen restoration network $RN_{\mathrm{\psi}}(\cdot, \cdot)$ as:

$$
\widehat{\mathbf{u}} = RN_{\mathrm{\psi}}(\mathbf{B}(\bm{\phi}, \mathbf{v}^{GT}), \mathbf{v}^{GT}).
$$

Using this estimate, we reblur the image by applying the blur operator followed by the previously defined saturation function $R(\cdot)$ (\cref{eq:saturation_function}), and define the reblur loss as
$$
\mathcal{L}_{\mathrm{reblur}}(\bm{\phi}) = \left\| R \left( \mathbf{B}(\bm{\phi}, \mathbf{v}^{GT}) \cdot \widehat{\mathbf{u}} \right) - \mathbf{v}^{GT} \right\|_2^2.
$$

In the same spirit of the deep image prior in \citep{Ulyanov2018DeepIP}, we optimize the TPN parameters $\bm{\phi}$ via gradient descent to minimize this loss. Although the restoration network is not updated, the restored image $\widehat{\mathbf{u}}$ still depends on $\bm{\phi}$ through the blur operator, so changes in the predicted trajectory can affect both the reblurred and restored outputs.

This process can be interpreted as a form of maximum likelihood refinement, where we search for trajectory parameters $\bm{\phi}$ that make the observed blurry image most consistent with the reconstructed latent image under the forward model. Importantly, the optimization remains regularized by the structure and initialization of the TPN: the updated trajectories continue to be plausible motions, as they are constrained by the network’s learned prior. In practice, this refinement leads to improved alignment between the deblurred and reblurred images, often enhancing both perceptual quality and fidelity.

\section{Experimental Results}

We evaluated our method on synthetic and real-world blurry images, comparing results with state-of-the-art deblurring techniques. Our complete pipeline includes trajectory prediction, latent image restoration, and trajectory refinement via reblur loss, as described in \Cref{sec:traj_optim}.

As comparison metrics, in addition to PSNR and SSIM, we also employed LPIPS~\citep{zhang2018perceptual} and the Sharpness Index (SI)~\citep{blanchet2012explicit}. LPIPS is a reference-based perceptual metric, whereas SI quantifies the sharpness of the restorations. This is particularly relevant since it is well known that metrics such as PSNR, due to the regression-to-the-mean effect \citep{bruna2015super,delbracio2023inversion}, may favor slightly blurred reconstructions. We highlight the \colorbox{Cyan1}{best} and \colorbox{Yellow1}{second-best} values for each metric. 
 
The models and code used to generate the results with the proposed Blur2Seq method are publicly available\footnote{\url{https://github.com/GuillermoCarbajal/Blur2Seq}}.

\subsection{Blind Deblurring using the Motion Trajectory estimation network}

The synthetic Lai nonuniform blur dataset proposed by \cite{lai2016comparative} was built to evaluate the performance of deblurring algorithms on blur due to camera shake. \Cref{tab:lai_non_uniform} presents the performance of state-of-the-art deblurring networks in this dataset, using either the models or the restored images provided by the authors. Our method ranks first in terms of distortion (PSNR and SSIM) and perceptual (LPIPS) metrics, with an excellent sharpness index.

\begin{table*}[ht]
    \caption[Results on the synthetic nonuniform blur dataset introduced by \cite{lai2016comparative}]{Results on the synthetic nonuniform blur dataset introduced by \cite{lai2016comparative}. The blur is simulated from camera trajectories obtained from a gyroscope attached to the camera.}
    \centering
    \footnotesize
    \begin{tabular}{l|c|c|c|c|c}
    \toprule
       Method & Model  &  PSNR & SSIM & LPIPS & SI \\
       \hline
      AnaSyn \citep{kaufman2020deblurring}      & Spatially uniform kernel-guided convolutions  & 21.46 & 0.695 & 0.285 & 2099  \\ 
      \hline 
      DeepDeblur \citep{Nah_2017_CVPR}  & \multirow{9}{*}{Fully data-driven, end-to-end}     & 20.26  & 0.640  & 0.335 & 1204 \\
      DGANv2-inception \citep{kupyn2019deblurgan}& & 21.37 & 0.671 & 0.271  & 1279 \\
      DGANv2-mobilenet \citep{kupyn2019deblurgan}& & 20.34 & 0.609 & 0.297  & 1080 \\
      SRN \citep{tao2018scale} &       & 21.25 & 0.668  & 0.282 & 1482 \\
      DMPHN \citep{Zhang_2019_CVPR} &       & 19.65  & 0.565 & 0.452 & 1175 \\
      MIMO-UNet \citep{cho2021rethinking} &  & 20.28  & 0.631 & 0.322 & 1213 \\
      MIMO-UNet+ \citep{cho2021rethinking} &  & 20.25  & 0.638 & 0.325 & 1243  \\
      MPRNet \citep{Zamir2021MPRNet}  &    & 20.75  & 0.661 & 0.297 & 1543 \\
      NAFNet \citep{NAFNet}  &    & 20.43  &  0.631 & 0.340 & 1481 \\
      \hline
      Motion-ETR \citep{zhang2021exposure}& Spatially-varying quadratic kernels & 20.77 & 0.651 & 0.316 & 1049 \\
      \cite{vasu2017local}& Projective Motion Blur Model &   20.73 & 0.652 & 0.263 &  \colorbox{Cyan1}{3862}\\
    Deblur MCEM \citep{li2023self} & Efficient Filter Flow & 22.05 & 0.684 & \colorbox{Yellow1}{0.201} & 2438 \\
    J-MKPD \citep{carbajal2023blind} &  Linear combination of kernel basis  &  \colorbox{Yellow1}{22.11}   &  \colorbox{Yellow1}{0.728}  & 0.250  & 2546  \\
    Blur2Seq (ours) &   Projective Motion Blur %
    Model & \colorbox{Cyan1}{23.87} & \colorbox{Cyan1}{0.791}   & \colorbox{Cyan1}{0.186} & \colorbox{Yellow1}{3179} \\
    \bottomrule
    \end{tabular}
    \label{tab:lai_non_uniform}
\end{table*}

\paragraph{Ablation Study: Effect of Trajectory Optimization}

\begin{figure}[h]
    \centering
    \includegraphics[width=0.45\textwidth]{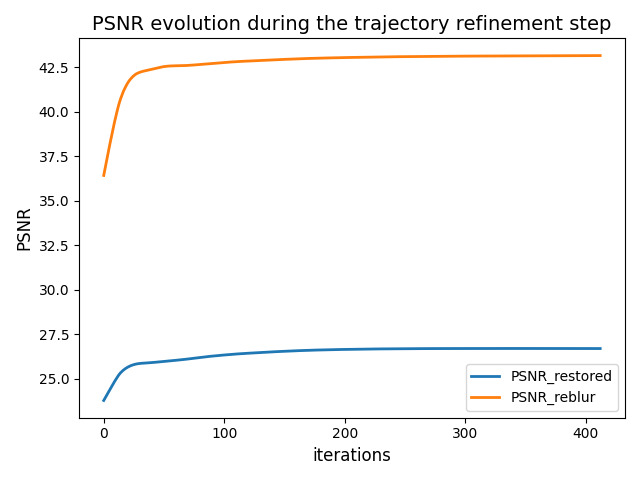}
    \caption{Evolution of reblur and restoration PSNRs during the trajectory optimization step for a sample image.}
    \label{fig:optim_evolution}
\end{figure}

To evaluate the contribution of the trajectory refinement stage, we conducted an ablation study comparing the complete pipeline with a variant that omits the final optimization step. In this variant, the predicted trajectory from the TPN is used directly for restoration, without re-estimation using the reblur loss. As shown in \Cref{tab:Lai_non_uniform_optim}, the optimization stage significantly improves all the metrics. These results highlight the challenge in accurately estimating global camera motion in a single forward pass, and underscore the benefits of refining the estimated trajectory using the reblur consistency loss guided by the Projective Motion Blur Model. These benefits are also illustrated in \cref{fig:optim_evolution}, showing a typical evolution of the reblur and the restoration PSNRs during the optimization of a sample image. Note how the PSNR of the restored image gets further increased by optimizing only the reblur loss, thanks to the regularization implicit in the TPN+RN network architecture.  

\Cref{fig:Lai_non_uniform_traj1,fig:Lai_non_uniform_traj2} show examples of estimated trajectories, kernels, and image restorations before and after the trajectory refinement step. %
The first row of each image shows the blurry image and selected patches with their corresponding motion blur kernels. The second row shows the initial restoration image among the restored patches and the kernels obtained from the trajectory predicted by the TPN. The third row shows the same information as the second row, but after the trajectory refinement step. Finally, the last row shows the initial trajectory estimated by the TPN, the trajectory after the refinement step, and the ground truth trajectory. We represent a trajectory as a sequence of 3D points (\textit{pitch}, \textit{yaw}, \textit{roll}), and also show, using different color curves,  the evolution of each rotation angle during the exposure time.  %

\paragraph{Video generation} We estimated the temporal order of the camera trajectory using a simple heuristic. Given a trajectory, we find an extreme point as the farthest from all the others. Then, we generate the trajectory by ordering the points according to the distance to the farthest one. Although this heuristic may not work for sinuous or noisy trajectories, it helps to visualize the results. For example, we can use the ordered trajectory to generate a video from the blurry image by warping the sharp image with the ordered homographies. Sample restorations and video examples obtained with Blur2Seq (w/o traj. optim.) and Blur2Seq are available \href{https://www.youtube.com/playlist?list=PLVXbcH0I4pbnkbG9y5tYyyZWX-aDsS9OA}{here}~\footnote{https://tinyurl.com/mry9jnrd} and \href {https://www.youtube.com/playlist?list=PLVXbcH0I4pbmfkJJFfquW8VeRUbAbVvIY}{here}~ \footnote{https://tinyurl.com/yc4ztej5} respectively.

\begin{table}[ht]
    \centering
    \setlength\tabcolsep{1pt}
    \scriptsize 
    \caption[Results on the Lai nonuniform synthetic dataset with the Blur2Seq method.]{Results on the Lai nonuniform synthetic dataset with the Blur2Seq method. The results improve significantly by applying consistency between the restored and the blurry image according to the Projective Motion Blur Model. \label{tab:Lai_non_uniform_optim}}
    \begin{tabular}{l|c|c|c|c}
    \toprule
                 &  PSNR  & SSIM & LPIPS $\downarrow$ & SI \\
         \midrule
         Blur2Seq w/o traj. optim. &   21.99  & 0.717 & 0.300 & 1807    \\
         Blur2Seq & 23.87 & 0.791 & 0.186  & 3179 \\
    \bottomrule
    \end{tabular}
\end{table}

\newlength{\imgwidthpt}
\setlength{\imgwidthpt}{\textwidth} %

\newlength{\pxsize}
\setlength{\pxsize}{\dimexpr \imgwidthpt * 140 / 1024  \relax} %

\begin{figure*}[ht]
\centering
\begin{tabular}{c}

\hspace{60pt} Blurry \hspace{110pt} Blurry Patches \hspace{50pt} Ground Truth kernels  \\
\begin{tikzpicture}[spy using outlines={rectangle, magnification=4.2, connect spies}]
  \node {\includegraphics[width=0.4\textwidth]{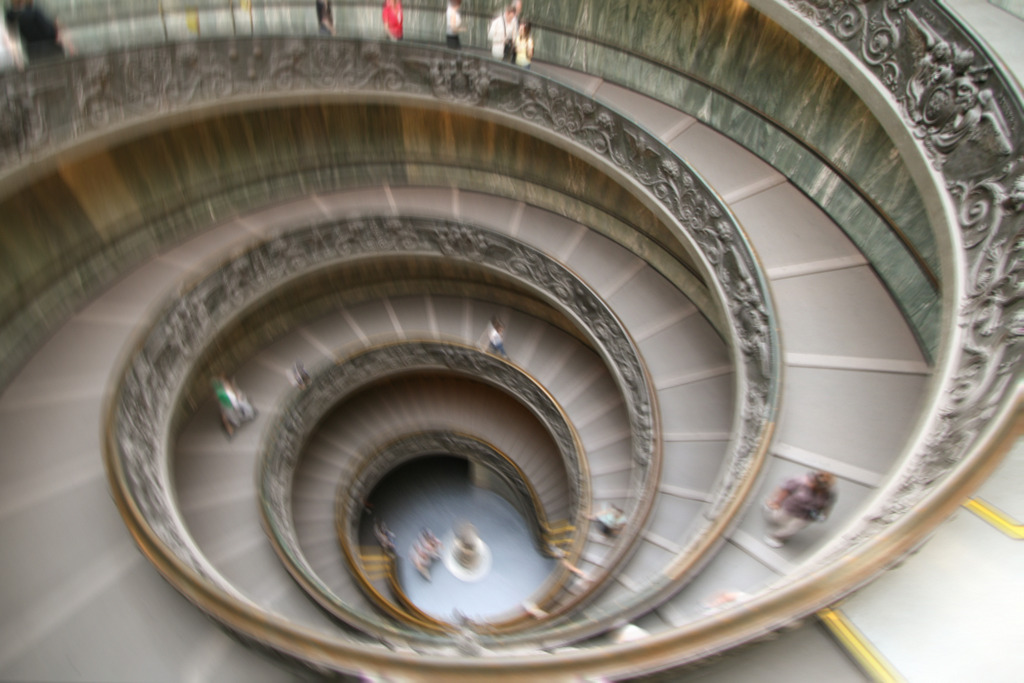}};

  \spy[draw=blue, thick, size=\pxsize]  on (-69.05pt,  47.43pt) in node [left] at (162pt,  33.33pt);
  \spy[draw=red,  line width=4pt, size=\pxsize]  on ( 69.23pt,  47.43pt) in node [left] at (230pt,  33.33pt);
  \spy[draw=green, line width=4pt, size=\pxsize]  on (-69.05pt, -47.59pt) in node [left] at (162pt, -33.33pt);
  \spy[draw=yellow, line width=4pt,size=\pxsize]  on ( 69.23pt, -47.59pt) in node [left] at (230pt, -33.33pt);

  \setlength{\fboxrule}{1pt} %
  \setlength{\fboxsep}{0pt}  %

  \node at (276pt,33pt) {\fcolorbox{blue}{white}{\includegraphics[trim=66 555 895 67, clip]{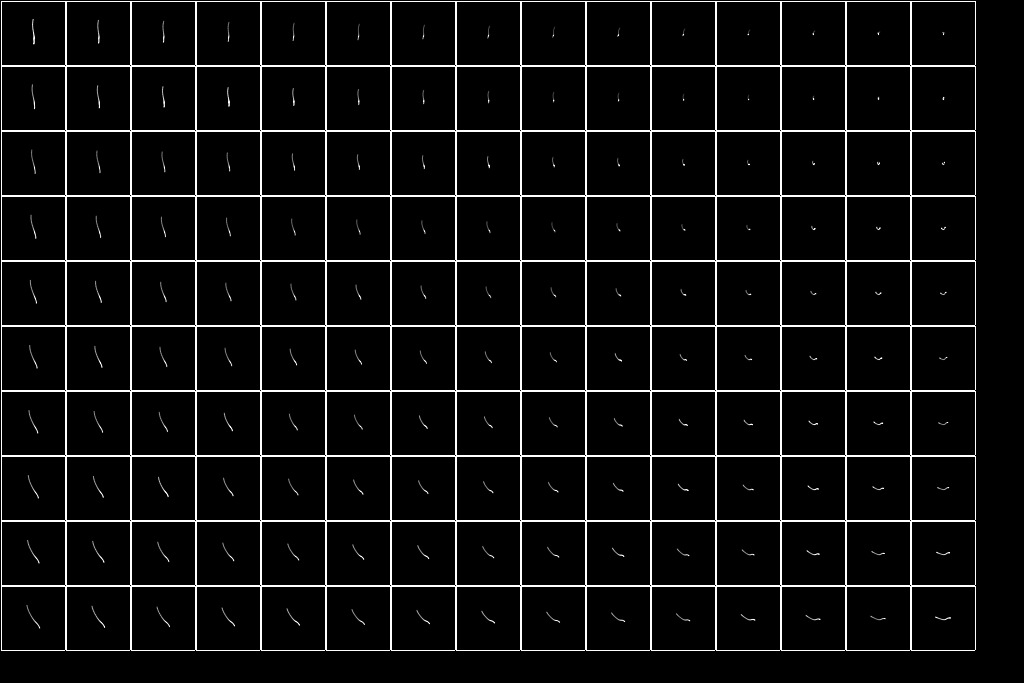}}};
  \node at (344pt,33pt) {\fcolorbox{red}{white}{\includegraphics[trim=912 555 50 67, clip]{Lai_non_uniform_GT_kernels_manmade_02_gyro_01.jpg}}};
  \node at (276pt,-33pt) {\fcolorbox{green}{white}{\includegraphics[trim=66 35 895 586, clip]{Lai_non_uniform_GT_kernels_manmade_02_gyro_01.jpg}}};
  \node at (344pt,-33pt) {\fcolorbox{yellow}{white}{\includegraphics[trim=912 35 50 586, clip]{Lai_non_uniform_GT_kernels_manmade_02_gyro_01.jpg}}};
\end{tikzpicture} \\

\hspace{35pt} Restored (w/o optim)\hspace{50pt} Restored Patches (w/o optim) \hspace{5pt} Restored kernels (w/o optim) \\
\begin{tikzpicture}[spy using outlines={rectangle, magnification=4.2, connect spies}]
  \node {\includegraphics[width=0.4\textwidth]{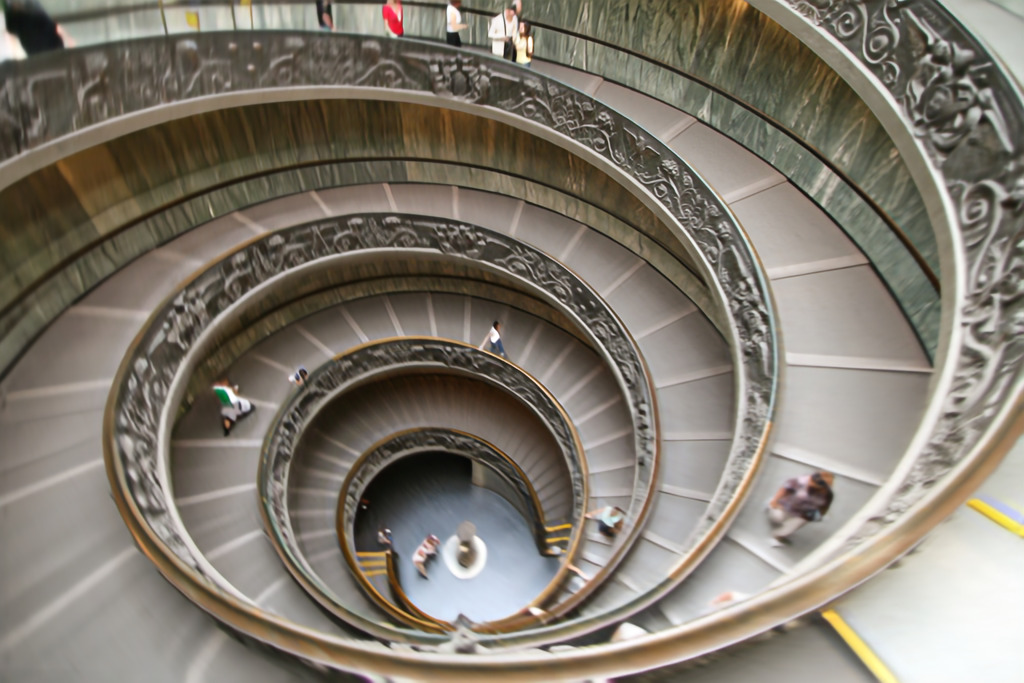}};

  \spy[draw=blue, thick, size=\pxsize]  on (-69.05pt,  47.43pt) in node [left] at (162pt,  33.33pt);
  \spy[draw=red,  line width=4pt, size=\pxsize]  on ( 69.23pt,  47.43pt) in node [left] at (230pt,  33.33pt);
  \spy[draw=green, line width=4pt, size=\pxsize]  on (-69.05pt, -47.59pt) in node [left] at (162pt, -33.33pt);
  \spy[draw=yellow, line width=4pt,size=\pxsize]  on ( 69.23pt, -47.59pt) in node [left] at (230pt, -33.33pt);

  \setlength{\fboxrule}{1pt} %
  \setlength{\fboxsep}{0pt}  %

  \node at (276pt,33pt) {\fcolorbox{blue}{white}{\includegraphics[trim=66 555 895 67, clip]{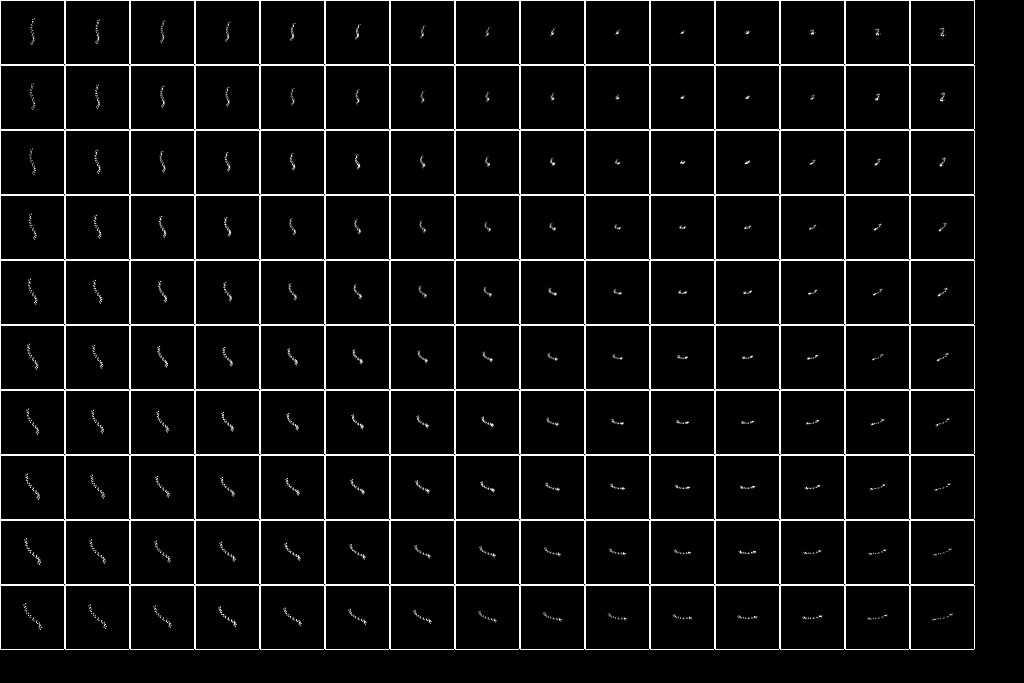}}};
  \node at (344pt,33pt) {\fcolorbox{red}{white}{\includegraphics[trim=912 555 50 67, clip]{traj_est_ill_aug_raw_manmade_02_gyro_01_kernels_found.jpg}}};
  \node at (276pt,-33pt) {\fcolorbox{green}{white}{\includegraphics[trim=66 35 895 586, clip]{traj_est_ill_aug_raw_manmade_02_gyro_01_kernels_found.jpg}}};
  \node at (344pt,-33pt) {\fcolorbox{yellow}{white}{\includegraphics[trim=912 35 50 586, clip]{traj_est_ill_aug_raw_manmade_02_gyro_01_kernels_found.jpg}}};
\end{tikzpicture} \\

\hspace{40pt} Restored \hspace{110pt} Restored Patches \hspace{50pt} Restored kernels  \\
\begin{tikzpicture}[spy using outlines={rectangle, magnification=4.2, connect spies}]
  \node {\includegraphics[width=0.4\textwidth]{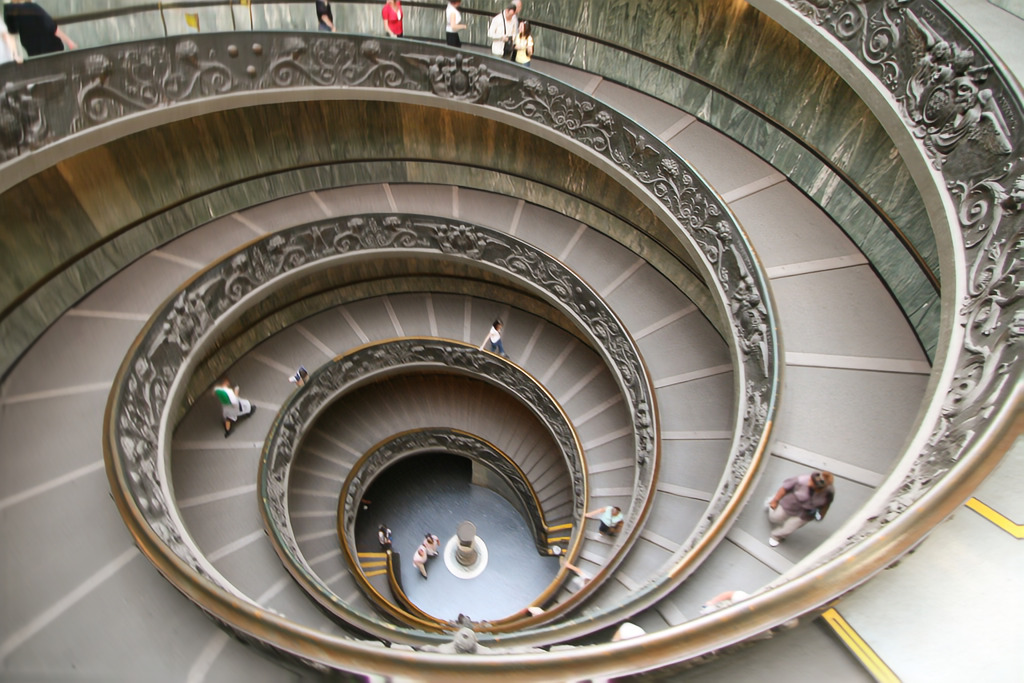}};

  \spy[draw=blue, thick, size=\pxsize]  on (-69.05pt,  47.43pt) in node [left] at (162pt,  33.33pt);
  \spy[draw=red,  line width=4pt, size=\pxsize]  on ( 69.23pt,  47.43pt) in node [left] at (230pt,  33.33pt);
  \spy[draw=green, line width=4pt, size=\pxsize]  on (-69.05pt, -47.59pt) in node [left] at (162pt, -33.33pt);
  \spy[draw=yellow, line width=4pt,size=\pxsize]  on ( 69.23pt, -47.59pt) in node [left] at (230pt, -33.33pt);

  \setlength{\fboxrule}{1pt} %
  \setlength{\fboxsep}{0pt}  %

  \node at (276pt,33pt) {\fcolorbox{blue}{white}{\includegraphics[trim=66 555 895 67, clip]{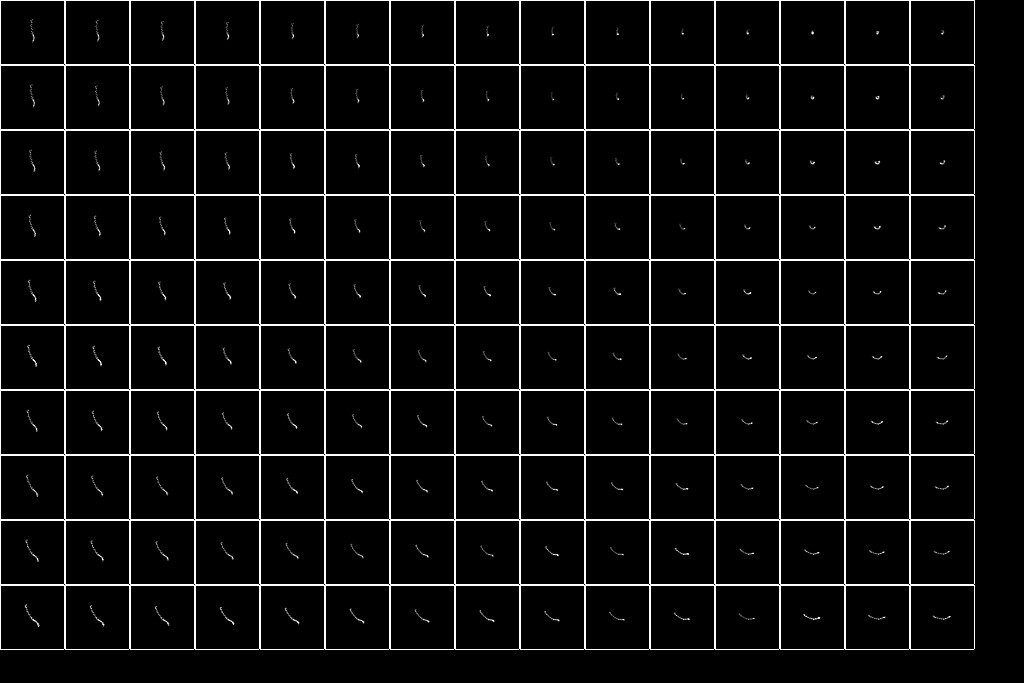}}};
  \node at (344pt,33pt) {\fcolorbox{red}{white}{\includegraphics[trim=912 555 50 67, clip]{traj_est_ill_aug_optim_manmade_02_gyro_01_kernels_found.jpg}}};
  \node at (276pt,-33pt) {\fcolorbox{green}{white}{\includegraphics[trim=66 35 895 586, clip]{traj_est_ill_aug_optim_manmade_02_gyro_01_kernels_found.jpg}}};
  \node at (344pt,-33pt) {\fcolorbox{yellow}{white}{\includegraphics[trim=912 35 50 586, clip]{traj_est_ill_aug_optim_manmade_02_gyro_01_kernels_found.jpg}}};
\end{tikzpicture} \\
Trajectory (w/o optim.) \hspace{80pt}  Trajectory  \hspace{110pt} GT Trajectory \\ 
     \includegraphics[width=0.33\textwidth, trim=45 40 45 45, clip]{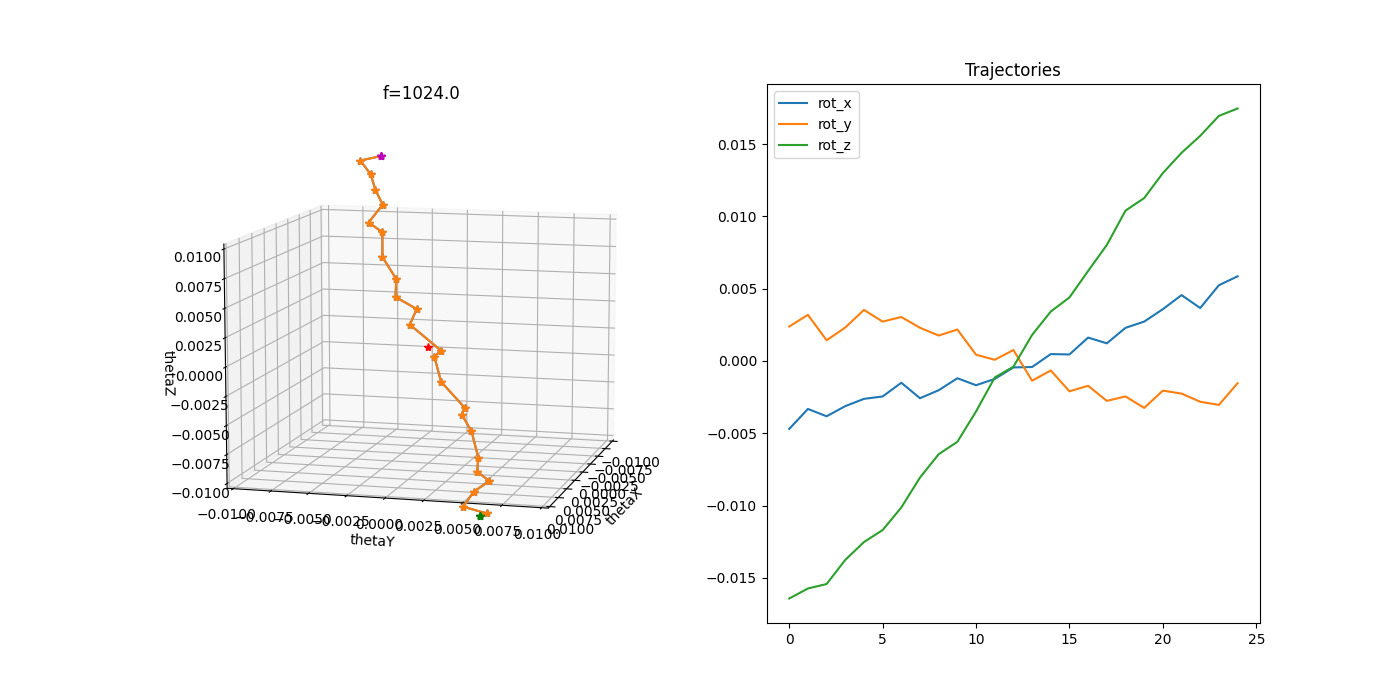} \includegraphics[width=0.33\textwidth, trim=45 40 45 45, clip]{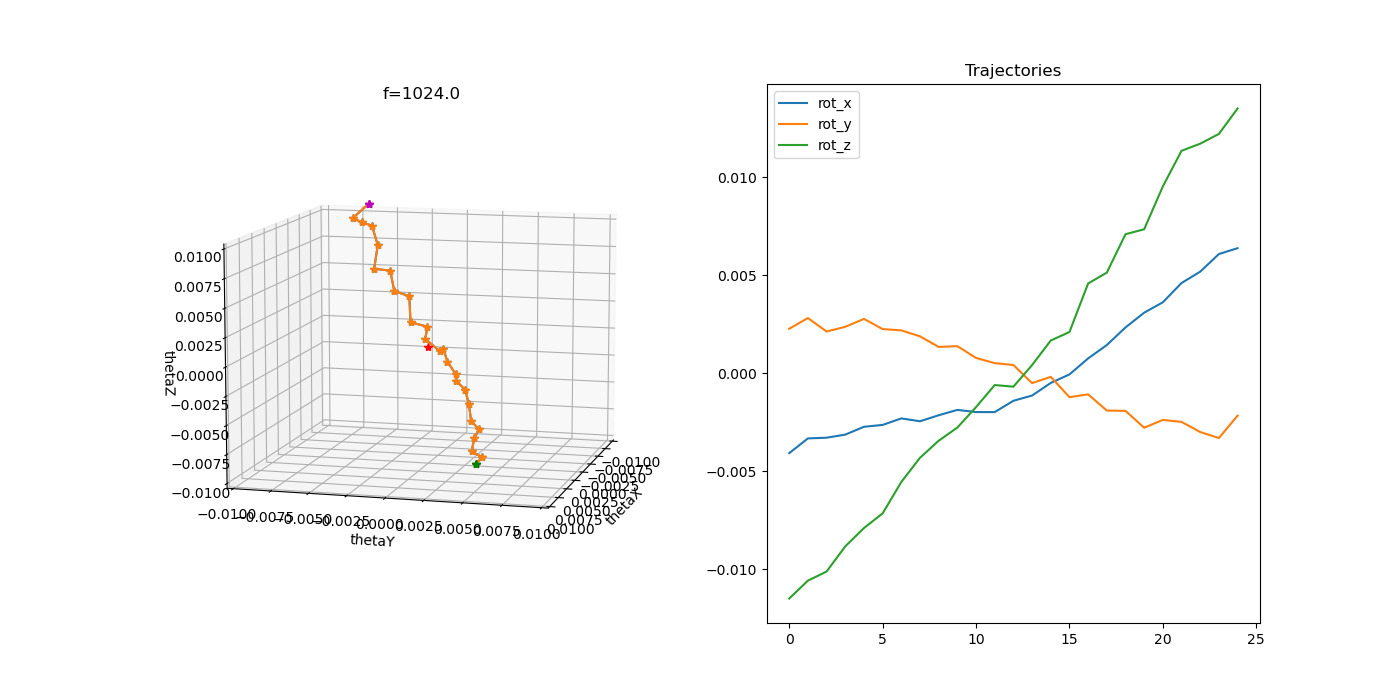}  
        \includegraphics[width=0.33\textwidth, trim=45 40 45 45, clip]{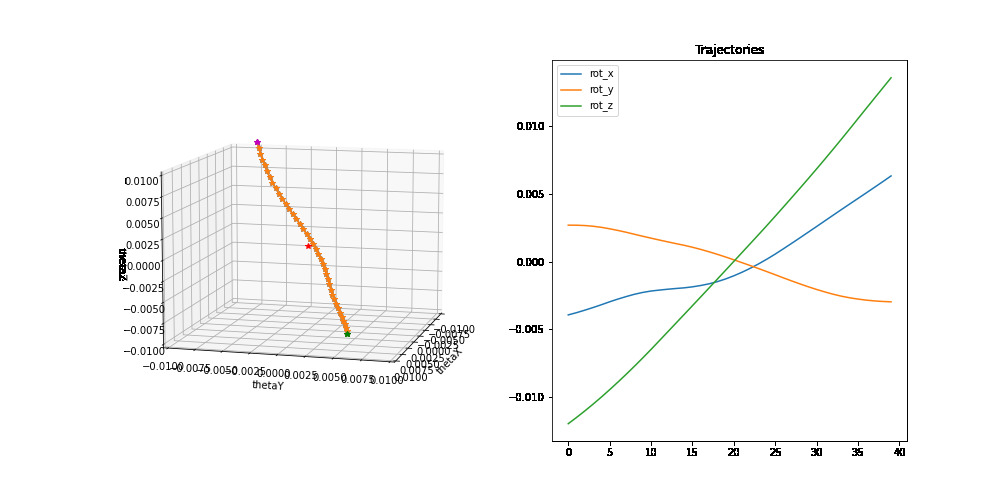} \\  
\end{tabular}
\caption{Kernels and restorations obtained with (3rd row) and without (2nd row) the trajectory refinement step. Kernel estimation is challenging for low textured regions since there is almost no information about the blur.  The global constraint imposed during the optimization improves the estimated kernels and the restored image. The last row shows the initial trajectory estimated by the TPN, the trajectory after the refinement step, and the ground-truth trajectory. We represent a trajectory as a sequence of 3D points (\textit{pitch}, \textit{yaw}, \textit{roll}), and also show, using different color curves,  the evolution of each rotation during the exposure time.  \label{fig:Lai_non_uniform_traj1}} 
\end{figure*}

\begin{figure*}[ht]
\centering
\begin{tabular}{c}

\hspace{60pt} Blurry \hspace{110pt} Blurry Patches \hspace{50pt} Ground Truth kernels  \\
\begin{tikzpicture}[spy using outlines={rectangle, magnification=4.2, connect spies}]
  \node {\includegraphics[width=0.4\textwidth]{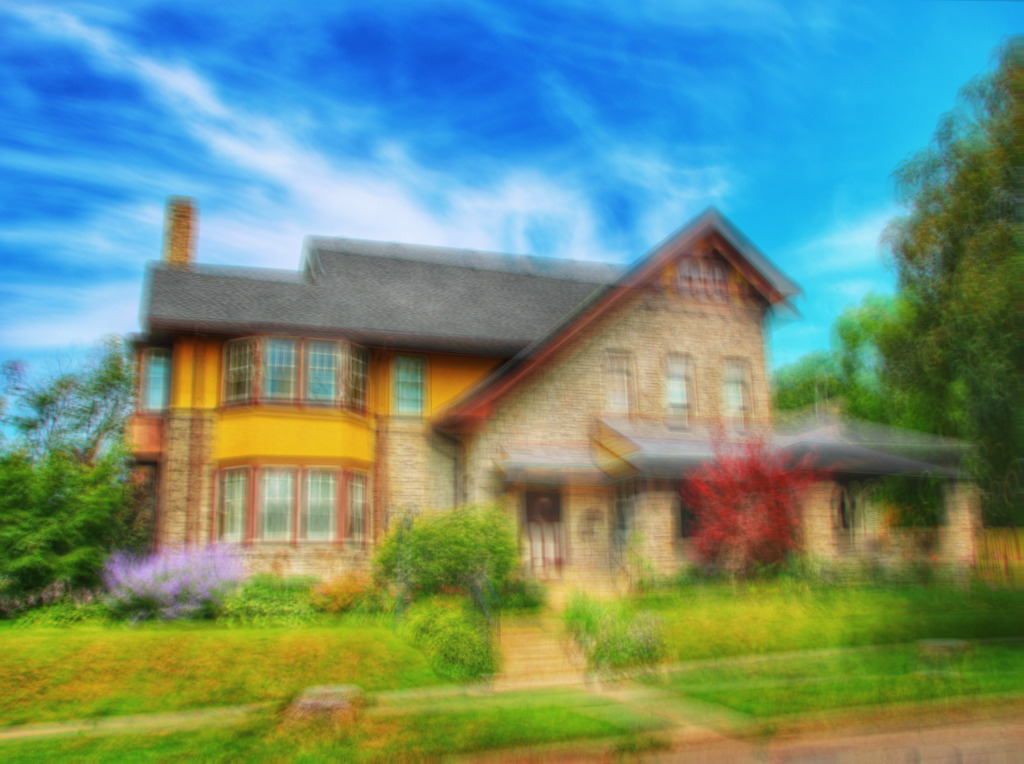}};

  \spy[draw=blue, thick, size=\pxsize]  on (-69.05pt,  47.43pt) in node [left] at (162pt,  34.33pt);
  \spy[draw=red,  line width=4pt, size=\pxsize]  on ( 69.23pt,  47.43pt) in node [left] at (230pt,  34.33pt);
  \spy[draw=green, line width=4pt, size=\pxsize]  on (-69.05pt, -47.59pt) in node [left] at (162pt, -34.33pt);
  \spy[draw=yellow, line width=4pt,size=\pxsize]  on ( 69.23pt, -47.59pt) in node [left] at (230pt, -34.33pt);

  \setlength{\fboxrule}{1pt} %
  \setlength{\fboxsep}{0pt}  %

  \node at (276pt,35pt) {\fcolorbox{blue}{white}{\includegraphics[trim=67 635 895 67, clip]{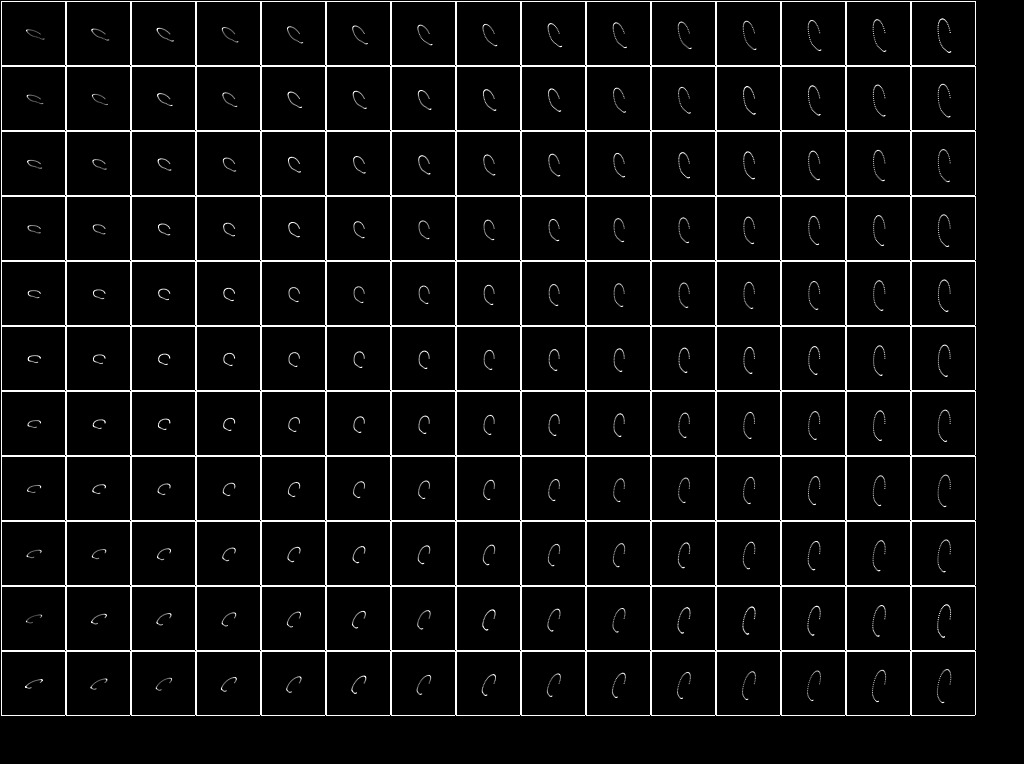}}};
  \node at (344pt,35pt) {\fcolorbox{red}{white}{\includegraphics[trim=912 635 50 67, clip]{Lai_non_uniform_GT_kernels_manmade_04_gyro_02.jpg}}};
  \node at (276pt,-35pt) {\fcolorbox{green}{white}{\includegraphics[trim=67 50 895 652, clip]{Lai_non_uniform_GT_kernels_manmade_04_gyro_02.jpg}}};
  \node at (344pt,-35pt) {\fcolorbox{yellow}{white}{\includegraphics[trim=912 50 50 652, clip]{Lai_non_uniform_GT_kernels_manmade_04_gyro_02.jpg}}};
\end{tikzpicture} \\

\hspace{35pt} Restored (w/o optim)\hspace{50pt} Restored Patches (w/o optim) \hspace{5pt} Restored kernels (w/o optim) \\
\begin{tikzpicture}[spy using outlines={rectangle, magnification=4.2, connect spies}]
  \node {\includegraphics[width=0.4\textwidth]{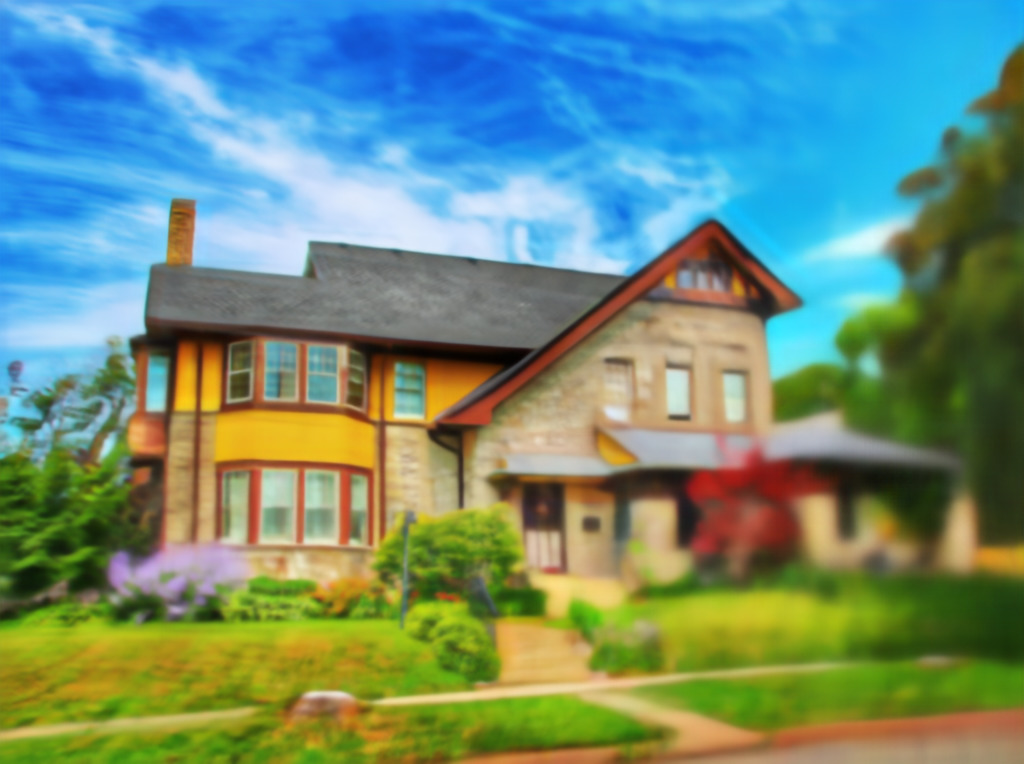}};

  \spy[draw=blue, thick, size=\pxsize]  on (-69.05pt,  47.43pt) in node [left] at (162pt,  34.33pt);
  \spy[draw=red,  line width=4pt, size=\pxsize]  on ( 69.23pt,  47.43pt) in node [left] at (230pt,  34.33pt);
  \spy[draw=green, line width=4pt, size=\pxsize]  on (-69.05pt, -47.59pt) in node [left] at (162pt, -34.33pt);
  \spy[draw=yellow, line width=4pt,size=\pxsize]  on ( 69.23pt, -47.59pt) in node [left] at (230pt, -34.33pt);

  \setlength{\fboxrule}{1pt} %
  \setlength{\fboxsep}{0pt}  %

  \node at (276pt,35pt) {\fcolorbox{blue}{white}{\includegraphics[trim=66 635 895 66, clip]{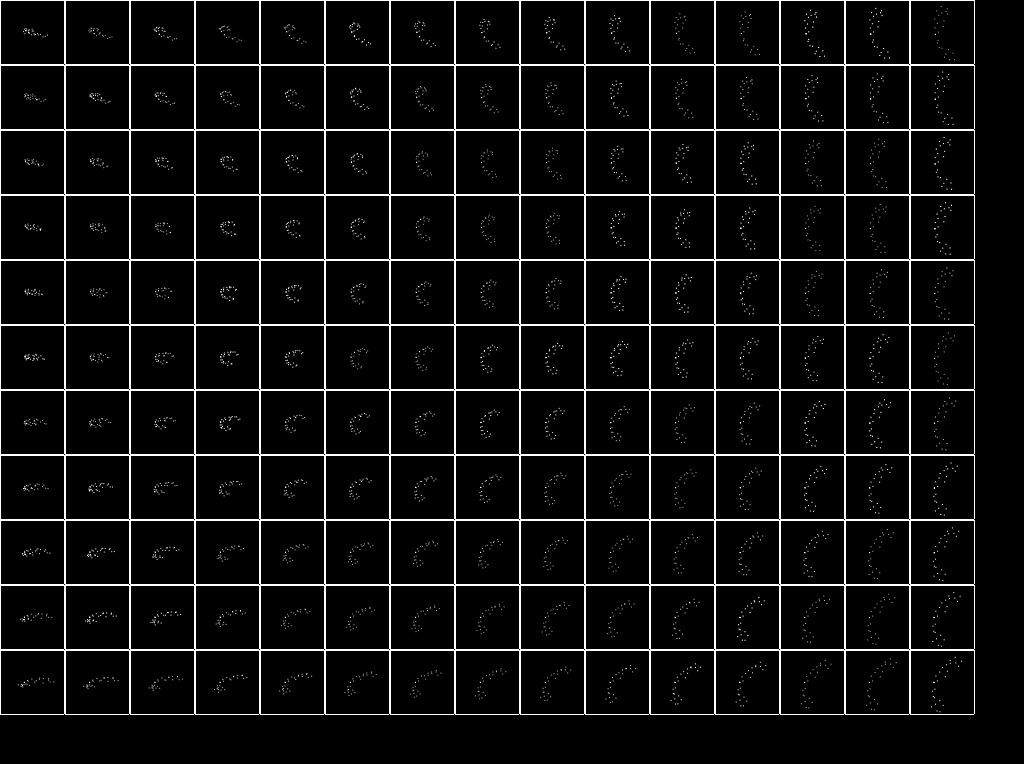}}};
  \node at (344pt,35pt) {\fcolorbox{red}{white}{\includegraphics[trim=912 635 50 65, clip]{traj_est_ill_aug_raw_manmade_04_gyro_02_kernels_found.jpg}}};
  \node at (276pt,-35pt) {\fcolorbox{green}{white}{\includegraphics[trim=66 50 895 651, clip]{traj_est_ill_aug_raw_manmade_04_gyro_02_kernels_found.jpg}}};
  \node at (344pt,-35pt) {\fcolorbox{yellow}{white}{\includegraphics[trim=912 50 50 651, clip]{traj_est_ill_aug_raw_manmade_04_gyro_02_kernels_found.jpg}}};
\end{tikzpicture} \\

\hspace{40pt} Restored \hspace{110pt} Restored Patches \hspace{50pt} Restored kernels  \\
\begin{tikzpicture}[spy using outlines={rectangle, magnification=4.2, connect spies}]
  \node {\includegraphics[width=0.4\textwidth]{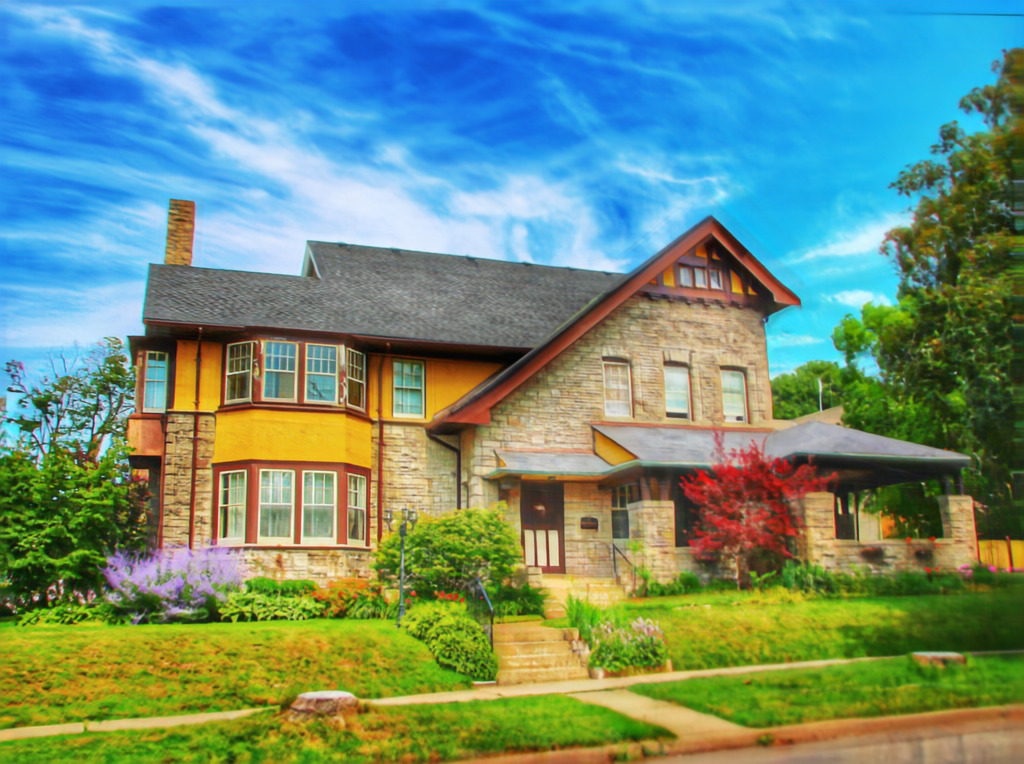}};

  \spy[draw=blue, thick, size=\pxsize]  on (-69.05pt,  47.43pt) in node [left] at (162pt,  34.33pt);
  \spy[draw=red,  line width=4pt, size=\pxsize]  on ( 69.23pt,  47.43pt) in node [left] at (230pt,  34.33pt);
  \spy[draw=green, line width=4pt, size=\pxsize]  on (-69.05pt, -47.59pt) in node [left] at (162pt, -34.33pt);
  \spy[draw=yellow, line width=4pt,size=\pxsize]  on ( 69.23pt, -47.59pt) in node [left] at (230pt, -34.33pt);

  \setlength{\fboxrule}{1pt} %
  \setlength{\fboxsep}{0pt}  %

  \node at (276pt,35pt) {\fcolorbox{blue}{white}{\includegraphics[trim=66 635 895 66, clip]{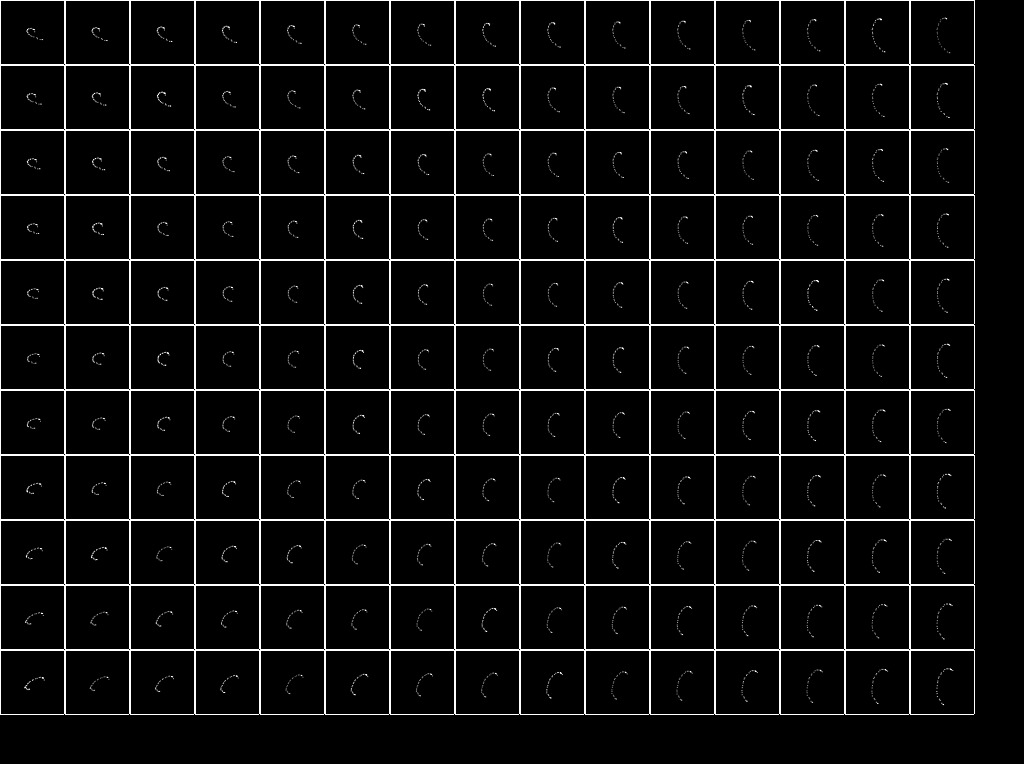}}};
  \node at (344pt,35pt) {\fcolorbox{red}{white}{\includegraphics[trim=912 635 50 65, clip]{traj_est_ill_aug_optim_manmade_04_gyro_02_kernels_found.jpg}}};
  \node at (276pt,-35pt) {\fcolorbox{green}{white}{\includegraphics[trim=66 50 895 651, clip]{traj_est_ill_aug_optim_manmade_04_gyro_02_kernels_found.jpg}}};
  \node at (344pt,-35pt) {\fcolorbox{yellow}{white}{\includegraphics[trim=912 50 50 651, clip]{traj_est_ill_aug_optim_manmade_04_gyro_02_kernels_found.jpg}}};
\end{tikzpicture} \\
Trajectory (w/o optim.) \hspace{80pt}  Trajectory  \hspace{110pt} GT Trajectory \\ 
     \includegraphics[width=0.33\textwidth, trim=45 40 45 45, clip]{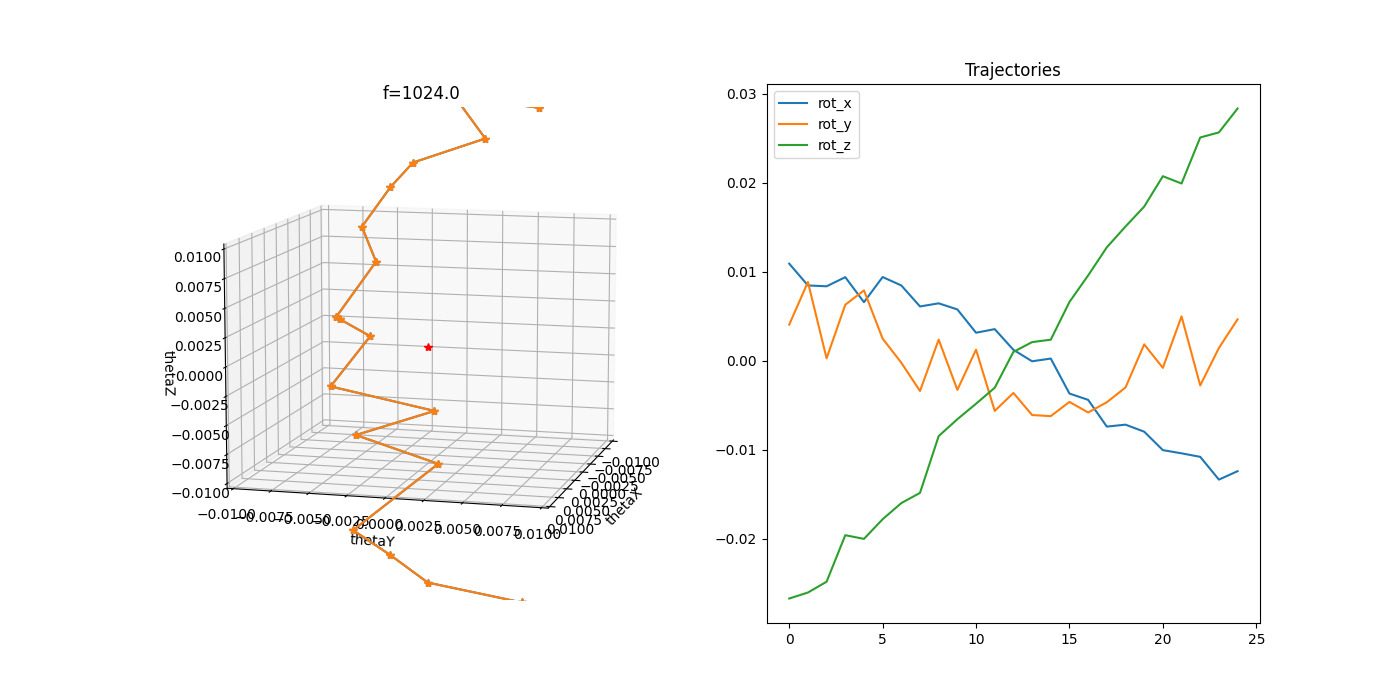} \includegraphics[width=0.33\textwidth, trim=45 40 45 45, clip]{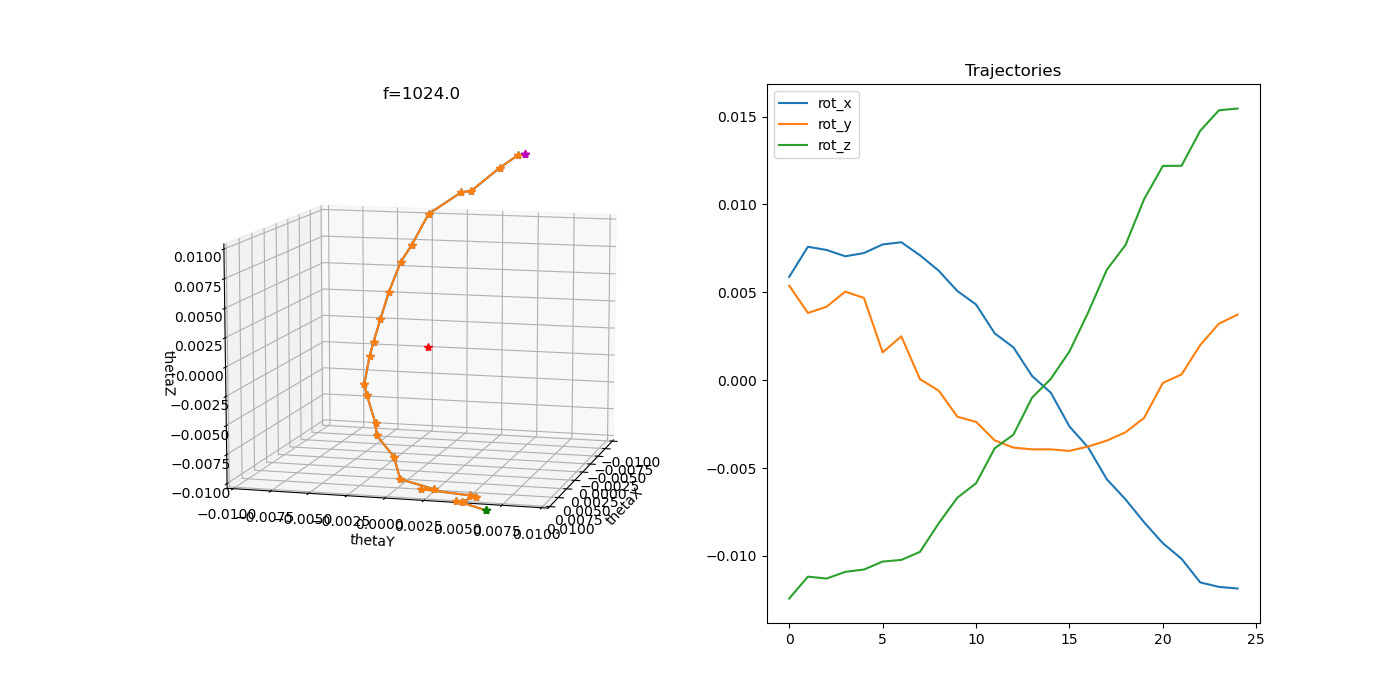}  
        \includegraphics[width=0.33\textwidth, trim=45 40 45 45, clip]{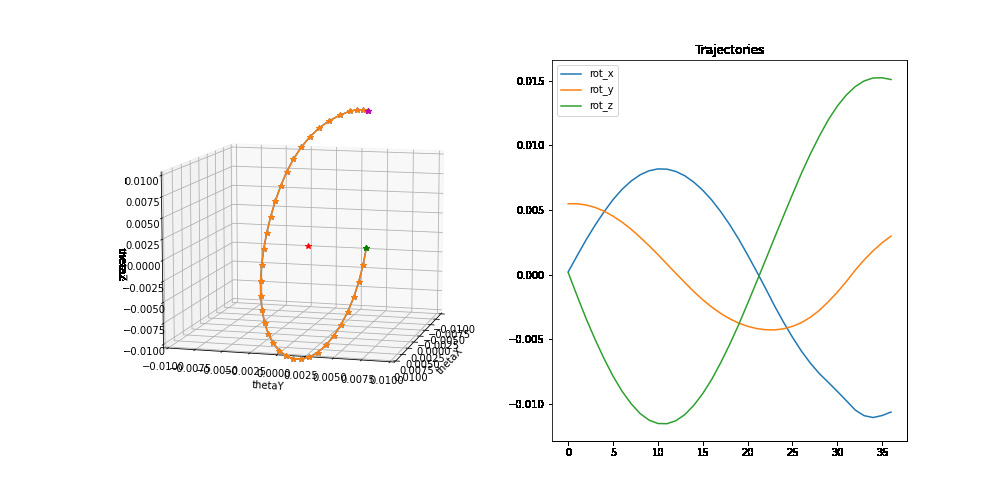} \\  
\end{tabular}
\caption{Kernels and restorations obtained with (3rd row) and without (2nd row) the trajectory refinement step. Kernel estimation is challenging for low textured regions since there is almost no information about the blur.  The global constraint imposed during the optimization improves the estimated kernels and the restored image. The last row shows the initial trajectory estimated by the TPN, the trajectory after the refinement step, and the ground-truth trajectory. We represent a trajectory as a sequence of 3D points (\textit{pitch}, \textit{yaw}, \textit{roll}), and also show, using different color curves,  the evolution of each rotation during the exposure time.  \label{fig:Lai_non_uniform_traj2}} 
\end{figure*}

\begin{figure*}[h]
    \centering
    \begin{tabular}{cc}
    Blurry - PSNR=15.72   & \cite{vasu2017local} - PSNR=17.23 \\
    \includegraphics[width=0.45\linewidth]{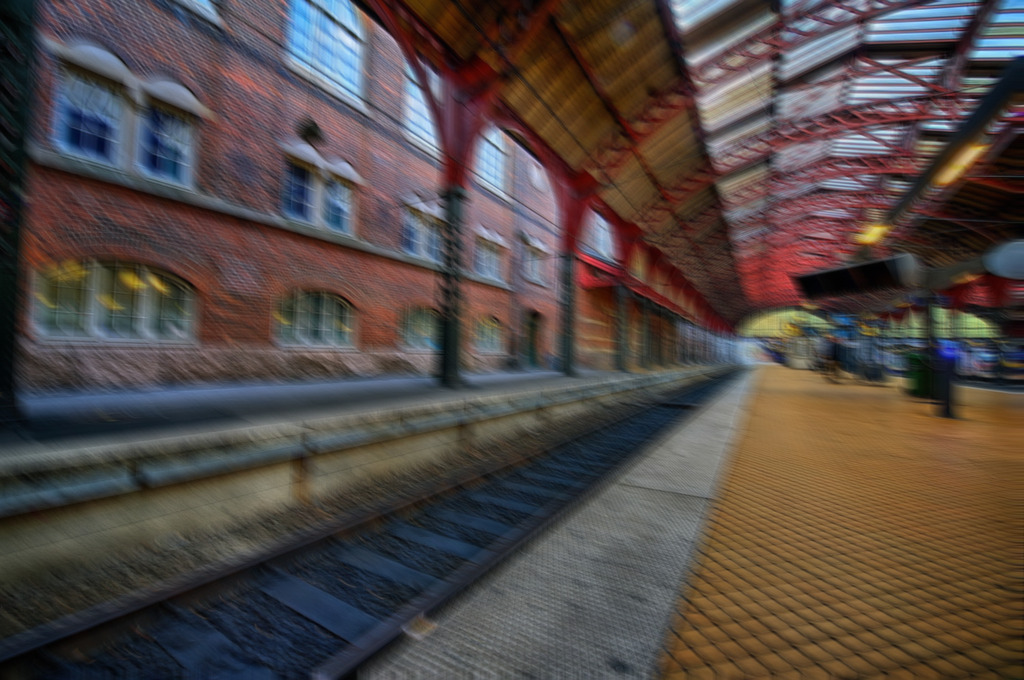}     &  
\includegraphics[width=0.45\linewidth]{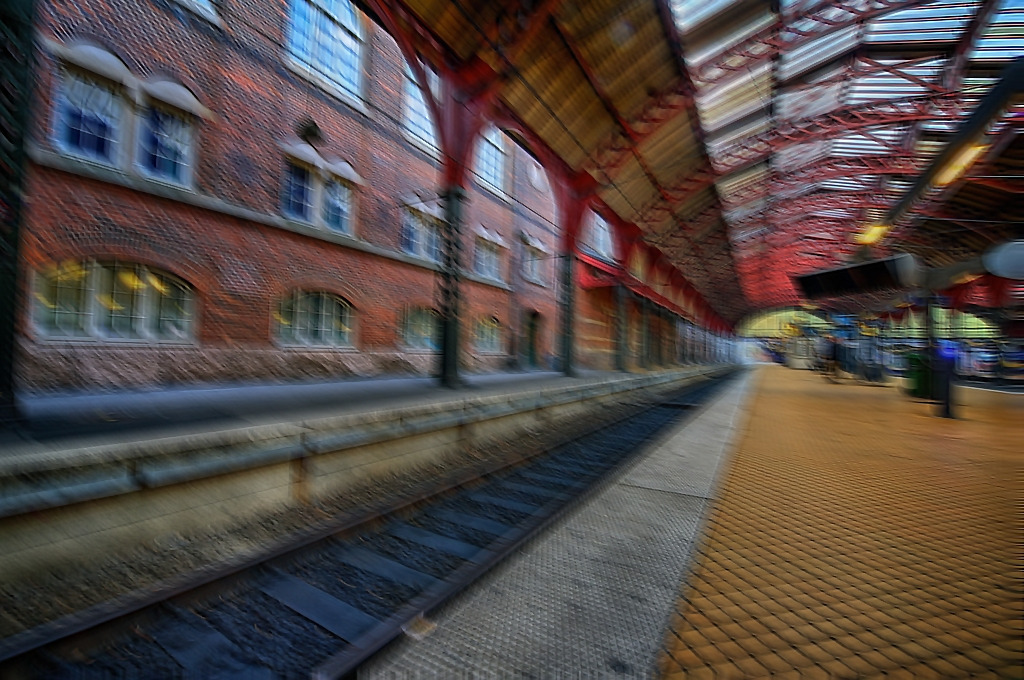} \\
    \cite{li2023self} - PSNR=17.91 &   Blur2Seq (ours) - PSNR=19.58  \\
\includegraphics[width=0.45\linewidth]{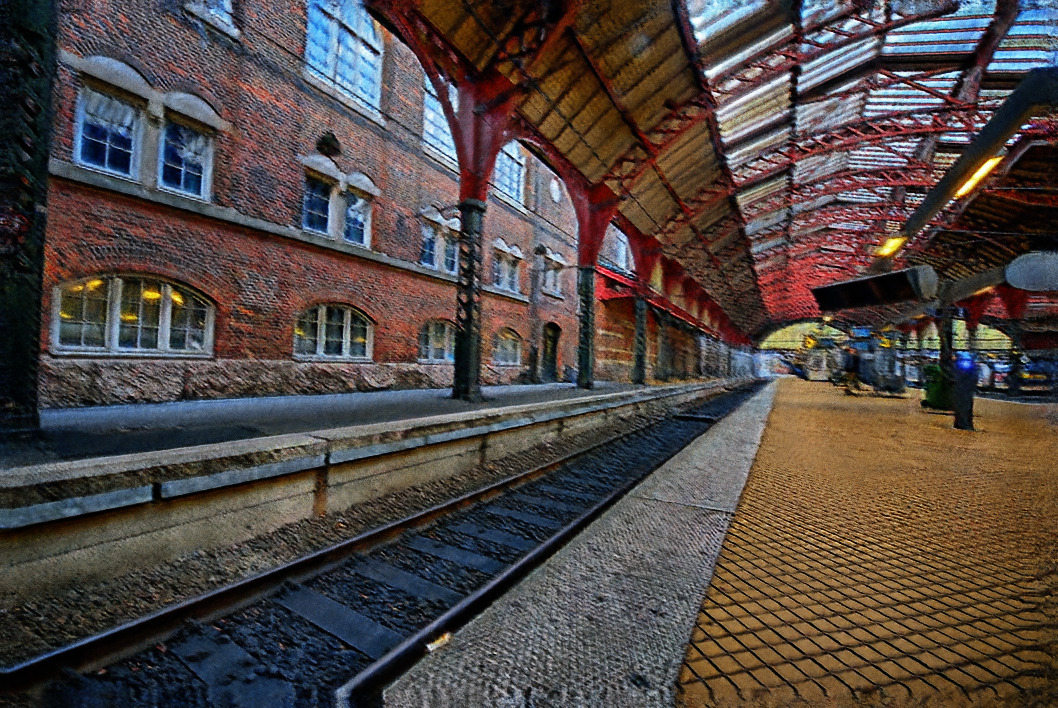}     & 
\includegraphics[width=0.45\linewidth]{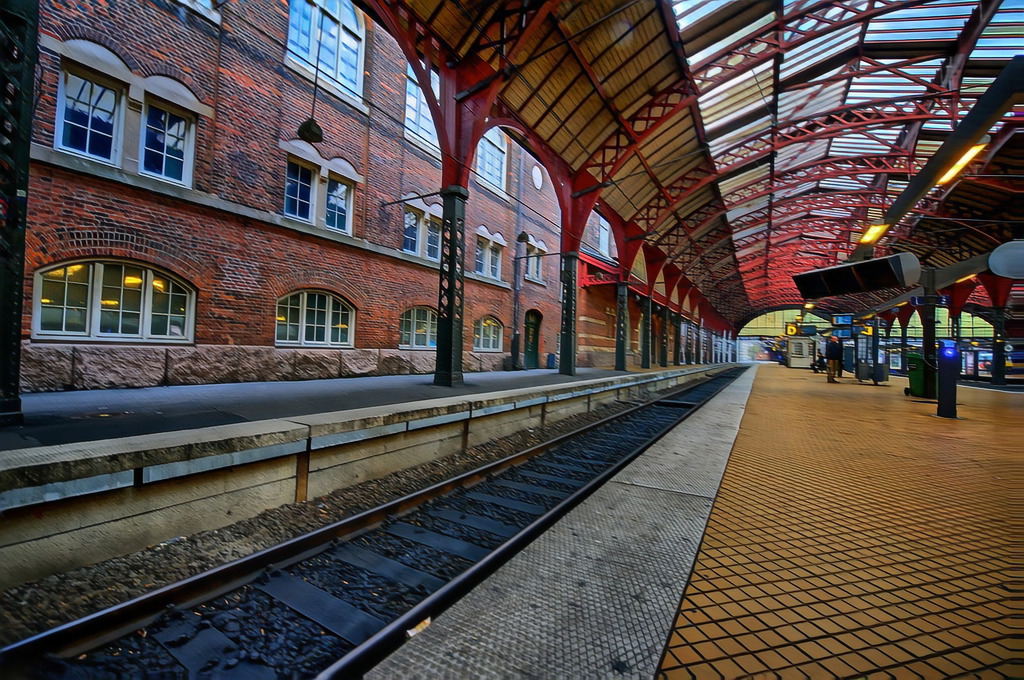} \\ 
   Kernels \cite{li2023self} &   Blur2Seq (ours)  kernels  \\
 \raisebox{5ex}{\includegraphics[width=0.45\linewidth]{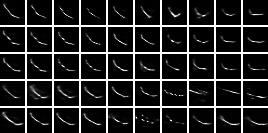}}     & 
\includegraphics[width=0.45\linewidth]{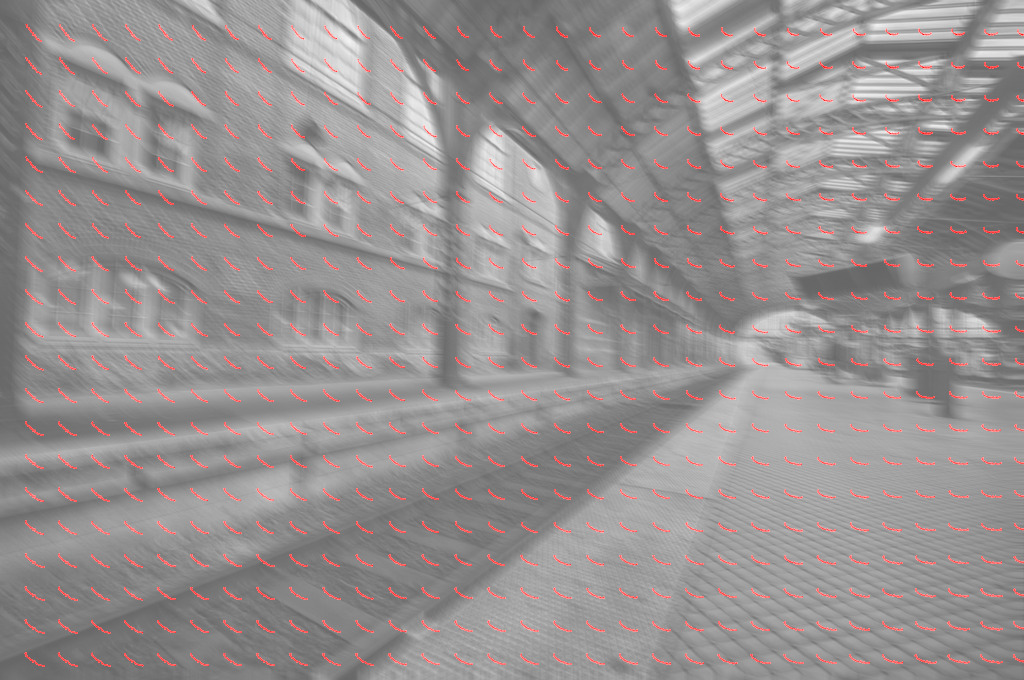}         
    \end{tabular}
    \caption[Comparison with the restorations provided by \cite{vasu2017local} and \cite{li2023self} on a sample image from Lai nonuniform dataset \citep{lai2016comparative}.]{Comparison with the restorations provided by \cite{vasu2017local} and \cite{li2023self} on a sample image from Lai nonuniform dataset \citep{lai2016comparative}. The video associated with our restoration can be seen \href{https://youtu.be/BneoSKJCzqE}{here}.}
    \label{fig:LNU_manmade_05_04}
\end{figure*}

\Cref{fig:LNU_manmade_05_04} compares the restorations obtained with Blur2Seq with other per-image optimization methods. For this highly textured image, our approach consistently recovers finer details across all regions. In contrast, the method proposed by \cite{vasu2017local} tends to converge to a delta kernel, whereas the approach by \cite{li2023self} does not impose any constraint on the kernel map, resulting in spatially irregular variations.

\paragraph{Effect of the Efficient Blur Creation Module}

\cref{tab:EffBCM_effect} shows the computational times of restoring an image and performing a training epoch for different implementations of the blur operator. The implementation of homographies using the Efficient Blur Creation Module enables a reduction of more than four times in image restoration compared to a base implementation using \textit{kornia} and a reduction of one-third compared to the Blur Creation Module \citep{zhang2021exposure}.

\begin{table}[h]
\centering
\caption{Average elapsed times on restoring an image, and performing a training epoch for 80 images of size 320x320    \label{tab:EffBCM_effect}}
\begin{tabular}{l|c|c}
\toprule
     Method & Restore & Train. epoch \\
     \hline
     Baseline\footnotemark  & 1.18 s & 28.61 s \\ 
     BCM     &   0.35 s  &  18.74 s   \\ 
     Eff. BCM    & 0.24 s  &  13.2 s   \\ 
\bottomrule
\end{tabular}
\end{table}
\footnotetext{Implementation that computes the homographies using \href{https://www.kornia.org/}{kornia}.}

\FloatBarrier

\subsection{Results on real datasets}

In the following, we evaluate the results of Blur2Seq on two datasets of real images: K\"{o}hler and RealBlur. We restrict our comparison to methods that can handle spatially varying blur, excluding approaches that explicitly assume spatially uniform blur. While the latter achieve excellent performance on the K\"{o}hler dataset—where blur is only slightly nonuniform—their performance deteriorates significantly when camera motion produces blur that can no longer be approximated as uniform.

\begin{table*}[h]
    \caption[Performance comparison of single-pass deblurring networks on datasets of real images.]{Performance comparison of single-pass deblurring networks capable of restoring blur that varies spatially across the image on datasets of real images.  We report the metrics average results for the  K\"{o}hler \citep{kohler2012recording} and RealBlur \citep{rim_2020_ECCV} datasets. \colorbox{Cyan1}{Best} and \colorbox{Yellow1}{second-best} values for each column are highlighted. 
      \label{tab:kohler_results}}  
  \setlength\tabcolsep{1pt} %
  \centering
  \begin{tabular}{l|c c c c|c c c c}
    \toprule 
           & \multicolumn{4}{|c|}{K\"{o}hler} & \multicolumn{4}{|c}{RealBlur} \\
               \hline
             & PSNR & SSIM & LPIPS $\downarrow$ & SI &  PSNR & SSIM & LPIPS $\downarrow$ & SI \\
    \hline   
    DeepDeblur \citep{Nah_2017_CVPR}  &   25.66 & 0.763 & 0.341 & 337 & 28.06 & 0.855 & 0.186 & 1577 \\
    SRN \citep{tao2018scale} & 26.91 & 0.789 & 0.327 & 314 & 28.56 & 0.867 & 0.151 & 1850  \\
    DMPHN 1-2-4 \citep{Zhang_2019_CVPR}  &  24.35& 0.697 & 0.698 & 257 & 27.78 & 0.841 & 0.197 & 1412  \\
    DeblurGANv2 Inc. \citep{kupyn2019deblurgan}  & 26.95 & 0.788 &0.295 & 432 &  28.70 & 0.866 & \colorbox{Cyan1}{0.139} & 1571 \\
    MIMO-UNet \citep{cho2021rethinking} &  25.21  & 0.746 & 0.361 & 222 & 27.76 & 0.836 & 0.189 & 1497 \\
    MIMO-UNet+ \citep{cho2021rethinking}  & 25.05  & 0.746 & 0.369 & 213 & 27.64 & 0.836 & 0.199 & 1483   \\
    MPRNet \citep{Zamir2021MPRNet}&  26.16 & 0.779 & 0.328  & 302 & 28.70 & 0.873 & 0.153 & 1748 \\
    NAFNet \citep{NAFNet} & 25.97 & 0.778 & 0.309 & 295 & 28.32 & 0.857 & 0.164 & 1592  \\         
     Motion-ETR \citep{zhang2021exposure}    &     25.42 & 0.754 & 0.358  & 211 & 28.08 & 0.849 & 0.180 & 1370 \\       
J-MKPD \citep{carbajal2023blind} &   \colorbox{Yellow1}{28.65} &  \colorbox{Yellow1}{0.832} &  \colorbox{Cyan1}{0.250} & \colorbox{Yellow1}{939} & 28.72 & \colorbox{Cyan1}{0.878} & \colorbox{Yellow1}{0.147} & \colorbox{Cyan1}{3323} \\ 
\hline
Blur2Seq (w/o traj. optim) (ours)    &   \colorbox{Cyan1}{28.87} & \colorbox{Cyan1}{0.834} & \colorbox{Yellow1}{0.284} & \colorbox{Cyan1}{965} & \colorbox{Cyan1}{29.08} & \colorbox{Yellow1}{0.874} & 0.162 & \colorbox{Yellow1}{2815} \\
    \bottomrule 
  \end{tabular}
\end{table*}

\begin{table*}[h]
    \caption{Performance comparison of methods that use the Projective Motion Blur Model and optimize its parameters per image. We report the metrics average results for the  K\"{o}hler (\cite{kohler2012recording}) and RealBlur (\cite{rim_2020_ECCV}) datasets of real images. %
      \label{tab:kohler_results_optim}}  
  \setlength\tabcolsep{1pt} %
  \centering
  \scriptsize
  \begin{tabular}{l|l|c c c| c c c}
    \toprule
    \multicolumn{2}{c}{} &  \multicolumn{3}{|c|}{K\"{o}hler } &  \multicolumn{3}{c}{RealBlur} \\
          \multicolumn{2}{c}{} &  \multicolumn{3}{|c|}{All images} & \multicolumn{3}{c}{}\\
          \hline
           & \hspace{1cm} Model  & PSNR & SSIM & LPIPS $\downarrow$ & PSNR & SSIM & LPIPS $\downarrow$ \\
    \hline   
    \citet{whyte2010nonuniform}& Projective Motion Blur Model   & 28.31   & 0.819 & 0.257  & -  & -  &  - \\
    \citet{xu2013unnatural} & Projective Motion Blur Model     & - &  - &  -  & 27.14 & 0.830 & - \\
    \citet{vasu2017local}  & Projective Motion Blur Model & \colorbox{Cyan1}{30.29} & \colorbox{Cyan1}{0.87} & \colorbox{Cyan1}{0.217}  & - &  - & -  \\
Blur2Seq (ours)  &   Projective Motion Blur Model     &   \colorbox{Yellow1}{29.62} & \colorbox{Yellow1}{0.854} & \colorbox{Yellow1}{0.247} & \colorbox{Cyan1}{29.01}  & \colorbox{Cyan1}{0.872}  &  \colorbox{Cyan1}{0.155} \\
    \bottomrule 
  \end{tabular}
\end{table*}

\Cref{tab:kohler_results} compares the results of single forward-pass deblurring networks with the proposed method without the trajectory optimization step. The proposed method Blur2Seq (w/o traj. optim.) ranks second in the K\"{o}hler dataset and first on the RealBlur dataset.

\Cref{tab:kohler_results_optim} compares PMBM-based methods that optimize their parameters per image. Comparison with these optimization methods is difficult when the restored images are unavailable due to long inference times, the need for image-specific hyperparameter tuning, and the lack of publicly available or up-to-date implementations in others. In particular, we were unable to conduct an extensive comparison with the methods that inspired this work \citep{gupta_mdf_deblurring, whyte2010nonuniform}, limiting it to the restoration samples provided by the authors. We show those examples in the Supplementary Material. Among the compared methods, the proposed method is the only non-variational and ranks second on the K\"{o}hler dataset and first on the RealBlur dataset. In addition to competitive deblurring performance, our method estimates the camera trajectories, whose usefulness we have illustrated by generating the corresponding motion videos of the scene. Moreover, it is the only approach that leverages a learned image prior, rather than a hand-crafted one, which is seamlessly integrated with the degradation model through a plug-and-play strategy.

\subsection{Impact of training on a PMBM-based dataset} The experiments presented so far compare the results of {\em Blur2Seq} with other approaches by running the models provided by the authors. Following the current trend, most end-to-end deblurring networks are trained on the GoPro dataset. Since our method is blind for testing but requires ground truth trajectories for training, we cannot effectively train on the GoPro dataset. To analyze the impact on generalization of training on the PMBM-based dataset, we selected one of the best-performing end-to-end deblurring networks (NAFNet (\cite{chen2022simple})), originally trained on GoPro, and retrained it on our dataset. Deblurring results on Lai nonuniform, K\"{o}hler and RealBlur datasets are presented in \cref{tab:NAFNet_our_dataset}. The first observation is that NAFNet achieves higher generalization performance when training on our dataset, rather than on the GoPro dataset. Regarding the comparison of NAFNet and our model when both are trained on our dataset, NAFNet performs slightly better on the K\"{o}hler test set; this results from the convolutional nature of the network, which is well adapted to locally uniform blur, as the one presented in K\"{o}hler's images. However, our restorations look sharper, which is consistently captured by the Sharpness Index metric.  Regarding Lai's nonuniform and Realblur datasets, which exhibit stronger spatial variation of the blur kernels, our method outperforms NafNet. In general, when the camera motion is large enough for the blur map to be considered nonuniform, our method performs better. %

\begin{table*}[h]
\centering
\scriptsize
\caption{Quantitative comparison on three benchmark datasets.}
\label{tab:NAFNet_our_dataset}
\setlength\tabcolsep{3pt}
\begin{tabular}{l ccc ccc ccc}
\toprule
\textbf{Method} & \multicolumn{3}{c}{\textbf{Lai Nonuniform}} & \multicolumn{3}{c}{\textbf{Köhler}} & \multicolumn{3}{c}{\textbf{RealBlur}} \\
\cmidrule(r){2-4} \cmidrule(r){5-7} \cmidrule(r){8-10}
 & PSNR ↑ & SSIM ↑ & SI ↑ & PSNR ↑ & SSIM ↑ & SI ↑ & PSNR ↑ & SSIM ↑ & SI ↑\\
\midrule
NAFNet (trained on GoPro) & 21.25 & 0.668 &  &25.97 & 0.778 & 280 & 28.32 & 0.857 & 1592 \\
NAFNet & 23.27 & 0.759 & 2747 & \textbf{30.82} & \textbf{0.862} & 607 & 28.71 & 0.871 & 2310 \\
Blur2Seq (ours) & \textbf{23.46} & \textbf{0.773} & \textbf{3242} & 29.62 & 0.854 & \textbf{965} & \textbf{29.01} & \textbf{0.872} & \textbf{2881} \\
\bottomrule
\end{tabular}
\end{table*}

\section{Conclusions}

We introduced a model-based blind deblurring method that explicitly leverages the camera trajectory as the underlying representation of motion blur. Our approach enables accurate image restoration in a single forward pass by jointly estimating this trajectory and performing non-blind deconvolution. A trajectory refinement stage further enforces consistency between the blurry and restored images, while the explicit trajectory estimation allows us to generate videos of the acquisition process, providing an additional output. Experimental results demonstrate the effectiveness of the method on synthetic and real images, particularly on challenging spatially variant blur such as roll motion. Overall, this work bridges the gap between classical variational model-based approaches and end-to-end neural methods. To the best of our knowledge, this is the first work that employs the PMBM as a forward degradation model within a neural network, while also regularizing the solution with a learned image prior combined with an unrolled plug-and-play strategy. Furthermore, the trajectory optimization stage naturally connects with self-supervised learning strategies, since no ground truth is available during this refinement process.

\section*{Acknowledgments}
The experiments presented in this paper were conducted using ClusterUY (https://cluster.uy) and GPU servers of MAP5 - CNRS - Université Paris Cité, funded by PEPR PDE AI project and kindly provided by Joan Glaunes. We thank Charles Laroche and Matías Tassano for fruitful discussions. AA thanks Joan Glaunes for fruitful discussions. 

\section*{Data Availability Statement}

The results presented in this article were generated using open and publicly available datasets:  
K\"{o}hler (benchmark for camera-shake deblurring) \url{https://webdav.tuebingen.mpg.de/pixel/benchmark4camerashake},  
RealBlur (real-world blur dataset) \url{https://github.com/rimchang/RealBlur},  
and Lai (naturally blurred images without ground truth) \url{https://www.wslai.net/publications/deblur_study}.  

The synthetic training set was created from the publicly available COCO (2017) training set \url{https://cocodataset.org/#download}.  
The code used to generate the synthetic training data is also publicly available at \url{https://github.com/GuillermoCarbajal/Blur2Seq}.  
\FloatBarrier

\bibliography{referencias}
\bibliographystyle{sn-basic}

\end{document}


\title{Blur2seq: Blind Deblurring and Camera Trajectory Estimation from a Single Camera Motion-blurred Image \\
        Supplementary Material}
\author{Guillermo Carbajal, Andrés Almansa, Pablo Musé}

\maketitle

\section{Camera poses parameterization}

In this work, we parameterize the camera poses with 3 degrees of freedom as in \citet{whyte2010nonuniform}. The pertinence of this approximation for modeling camera shake has been previously examined by \cite{whyte2010nonuniform} and  \cite{kohler2012recording}. At a given time $t_s$, the camera pose is represented by a 6D vector $\mathbf{p}^{(t_s)} = [\theta_x,\theta_y,\theta_z, t_x, t_y, t_z]^T$, where $\theta_{(\cdot)}$ and $t_{(\cdot)}$ denote the rotation and translation components, respectively. The ``rotations only'' approximation reduces this vector by compensating small translations as equivalent rotations:
\begin{equation}
    \mathbf{p}^{(t_s)} = \left[ \begin{matrix}
        \theta_x \\
        \theta_y \\
        \theta_z \\
        t_x \\
        t_y \\
        t_z \\
    \end{matrix}\right] \rightarrow \left[ \begin{matrix}
        \theta_x + \arcsin(t_y/d) \\
        \theta_y + \arcsin(-t_x/d) \\
        \theta_z \\
        0 \\
        0 \\
        0 \\
    \end{matrix}\right], 
\end{equation}
where $d$ denotes the distance from the camera to a planar or distant scene. 

\begin{figure}[ht]
\centering
\includegraphics[width=0.49\linewidth]{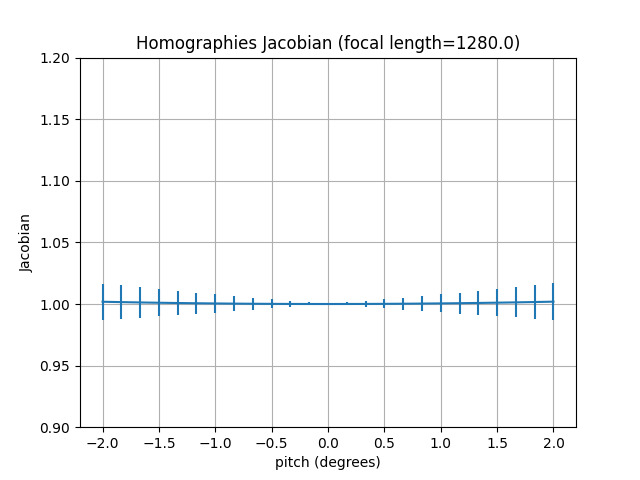}
\includegraphics[width=0.49\linewidth]{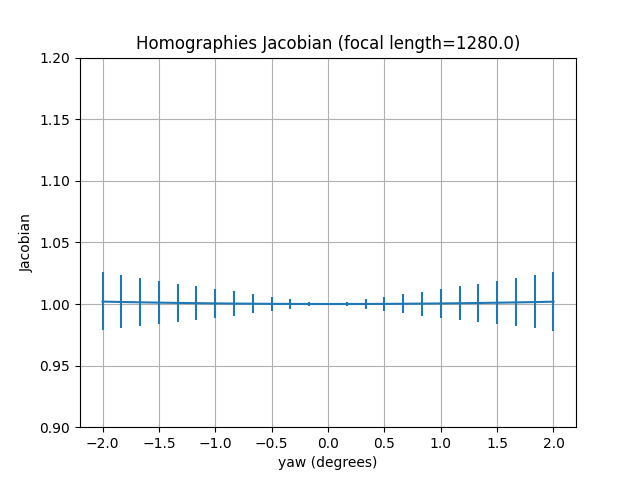}
    \caption{Variation in the Jacobian determinant for pure pitch and yaw rotations. The bar plots show the mean determinant of the Jacobian and the minimum and maximum deviations from the mean for pure pitch (left image) and pure yaw (right image) rotations in the (-2,2) degrees range. Deviations from $1$ are not significant for the range of interest. The determinant of the Jacobian for a roll motion is $1$ because the rotation is in-plane, and therefore, there is no grid deformation for any angle.}
    \label{fig:deviations_from_det1}
\end{figure}

With this parametrization, there is no compression or dilation for in-plane rotations ({rolls}),  and the effect can be safely neglected for {pitch} and {yaw}  rotations within the range of interest. \Cref{fig:deviations_from_det1} illustrates that for pure {pitch} and {yaw} rotations within the interval $(-2^\circ,2^\circ)$,  the deviation of the Jacobian determinant from 1 (related to image scaling) is not significant for an image taken with a short focal length. Moreover, as the focal length increases, the maximum rotation angle of interest decreases, further mitigating any potential compression or dilation effects.

\begin{figure}[ht]
    \centering
\includegraphics[width=0.45\linewidth]{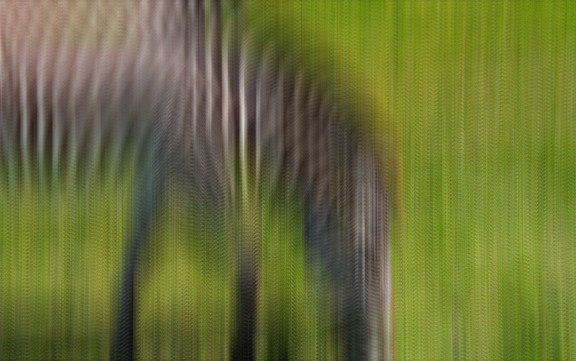}
\includegraphics[width=0.45\linewidth]{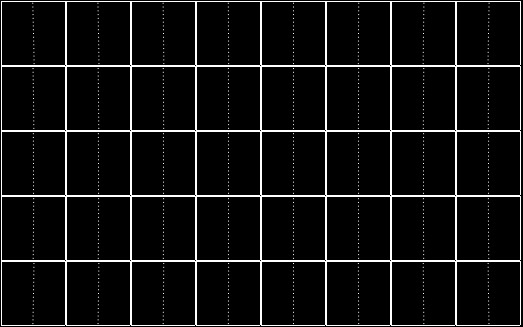}
\includegraphics[width=\linewidth]{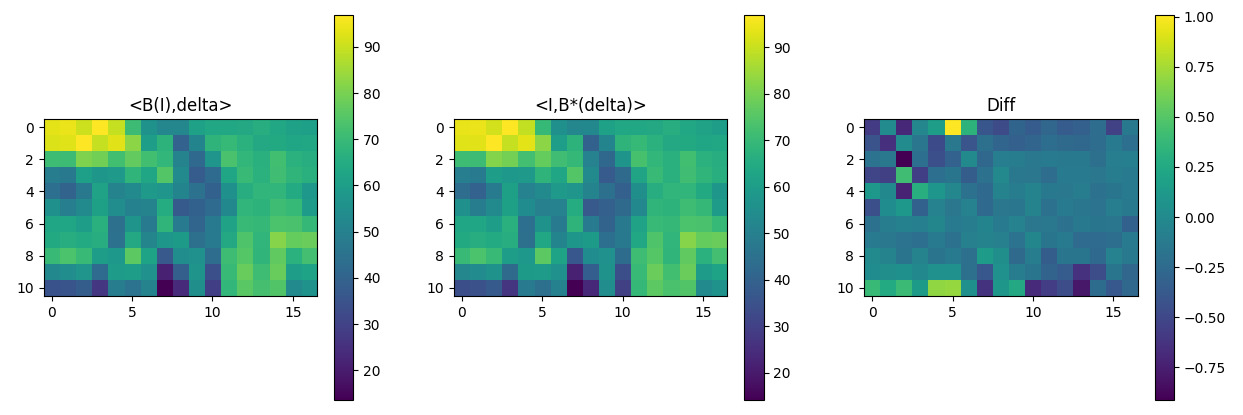}
    \caption{Blur due to a 2 degrees pitch camera motion. The upper row shows the blurry image and the blur kernels for pixels on a grid. The bottom row shows the consistency between applying the dots products $\langle B(\mathbf{I}),\bm{\delta} \rangle$  and $\langle \mathbf{I}, B^*(\bm{\delta}) \rangle$, where $\mathbf{I}$ is the zebra image, $B(\cdot)$ is the blur operator (from {pitch} rotation in this example) and $\bm{\delta}$ are small variance Gaussian functions placed on different regions of the image. The maximum difference between computing the dot products using $B(\cdot)$ and $B^*(\cdot)$ is close to $1$ for images represented with 256 levels.  }
    \label{fig:pitch_consistency}
\end{figure}

\begin{figure}[ht]
    \centering
\includegraphics[width=0.45\linewidth]{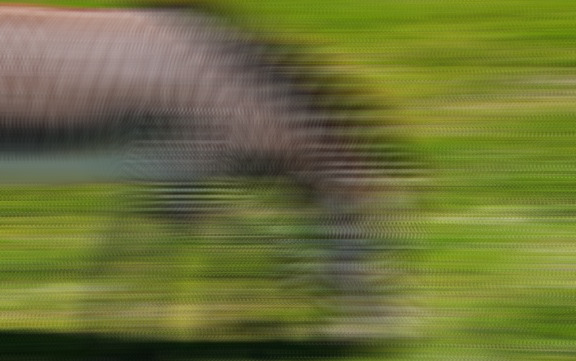}
\includegraphics[width=0.45\linewidth]{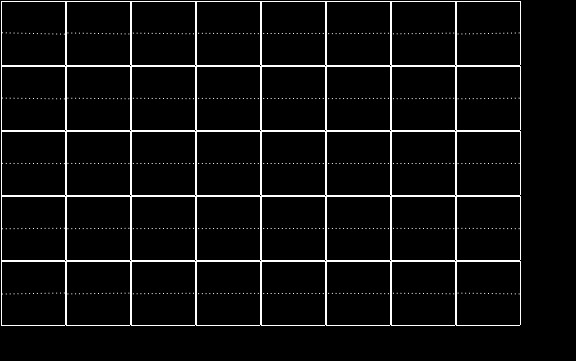}
\includegraphics[width=\linewidth]{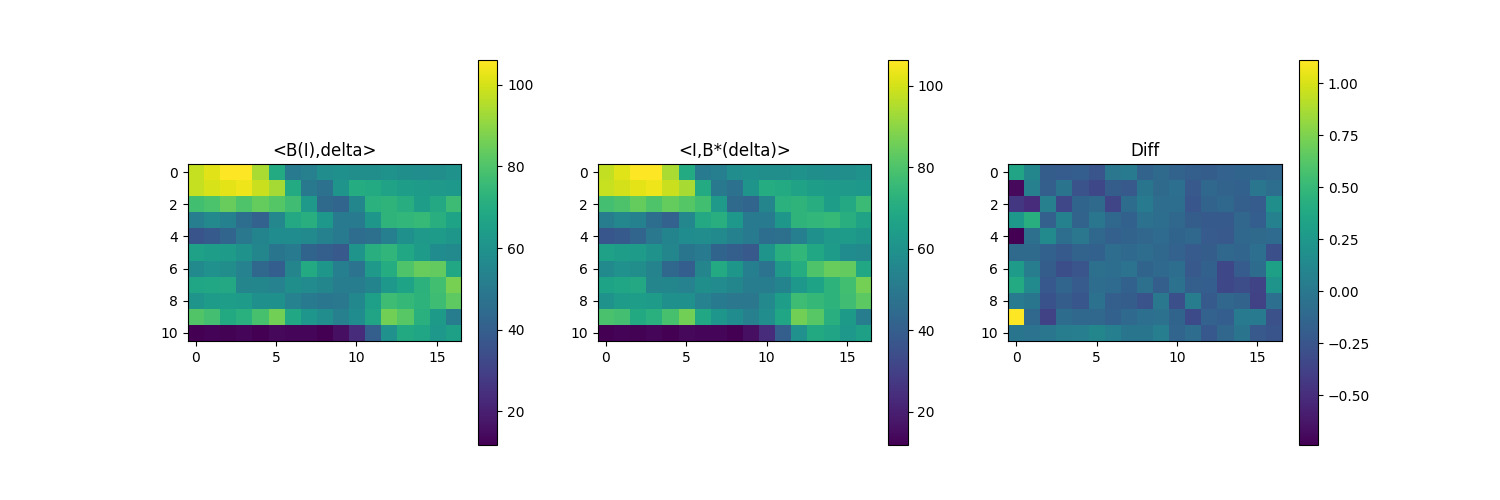}
    \caption[Sample of blur due to 2 degrees\textit{yaw} movement.]{Blur due to 2 degrees \textit{yaw} motion. The upper row shows the blurry image and the blur kernels for pixels on a grid. The bottom row shows the consistency between applying the dots products $\langle B(\mathbf{I}),\bm{\delta} \rangle$  and $\langle \mathbf{I}, B^*(\bm{\delta}) \rangle$, where $\mathbf{I}$ is the zebra image, $B(\cdot)$ is the blur operator (\textit{yaw} rotation in this example) and $\bm{\delta}$ are small variance Gaussian functions placed on different regions of the image. The maximum difference between computing the dot products using $B(\cdot)$ and $B^*(\cdot)$ is close to $1$ for images represented with 256 levels.   }
    \label{fig:yaw_consistency}
\end{figure}

\begin{figure}[ht]
    \centering
\includegraphics[width=0.45\linewidth]{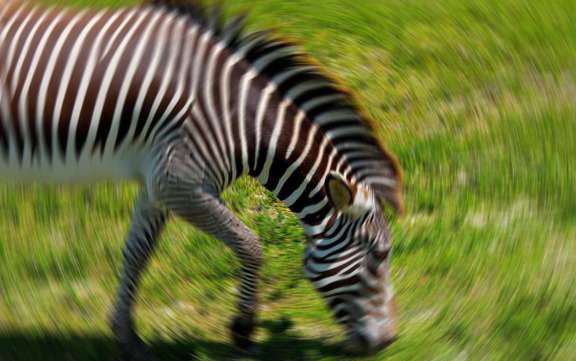}
\includegraphics[width=0.45\linewidth]{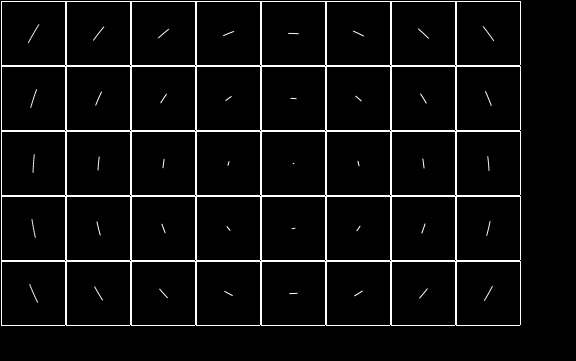}
\includegraphics[width=\linewidth]{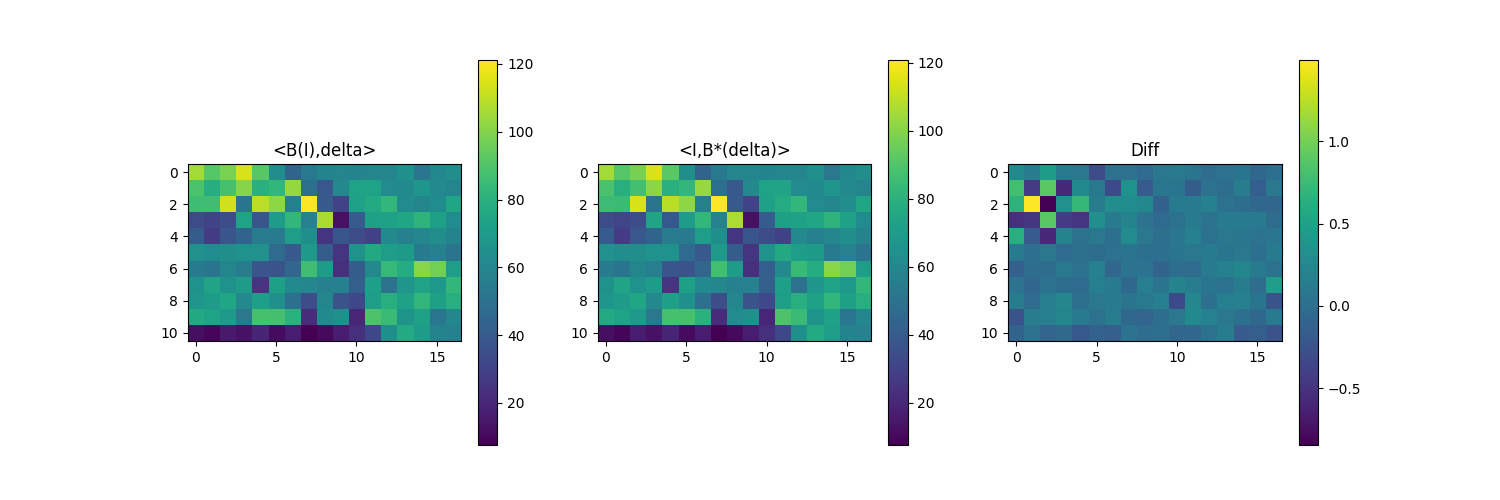}
    \caption[Sample of blur due to 2 degrees roll movement.]{Blur due to 2 degrees \textit{roll} movement. The upper row shows the blurry image and the blur kernels for pixels on a grid. The bottom row shows the consistency between applying the dots products $\langle B(\mathbf{I}),\bm{\delta} \rangle$  and $\langle \mathbf{I}, B^*(\bm{\delta}) \rangle$, where $\mathbf{I}$ is the zebra image, $B(\cdot)$ is the blur operator (\textit{roll} rotation in this example) and $\bm{\delta}$ are small variance Gaussian functions placed on different regions of the image. The maximum difference between computing the dot products using $B(\cdot)$ and $B^*(\cdot)$ is close to $1$ for images represented with 256 levels.  }
    \label{fig:roll_consistency}
\end{figure}

To further assess the validity of assuming $|J_{\mathcal{H}_t}(y)| = 1$, we compare the scalar products $\langle w,B(\{\mathcal{H}_t\})(u) \rangle $ with \newline $\langle B^*(\{\mathcal{H}_t\})(w), u \rangle$. Specifically, given a fixed image $u$,  we generate multiple images $w$, each being a narrow Gaussian (delta function) placed in a different region. As an example, we show the result for pure {pitch} transformations in \Cref{fig:pitch_consistency}. Despite the severe blur in the image, the maximum discrepancies remain below 1\%. \cref{fig:yaw_consistency} and \cref{fig:roll_consistency} show a similar experiment but for severe \textit{yaw} and \textit{roll} respectively.  %

\FloatBarrier

\section{Efficient computation of 
\texorpdfstring{$\mathbf{B}^{T}=\{\mathbf{T}_{{\mathbf{\mathcal{H}}}_{t}^{-1}}\}$}
{BT}}

The computation of $\mathbf{B}^T\mathbf{x}$ using the sequence of inverse homographies $\{\mathbf{T}_{{\mathbf{\mathcal{H}}}_t^{-1}}\}$ can be achieved similarly as the calculation of $\mathbf{B}\mathbf{x}$ from $\{\mathbf{T}_{{\mathbf{\mathcal{H}}}_t}\}$. However, we propose an efficient approximation that avoids the computation of offsets from the sequence of inverse homographies. 

The homography $\mathbf{\mathcal{H}}^{-1}_t$ at time step $t$, applied to a blurry image pixel $\mathbf{i}$, can be written as

\begin{equation}
\mathbf{\mathcal{H}}_t^{-1}(\mathbf{i}) = \mathbf{i} + \Delta^{(t)}(\mathbf{i}),    
\end{equation}
where $\Delta^{(t)}(\mathbf{i})$ represents the 2D motion vector applied to the pixel $\mathbf{i}$. We approximate the homography as $\tilde{\mathbf{\mathcal{H}}}_t(\mathbf{i}) = \mathbf{i} - \Delta^{(t)}(\mathbf{i})$, and verify that this approximation is effective by analyzing how close the composition $\mathbf{\mathcal{H}}_t^{-1}\tilde{\mathbf{\mathcal{H}}}_t(\mathbf{j})$ is to identity, that is, $\mathbf{\mathcal{H}}_t^{-1}\tilde{\mathbf{\mathcal{H}}}_t(\mathbf{j})\simeq \mathbf{j}$.

By approximating the homography as
\begin{equation}
\tilde{\mathbf{\mathcal{H}}}_t(\mathbf{j}) = \mathbf{j} - \Delta^{(t)}(\mathbf{j}),    \label{eq:inv_homo_approx}
\end{equation}
and substituting into the composition $\mathbf{\mathcal{H}}_t^{-1}\tilde{\mathbf{\mathcal{H}}}_t(\mathbf{j})$, we get 
\begin{align}   \mathbf{\mathcal{H}}_t^{-1}\tilde{\mathbf{\mathcal{H}}}_t(\mathbf{j}) &= \mathcal{H}_t^{-1} (\mathbf{j} - \Delta^{(t)}(\mathbf{j}))   \\
     &= \mathcal{H}_t^{-1}  (\mathbf{j}) - \frac{\partial \mathbf{\mathcal{H}}^{-1}_t(\mathbf{j})}{\partial \mathbf{j}} \Delta^{(t)}(\mathbf{j}) + o(\Delta^{(t)}(\mathbf{j})^2) \\
     &\simeq \mathbf{j} + \Delta^{(t)}(\mathbf{j}) - \left( \frac{\partial \mathbf{j} }{\partial \mathbf{j}} + \frac{\partial \Delta^{(t)}(\mathbf{j})}{\partial \mathbf{j}} \right) \Delta^{(t)}(\mathbf{j})  \\
     &= \mathbf{j} - \frac{\partial \Delta^{(t)}(\mathbf{j})}{\partial \mathbf{j}}\Delta^{(t)}(\mathbf{j})  \\
     &= \mathbf{j} - \left( \frac{\partial \mathcal{H}_{t}^{-1}(\mathbf{j})}{\partial \mathbf{j}} - \mathbf{Id} \right)\Delta^{(t)}(\mathbf{j}) .  \label{eq:inverse_approx}
\end{align}

\Cref{eq:inverse_approx} shows that \cref{eq:inv_homo_approx} is a good approximation when the homography gradient at a given pixel is close to the identity. While this condition may seem too restrictive, it is appropriate for the camera rotation angles that appear in practice.

To illustrate the quality of the approximation, we compare $\mathbf{B}^T$ calculated from $\{ \mathbf{\mathcal{H}}^{-1}_t \} $ with $\mathbf{B}^T$ obtained by the approximation. To this end, we compare the $i$-th column of $\mathbf{B}$ seen as an image with its approximation given by \cref{eq:inv_homo_approx}, which is a vertical and horizontal flip of the $i$-th row of $\mathbf{B}$ (also seen as an image).

\begin{figure}[ht]
    \centering
    \setlength{\tabcolsep}{1pt} %
    \begin{tabular}{ccc}  %
     \multicolumn{3}{c}{Blurry} \\
 \multicolumn{3}{c}{
         \begin{tikzpicture}
            \node[anchor=south west,inner sep=0pt] (img) {\includegraphics[width=0.5\textwidth]{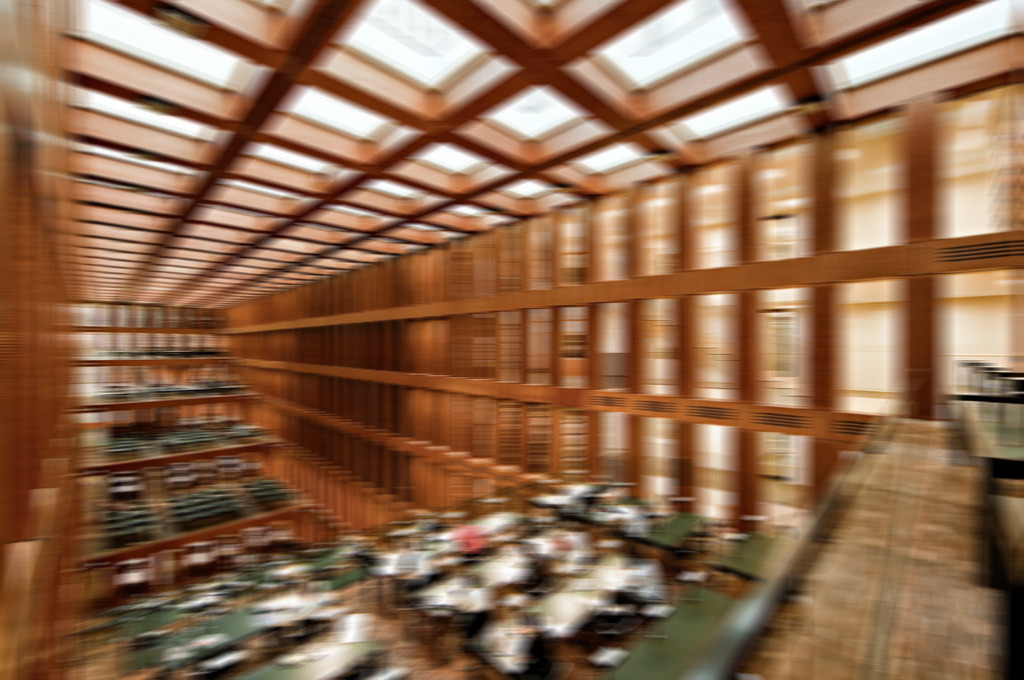}};
  \begin{scope}[x={(img.south east)},y={(img.north west)}]
    \draw[red,very thick] (0/1024,616/680) rectangle (64/1024,680/680);
    \draw[blue,very thick] (960/1024,616/680) rectangle (1024/1024,680/680);
    \draw[green,very thick] (0/1024,0/680) rectangle (64/1024,64/680);
    \draw[orange,very thick] (960/1024,0/680) rectangle (1024/1024,64/680);
    \draw[yellow,very thick] (480/1024,308/680) rectangle (544/1024,372/680);
  \end{scope}
        \end{tikzpicture}
        }  \\
    
     \multicolumn{3}{c}{$\mathbf{B}$ rows and columns for corners and the central pixel} \\
     \fcolorbox{red}{white}{\includegraphics[trim=0 616 960 0,clip,width=0.15\textwidth]{B_BT_pure_yaw_manmade_01.jpg}} &
     \multicolumn{2}{c}{ {\includegraphics[trim=10 10 0 0,clip,width=0.52\textwidth]{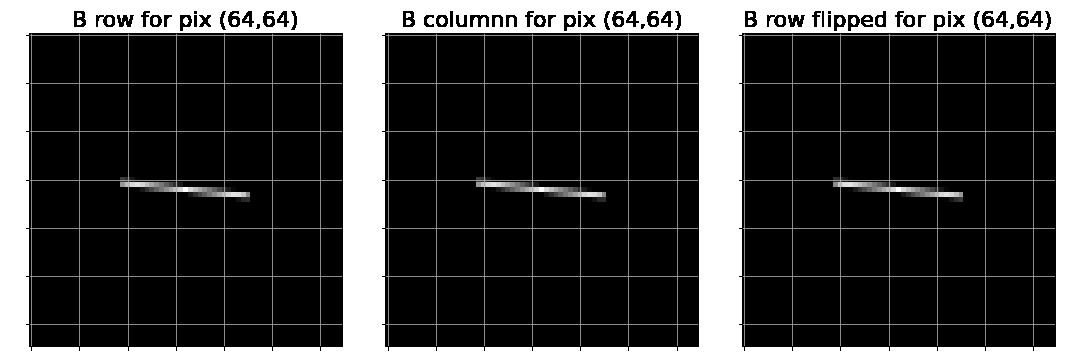}}} \\
     \fcolorbox{blue}{white}{\includegraphics[trim=960 616 0 0,clip,width=0.15\textwidth]{B_BT_pure_yaw_manmade_01.jpg}} &
     \multicolumn{2}{c}{{\includegraphics[trim=10 10 0 0,clip,width=0.52\textwidth]{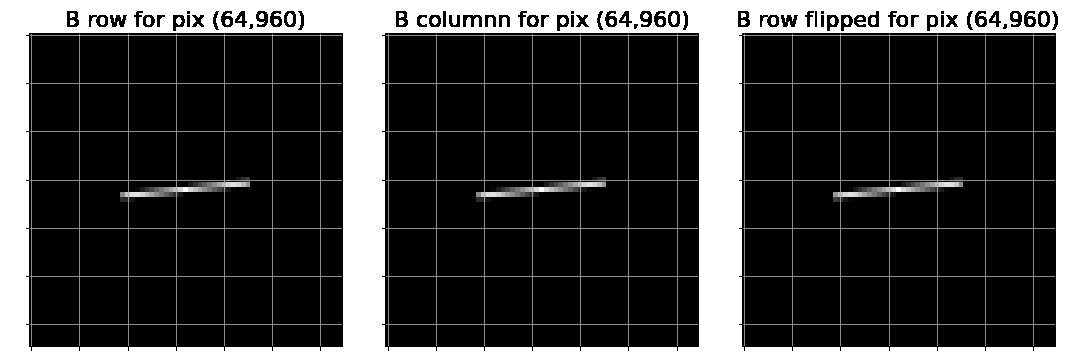}}} \\
     \fcolorbox{yellow}{white}{\includegraphics[trim=480 308 480 308,clip,width=0.15\textwidth]{B_BT_pure_yaw_manmade_01.jpg}} &
    \multicolumn{2}{c}{{\includegraphics[trim=10 10 0 0,clip,width=0.52\textwidth]{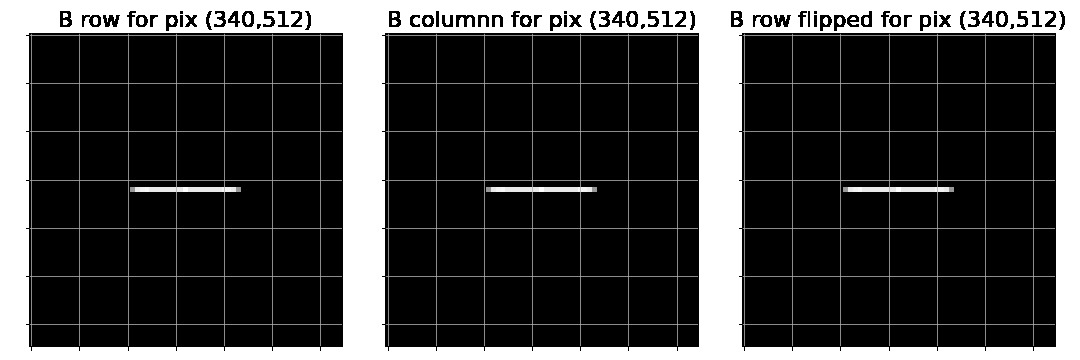}}}  \\
\fcolorbox{green}{white}{\includegraphics[trim=0 0 960 616,clip,width=0.15\textwidth]{B_BT_pure_yaw_manmade_01.jpg}} &
\multicolumn{2}{c}{ {\includegraphics[trim=10 10 0 0,clip,width=0.52\textwidth]{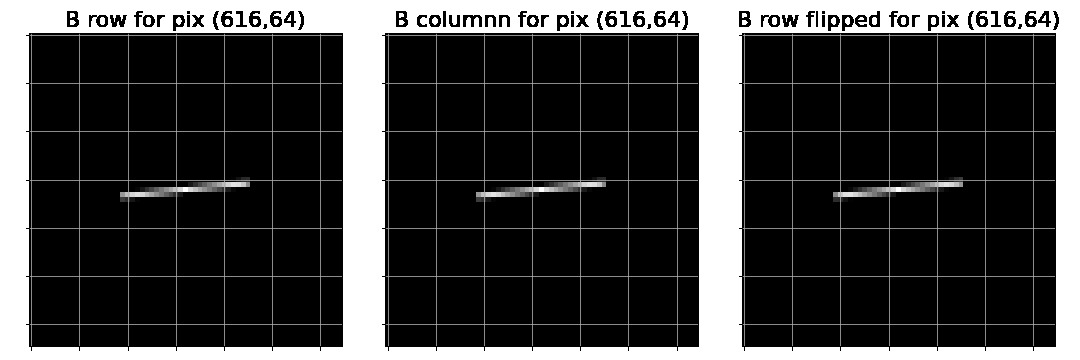}}} \\ 
\fcolorbox{orange}{white}{\includegraphics[trim=960 0 0  616,clip,width=0.15\textwidth]{B_BT_pure_yaw_manmade_01.jpg}} &
\multicolumn{2}{c}{ {\includegraphics[trim=10 10 0 0,clip,width=0.52\textwidth]{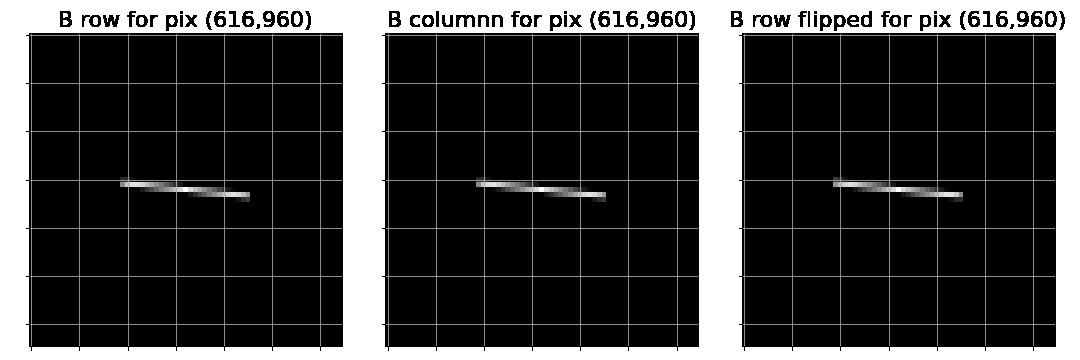}} }\\
    \end{tabular}
    \caption[Blurry image and kernel field for a pure \textit{yaw} rotation.]{Blurry image and kernel field for a pure \textit{yaw} rotation. When the motion trajectory is centered, the angle range of the rotation is small, and the angles vary linearly, the kernels are line segments and $\mathbf{B}=\mathbf{B}^T$. By flipping the rows of $\mathbf{B}$ (seen as an image) we have an accurate approximation of the $\mathbf{B}$ columns in all the image regions.\label{fig:B_BT_pure_yaw}}
\end{figure}

\begin{figure}[ht]
    \centering
    \setlength{\tabcolsep}{1pt} %
    \begin{tabular}{ccc}  %
     \multicolumn{3}{c}{Blurry} \\
 \multicolumn{3}{c}{
         \begin{tikzpicture}
            \node[anchor=south west,inner sep=0pt] (img) {\includegraphics[width=0.5\textwidth]{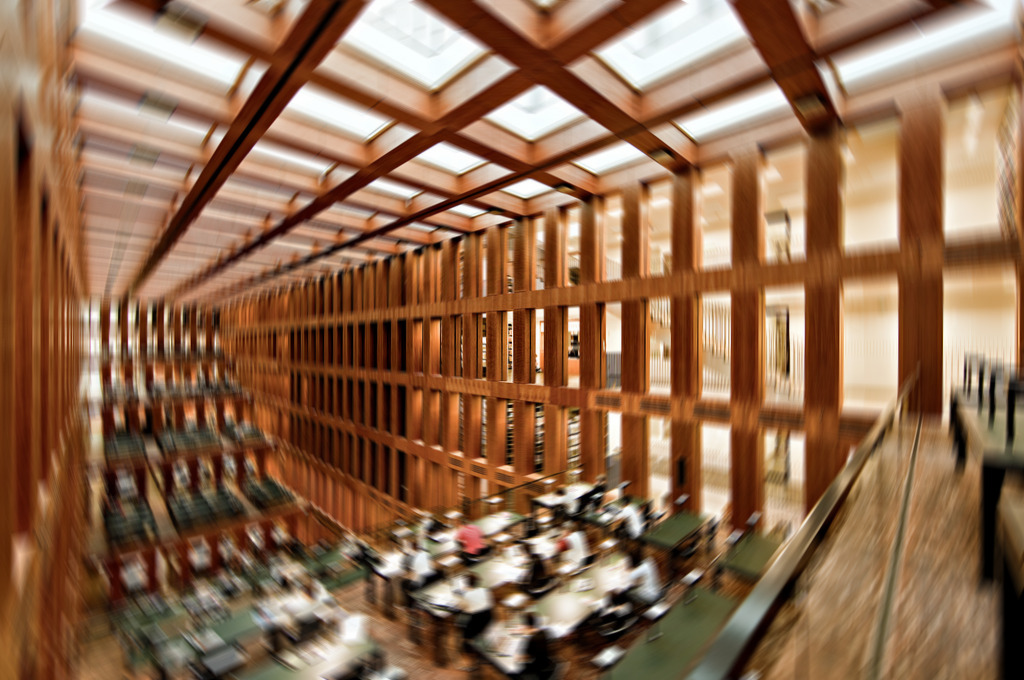}};
  \begin{scope}[x={(img.south east)},y={(img.north west)}]
    \draw[red,very thick] (0/1024,616/680) rectangle (64/1024,680/680);
    \draw[blue,very thick] (960/1024,616/680) rectangle (1024/1024,680/680);
    \draw[green,very thick] (0/1024,0/680) rectangle (64/1024,64/680);
    \draw[orange,very thick] (960/1024,0/680) rectangle (1024/1024,64/680);
    \draw[yellow,very thick] (480/1024,308/680) rectangle (544/1024,372/680);
  \end{scope}
        \end{tikzpicture}
        }  \\
    
     \multicolumn{3}{c}{$\mathbf{B}$ rows and columns for corners and the central pixel} \\
     \fcolorbox{red}{white}{\includegraphics[trim=0 616 960 0,clip,width=0.15\textwidth]{B_BT_pure_roll_manmade_01.jpg}} &
     \multicolumn{2}{c}{ {\includegraphics[trim=10 10 0 0,clip,width=0.52\textwidth]{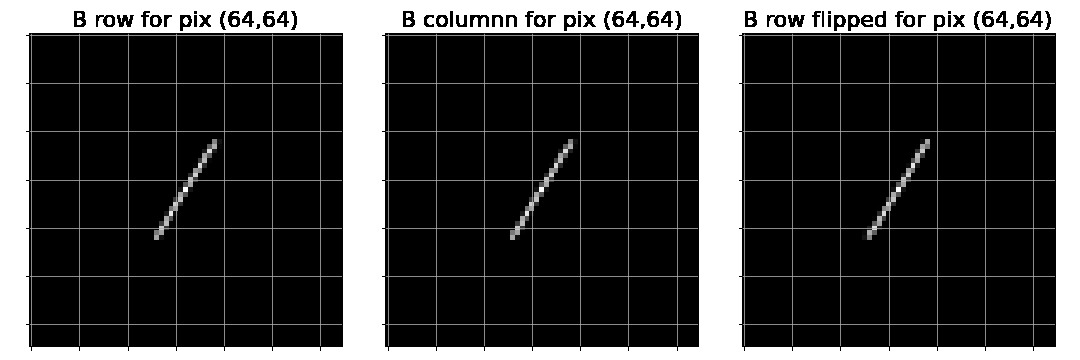}}} \\
     \fcolorbox{blue}{white}{\includegraphics[trim=960 616 0 0,clip,width=0.15\textwidth]{B_BT_pure_roll_manmade_01.jpg}} &
     \multicolumn{2}{c}{{\includegraphics[trim=10 10 0 0,clip,width=0.52\textwidth]{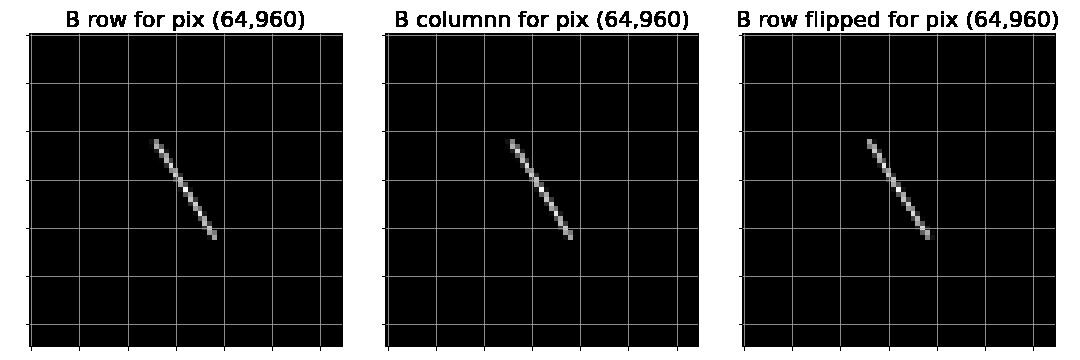}}} \\
     \fcolorbox{yellow}{white}{\includegraphics[trim=480 308 480 308,clip,width=0.15\textwidth]{B_BT_pure_roll_manmade_01.jpg}} &
    \multicolumn{2}{c}{{\includegraphics[trim=10 10 0 0,clip,width=0.52\textwidth]{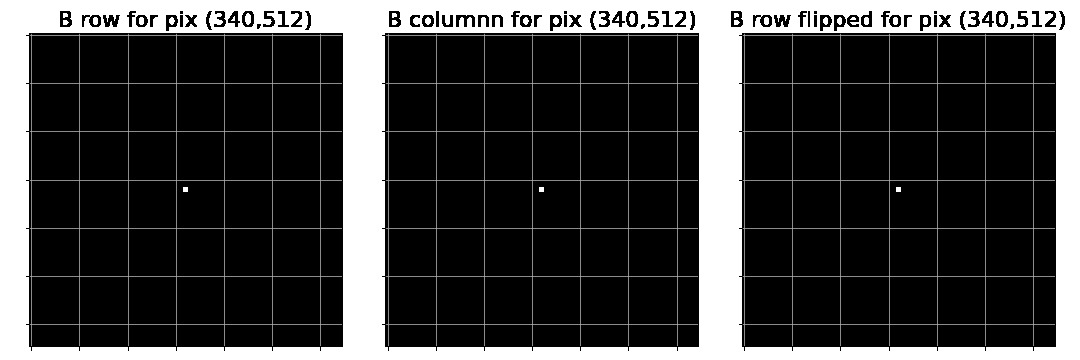}}}  \\
\fcolorbox{green}{white}{\includegraphics[trim=0 0 960 616,clip,width=0.15\textwidth]{B_BT_pure_roll_manmade_01.jpg}} &
\multicolumn{2}{c}{ {\includegraphics[trim=10 10 0 0,clip,width=0.52\textwidth]{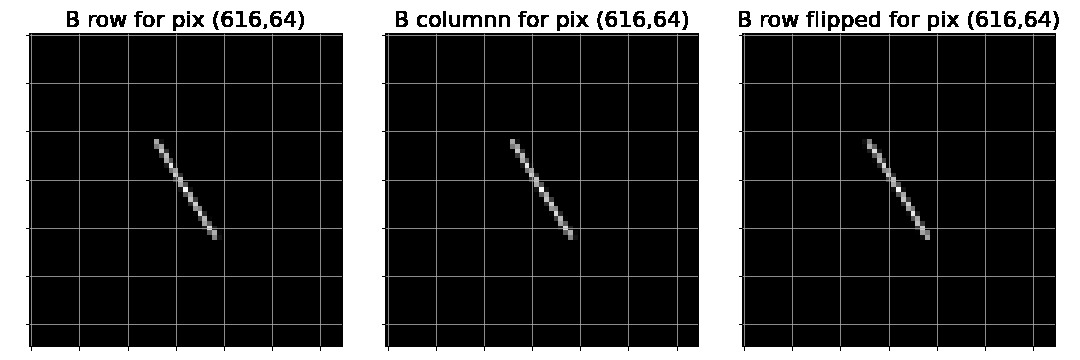}}} \\ 
\fcolorbox{orange}{white}{\includegraphics[trim=960 0 0  616,clip,width=0.15\textwidth]{B_BT_pure_roll_manmade_01.jpg}} &
\multicolumn{2}{c}{ {\includegraphics[trim=10 10 0 0,clip,width=0.52\textwidth]{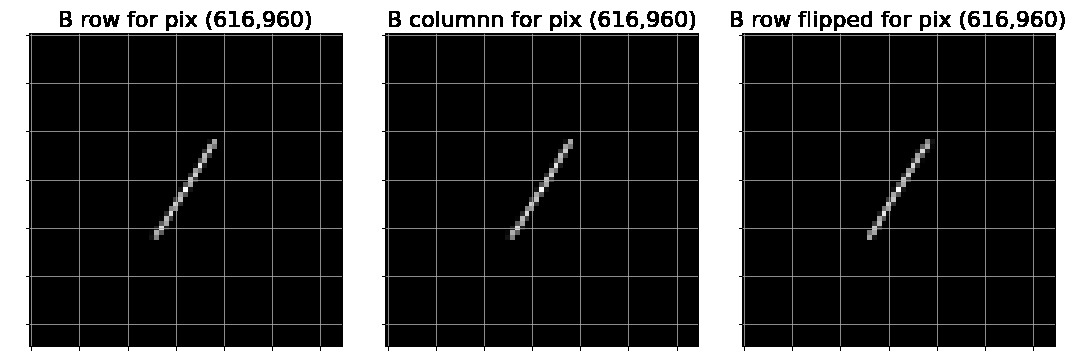}} }\\
    \end{tabular}
\caption[Blurry image and kernel field for a pure \textit{roll} rotation.]{Blurry image and kernel field for a pure \textit{roll} rotation. When the motion trajectory is centered, the angle range of the rotation is small, and the angles vary linearly, the kernels are line segments and $\mathbf{B}=\mathbf{B}^T$. By flipping the rows of $\mathbf{B}$ (seen as an image) we have an accurate approximation of the $\mathbf{B}$ columns in all the image regions.\label{fig:B_BT_pure_roll}}
\end{figure}

\begin{figure*}[ht]
    \centering
    \setlength{\tabcolsep}{1pt} %
    \begin{tabular}{ccc}  %
     \multicolumn{3}{c}{Blurry} \\
 \multicolumn{3}{c}{
         \begin{tikzpicture}
            \node[anchor=south west,inner sep=0pt] (img) {\includegraphics[width=0.5\textwidth]{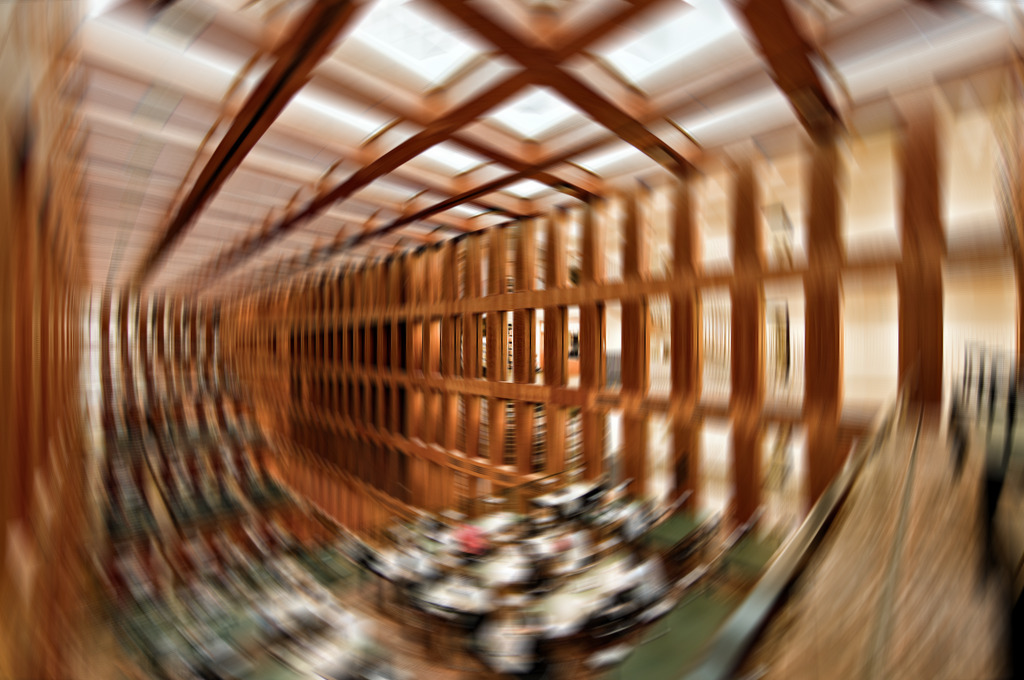}};
  \begin{scope}[x={(img.south east)},y={(img.north west)}]
    \draw[red,very thick] (0/1024,616/680) rectangle (64/1024,680/680);
    \draw[blue,very thick] (960/1024,616/680) rectangle (1024/1024,680/680);
    \draw[green,very thick] (0/1024,0/680) rectangle (64/1024,64/680);
    \draw[orange,very thick] (960/1024,0/680) rectangle (1024/1024,64/680);
    \draw[yellow,very thick] (480/1024,308/680) rectangle (544/1024,372/680);
  \end{scope}
        \end{tikzpicture}
        }  \\
    
     \multicolumn{3}{c}{$\mathbf{B}$ rows and columns for corners and the central pixel} \\
     \fcolorbox{red}{white}{\includegraphics[trim=0 616 960 0,clip,width=0.15\textwidth]{B_BT_pure_roll_5_manmade_01.jpg}} &
     \multicolumn{2}{c}{ {\includegraphics[trim=10 10 0 0,clip,width=0.52\textwidth]{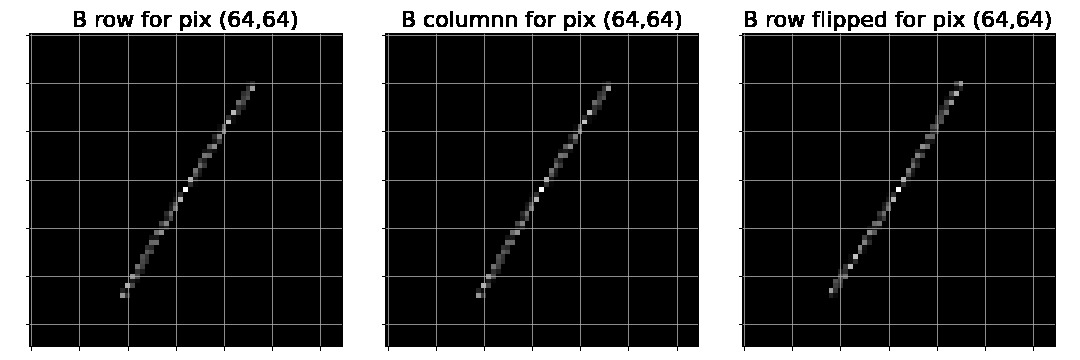}}} \\
     \fcolorbox{blue}{white}{\includegraphics[trim=960 616 0 0,clip,width=0.15\textwidth]{B_BT_pure_roll_5_manmade_01.jpg}} &
     \multicolumn{2}{c}{{\includegraphics[trim=10 10 0 0,clip,width=0.52\textwidth]{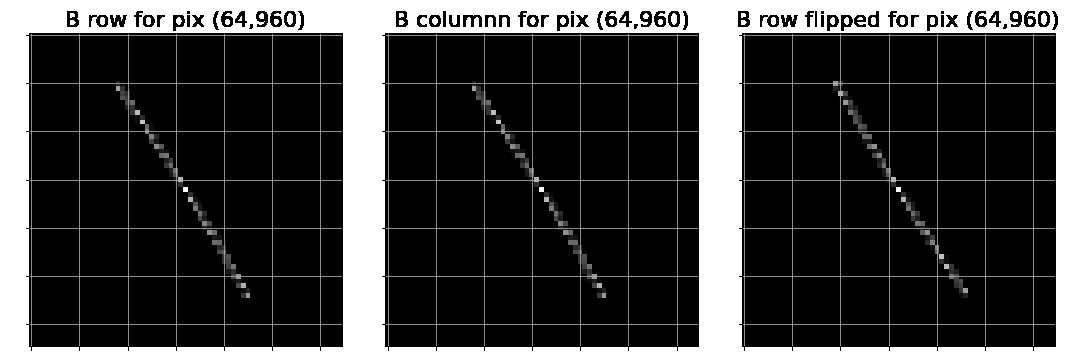}}} \\
     \fcolorbox{yellow}{white}{\includegraphics[trim=480 308 480 308,clip,width=0.15\textwidth]{B_BT_pure_roll_5_manmade_01.jpg}} &
    \multicolumn{2}{c}{{\includegraphics[trim=10 10 0 0,clip,width=0.52\textwidth]{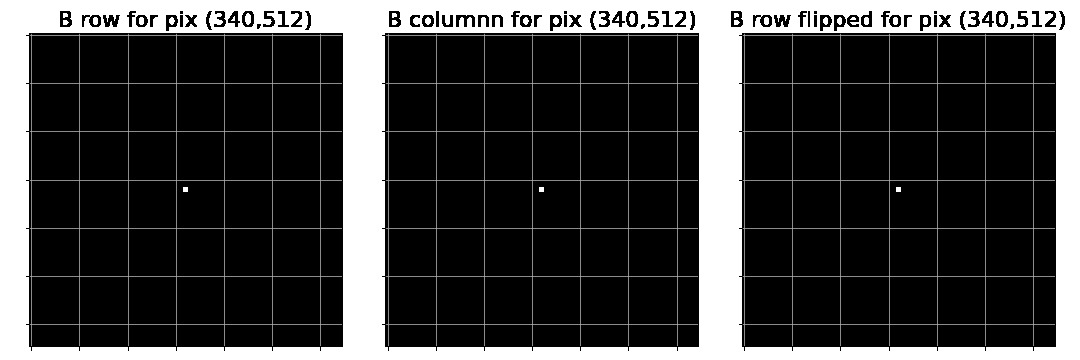}}}  \\
\fcolorbox{green}{white}{\includegraphics[trim=0 0 960 616,clip,width=0.15\textwidth]{B_BT_pure_roll_5_manmade_01.jpg}} &
\multicolumn{2}{c}{ {\includegraphics[trim=10 10 0 0,clip,width=0.52\textwidth]{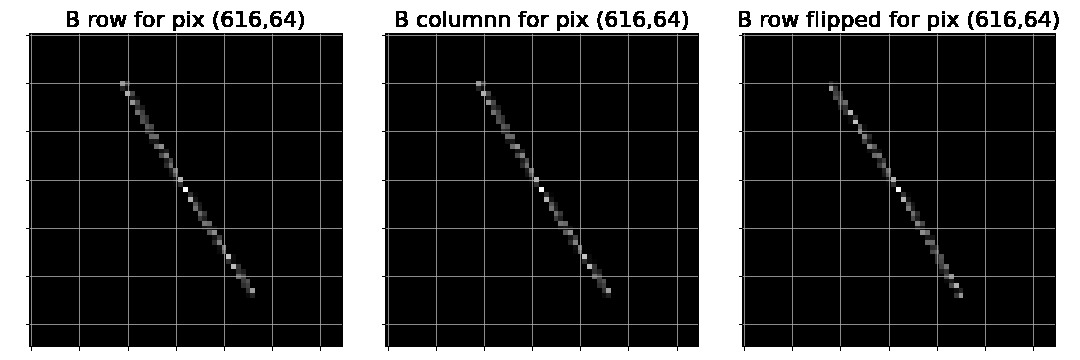}}} \\ 
\fcolorbox{orange}{white}{\includegraphics[trim=960 0 0  616,clip,width=0.15\textwidth]{B_BT_pure_roll_5_manmade_01.jpg}} &
\multicolumn{2}{c}{ {\includegraphics[trim=10 10 0 0,clip,width=0.52\textwidth]{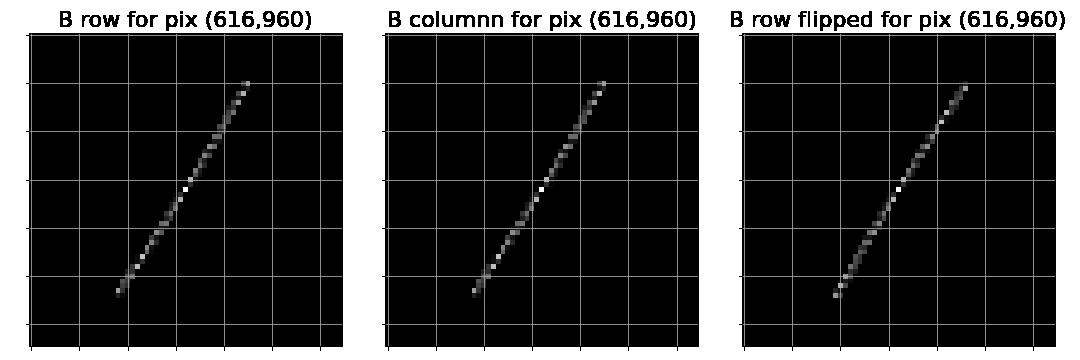}} }\\
    \end{tabular}
 \caption[Blurry images and kernel fields for a large \textit{roll} ($5^{\circ}$).]{Blurry images and kernel fields corresponding to a large \textit{roll} ($5^\circ$). When the motion trajectory is centered, and the camera maintains a constant velocity, the blur matrix satisfies $\mathbf{B} = \mathbf{B}^T$. Approximating the columns of $\mathbf{B}$ by flipping its rows (interpreted as an image) yields an accurate result, except when the kernel exhibits curvature. Noticeable differences arise primarily at the corners of the image. \label{fig:B_BT_pure_roll_5}}
\end{figure*}

\begin{figure}[ht]
    \centering
    \setlength{\tabcolsep}{1pt} %
    \begin{tabular}{ccc}  %
     \multicolumn{3}{c}{Blurry} \\

  \multicolumn{3}{c}{
         \begin{tikzpicture}
            \node[anchor=south west,inner sep=0pt] (img) {\includegraphics[width=0.5\textwidth]{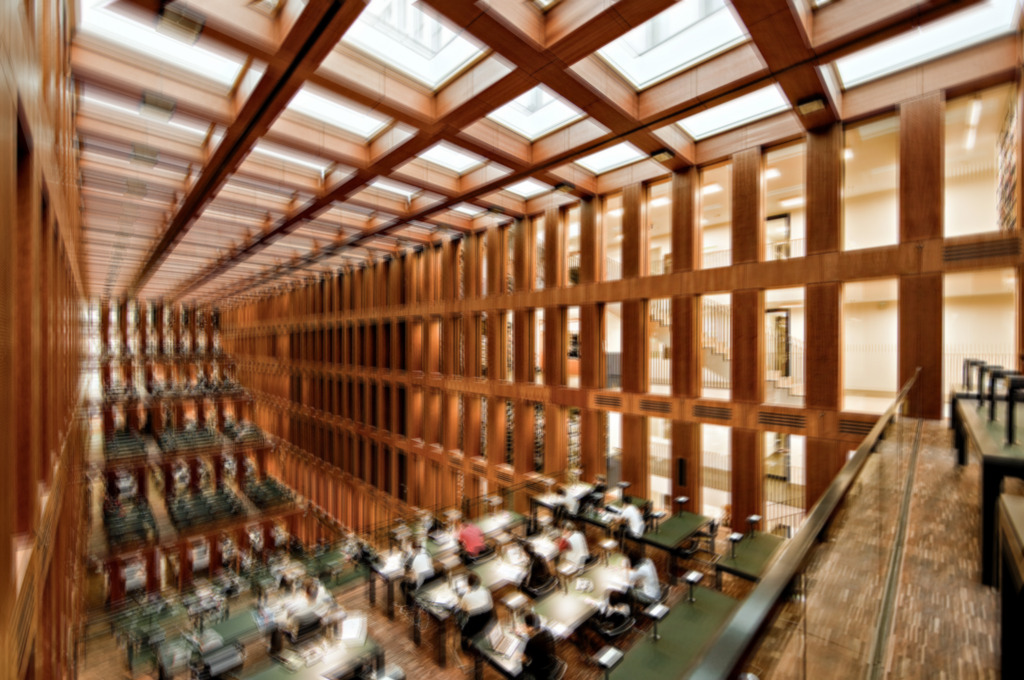}};
  \begin{scope}[x={(img.south east)},y={(img.north west)}]
    \draw[red,very thick] (0/1024,616/680) rectangle (64/1024,680/680);
    \draw[blue,very thick] (960/1024,616/680) rectangle (1024/1024,680/680);
    \draw[green,very thick] (0/1024,0/680) rectangle (64/1024,64/680);
    \draw[orange,very thick] (960/1024,0/680) rectangle (1024/1024,64/680);
    \draw[yellow,very thick] (480/1024,308/680) rectangle (544/1024,372/680);
  \end{scope}
        \end{tikzpicture}
        }  \\
        
     \multicolumn{3}{c}{$\mathbf{B}$ rows and columns for corners and the central pixel} \\
     \fcolorbox{red}{white}{\includegraphics[trim=0 616 960 0,clip,width=0.15\textwidth]{B_BT_tremor1_manmade_01.jpg}} &
     \multicolumn{2}{c}{ {\includegraphics[trim=10 10 0 0,clip,width=0.52\textwidth]{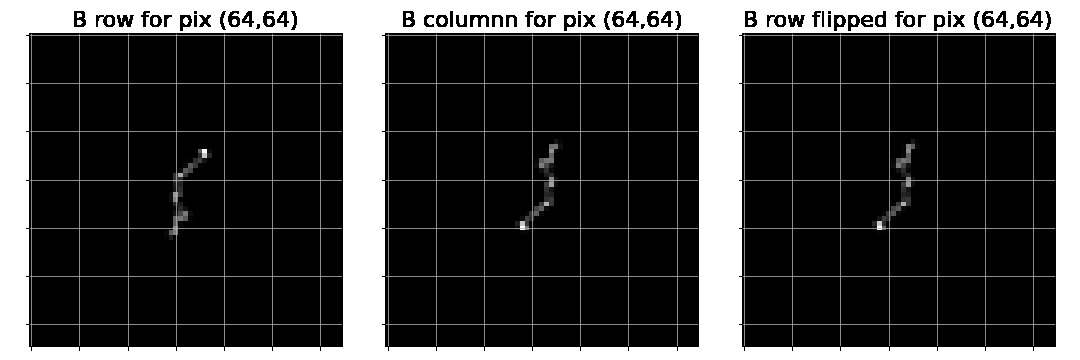}}} \\
     \fcolorbox{blue}{white}{\includegraphics[trim=960 616 0 0,clip,width=0.15\textwidth]{B_BT_tremor1_manmade_01.jpg}} &
     \multicolumn{2}{c}{{\includegraphics[trim=10 10 0 0,clip,width=0.52\textwidth]{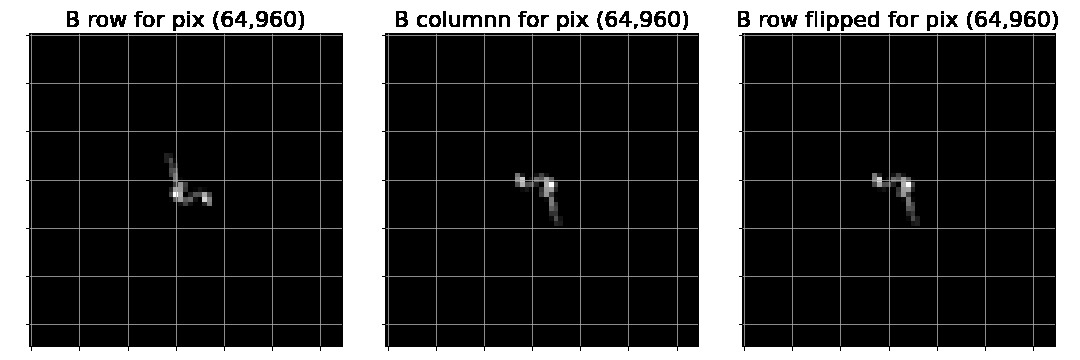}}} \\
     \fcolorbox{yellow}{white}{\includegraphics[trim=448 276 448 276,clip,width=0.15\textwidth]{B_BT_tremor1_manmade_01.jpg}} &
    \multicolumn{2}{c}{{\includegraphics[trim=10 10 0 0,clip,width=0.52\textwidth]{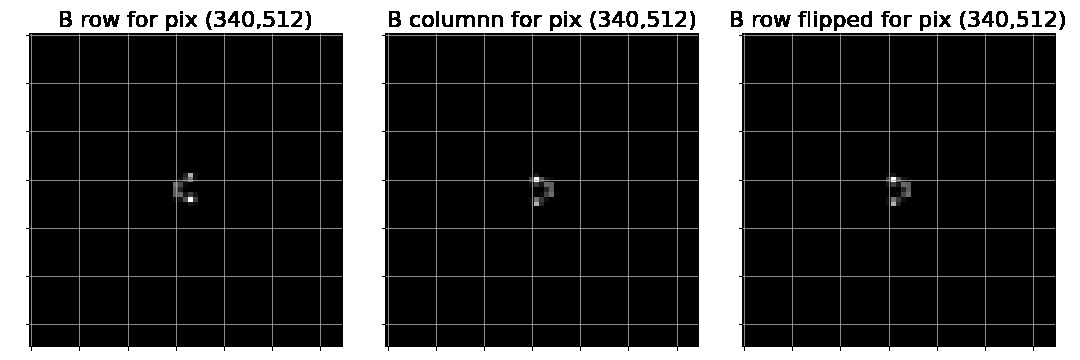}}}  \\
\fcolorbox{green}{white}{\includegraphics[trim=0 0 960 616,clip,width=0.15\textwidth]{B_BT_tremor1_manmade_01.jpg}} &
\multicolumn{2}{c}{ {\includegraphics[trim=10 10 0 0,clip,width=0.52\textwidth]{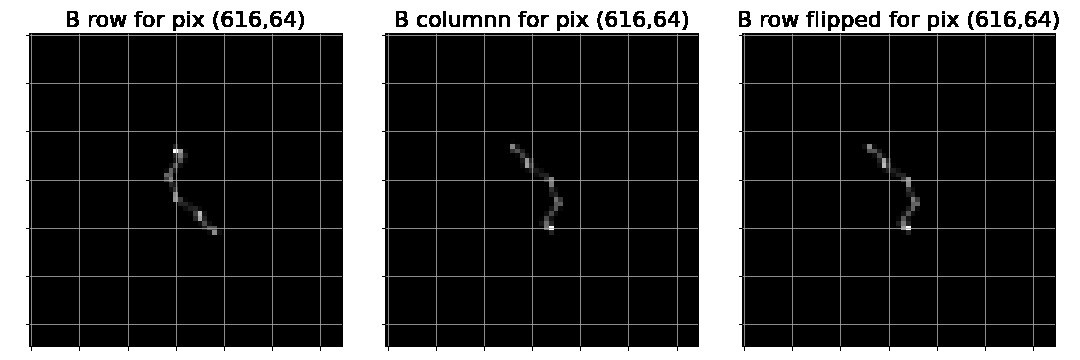}}} \\ 
\fcolorbox{orange}{white}{\includegraphics[trim=960 0 0  616,clip,width=0.15\textwidth]{B_BT_tremor1_manmade_01.jpg}} &
\multicolumn{2}{c}{ {\includegraphics[trim=10 10 0 0,clip,width=0.52\textwidth]{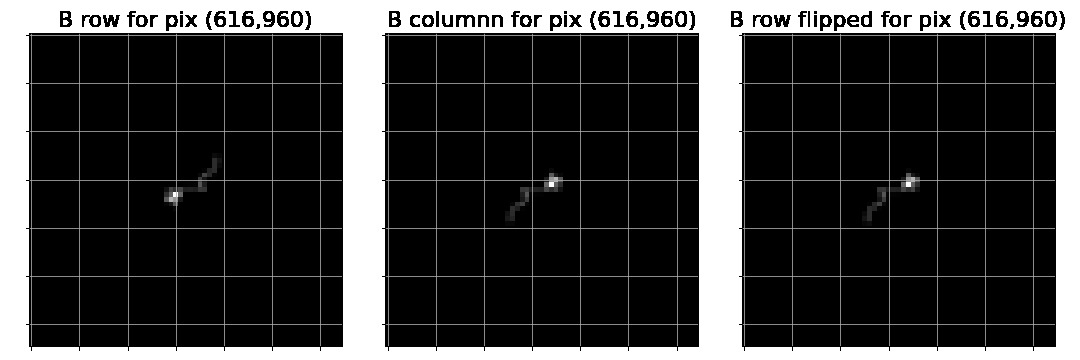}} }\\
    \end{tabular}
    \caption[Blur kernels maps for arbitrary camera shakes 1.]{Blur kernels maps for arbitrary camera shakes are spatially variants and $\mathbf{B}$ differs form $\mathbf{B}^T$. By flipping the rows of $\mathbf{B}$ (seen as an image) we have an accurate approximation of the $\mathbf{B}$ columns in all the image regions.\label{fig:B_BT_tremor1}}
\end{figure}

\begin{figure}[ht]
    \centering
    \setlength{\tabcolsep}{1pt} %
    \begin{tabular}{ccc}  %
     \multicolumn{3}{c}{Blurry} \\

  \multicolumn{3}{c}{
         \begin{tikzpicture}
            \node[anchor=south west,inner sep=0pt] (img) {\includegraphics[width=0.5\textwidth]{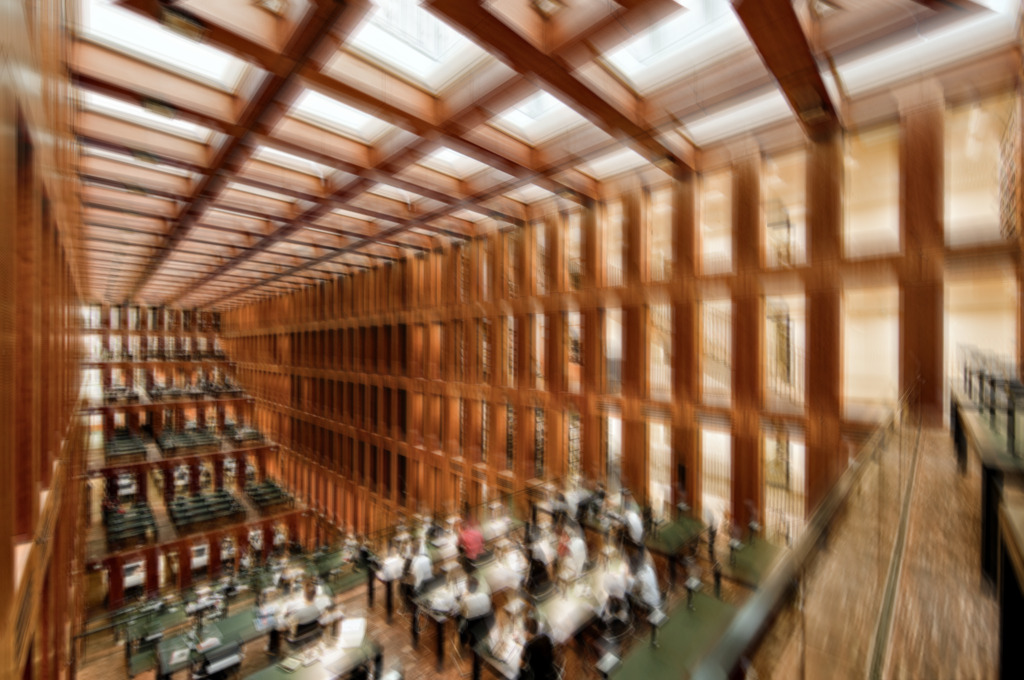}};
  \begin{scope}[x={(img.south east)},y={(img.north west)}]
    \draw[red,very thick] (0/1024,616/680) rectangle (64/1024,680/680);
    \draw[blue,very thick] (960/1024,616/680) rectangle (1024/1024,680/680);
    \draw[green,very thick] (0/1024,0/680) rectangle (64/1024,64/680);
    \draw[orange,very thick] (960/1024,0/680) rectangle (1024/1024,64/680);
    \draw[yellow,very thick] (480/1024,308/680) rectangle (544/1024,372/680);
  \end{scope}
        \end{tikzpicture}
        }  \\
        
     \multicolumn{3}{c}{$\mathbf{B}$ rows and columns for corners and the central pixel} \\
     \fcolorbox{red}{white}{\includegraphics[trim=0 616 960 0,clip,width=0.15\textwidth]{B_BT_tremor2_manmade_01.jpg}} &
     \multicolumn{2}{c}{ {\includegraphics[trim=10 10 0 0,clip,width=0.52\textwidth]{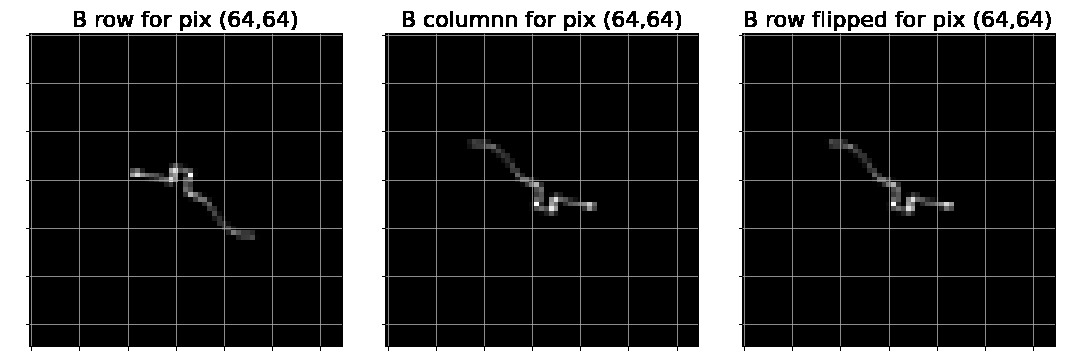}}} \\
     \fcolorbox{blue}{white}{\includegraphics[trim=960 616 0 0,clip,width=0.15\textwidth]{B_BT_tremor2_manmade_01.jpg}} &
     \multicolumn{2}{c}{{\includegraphics[trim=10 10 0 0,clip,width=0.52\textwidth]{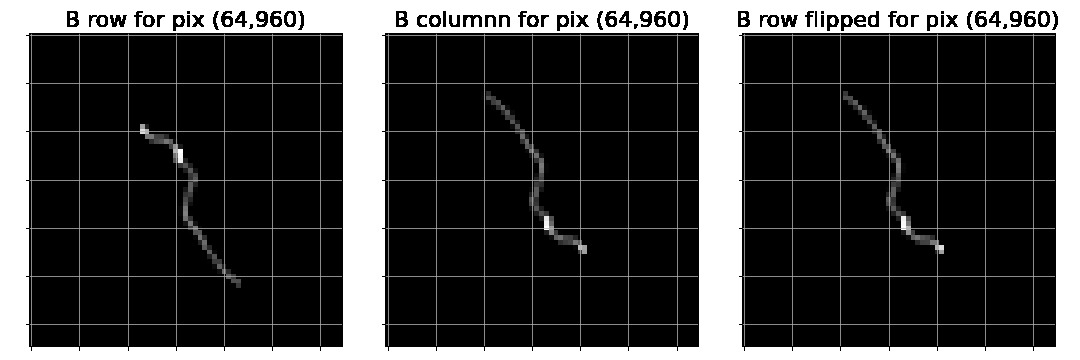}}} \\
     \fcolorbox{yellow}{white}{\includegraphics[trim=448 276 448 276,clip,width=0.15\textwidth]{B_BT_tremor2_manmade_01.jpg}} &
    \multicolumn{2}{c}{{\includegraphics[trim=10 10 0 0,clip,width=0.52\textwidth]{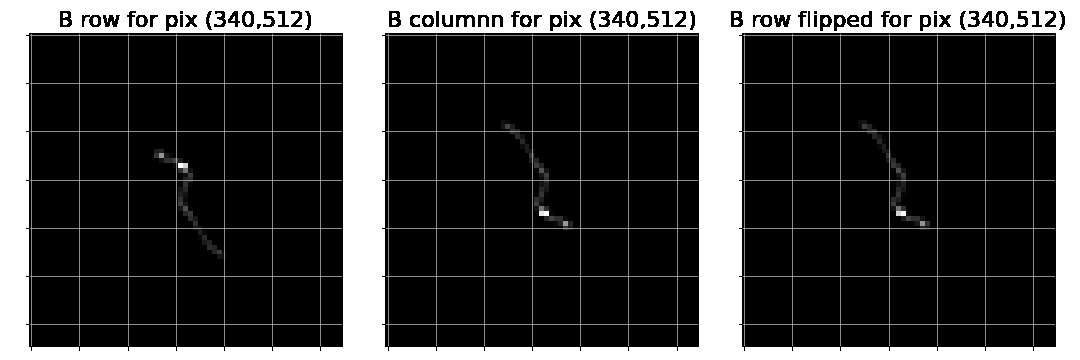}}}  \\
\fcolorbox{green}{white}{\includegraphics[trim=0 0 960 616,clip,width=0.15\textwidth]{B_BT_tremor2_manmade_01.jpg}} &
\multicolumn{2}{c}{ {\includegraphics[trim=10 10 0 0,clip,width=0.52\textwidth]{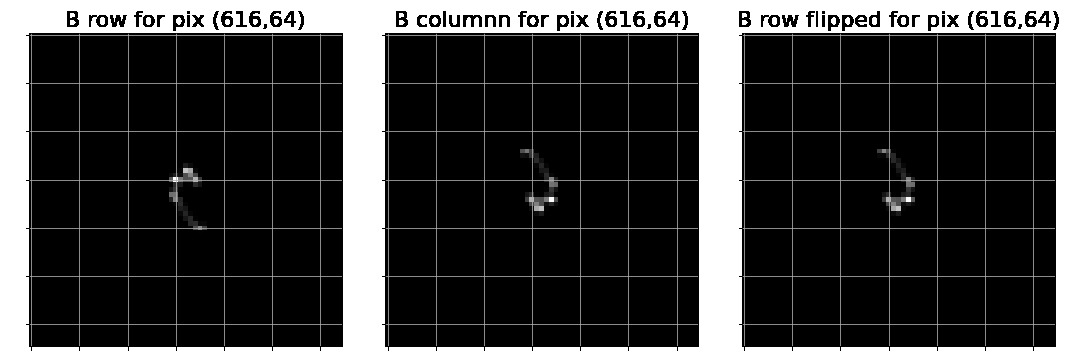}}} \\ 
\fcolorbox{orange}{white}{\includegraphics[trim=960 0 0  616,clip,width=0.15\textwidth]{B_BT_tremor2_manmade_01.jpg}} &
\multicolumn{2}{c}{ {\includegraphics[trim=10 10 0 0,clip,width=0.52\textwidth]{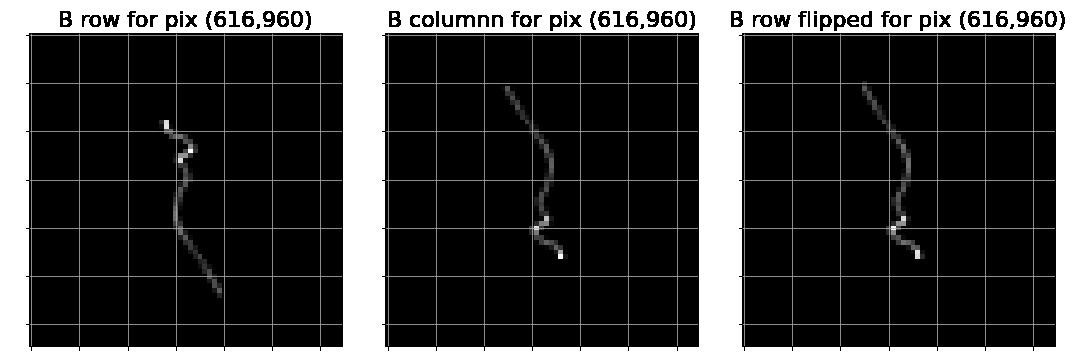}} }\\
    \end{tabular}
    \caption[Blur kernels maps for arbitrary camera shakes 2.]{Blur kernels maps for arbitrary camera shakes are spatially variants and $\mathbf{B}$ differs form $\mathbf{B}^T$. By flipping the rows of $\mathbf{B}$ (seen as an image) we have an accurate approximation of the $\mathbf{B}$ columns in all the image regions.\label{fig:B_BT_tremor2}}
\end{figure}

\begin{figure}[ht]
    \centering
    \setlength{\tabcolsep}{1pt} %
    \begin{tabular}{ccc}  %
     \multicolumn{3}{c}{Blurry} \\

  \multicolumn{3}{c}{
         \begin{tikzpicture}
            \node[anchor=south west,inner sep=0pt] (img) {\includegraphics[width=0.5\textwidth]{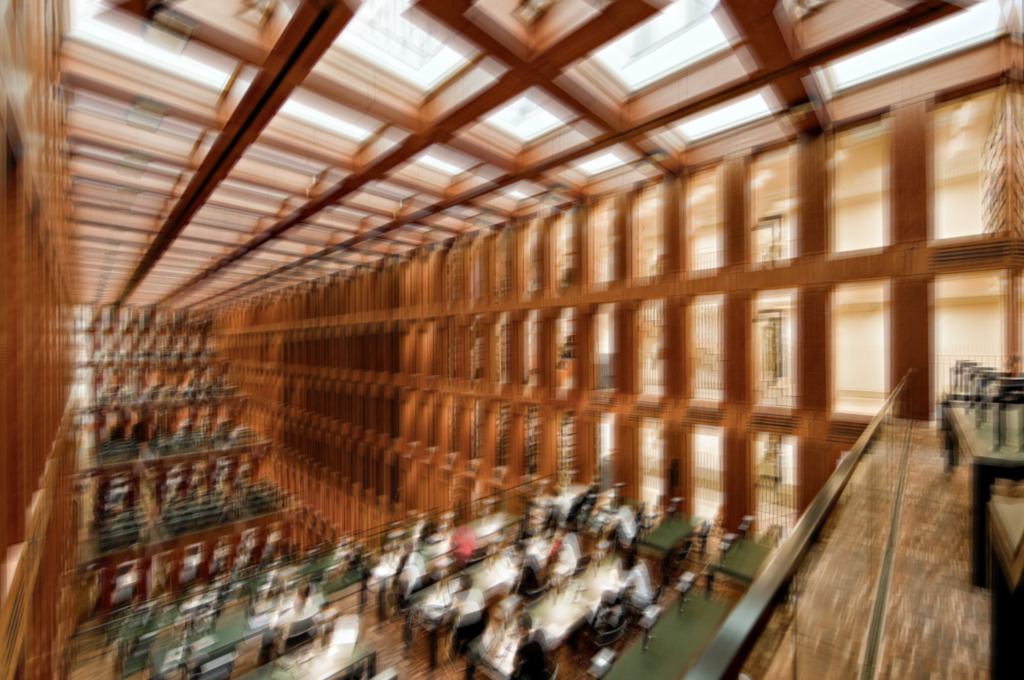}};
  \begin{scope}[x={(img.south east)},y={(img.north west)}]
    \draw[red,very thick] (0/1024,616/680) rectangle (64/1024,680/680);
    \draw[blue,very thick] (960/1024,616/680) rectangle (1024/1024,680/680);
    \draw[green,very thick] (0/1024,0/680) rectangle (64/1024,64/680);
    \draw[orange,very thick] (960/1024,0/680) rectangle (1024/1024,64/680);
    \draw[yellow,very thick] (480/1024,308/680) rectangle (544/1024,372/680);
  \end{scope}
        \end{tikzpicture}
        }  \\
        
     \multicolumn{3}{c}{$\mathbf{B}$ rows and columns for corners and the central pixel} \\
     \fcolorbox{red}{white}{\includegraphics[trim=0 616 960 0,clip,width=0.15\textwidth]{B_BT_tremor1_no_centered_manmade_01.jpg}} &
     \multicolumn{2}{c}{ {\includegraphics[trim=10 10 0 0,clip,width=0.52\textwidth]{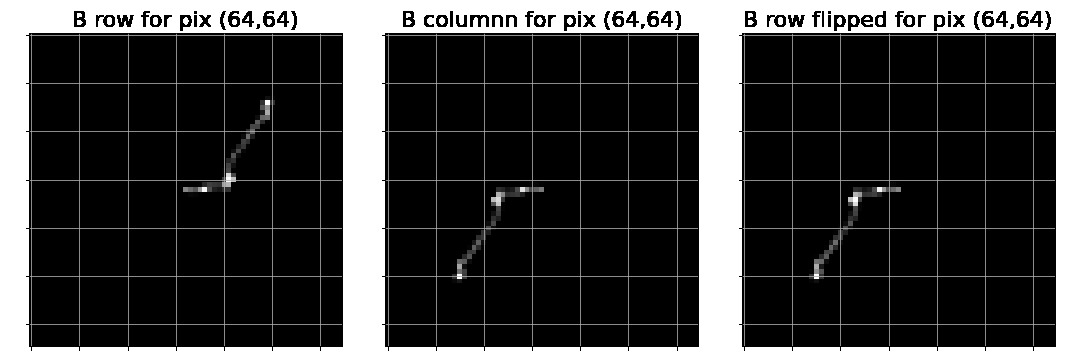}}} \\
     \fcolorbox{blue}{white}{\includegraphics[trim=960 616 0 0,clip,width=0.15\textwidth]{B_BT_tremor1_no_centered_manmade_01.jpg}} &
     \multicolumn{2}{c}{{\includegraphics[trim=10 10 0 0,clip,width=0.52\textwidth]{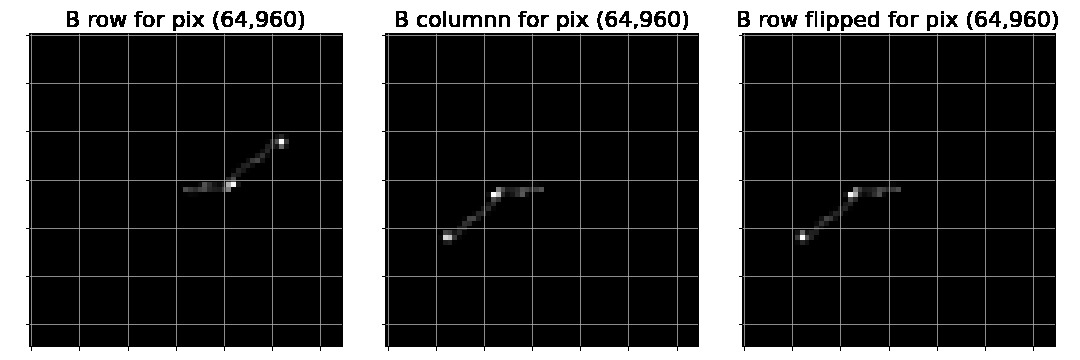}}} \\
     \fcolorbox{yellow}{white}{\includegraphics[trim=448 276 448 276,clip,width=0.15\textwidth]{B_BT_tremor1_no_centered_manmade_01.jpg}} &
    \multicolumn{2}{c}{{\includegraphics[trim=10 10 0 0,clip,width=0.52\textwidth]{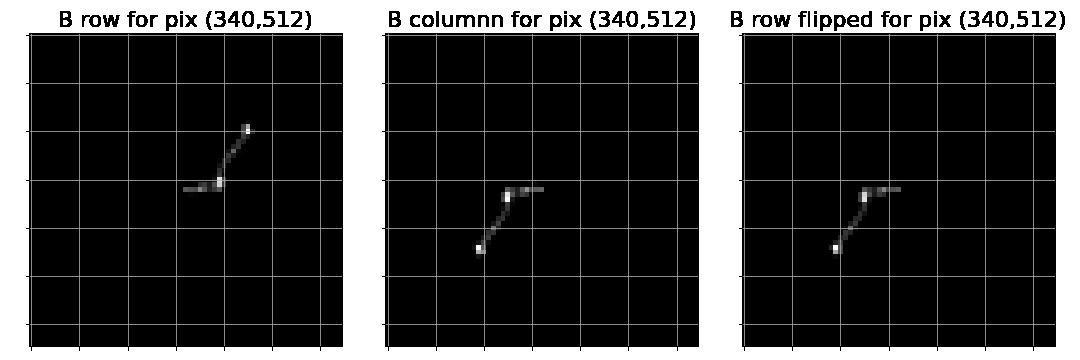}}}  \\
\fcolorbox{green}{white}{\includegraphics[trim=0 0 960 616,clip,width=0.15\textwidth]{B_BT_tremor1_no_centered_manmade_01.jpg}} &
\multicolumn{2}{c}{ {\includegraphics[trim=10 10 0 0,clip,width=0.52\textwidth]{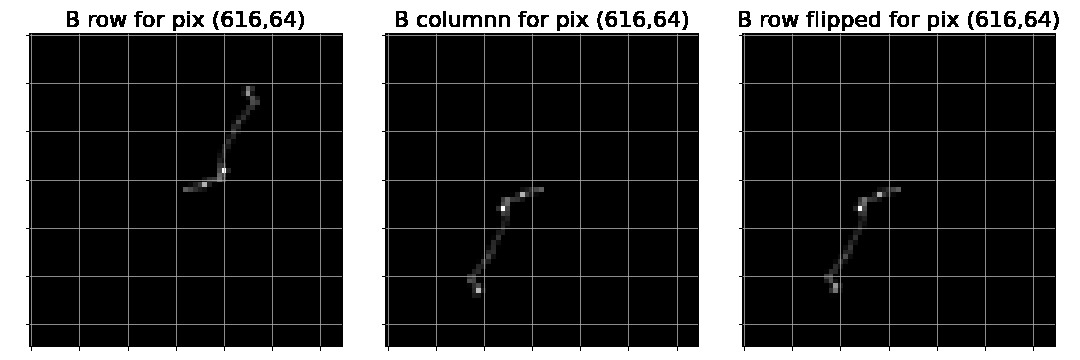}}} \\ 
\fcolorbox{orange}{white}{\includegraphics[trim=960 0 0  616,clip,width=0.15\textwidth]{B_BT_tremor1_no_centered_manmade_01.jpg}} &
\multicolumn{2}{c}{ {\includegraphics[trim=10 10 0 0,clip,width=0.52\textwidth]{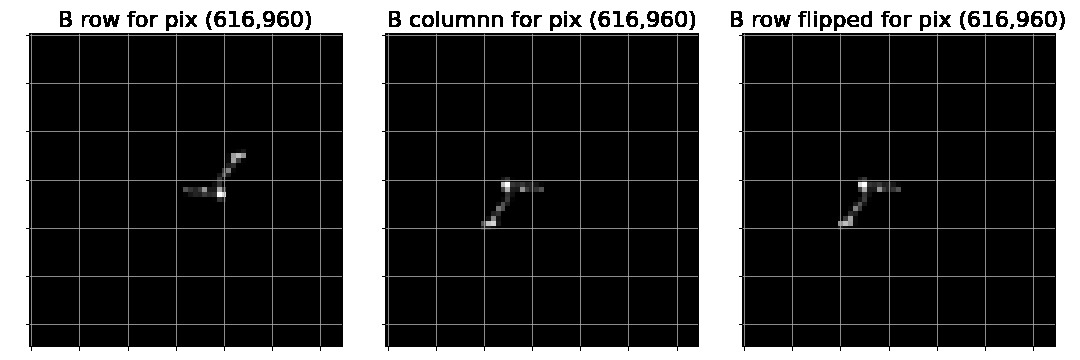}} }\\
    \end{tabular}
 \caption[Blur kernels maps for arbitrary camera shakes 1.]{Blur kernels maps for arbitrary camera shakes are spatially variants and $\mathbf{B}$ differs form $\mathbf{B}^T$. By flipping the rows of $\mathbf{B}$ (seen as an image) we have an accurate approximation of the $\mathbf{B}$ columns in all the image regions.\label{fig:B_BT_tremor1_no_centered}}
\end{figure}

\begin{figure}[ht]
    \centering
    \setlength{\tabcolsep}{1pt} %
    \begin{tabular}{ccc}  %
     \multicolumn{3}{c}{Blurry} \\

  \multicolumn{3}{c}{
         \begin{tikzpicture}
            \node[anchor=south west,inner sep=0pt] (img) {\includegraphics[width=0.5\textwidth]{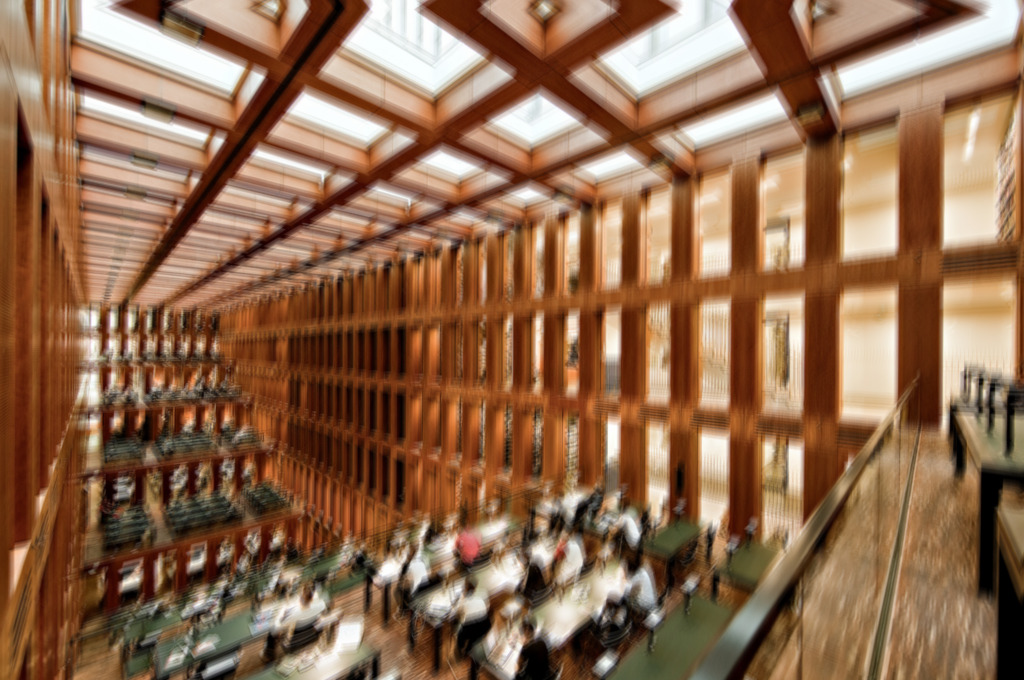}};
  \begin{scope}[x={(img.south east)},y={(img.north west)}]
    \draw[red,very thick] (0/1024,616/680) rectangle (64/1024,680/680);
    \draw[blue,very thick] (960/1024,616/680) rectangle (1024/1024,680/680);
    \draw[green,very thick] (0/1024,0/680) rectangle (64/1024,64/680);
    \draw[orange,very thick] (960/1024,0/680) rectangle (1024/1024,64/680);
    \draw[yellow,very thick] (480/1024,308/680) rectangle (544/1024,372/680);
  \end{scope}
        \end{tikzpicture}
        }  \\
        
     \multicolumn{3}{c}{$\mathbf{B}$ rows and columns for corners and the central pixel} \\
     \fcolorbox{red}{white}{\includegraphics[trim=0 616 960 0,clip,width=0.15\textwidth]{B_BT_tremor2_no_centered_manmade_01.jpg}} &
     \multicolumn{2}{c}{ {\includegraphics[trim=10 10 0 0,clip,width=0.52\textwidth]{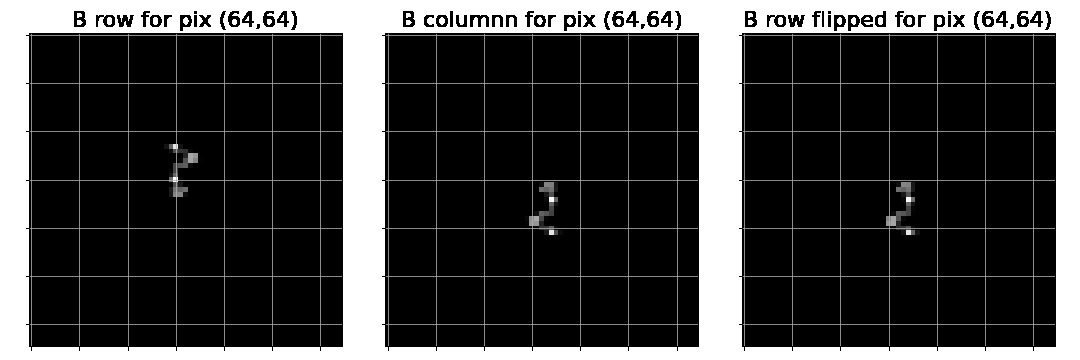}}} \\
     \fcolorbox{blue}{white}{\includegraphics[trim=960 616 0 0,clip,width=0.15\textwidth]{B_BT_tremor2_no_centered_manmade_01.jpg}} &
     \multicolumn{2}{c}{{\includegraphics[trim=10 10 0 0,clip,width=0.52\textwidth]{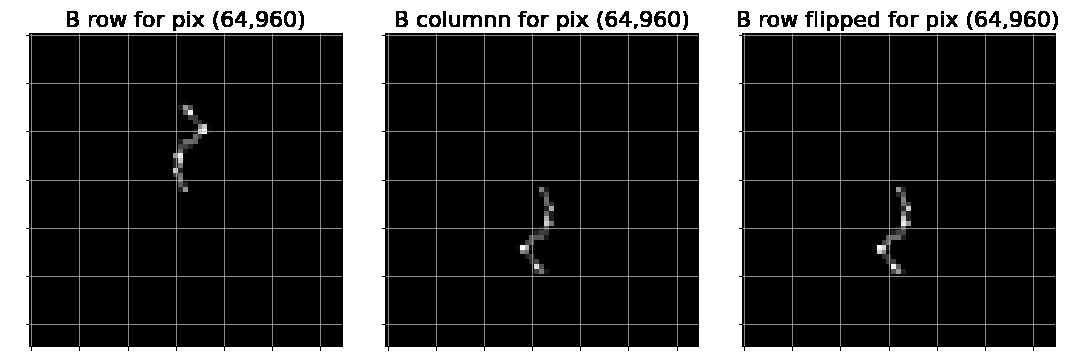}}} \\
     \fcolorbox{yellow}{white}{\includegraphics[trim=448 276 448 276,clip,width=0.15\textwidth]{B_BT_tremor2_no_centered_manmade_01.jpg}} &
    \multicolumn{2}{c}{{\includegraphics[trim=10 10 0 0,clip,width=0.52\textwidth]{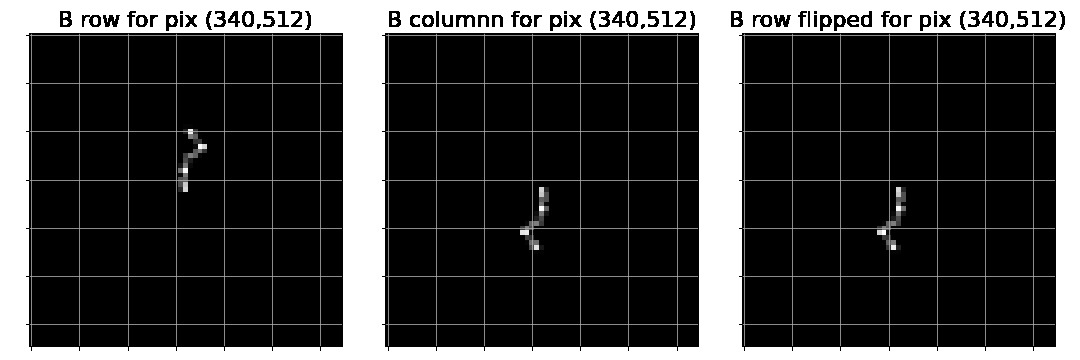}}}  \\
\fcolorbox{green}{white}{\includegraphics[trim=0 0 960 616,clip,width=0.15\textwidth]{B_BT_tremor2_no_centered_manmade_01.jpg}} &
\multicolumn{2}{c}{ {\includegraphics[trim=10 10 0 0,clip,width=0.52\textwidth]{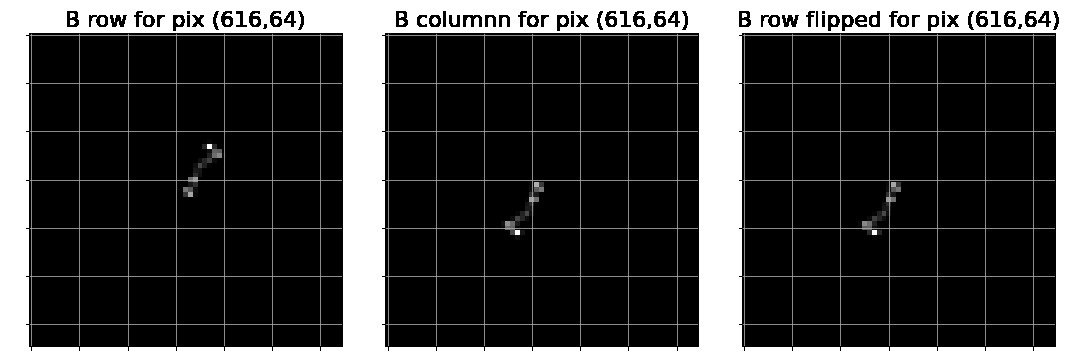}}} \\ 
\fcolorbox{orange}{white}{\includegraphics[trim=960 0 0  616,clip,width=0.15\textwidth]{B_BT_tremor2_no_centered_manmade_01.jpg}} &
\multicolumn{2}{c}{ {\includegraphics[trim=10 10 0 0,clip,width=0.52\textwidth]{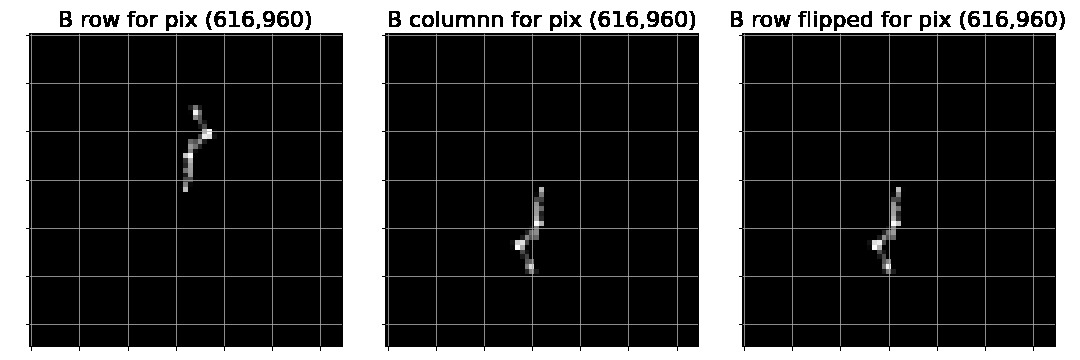}} }\\
    \end{tabular}
 \caption[Blur kernels maps for arbitrary camera shakes 1.]{Blur kernels maps for arbitrary camera shakes are spatially variants and $\mathbf{B}$ differs form $\mathbf{B}^T$. By flipping the rows of $\mathbf{B}$ (seen as an image) we have an accurate approximation of the $\mathbf{B}$ columns in all the image regions.\label{fig:B_BT_tremor2_no_centered}}
\end{figure}

\begin{figure*}[ht]
    \centering
    \setlength{\tabcolsep}{1pt} %
    \begin{tabular}{ccc}  %
     \multicolumn{3}{c}{Blurry} \\

  \multicolumn{3}{c}{
         \begin{tikzpicture}
            \node[anchor=south west,inner sep=0pt] (img) {\includegraphics[width=0.5\textwidth]{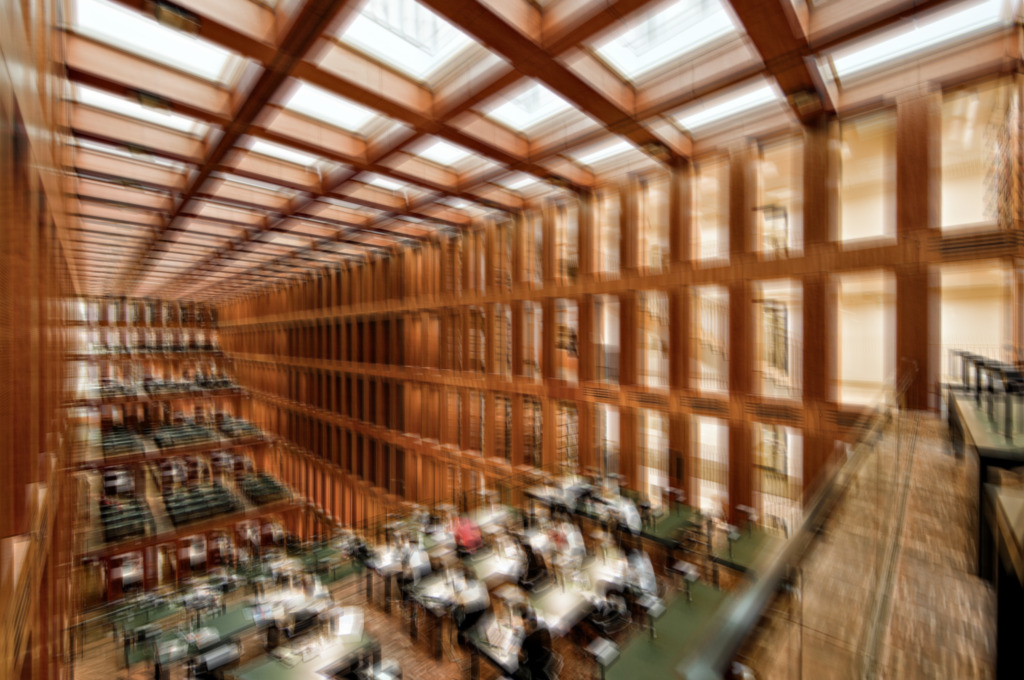}};
  \begin{scope}[x={(img.south east)},y={(img.north west)}]
    \draw[red,very thick] (0/1024,616/680) rectangle (64/1024,680/680);
    \draw[blue,very thick] (960/1024,616/680) rectangle (1024/1024,680/680);
    \draw[green,very thick] (0/1024,0/680) rectangle (64/1024,64/680);
    \draw[orange,very thick] (960/1024,0/680) rectangle (1024/1024,64/680);
    \draw[yellow,very thick] (480/1024,308/680) rectangle (544/1024,372/680);
  \end{scope}
        \end{tikzpicture}
        }  \\
        
     \multicolumn{3}{c}{$\mathbf{B}$ rows and columns for corners and the central pixel} \\
     \fcolorbox{red}{white}{\includegraphics[trim=0 616 960 0,clip,width=0.15\textwidth]{B_BT_tremor3_no_centered_manmade_01.jpg}} &
     \multicolumn{2}{c}{ {\includegraphics[trim=10 10 0 0,clip,width=0.52\textwidth]{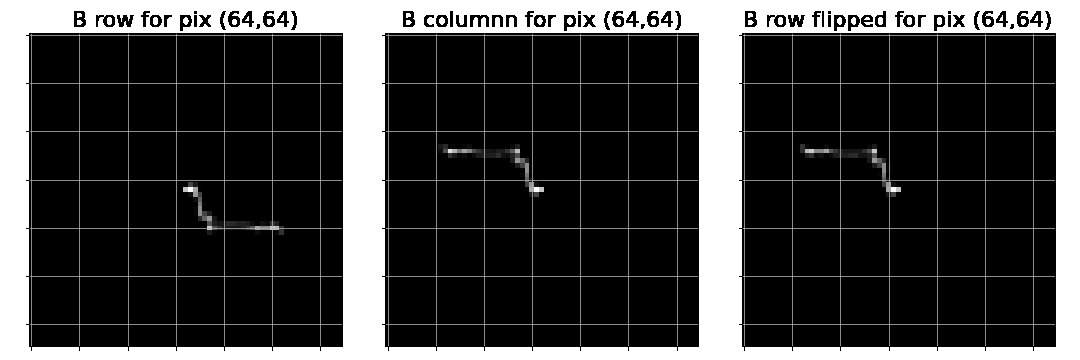}}} \\
     \fcolorbox{blue}{white}{\includegraphics[trim=960 616 0 0,clip,width=0.15\textwidth]{B_BT_tremor3_no_centered_manmade_01.jpg}} &
     \multicolumn{2}{c}{{\includegraphics[trim=10 10 0 0,clip,width=0.52\textwidth]{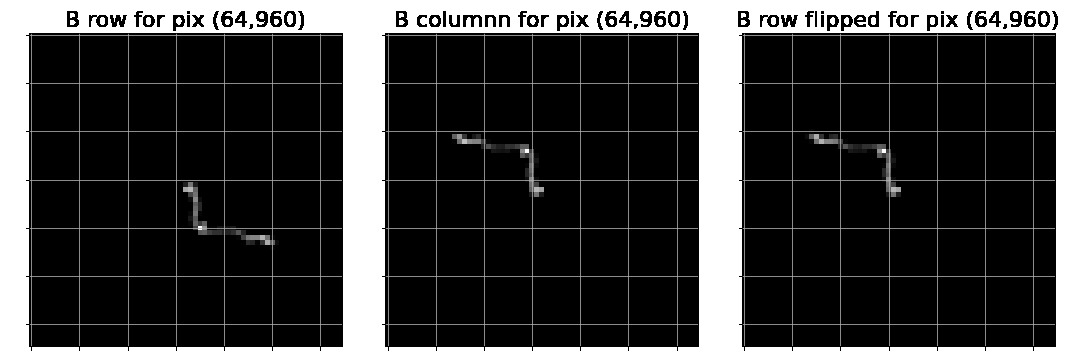}}} \\
     \fcolorbox{yellow}{white}{\includegraphics[trim=448 276 448 276,clip,width=0.15\textwidth]{B_BT_tremor3_no_centered_manmade_01.jpg}} &
    \multicolumn{2}{c}{{\includegraphics[trim=10 10 0 0,clip,width=0.52\textwidth]{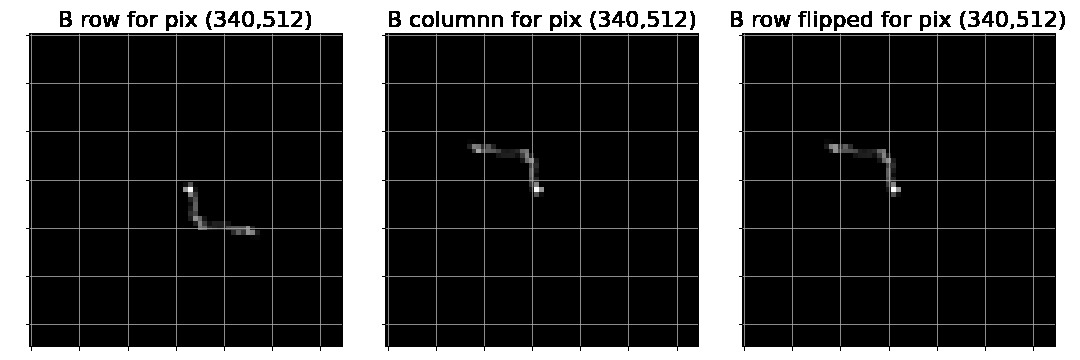}}}  \\
\fcolorbox{green}{white}{\includegraphics[trim=0 0 960 616,clip,width=0.15\textwidth]{B_BT_tremor3_no_centered_manmade_01.jpg}} &
\multicolumn{2}{c}{ {\includegraphics[trim=10 10 0 0,clip,width=0.52\textwidth]{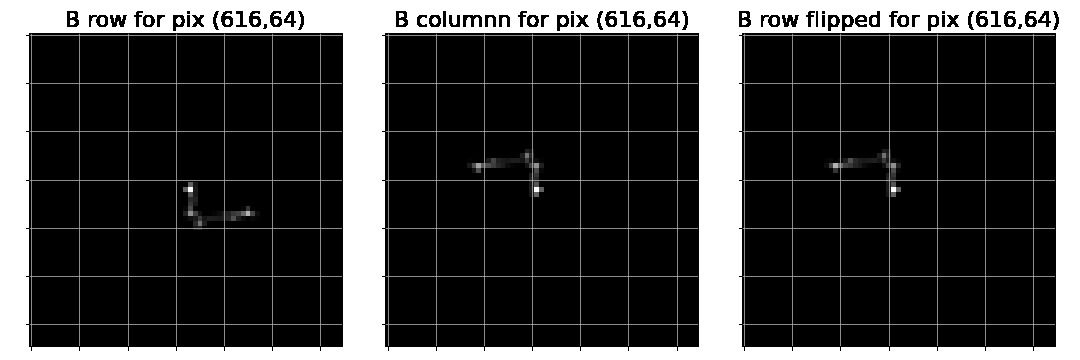}}} \\ 
\fcolorbox{orange}{white}{\includegraphics[trim=960 0 0  616,clip,width=0.15\textwidth]{B_BT_tremor3_no_centered_manmade_01.jpg}} &
\multicolumn{2}{c}{ {\includegraphics[trim=10 10 0 0,clip,width=0.52\textwidth]{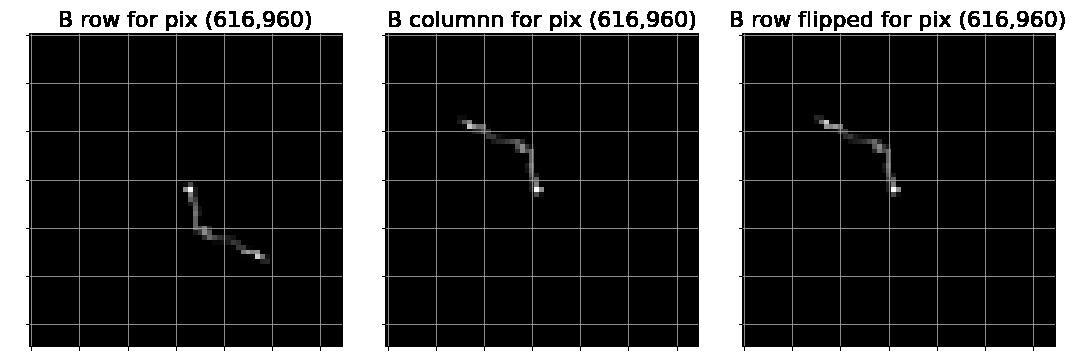}} }\\
    \end{tabular}
    \caption[Blur kernels maps for arbitrary camera shakes.]{Blur kernels maps for arbitrary camera shakes are spatially variant, and $\mathbf{B}$ differs from $\mathbf{B}^T$. By flipping the rows of $\mathbf{B}$ (seen as an image), we have an accurate approximation of the columns of $\mathbf{B}$.\label{fig:B_BT_tremor3_no_centered}
    }
\end{figure*}

\Cref{fig:B_BT_pure_yaw,fig:B_BT_pure_roll}  show that the approximation is effective for pure \textit{yaw} and \textit{roll} trajectories. The behavior of the \textit{pitch} is similar to the \textit{yaw} but in the other image axis. The simulated trajectories are centered at zero and assume constant velocity. Since the kernels are linear and symmetric, the kernel fields of $\mathbf{B}$ and $\mathbf{B}^T$ are identical.   \Cref{fig:B_BT_pure_roll_5} shows that the symmetry of the kernels (and the approximation) is broken for extreme \textit{rolls}, but holds well for moderated multi-axis camera shakes. We observe that for \textit{pitch} and \textit{yaw}, differences are barely noticeable, while for \textit{roll} subtle differences can be observed around the image borders when rotations are severe, exceeding 2 degrees. %
In \Cref{fig:B_BT_tremor1,fig:B_BT_tremor2} we show examples of camera shakes generated by simulating tremors \cite{gavant2011physiological}. Since the motion velocity is not constant for real camera shakes, the kernels fields of $\mathbf{B}$ and $\mathbf{B}^T$ look very different. Nonetheless, the approximation holds within the same range of angles observed for constant velocity motions. 
Additionally, \Cref{fig:B_BT_tremor1_no_centered,fig:B_BT_tremor2_no_centered,fig:B_BT_tremor3_no_centered} show that this approximation holds even for trajectories that are not centered. 

\clearpage

\section{More results on Lai nonuniform \cite{lai2016comparative}} 

\begin{figure*}[ht]
\centering
\begin{tabular}{c}

\hspace{60pt} Blurry \hspace{110pt} Blurry Patches \hspace{50pt} Ground Truth kernels  \\
\begin{tikzpicture}[spy using outlines={rectangle, magnification=4.2, connect spies}]
  \node {\includegraphics[width=0.4\textwidth]{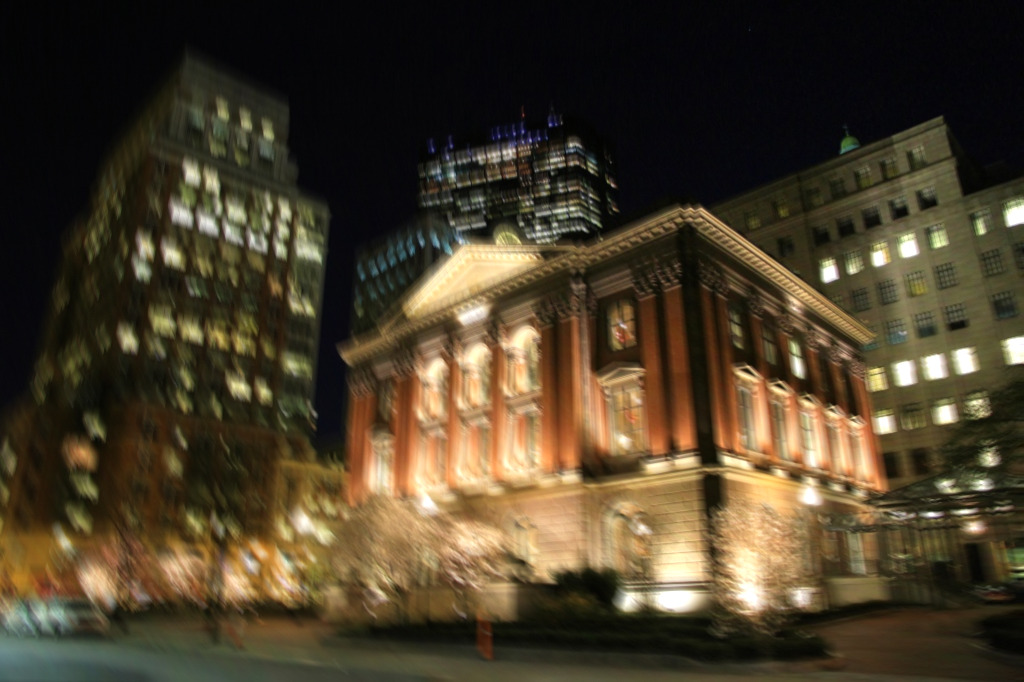}};

  \spy[draw=blue, thick, size=\pxsize]  on (-74.05pt,  42.43pt) in node [left] at (162pt,  34.33pt);
  \spy[draw=red,  line width=4pt, size=\pxsize]  on ( 74.23pt,  42.43pt) in node [left] at (230pt,  34.33pt);
  \spy[draw=green, line width=4pt, size=\pxsize]  on (-74.05pt, -42.59pt) in node [left] at (162pt, -34.33pt);
  \spy[draw=yellow, line width=4pt,size=\pxsize]  on ( 74.23pt, -42.59pt) in node [left] at (230pt, -34.33pt);

  \setlength{\fboxrule}{1pt} %
  \setlength{\fboxsep}{0pt}  %

  \node at (276pt,35pt) {\fcolorbox{blue}{white}{\includegraphics[trim=67 553 895 67, clip]{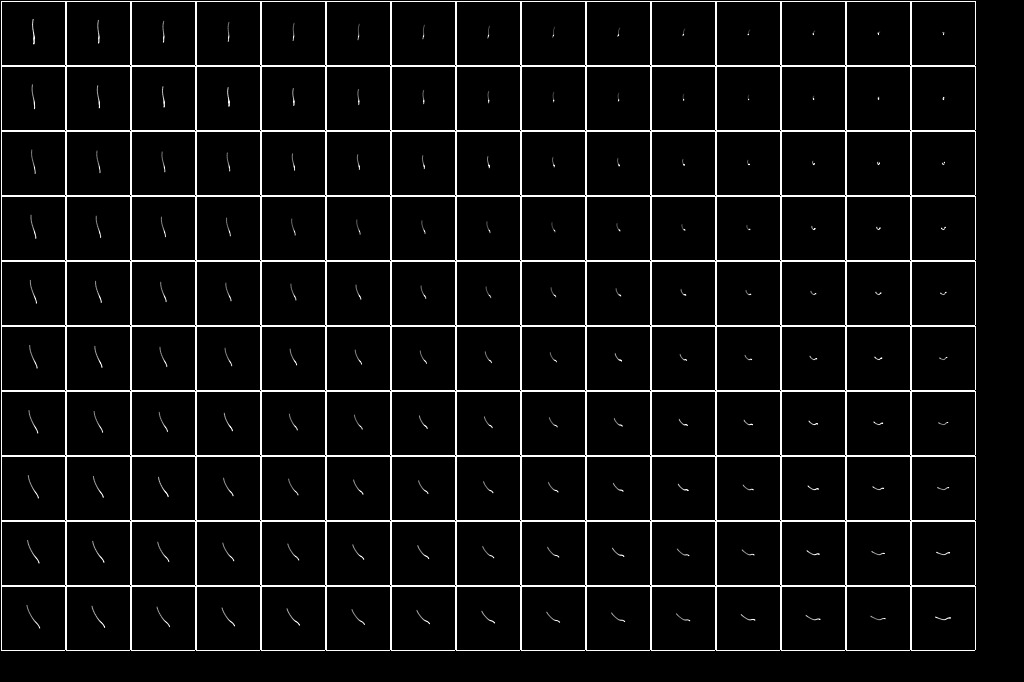}}};
  \node at (344pt,35pt) {\fcolorbox{red}{white}{\includegraphics[trim=912 553 50 67, clip]{Lai_non_uniform_GT_kernels_saturated_02_gyro_01.jpg}}};
  \node at (276pt,-35pt) {\fcolorbox{green}{white}{\includegraphics[trim=67 33 895 587, clip]{Lai_non_uniform_GT_kernels_saturated_02_gyro_01.jpg}}};
  \node at (344pt,-35pt) {\fcolorbox{yellow}{white}{\includegraphics[trim=912 33 50 587, clip]{Lai_non_uniform_GT_kernels_saturated_02_gyro_01.jpg}}};
\end{tikzpicture} \\

\hspace{35pt} Restored (w/o optim)\hspace{50pt} Restored Patches (w/o optim) \hspace{5pt} Restored kernels (w/o optim) \\
\begin{tikzpicture}[spy using outlines={rectangle, magnification=4.2, connect spies}]
  \node {\includegraphics[width=0.4\textwidth]{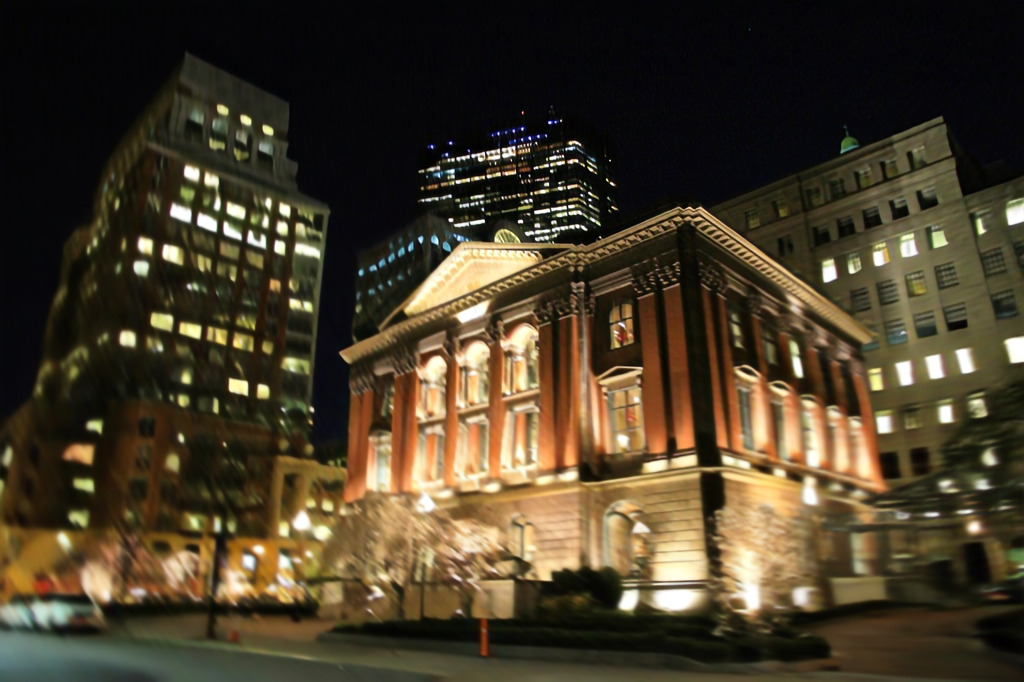}};

  \spy[draw=blue, thick, size=\pxsize]  on (-74.05pt,  42.43pt) in node [left] at (162pt,  34.33pt);
  \spy[draw=red,  line width=4pt, size=\pxsize]  on ( 74.23pt,  42.43pt) in node [left] at (230pt,  34.33pt);
  \spy[draw=green, line width=4pt, size=\pxsize]  on (-74.05pt, -42.59pt) in node [left] at (162pt, -34.33pt);
  \spy[draw=yellow, line width=4pt,size=\pxsize]  on ( 74.23pt, -42.59pt) in node [left] at (230pt, -34.33pt);

  \setlength{\fboxrule}{1pt} %
  \setlength{\fboxsep}{0pt}  %

  \node at (276pt,35pt) {\fcolorbox{blue}{white}{\includegraphics[trim=66 553 895 66, clip]{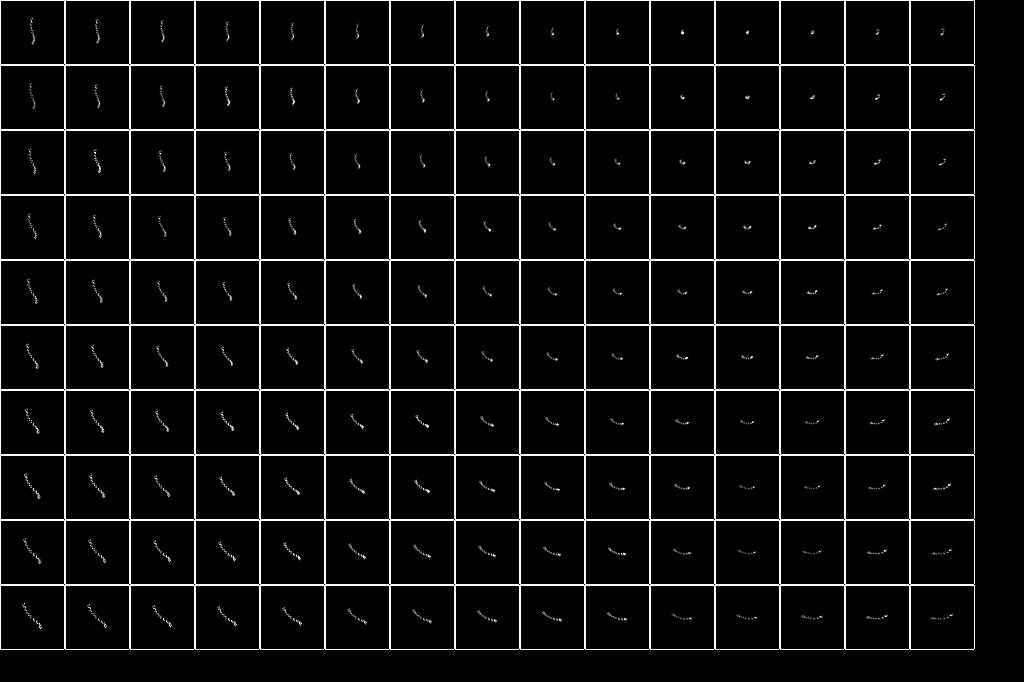}}};
  \node at (344pt,35pt) {\fcolorbox{red}{white}{\includegraphics[trim=912 553 50 66, clip]{traj_est_ill_aug_raw_saturated_02_gyro_01_kernels_found.jpg}}};
  \node at (276pt,-35pt) {\fcolorbox{green}{white}{\includegraphics[trim=66 33 895 587, clip]{traj_est_ill_aug_raw_saturated_02_gyro_01_kernels_found.jpg}}};
  \node at (344pt,-35pt) {\fcolorbox{yellow}{white}{\includegraphics[trim=912 33 50 587, clip]{traj_est_ill_aug_raw_saturated_02_gyro_01_kernels_found.jpg}}};
\end{tikzpicture} \\

\hspace{40pt} Restored \hspace{110pt} Restored Patches \hspace{50pt} Restored kernels  \\
\begin{tikzpicture}[spy using outlines={rectangle, magnification=4.2, connect spies}]
  \node {\includegraphics[width=0.4\textwidth]{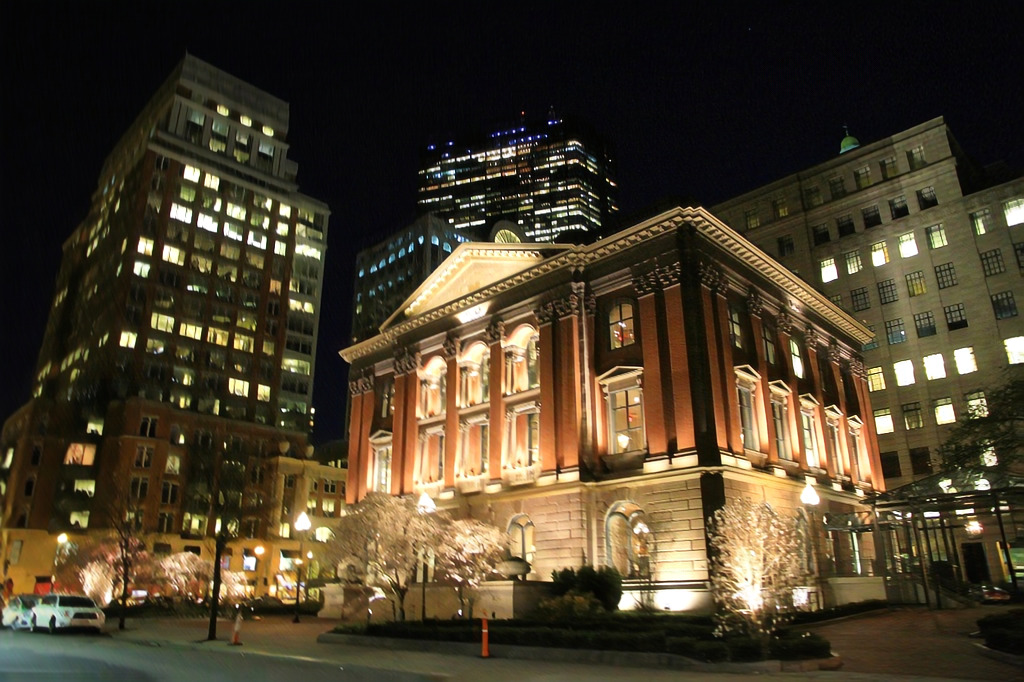}};

  \spy[draw=blue, thick, size=\pxsize]  on (-74.05pt,  42.43pt) in node [left] at (162pt,  34.33pt);
  \spy[draw=red,  line width=4pt, size=\pxsize]  on ( 74.23pt,  42.43pt) in node [left] at (230pt,  34.33pt);
  \spy[draw=green, line width=4pt, size=\pxsize]  on (-74.05pt, -42.59pt) in node [left] at (162pt, -34.33pt);
  \spy[draw=yellow, line width=4pt,size=\pxsize]  on ( 74.23pt, -42.59pt) in node [left] at (230pt, -34.33pt);

  \setlength{\fboxrule}{1pt} %
  \setlength{\fboxsep}{0pt}  %

  \node at (276pt,35pt) {\fcolorbox{blue}{white}{\includegraphics[trim=66 553 895 66, clip]{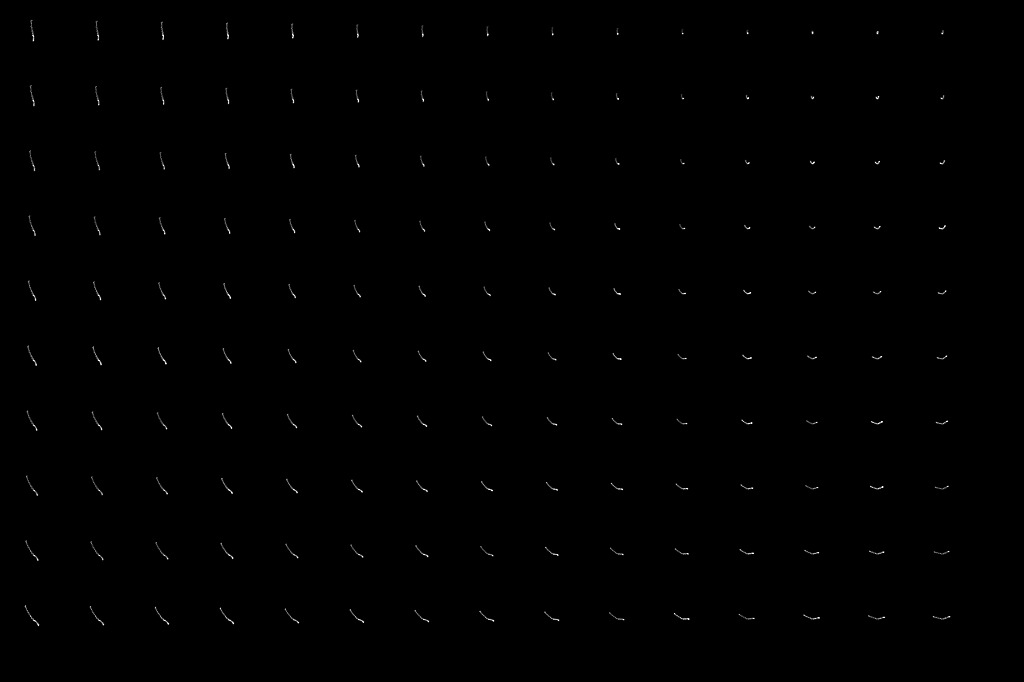}}};
  \node at (344pt,35pt) {\fcolorbox{red}{white}{\includegraphics[trim=912 553 50 66, clip]{traj_est_ill_aug_optim_saturated_02_gyro_01_kernels_found.jpg}}};
  \node at (276pt,-35pt) {\fcolorbox{green}{white}{\includegraphics[trim=66 33 895 583, clip]{traj_est_ill_aug_optim_saturated_02_gyro_01_kernels_found.jpg}}};
  \node at (344pt,-35pt) {\fcolorbox{yellow}{white}{\includegraphics[trim=912 33 50 583, clip]{traj_est_ill_aug_optim_saturated_02_gyro_01_kernels_found.jpg}}};
\end{tikzpicture} \\
Trajectory (w/o optim.) \hspace{80pt}  Trajectory  \hspace{110pt} GT Trajectory \\ 
     \includegraphics[width=0.33\textwidth, trim=45 40 45 45, clip]{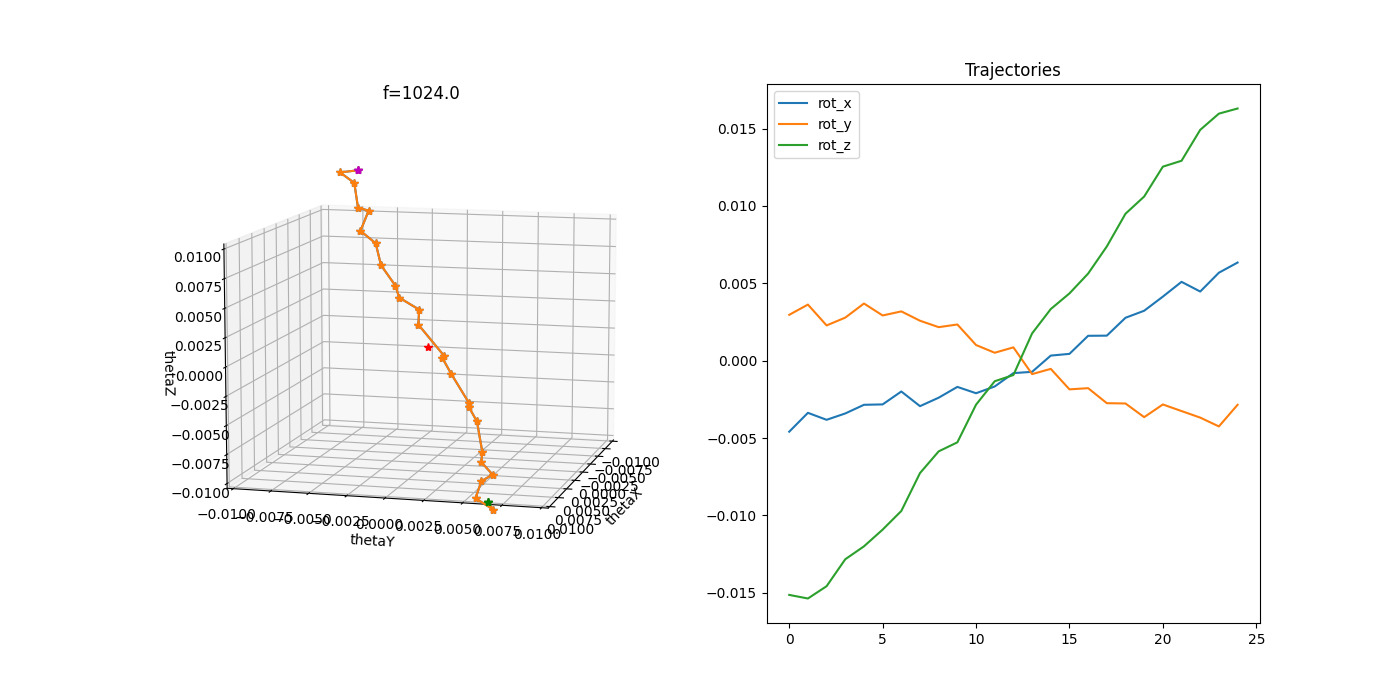} \includegraphics[width=0.33\textwidth, trim=45 40 45 45, clip]{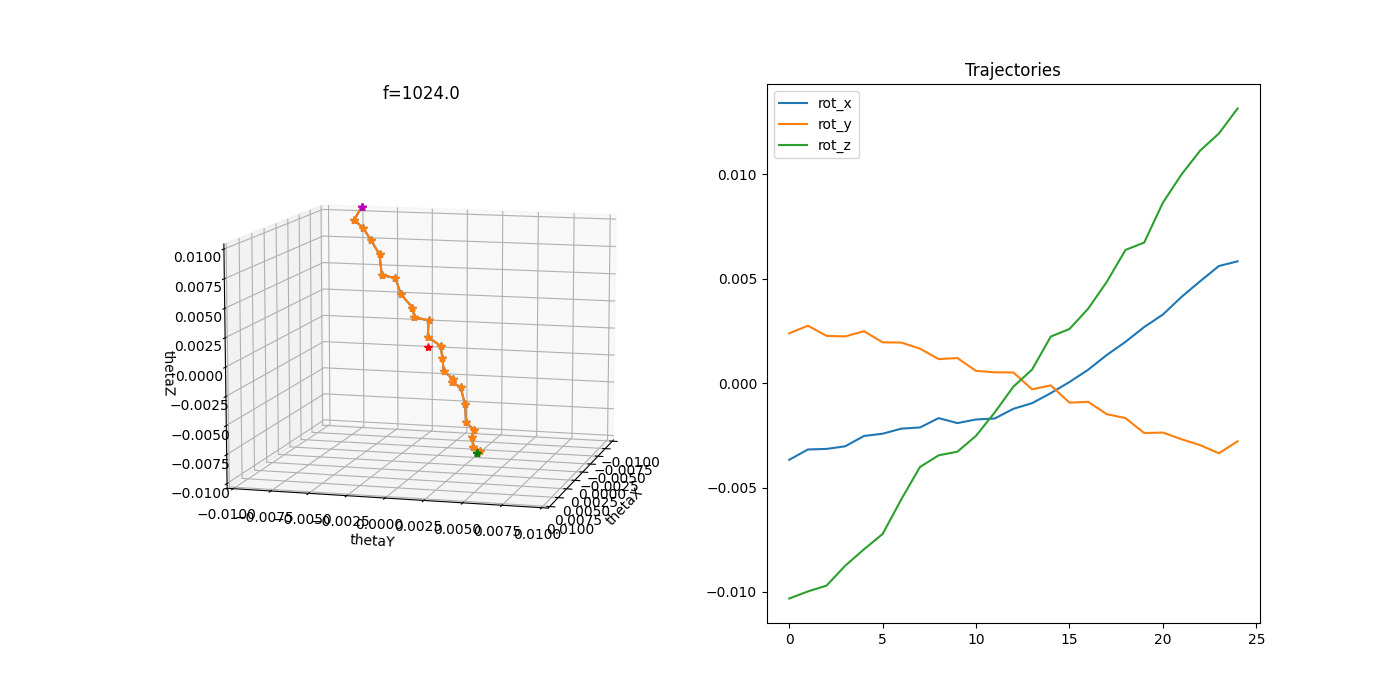}  
        \includegraphics[width=0.33\textwidth, trim=45 40 45 45, clip]{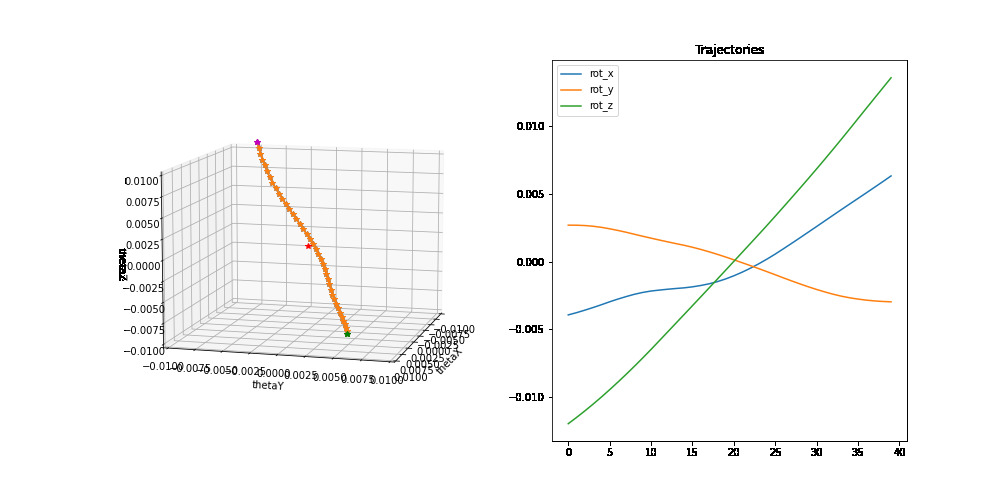} \\  
\end{tabular}
\caption{Kernels and restorations obtained with (3rd row) and without (2nd row) the trajectory refinement step. Kernel estimation is challenging for low textured regions since there is almost no information about the blur.  The global constraint imposed during the optimization improves the estimated kernels and the restored image. The last row shows the initial trajectory estimated by the TPN, the trajectory after the refinement step, and the ground-truth trajectory. We represent a trajectory as a sequence of 3D points (\textit{pitch}, \textit{yaw}, \textit{roll}), and also show, using different color curves,  the evolution of each rotation during the exposure time.  \label{fig:Lai_non_uniform_traj1}} 
\end{figure*}

\begin{figure*}[ht]
\centering
\begin{tabular}{c}

\hspace{60pt} Blurry \hspace{110pt} Blurry Patches \hspace{50pt} Ground Truth kernels  \\
\begin{tikzpicture}[spy using outlines={rectangle, magnification=4.2, connect spies}]
  \node {\includegraphics[width=0.4\textwidth]{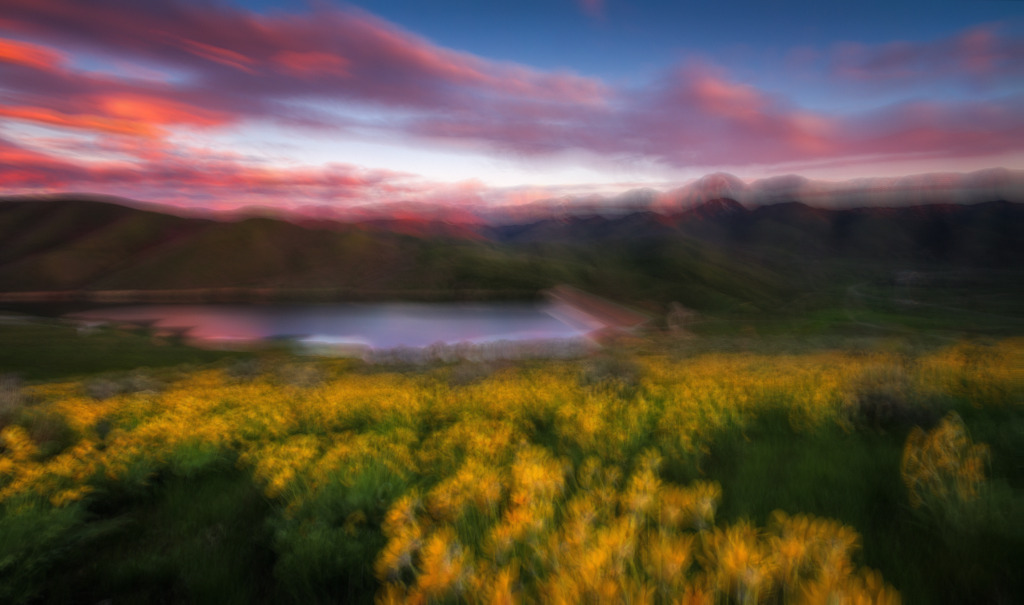}};

  \spy[draw=blue, thick, size=\pxsize]  on (-74.05pt,  42.43pt) in node [left] at (162pt,  34.33pt);
  \spy[draw=red,  line width=4pt, size=\pxsize]  on ( 74.23pt,  42.43pt) in node [left] at (230pt,  34.33pt);
  \spy[draw=green, line width=4pt, size=\pxsize]  on (-74.05pt, -42.59pt) in node [left] at (162pt, -34.33pt);
  \spy[draw=yellow, line width=4pt,size=\pxsize]  on ( 74.23pt, -42.59pt) in node [left] at (230pt, -34.33pt);

  \setlength{\fboxrule}{1pt} %
  \setlength{\fboxsep}{0pt}  %

  \node at (276pt,35pt) {\fcolorbox{blue}{white}{\includegraphics[trim=67 480 895 67, clip]{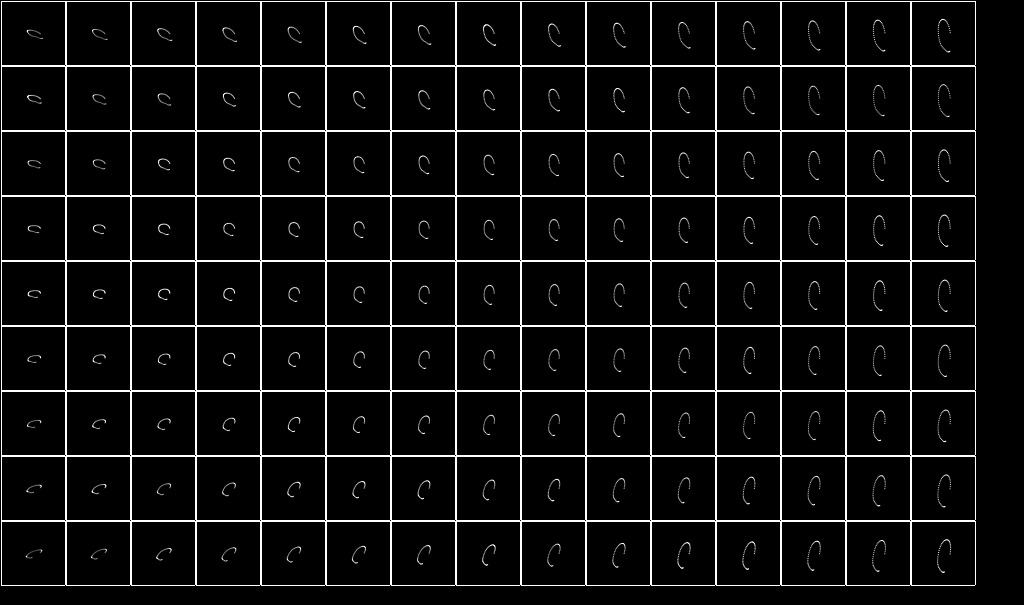}}};
  \node at (344pt,35pt) {\fcolorbox{red}{white}{\includegraphics[trim=912 480 50 67, clip]{Lai_non_uniform_GT_kernels_natural_03_gyro_02.jpg}}};
  \node at (276pt,-35pt) {\fcolorbox{green}{white}{\includegraphics[trim=67 20 895 525, clip]{Lai_non_uniform_GT_kernels_natural_03_gyro_02.jpg}}};
  \node at (344pt,-35pt) {\fcolorbox{yellow}{white}{\includegraphics[trim=912 20 50 525, clip]{Lai_non_uniform_GT_kernels_natural_03_gyro_02.jpg}}};
\end{tikzpicture} \\

\hspace{35pt} Restored (w/o optim)\hspace{50pt} Restored Patches (w/o optim) \hspace{5pt} Restored kernels (w/o optim) \\
\begin{tikzpicture}[spy using outlines={rectangle, magnification=4.2, connect spies}]
  \node {\includegraphics[width=0.4\textwidth]{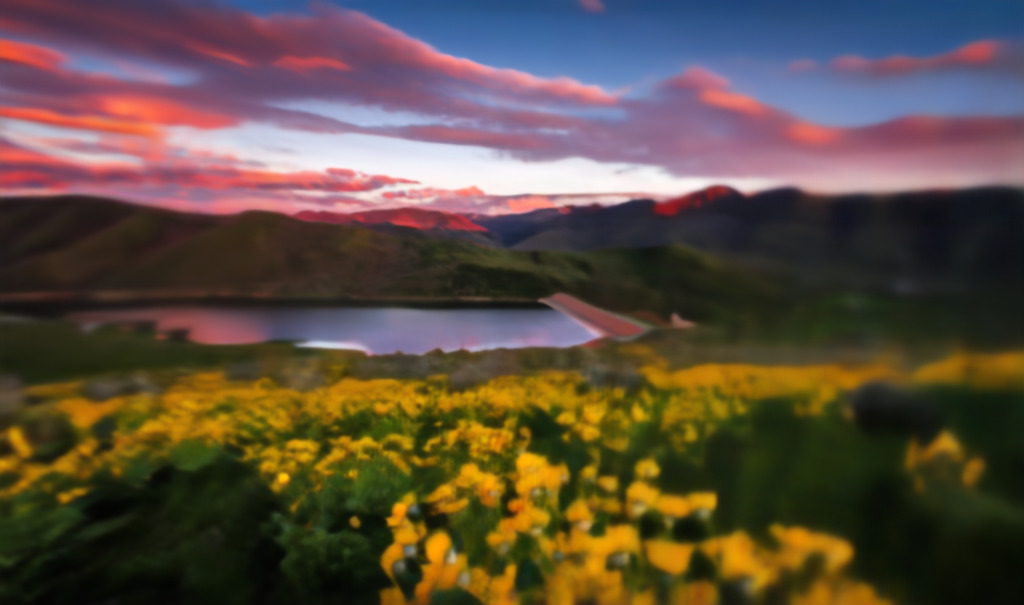}};

  \spy[draw=blue, thick, size=\pxsize]  on (-74.05pt,  42.43pt) in node [left] at (162pt,  34.33pt);
  \spy[draw=red,  line width=4pt, size=\pxsize]  on ( 74.23pt,  42.43pt) in node [left] at (230pt,  34.33pt);
  \spy[draw=green, line width=4pt, size=\pxsize]  on (-74.05pt, -42.59pt) in node [left] at (162pt, -34.33pt);
  \spy[draw=yellow, line width=4pt,size=\pxsize]  on ( 74.23pt, -42.59pt) in node [left] at (230pt, -34.33pt);

  \setlength{\fboxrule}{1pt} %
  \setlength{\fboxsep}{0pt}  %

  \node at (276pt,35pt) {\fcolorbox{blue}{white}{\includegraphics[trim=66 480 895 66, clip]{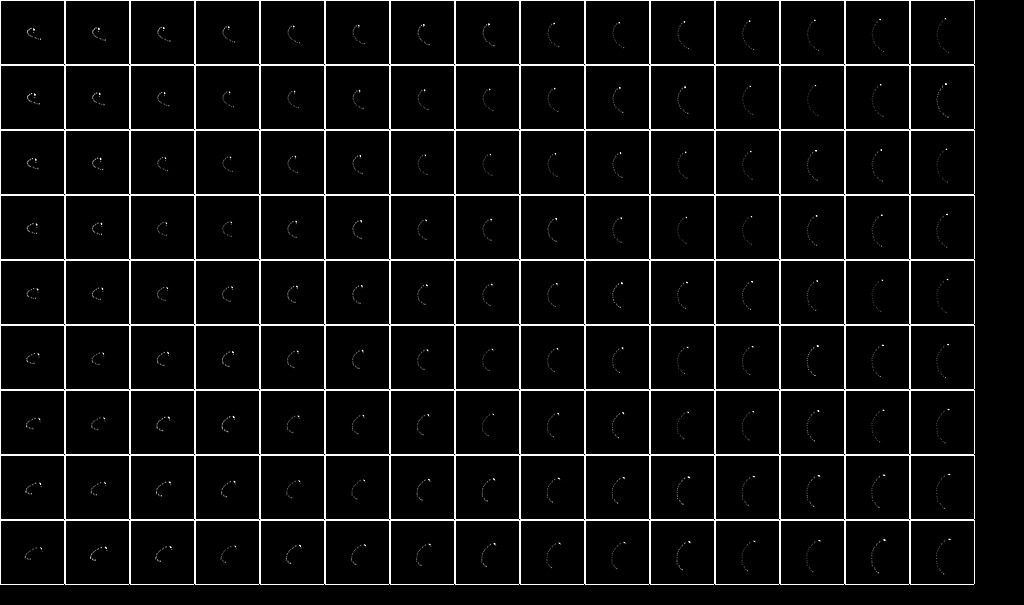}}};
  \node at (344pt,35pt) {\fcolorbox{red}{white}{\includegraphics[trim=912 480 50 65, clip]{traj_est_ill_aug_raw_natural_03_gyro_02_kernels_found.jpg}}};
  \node at (276pt,-35pt) {\fcolorbox{green}{white}{\includegraphics[trim=66 20 895 525, clip]{traj_est_ill_aug_raw_natural_03_gyro_02_kernels_found.jpg}}};
  \node at (344pt,-35pt) {\fcolorbox{yellow}{white}{\includegraphics[trim=912 20 50 525, clip]{traj_est_ill_aug_raw_natural_03_gyro_02_kernels_found.jpg}}};
\end{tikzpicture} \\

\hspace{40pt} Restored \hspace{110pt} Restored Patches \hspace{50pt} Restored kernels  \\
\begin{tikzpicture}[spy using outlines={rectangle, magnification=4.2, connect spies}]
  \node {\includegraphics[width=0.4\textwidth]{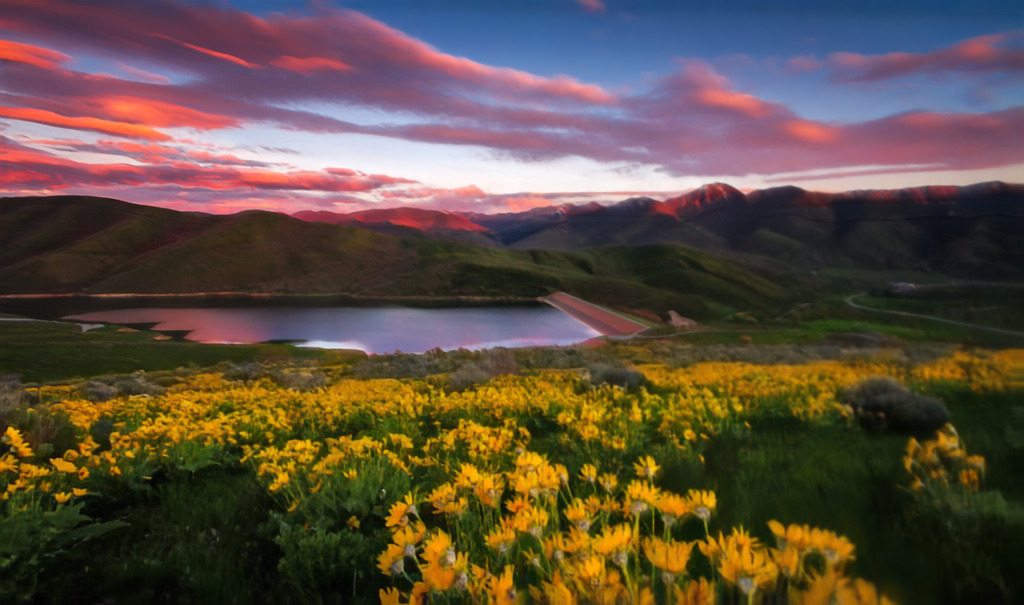}};

  \spy[draw=blue, thick, size=\pxsize]  on (-74.05pt,  42.43pt) in node [left] at (162pt,  34.33pt);
  \spy[draw=red,  line width=4pt, size=\pxsize]  on ( 74.23pt,  42.43pt) in node [left] at (230pt,  34.33pt);
  \spy[draw=green, line width=4pt, size=\pxsize]  on (-74.05pt, -42.59pt) in node [left] at (162pt, -34.33pt);
  \spy[draw=yellow, line width=4pt,size=\pxsize]  on ( 74.23pt, -42.59pt) in node [left] at (230pt, -34.33pt);

  \setlength{\fboxrule}{1pt} %
  \setlength{\fboxsep}{0pt}  %

  \node at (276pt,35pt) {\fcolorbox{blue}{white}{\includegraphics[trim=66 480 895 66, clip]{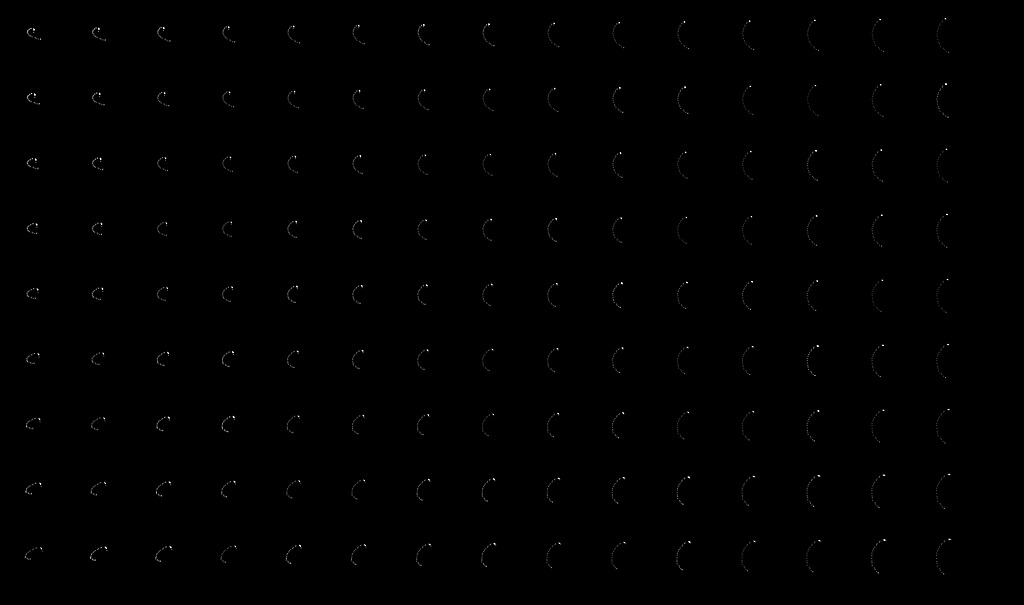}}};
  \node at (344pt,35pt) {\fcolorbox{red}{white}{\includegraphics[trim=912 480 50 65, clip]{traj_est_ill_aug_optim_natural_03_gyro_02_kernels_found.jpg}}};
  \node at (276pt,-35pt) {\fcolorbox{green}{white}{\includegraphics[trim=66 20 895 525, clip]{traj_est_ill_aug_optim_natural_03_gyro_02_kernels_found.jpg}}};
  \node at (344pt,-35pt) {\fcolorbox{yellow}{white}{\includegraphics[trim=912 20 50 525, clip]{traj_est_ill_aug_optim_natural_03_gyro_02_kernels_found.jpg}}};
\end{tikzpicture} \\
Trajectory (w/o optim.) \hspace{80pt}  Trajectory  \hspace{110pt} GT Trajectory \\ 
     \includegraphics[width=0.33\textwidth, trim=45 40 45 45, clip]{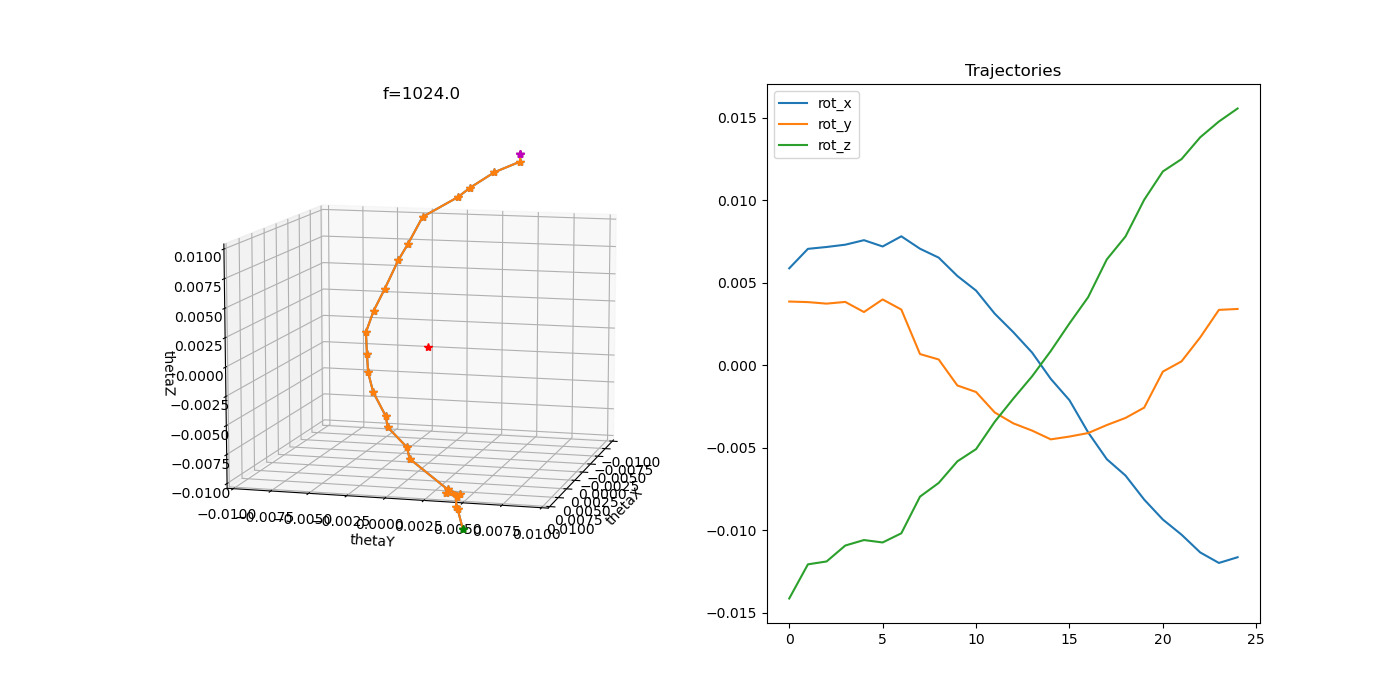} \includegraphics[width=0.33\textwidth, trim=45 40 45 45, clip]{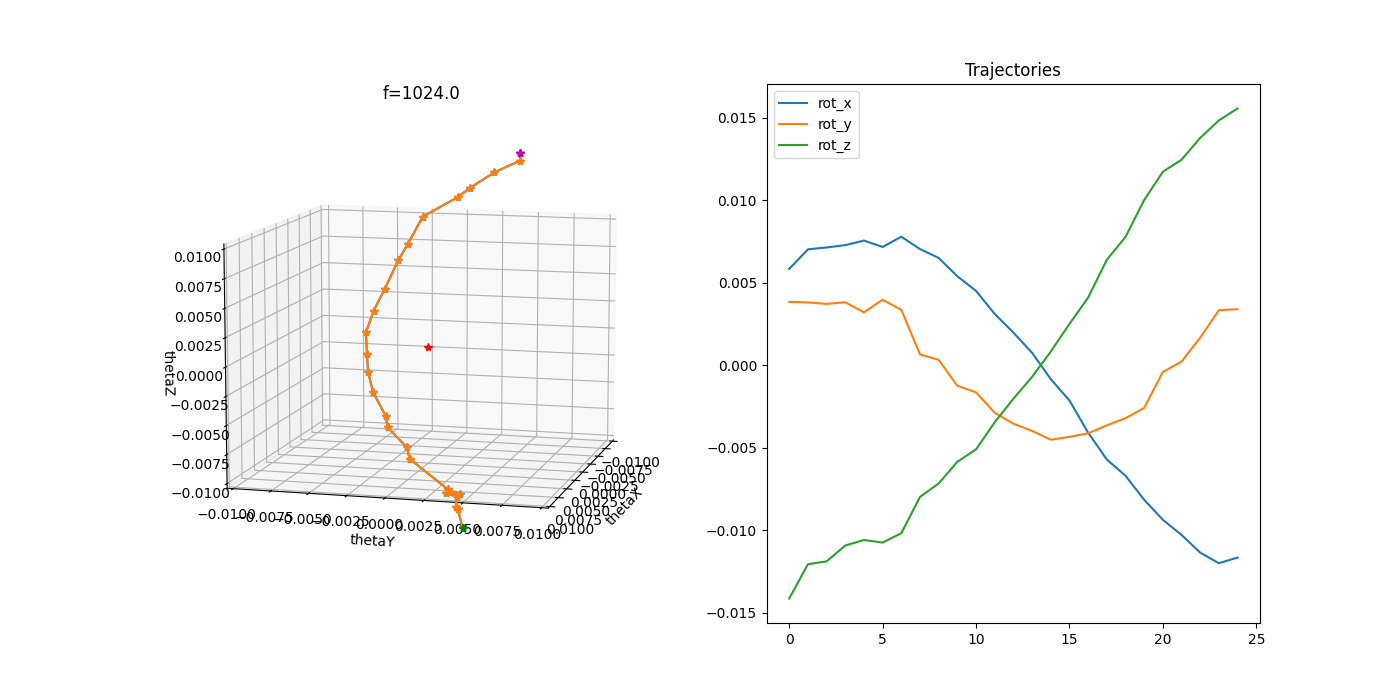}  
        \includegraphics[width=0.33\textwidth, trim=45 40 45 45, clip]{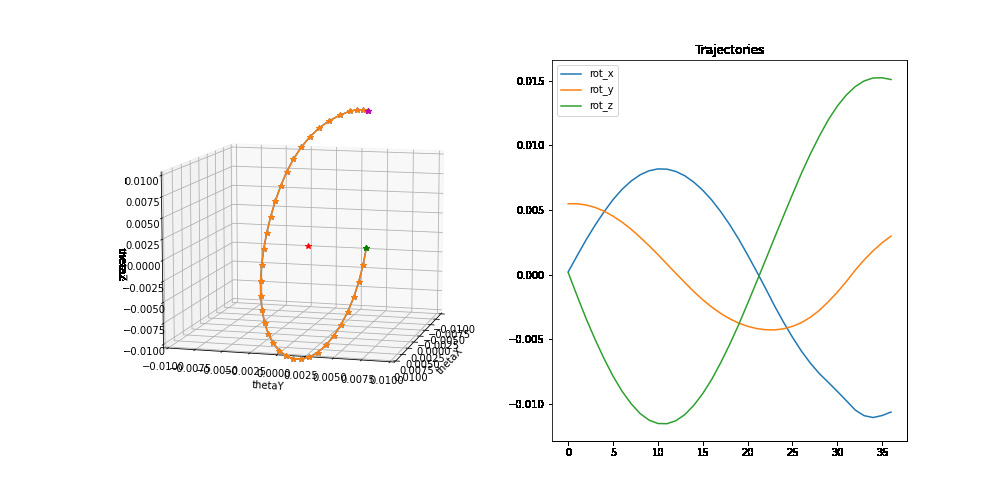} \\  
\end{tabular}
\caption{Kernels and restorations obtained with (3rd row) and without (2nd row) the trajectory refinement step. Kernel estimation is challenging for low textured regions since there is almost no information about the blur.  The global constraint imposed during the optimization improves the estimated kernels and the restored image. The last row shows the initial trajectory estimated by the TPN, the trajectory after the refinement step, and the ground-truth trajectory. We represent a trajectory as a sequence of 3D points (\textit{pitch}, \textit{yaw}, \textit{roll}), and also show, using different color curves,  the evolution of each rotation during the exposure time.  \label{fig:Lai_non_uniform_traj2}} 
\end{figure*}

\begin{figure*}[ht]
\centering
\begin{tabular}{c}

\hspace{60pt} Blurry \hspace{110pt} Blurry Patches \hspace{50pt} Ground Truth kernels  \\
\begin{tikzpicture}[spy using outlines={rectangle, magnification=4.2, connect spies}]
  \node {\includegraphics[width=0.4\textwidth, trim=0 50 0 50, clip]{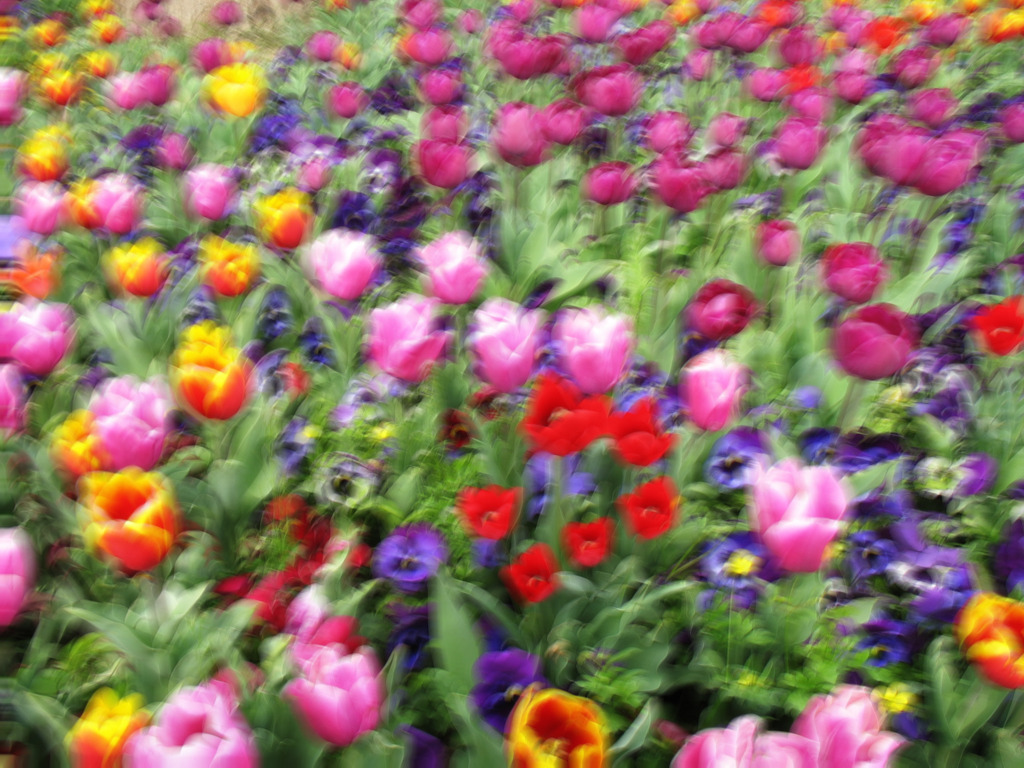}};

  \spy[draw=blue, thick, size=\pxsize]  on (-69.05pt,  47.43pt) in node [left] at (162pt,  33.33pt);
  \spy[draw=red,  line width=4pt, size=\pxsize]  on ( 69.23pt,  47.43pt) in node [left] at (230pt,  33.33pt);
  \spy[draw=green, line width=4pt, size=\pxsize]  on (-69.05pt, -47.59pt) in node [left] at (162pt, -33.33pt);
  \spy[draw=yellow, line width=4pt,size=\pxsize]  on ( 69.23pt, -47.59pt) in node [left] at (230pt, -33.33pt);

  \setlength{\fboxrule}{1pt} %
  \setlength{\fboxsep}{0pt}  %

  \node at (276pt,33pt) {\fcolorbox{blue}{white}{\includegraphics[trim=66 645 895 67, clip]{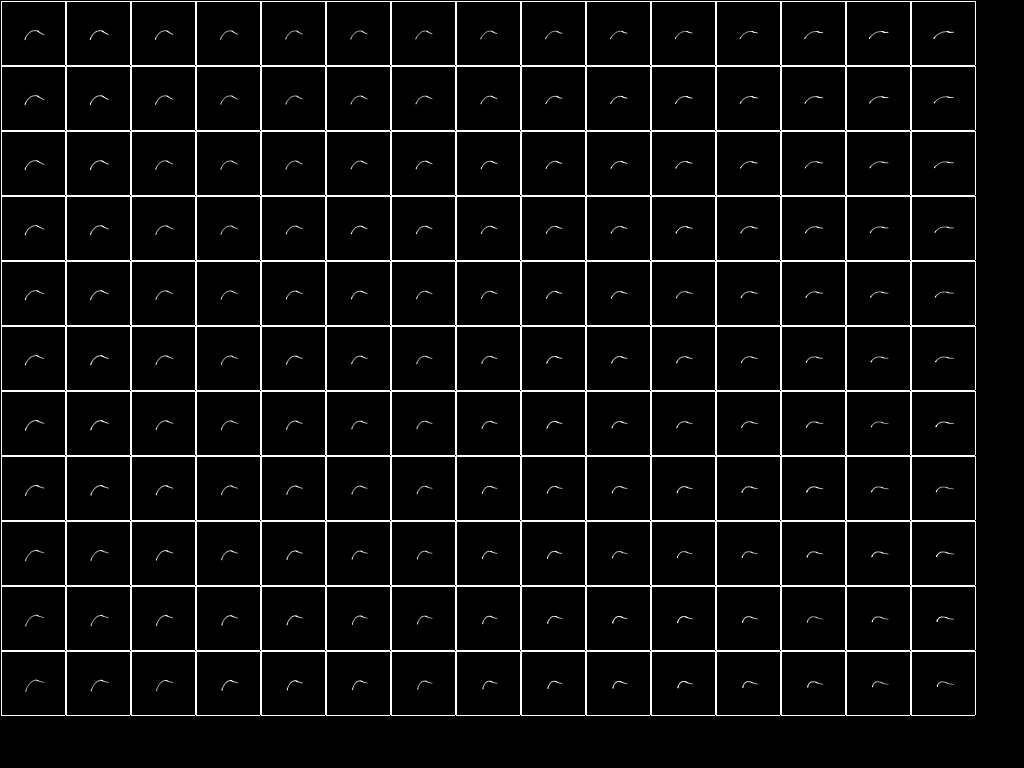}}};
  \node at (344pt,33pt) {\fcolorbox{red}{white}{\includegraphics[trim=912 645 50 67, clip]{Lai_non_uniform_GT_kernels_natural_05_gyro_03.jpg}}};
  \node at (276pt,-33pt) {\fcolorbox{green}{white}{\includegraphics[trim=66 55 895 655, clip]{Lai_non_uniform_GT_kernels_natural_05_gyro_03.jpg}}};
  \node at (344pt,-33pt) {\fcolorbox{yellow}{white}{\includegraphics[trim=912 55 50 655, clip]{Lai_non_uniform_GT_kernels_natural_05_gyro_03.jpg}}};
\end{tikzpicture} \\

\hspace{35pt} Restored (w/o optim)\hspace{50pt} Restored Patches (w/o optim) \hspace{5pt} Restored kernels (w/o optim) \\
\begin{tikzpicture}[spy using outlines={rectangle, magnification=4.2, connect spies}]
  \node {\includegraphics[width=0.4\textwidth, trim=0 50 0 50, clip]{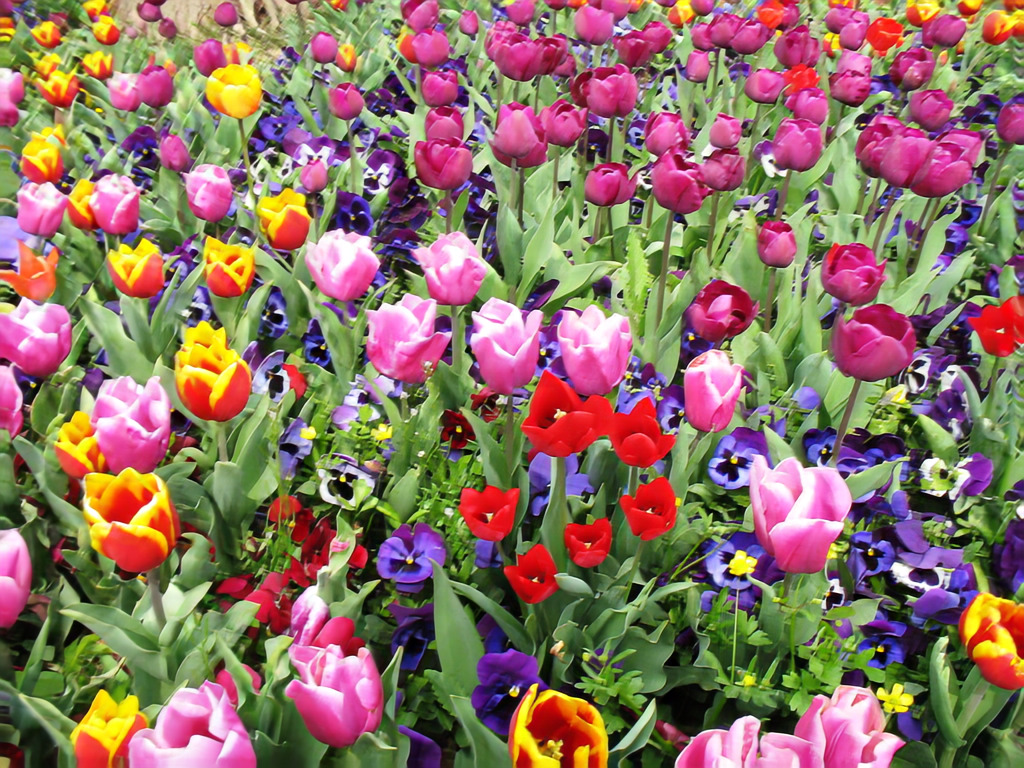}};

  \spy[draw=blue, thick, size=\pxsize]  on (-69.05pt,  47.43pt) in node [left] at (162pt,  33.33pt);
  \spy[draw=red,  line width=4pt, size=\pxsize]  on ( 69.23pt,  47.43pt) in node [left] at (230pt,  33.33pt);
  \spy[draw=green, line width=4pt, size=\pxsize]  on (-69.05pt, -47.59pt) in node [left] at (162pt, -33.33pt);
  \spy[draw=yellow, line width=4pt,size=\pxsize]  on ( 69.23pt, -47.59pt) in node [left] at (230pt, -33.33pt);

  \setlength{\fboxrule}{1pt} %
  \setlength{\fboxsep}{0pt}  %

  \node at (276pt,33pt) {\fcolorbox{blue}{white}{\includegraphics[trim=66 645 895 67, clip]{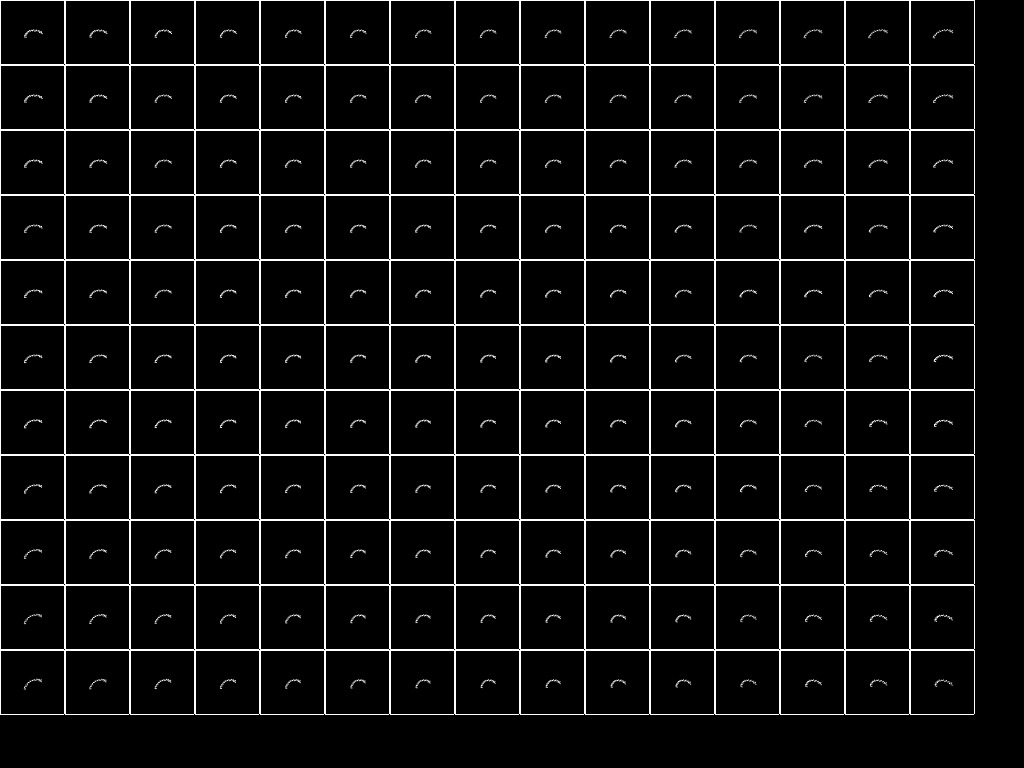}}};
  \node at (344pt,33pt) {\fcolorbox{red}{white}{\includegraphics[trim=912 645 50 67, clip]{traj_est_ill_aug_raw_natural_05_gyro_03_kernels_found.jpg}}};
  \node at (276pt,-33pt) {\fcolorbox{green}{white}{\includegraphics[trim=66 55 895 656, clip]{traj_est_ill_aug_raw_natural_05_gyro_03_kernels_found.jpg}}};
  \node at (344pt,-33pt) {\fcolorbox{yellow}{white}{\includegraphics[trim=912 55 50 656, clip]{traj_est_ill_aug_raw_natural_05_gyro_03_kernels_found.jpg}}};
\end{tikzpicture} \\

\hspace{40pt} Restored \hspace{110pt} Restored Patches \hspace{50pt} Restored kernels  \\
\begin{tikzpicture}[spy using outlines={rectangle, magnification=4.2, connect spies}]
  \node {\includegraphics[width=0.4\textwidth, trim=0 50 0 50, clip]{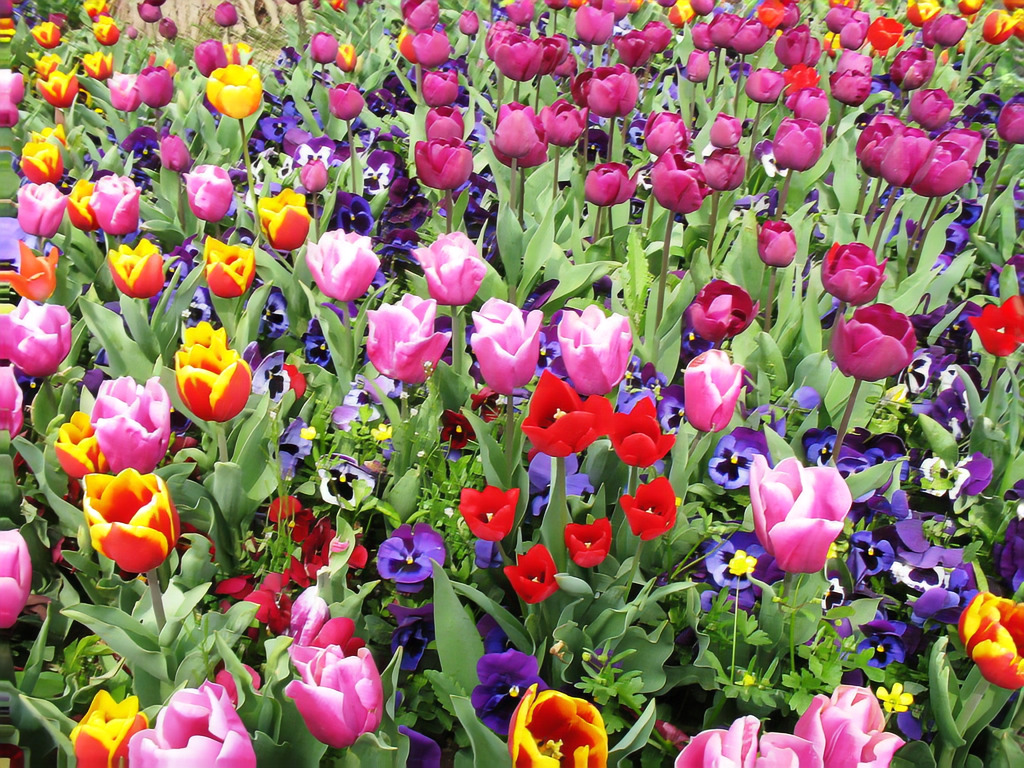}};

  \spy[draw=blue, thick, size=\pxsize]  on (-69.05pt,  47.43pt) in node [left] at (162pt,  33.33pt);
  \spy[draw=red,  line width=4pt, size=\pxsize]  on ( 69.23pt,  47.43pt) in node [left] at (230pt,  33.33pt);
  \spy[draw=green, line width=4pt, size=\pxsize]  on (-69.05pt, -47.59pt) in node [left] at (162pt, -33.33pt);
  \spy[draw=yellow, line width=4pt,size=\pxsize]  on ( 69.23pt, -47.59pt) in node [left] at (230pt, -33.33pt);

  \setlength{\fboxrule}{1pt} %
  \setlength{\fboxsep}{0pt}  %

  \node at (276pt,33pt) {\fcolorbox{blue}{white}{\includegraphics[trim=66 635 895 67, clip]{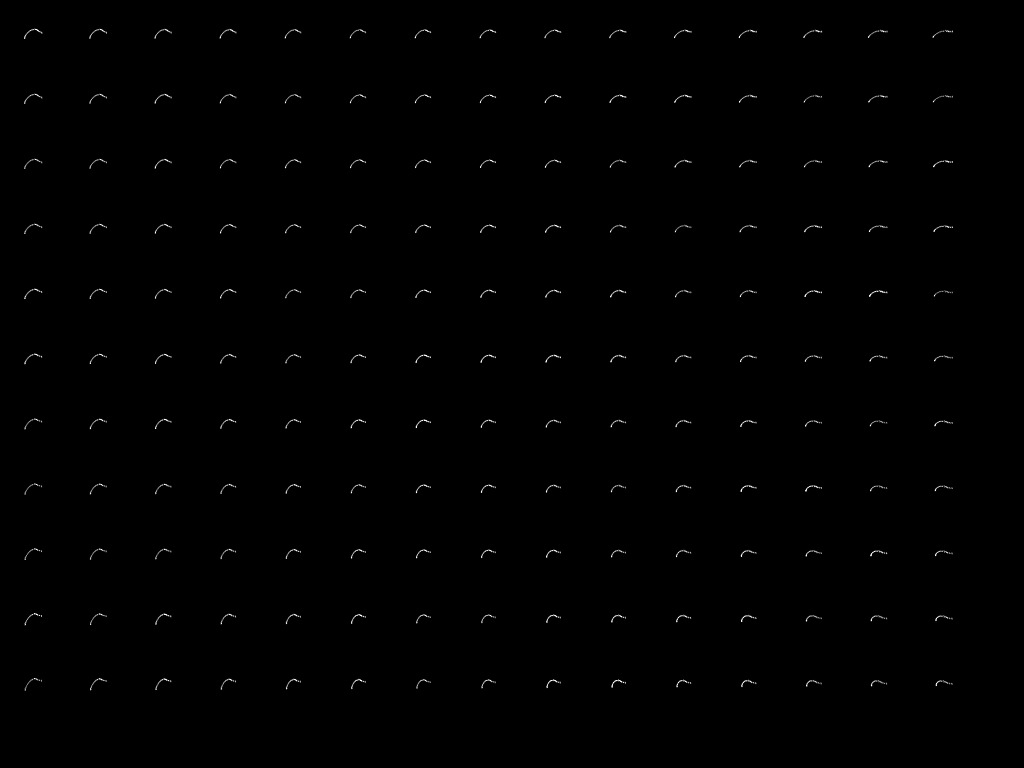}}};
  \node at (344pt,33pt) {\fcolorbox{red}{white}{\includegraphics[trim=912 635 50 67, clip]{traj_est_ill_aug_optim_natural_05_gyro_03_kernels_found.jpg}}};
  \node at (276pt,-33pt) {\fcolorbox{green}{white}{\includegraphics[trim=66 55 895 656, clip]{traj_est_ill_aug_optim_natural_05_gyro_03_kernels_found.jpg}}};
  \node at (344pt,-33pt) {\fcolorbox{yellow}{white}{\includegraphics[trim=912 55 50 656, clip]{traj_est_ill_aug_optim_natural_05_gyro_03_kernels_found.jpg}}};
\end{tikzpicture} \\
Trajectory (w/o optim.) \hspace{80pt}  Trajectory  \hspace{110pt} GT Trajectory \\ 
     \includegraphics[width=0.33\textwidth, trim=45 40 45 45, clip]{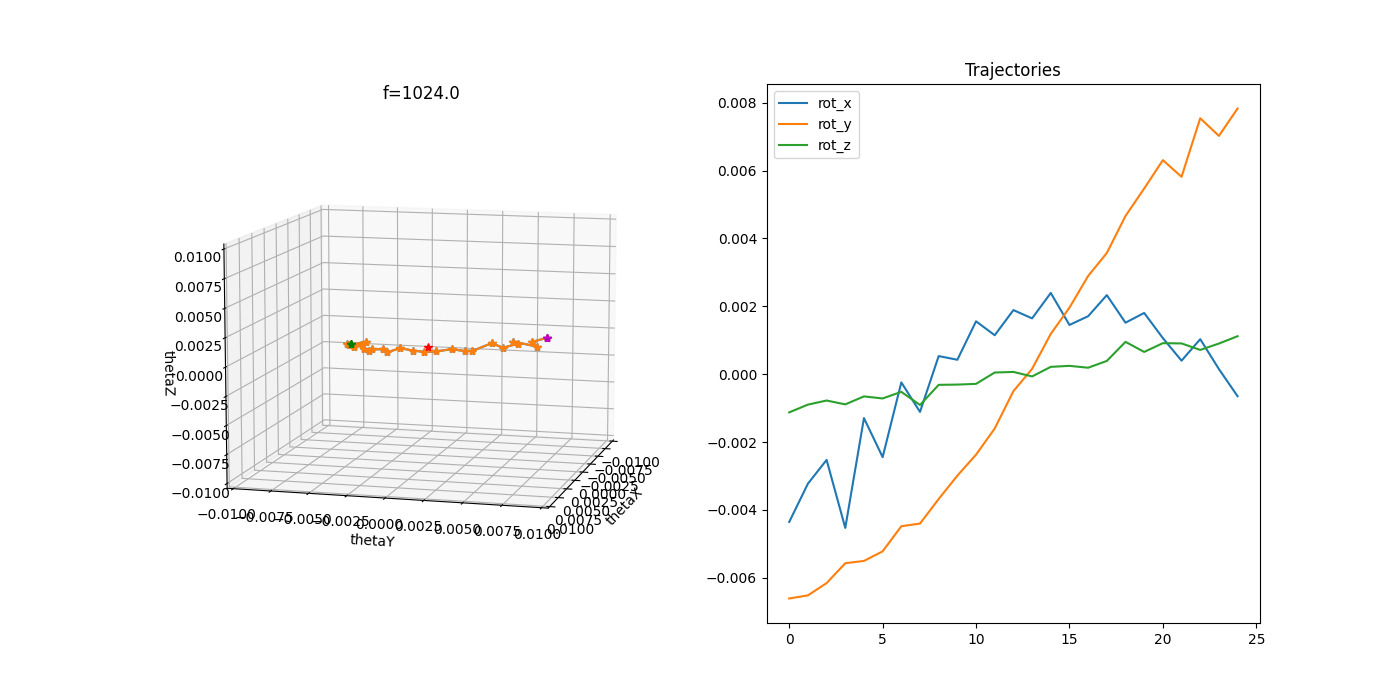} \includegraphics[width=0.33\textwidth, trim=45 40 45 45, clip]{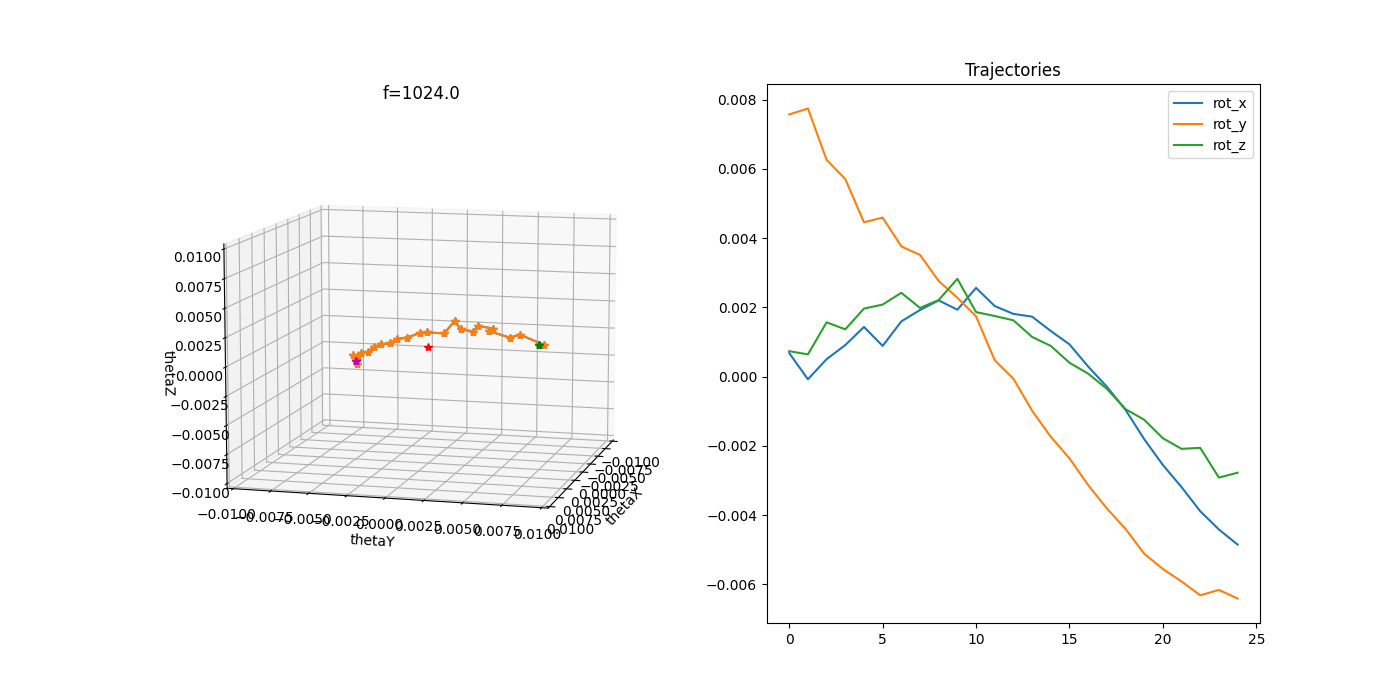}  
        \includegraphics[width=0.33\textwidth, trim=45 40 45 45, clip]{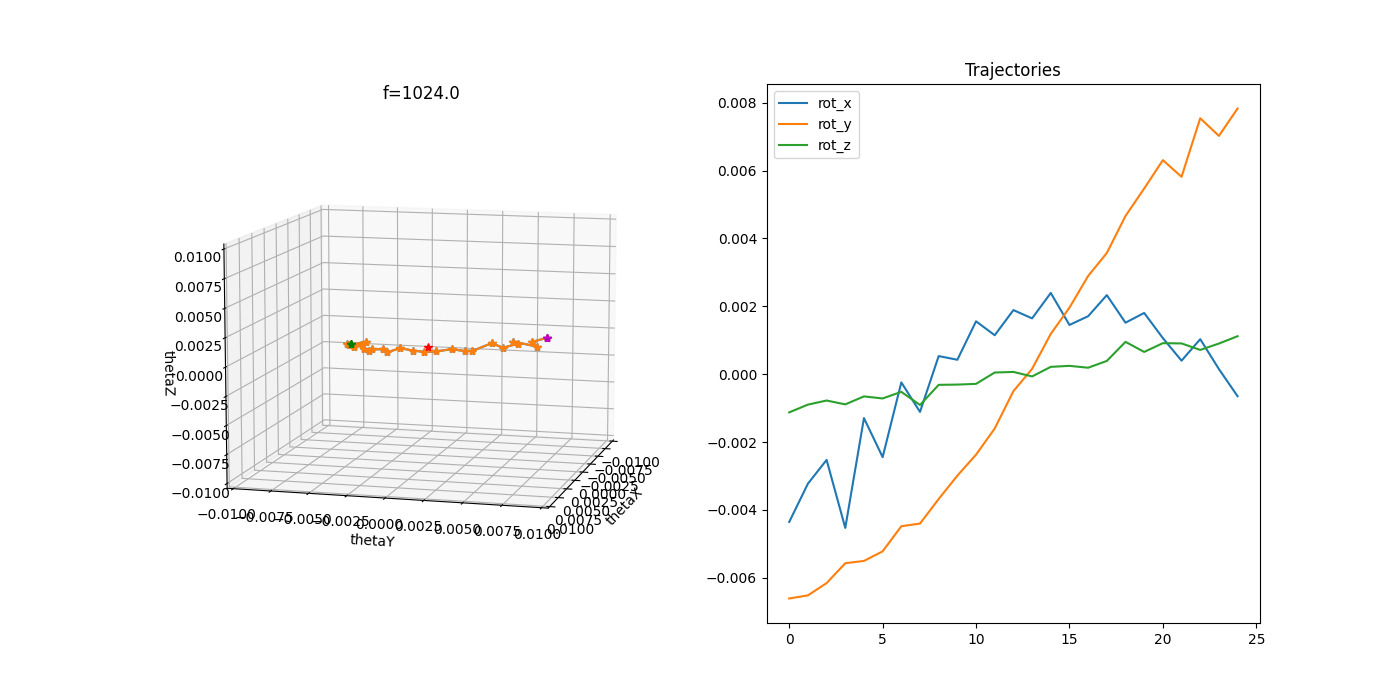} \\  
\end{tabular}
\caption{Kernels and restorations obtained with (3rd row) and without (2nd row) the trajectory refinement step. Single-pass kernel estimation produces better results for highly textured images. Still, the global constraint imposed during the optimization refines the estimated kernels and the restored image. The last row shows the initial trajectory estimated by the TPN, the trajectory after the refinement step, and the ground-truth trajectory. We represent a trajectory as a sequence of 3D points (\textit{pitch}, \textit{yaw}, \textit{roll}), and also show, using different color curves,  the evolution of each rotation during the exposure time.  \label{fig:Lai_non_uniform_traj3}} 
\end{figure*}

\begin{figure*}[ht]
\centering
\begin{tabular}{c}

\hspace{60pt} Blurry \hspace{110pt} Blurry Patches \hspace{50pt} Ground Truth kernels  \\
\begin{tikzpicture}[spy using outlines={rectangle, magnification=4.2, connect spies}]
  \node {\includegraphics[width=0.4\textwidth]{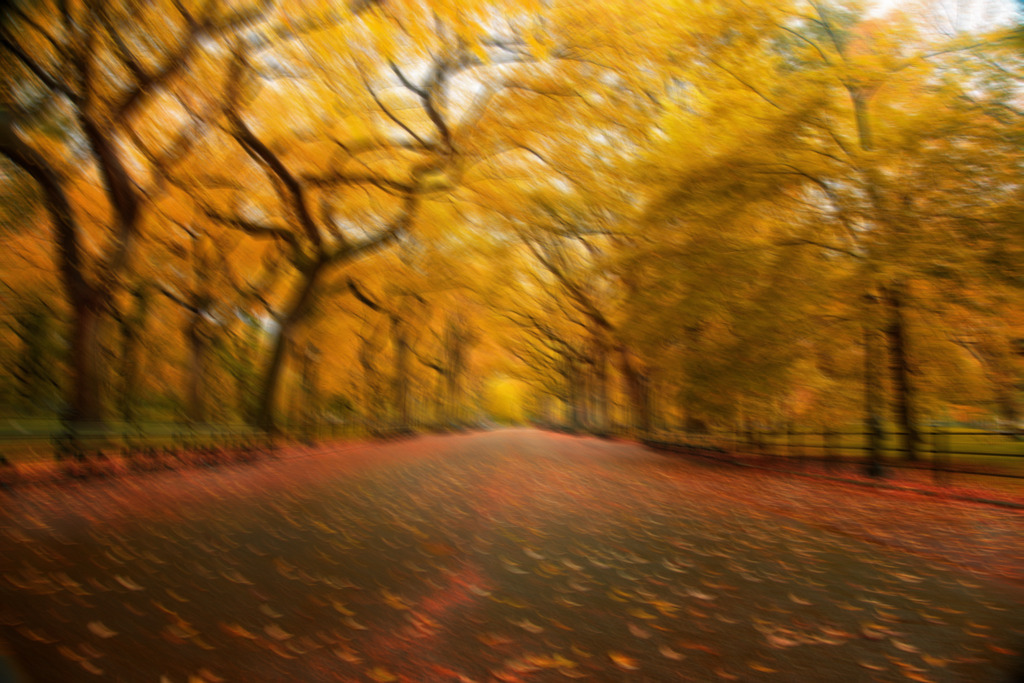}};

  \spy[draw=blue, thick, size=\pxsize]  on (-69.05pt,  47.43pt) in node [left] at (162pt,  33.33pt);
  \spy[draw=red,  line width=4pt, size=\pxsize]  on ( 69.23pt,  47.43pt) in node [left] at (230pt,  33.33pt);
  \spy[draw=green, line width=4pt, size=\pxsize]  on (-69.05pt, -47.59pt) in node [left] at (162pt, -33.33pt);
  \spy[draw=yellow, line width=4pt,size=\pxsize]  on ( 69.23pt, -47.59pt) in node [left] at (230pt, -33.33pt);

  \setlength{\fboxrule}{1pt} %
  \setlength{\fboxsep}{0pt}  %

  \node at (276pt,33pt) {\fcolorbox{blue}{white}{\includegraphics[trim=66 555 895 67, clip]{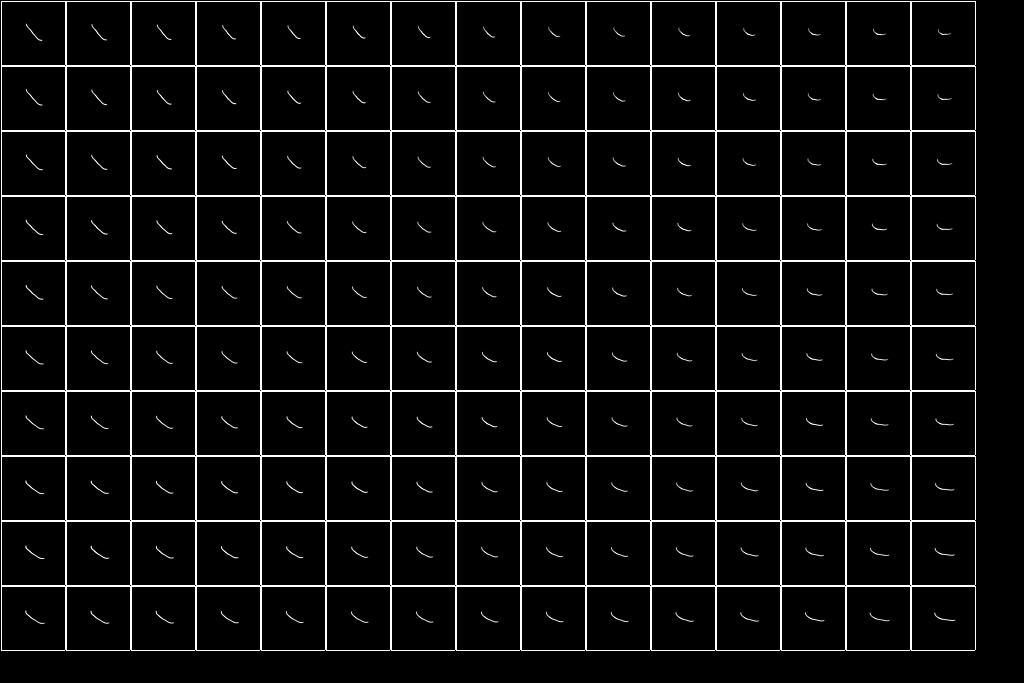}}};
  \node at (344pt,33pt) {\fcolorbox{red}{white}{\includegraphics[trim=912 555 50 67, clip]{Lai_non_uniform_GT_kernels_natural_04_gyro_04.jpg}}};
  \node at (276pt,-33pt) {\fcolorbox{green}{white}{\includegraphics[trim=66 35 895 586, clip]{Lai_non_uniform_GT_kernels_natural_04_gyro_04.jpg}}};
  \node at (344pt,-33pt) {\fcolorbox{yellow}{white}{\includegraphics[trim=912 35 50 586, clip]{Lai_non_uniform_GT_kernels_natural_04_gyro_04.jpg}}};
\end{tikzpicture} \\

\hspace{35pt} Restored (w/o optim)\hspace{50pt} Restored Patches (w/o optim) \hspace{5pt} Restored kernels (w/o optim) \\
\begin{tikzpicture}[spy using outlines={rectangle, magnification=4.2, connect spies}]
  \node {\includegraphics[width=0.4\textwidth]{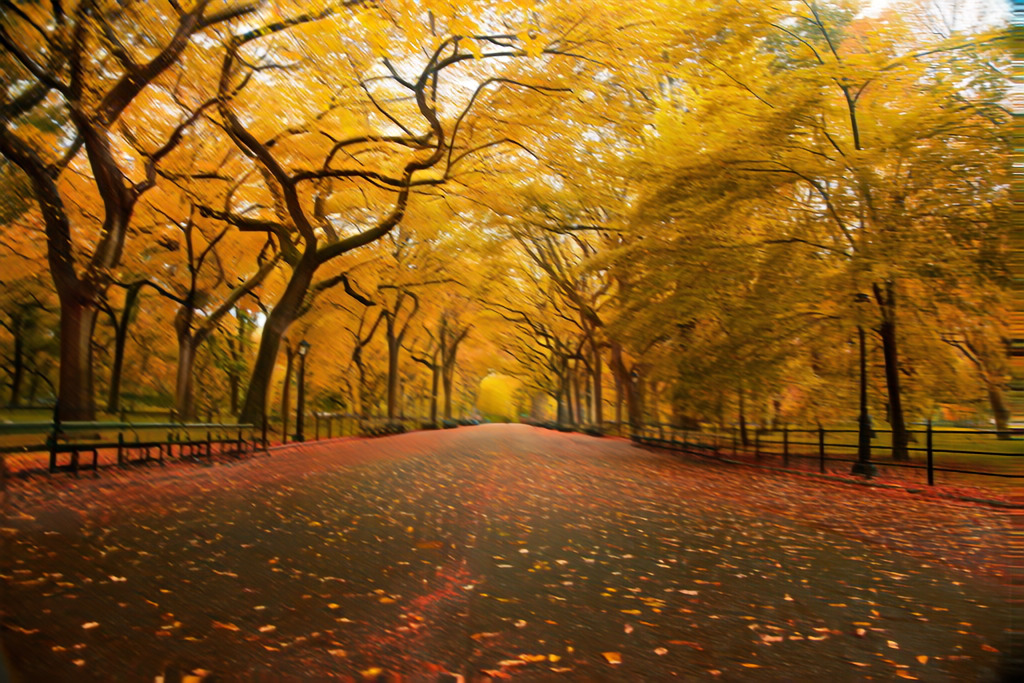}};

  \spy[draw=blue, thick, size=\pxsize]  on (-69.05pt,  47.43pt) in node [left] at (162pt,  33.33pt);
  \spy[draw=red,  line width=4pt, size=\pxsize]  on ( 69.23pt,  47.43pt) in node [left] at (230pt,  33.33pt);
  \spy[draw=green, line width=4pt, size=\pxsize]  on (-69.05pt, -47.59pt) in node [left] at (162pt, -33.33pt);
  \spy[draw=yellow, line width=4pt,size=\pxsize]  on ( 69.23pt, -47.59pt) in node [left] at (230pt, -33.33pt);

  \setlength{\fboxrule}{1pt} %
  \setlength{\fboxsep}{0pt}  %

  \node at (276pt,33pt) {\fcolorbox{blue}{white}{\includegraphics[trim=66 555 895 67, clip]{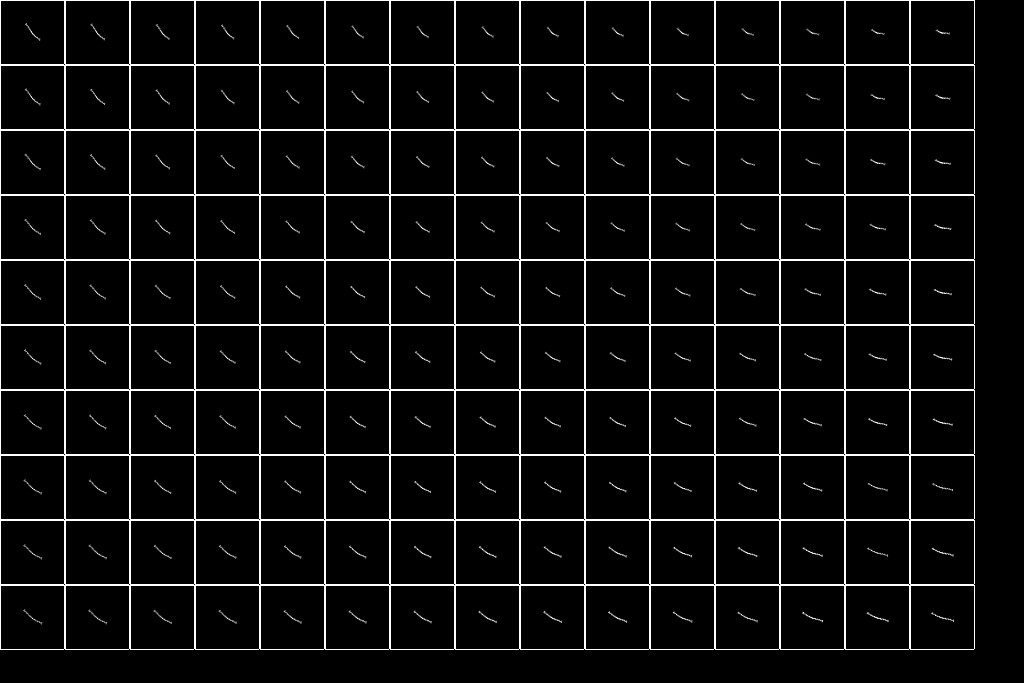}}};
  \node at (344pt,33pt) {\fcolorbox{red}{white}{\includegraphics[trim=912 555 50 67, clip]{traj_est_ill_aug_raw_natural_04_gyro_04_kernels_found.jpg}}};
  \node at (276pt,-33pt) {\fcolorbox{green}{white}{\includegraphics[trim=66 35 895 586, clip]{traj_est_ill_aug_raw_natural_04_gyro_04_kernels_found.jpg}}};
  \node at (344pt,-33pt) {\fcolorbox{yellow}{white}{\includegraphics[trim=912 35 50 586, clip]{traj_est_ill_aug_raw_natural_04_gyro_04_kernels_found.jpg}}};
\end{tikzpicture} \\

\hspace{40pt} Restored \hspace{110pt} Restored Patches \hspace{50pt} Restored kernels  \\
\begin{tikzpicture}[spy using outlines={rectangle, magnification=4.2, connect spies}]
  \node {\includegraphics[width=0.4\textwidth]{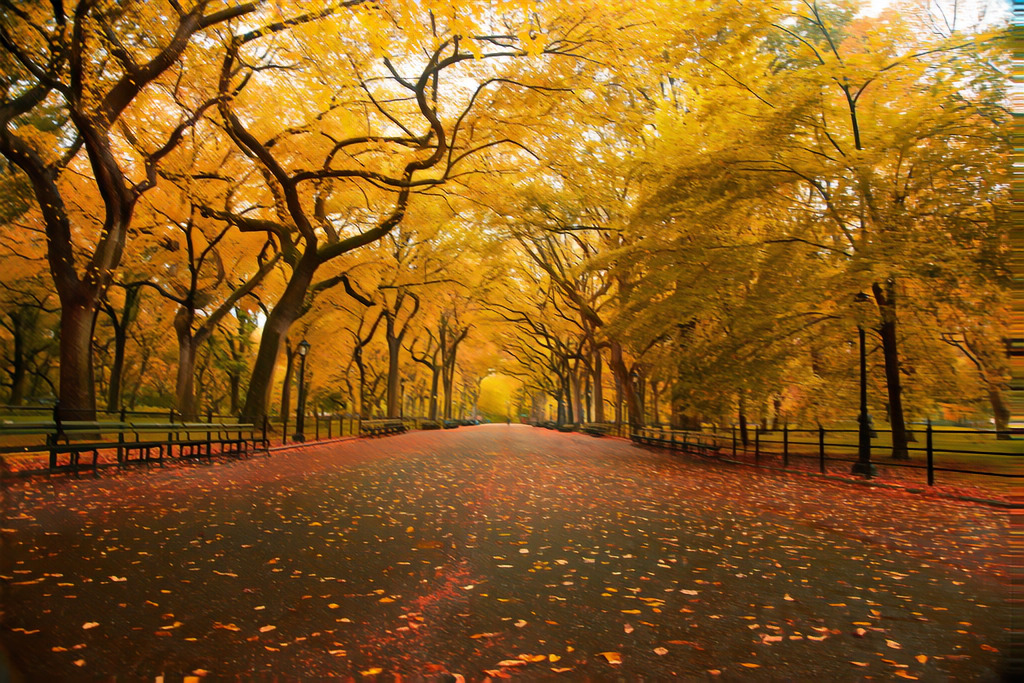}};

  \spy[draw=blue, thick, size=\pxsize]  on (-69.05pt,  47.43pt) in node [left] at (162pt,  33.33pt);
  \spy[draw=red,  line width=4pt, size=\pxsize]  on ( 69.23pt,  47.43pt) in node [left] at (230pt,  33.33pt);
  \spy[draw=green, line width=4pt, size=\pxsize]  on (-69.05pt, -47.59pt) in node [left] at (162pt, -33.33pt);
  \spy[draw=yellow, line width=4pt,size=\pxsize]  on ( 69.23pt, -47.59pt) in node [left] at (230pt, -33.33pt);

  \setlength{\fboxrule}{1pt} %
  \setlength{\fboxsep}{0pt}  %

  \node at (276pt,33pt) {\fcolorbox{blue}{white}{\includegraphics[trim=66 555 895 67, clip]{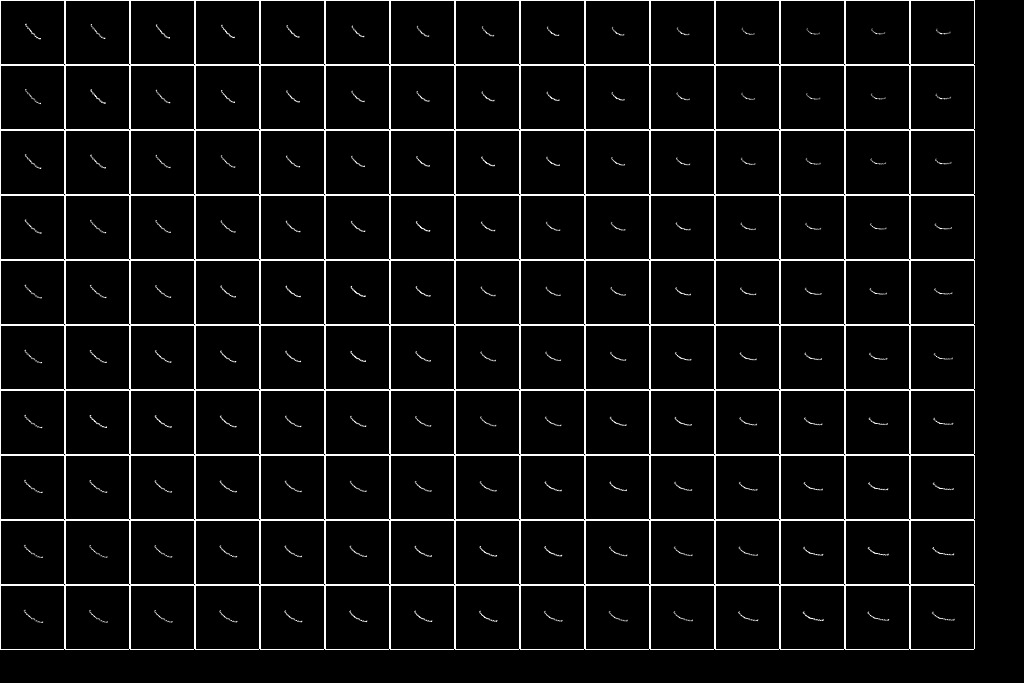}}};
  \node at (344pt,33pt) {\fcolorbox{red}{white}{\includegraphics[trim=912 555 50 67, clip]{traj_est_ill_aug_optim_natural_04_gyro_04_kernels_found.jpg}}};
  \node at (276pt,-33pt) {\fcolorbox{green}{white}{\includegraphics[trim=66 35 895 586, clip]{traj_est_ill_aug_optim_natural_04_gyro_04_kernels_found.jpg}}};
  \node at (344pt,-33pt) {\fcolorbox{yellow}{white}{\includegraphics[trim=912 35 50 586, clip]{traj_est_ill_aug_optim_natural_04_gyro_04_kernels_found.jpg}}};
\end{tikzpicture} \\
Trajectory (w/o optim.) \hspace{80pt}  Trajectory  \hspace{110pt} GT Trajectory \\ 
     \includegraphics[width=0.33\textwidth, trim=45 40 45 45, clip]{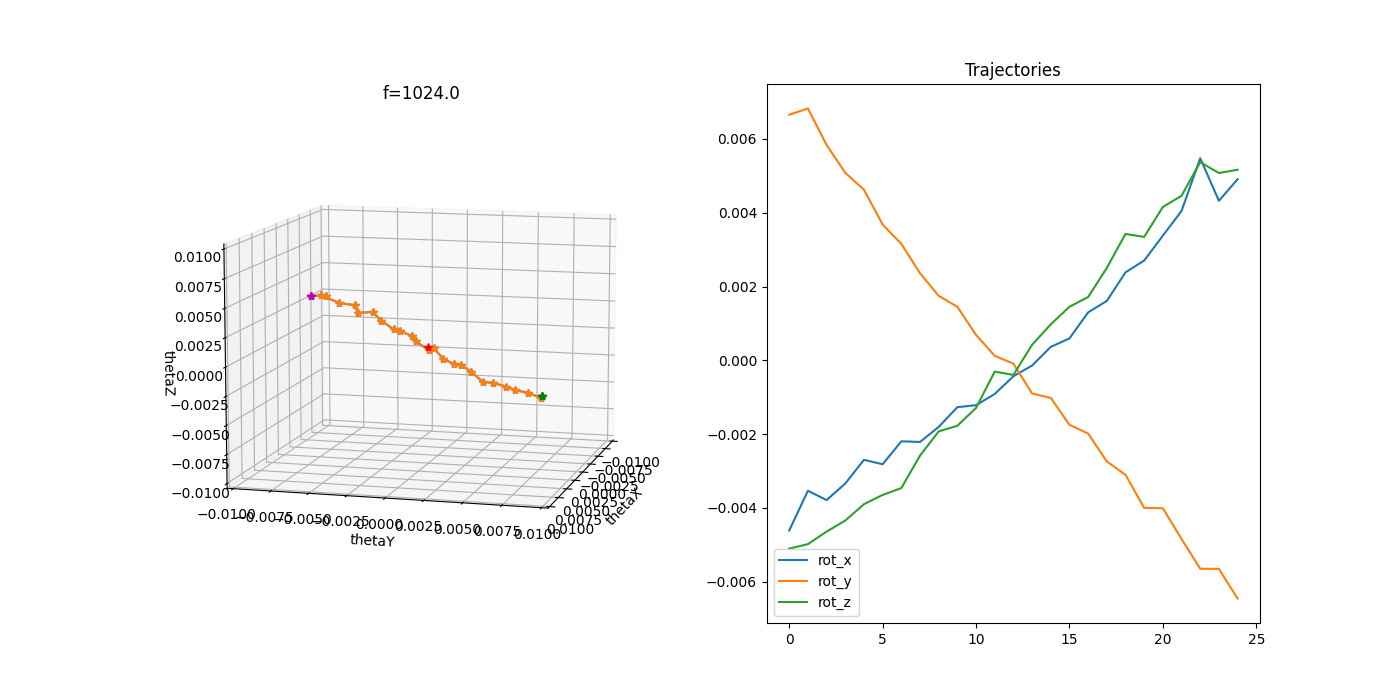} \includegraphics[width=0.33\textwidth, trim=45 40 45 45, clip]{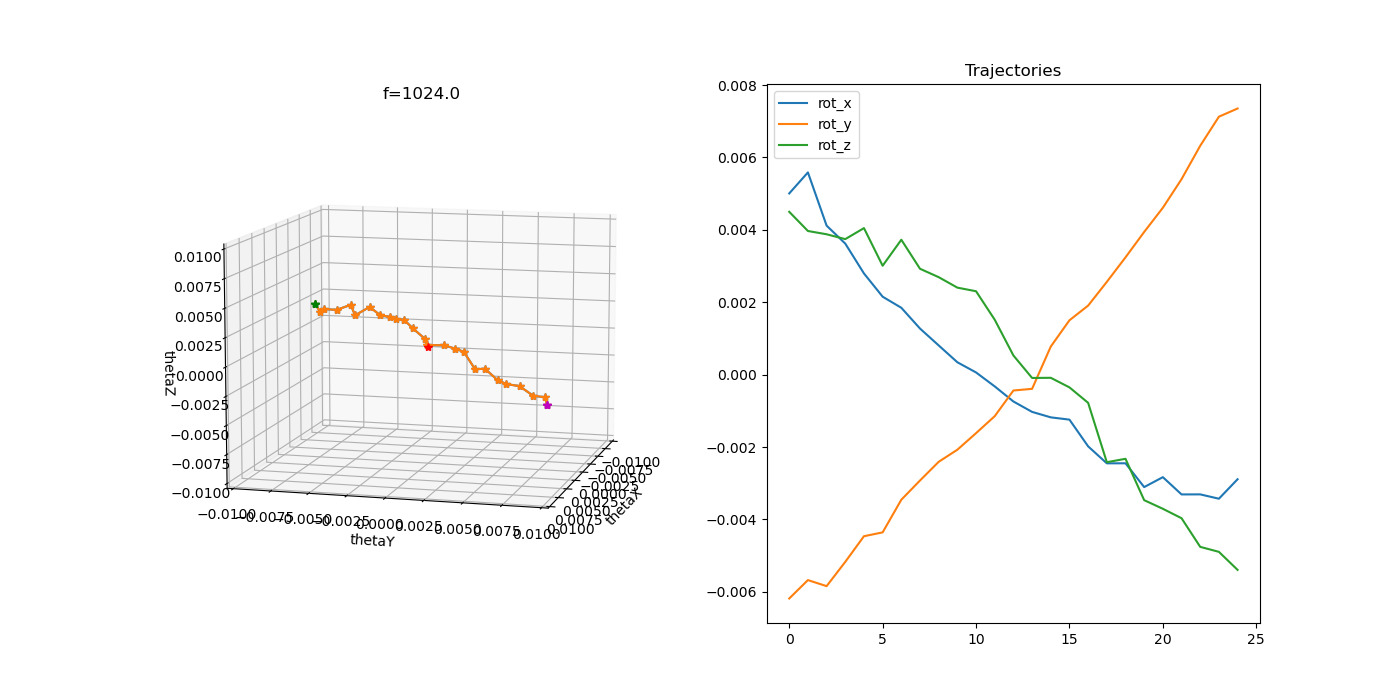}  
        \includegraphics[width=0.33\textwidth, trim=45 40 45 45, clip]{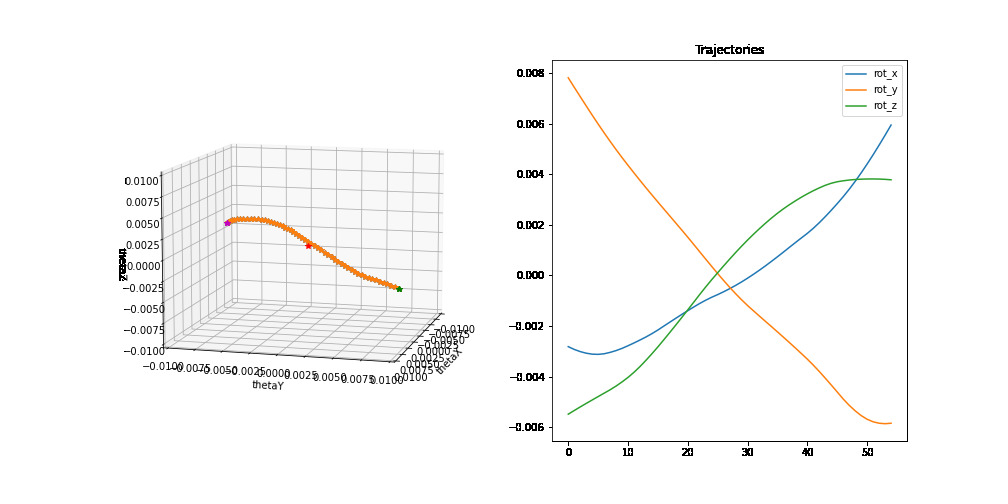} \\  
\end{tabular}
\caption{Kernels and restorations obtained with (3rd row) and without (2nd row) the trajectory refinement step. Single-pass kernel estimation produces better results for highly textured images. Still, the global constraint imposed during the optimization refines the estimated kernels and the restored image. The last row shows the initial trajectory estimated by the TPN, the trajectory after the refinement step, and the ground-truth trajectory. We represent a trajectory as a sequence of 3D points (\textit{pitch}, \textit{yaw}, \textit{roll}), and also show, using different color curves,  the evolution of each rotation during the exposure time.  \label{fig:Lai_non_uniform_traj4}} 
\end{figure*}

\cref{fig:Lai_non_uniform_traj1,fig:Lai_non_uniform_traj2,fig:Lai_non_uniform_traj3,fig:Lai_non_uniform_traj4} show examples of trajectories, kernels, and restorations obtained with Blur2Seq on the Lai nonuniform dataset. We show the results with and without the trajectory refinement step.

\begin{figure*}[ht]
\setlength{\tabcolsep}{1pt}
    \centering
    \begin{tabular}{cc}
    Blurry - PSNR=19.11   &  \cite{vasu2017local} - PSNR=21.86 \\
    \includegraphics[width=0.45\linewidth]{Blurry_manmade_02_gyro_01.jpg}     &  
\includegraphics[width=0.45\linewidth]{2017\_subeesh\_manmade\_02\_gyro\_01\_17\_Subeesh-Xu-Ep.jpg} \\
    \cite{li2023self} - PSNR=23.56 &   Blur2Seq (ours) - PSNR=25.11  \\
\includegraphics[width=0.45\linewidth]{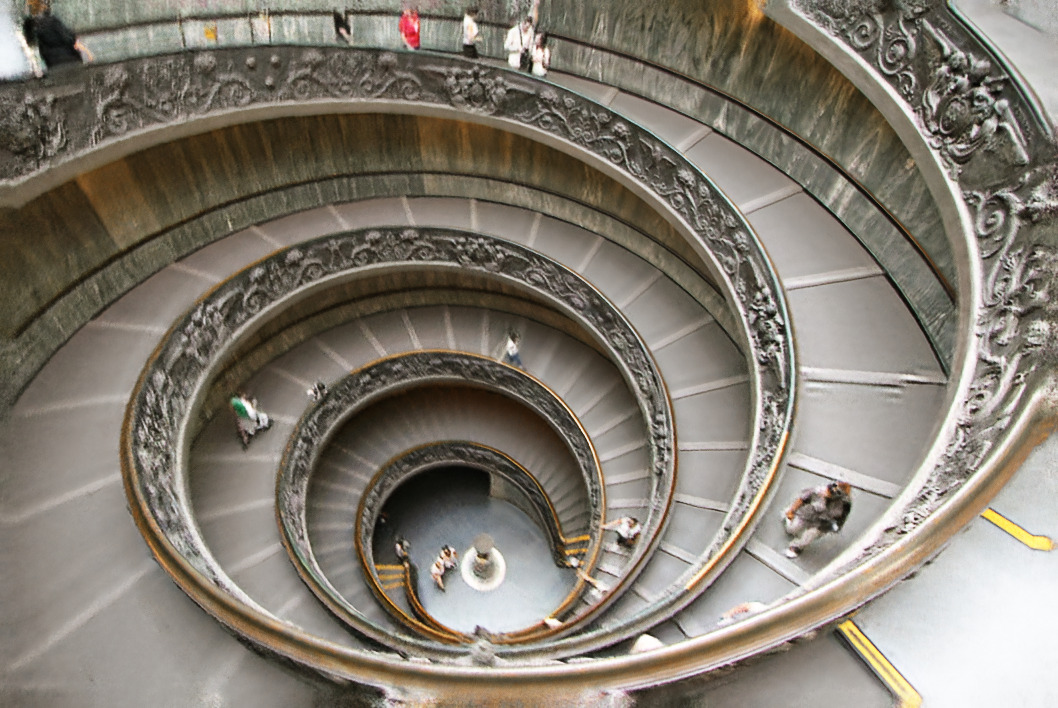}     & 
\includegraphics[width=0.45\linewidth]{traj_est_ill_aug_optim_manmade_02_gyro_01_img_restored.jpg} \\
    Kernels \cite{li2023self}  &   Blur2Seq (ours) kernels  \\
\includegraphics[width=0.45\linewidth]{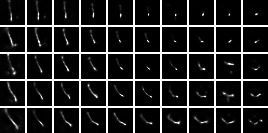}     & 
\includegraphics[width=0.45\linewidth]{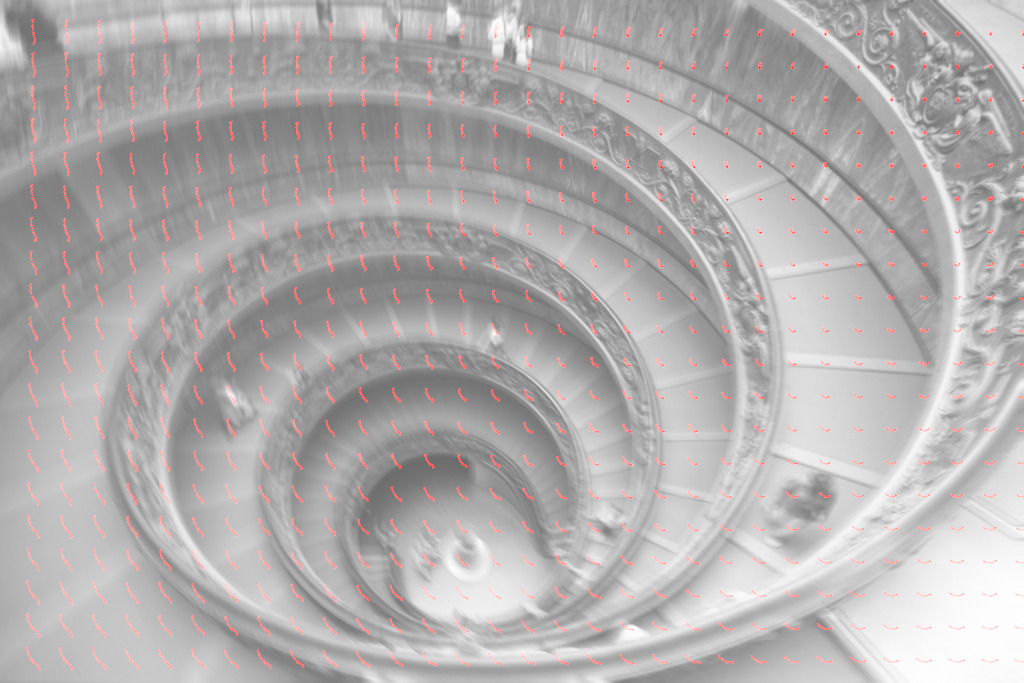}     
    \end{tabular}
    \caption[Comparison with the restorations provided by \cite{vasu2017local} and \cite{li2023self} on a sample image from Lai \cite{lai2016comparative}.]{Comparison with the restorations provided by \cite{vasu2017local} and \cite{li2023self} on a sample image from \cite{lai2016comparative}. The video associated with our restoration can be seen \href{https://youtu.be/BKZihBE8qfU}{here}. }
    \label{fig:LNU_manmade_02_01}
\end{figure*}

\FloatBarrier
\section{Qualitative Comparison on Sample Images Provided by some Classic Methods}

\begin{figure*}[ht]
    \centering
    \begin{tabular}{cc}
    Blurry & \cite{gupta_mdf_deblurring} \\
    \includegraphics[width=0.45\linewidth]{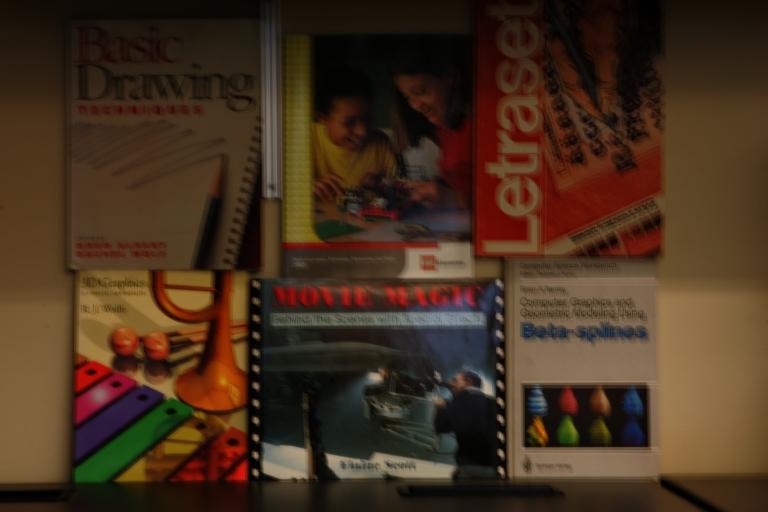}     &  \includegraphics[width=0.45\linewidth]{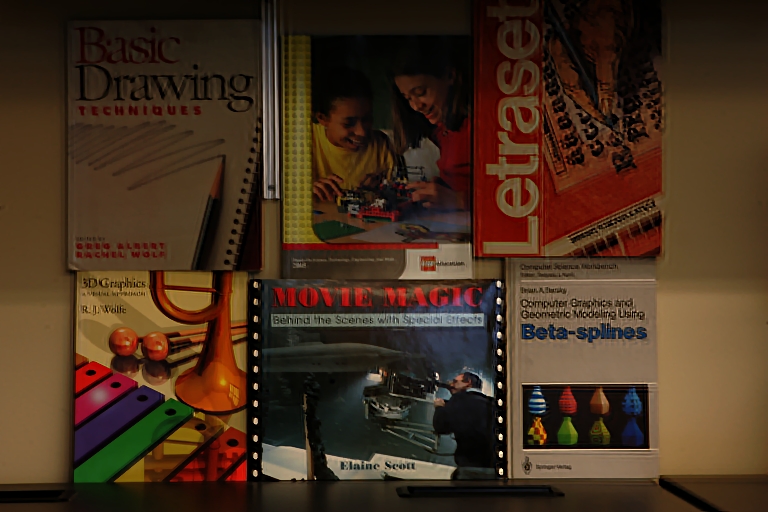}   \\
    Blur2Seq (ours) kernels & Blur2Seq (ours)   \\
    \includegraphics[width=0.45\linewidth]{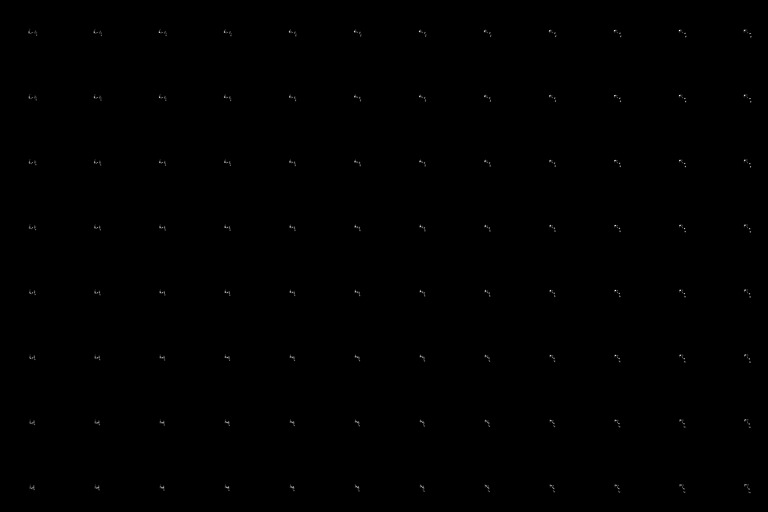}  &
    \includegraphics[width=0.45\linewidth]{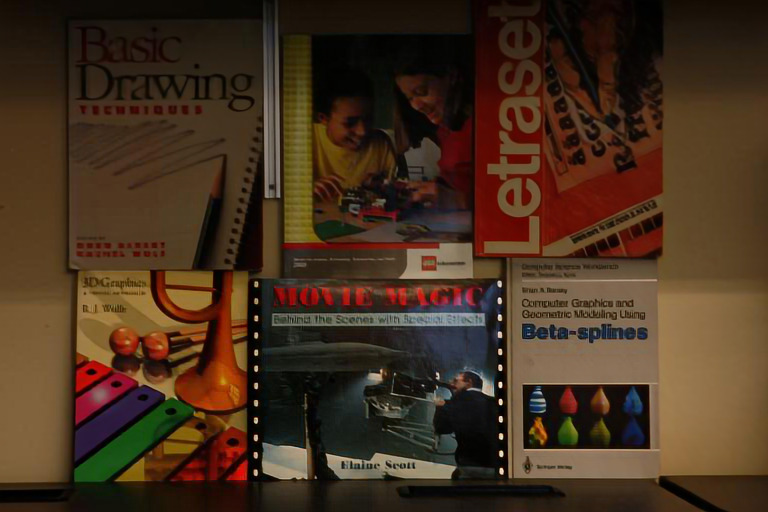} 
    \end{tabular}
    \caption[Restorations on a sample image provided by \cite{gupta_mdf_deblurring}.]{Restorations on a sample image provided by \cite{gupta_mdf_deblurring}. The video generated by our method can be seen \href{https://youtu.be/ElIWIZLgiDw}{here}. }
    \label{fig:sample_Gupta_book1}
\end{figure*}

\begin{figure*}[ht]
    \centering
    \begin{tabular}{cc}
    Blurry & \cite{gupta_mdf_deblurring} \\
    \includegraphics[width=0.45\linewidth]{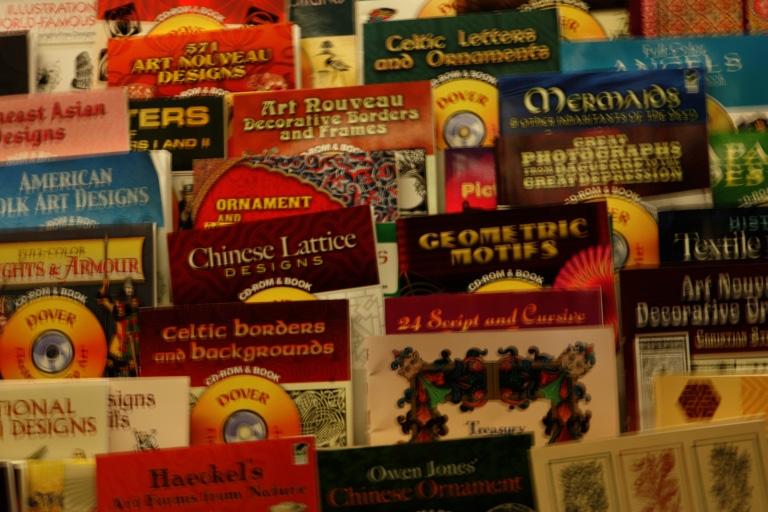}     &  \includegraphics[width=0.45\linewidth]{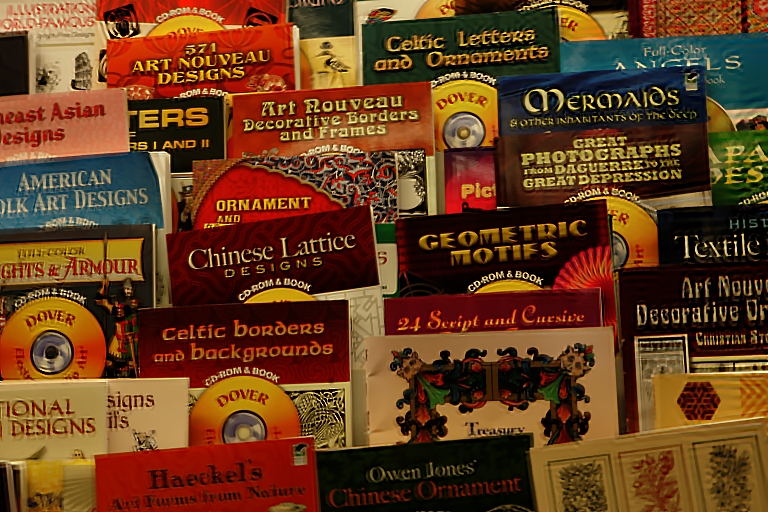}   \\
    Blur2Seq (ours) kernels & Blur2Seq (ours)  \\
    \includegraphics[width=0.45\linewidth]{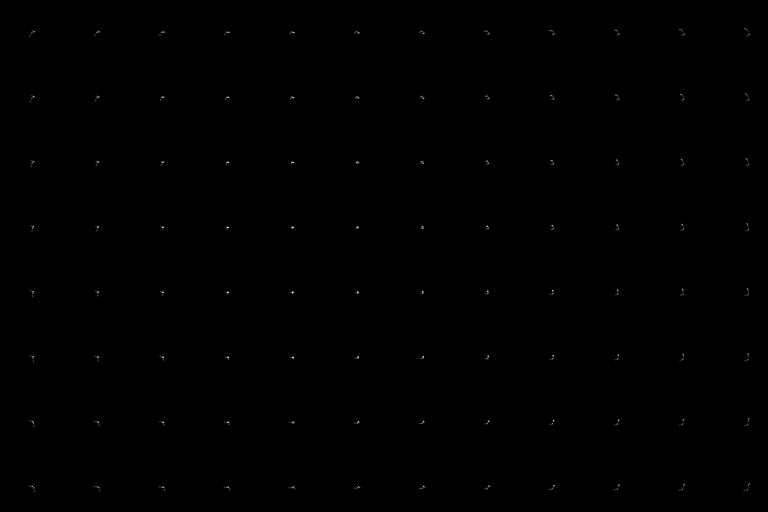}  &
    \includegraphics[width=0.45\linewidth]{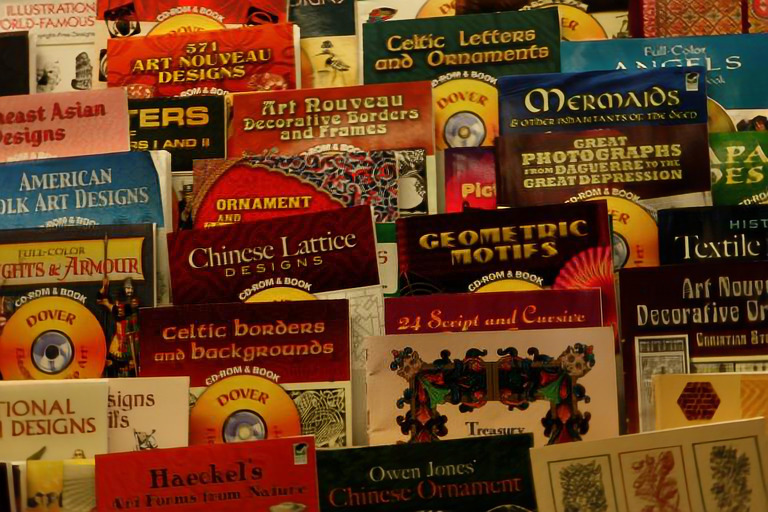} 
    \end{tabular}
    \caption[Restorations on a sample image provided by \cite{gupta_mdf_deblurring}.]{Restorations on a sample image provided by \cite{gupta_mdf_deblurring}. The video generated by our method can be seen \href{https://youtu.be/fu-3kH6HB3g}{here}.  }
    \label{fig:sample_Gupta_book2}
\end{figure*}

\begin{figure*}[ht]
    \centering
    \begin{tabular}{cc}
    Blurry & \cite{whyte2010nonuniform} \\
    \includegraphics[width=0.45\linewidth]{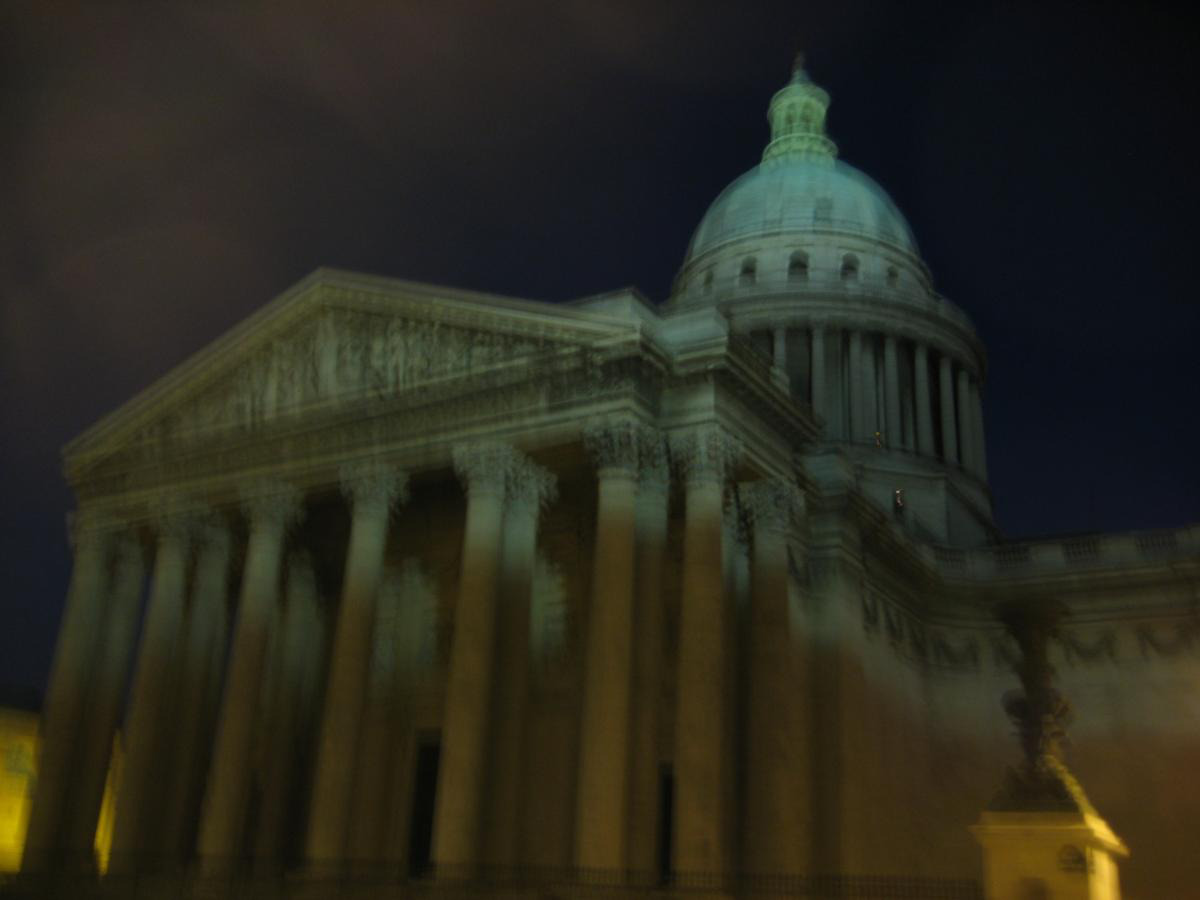}     &  \includegraphics[width=0.45\linewidth]{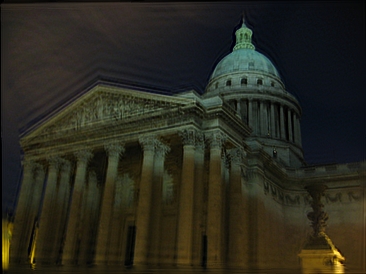}   \\
    Blur2Seq (ours) kernels & Blur2Seq (ours)   \\
    \includegraphics[width=0.45\linewidth]{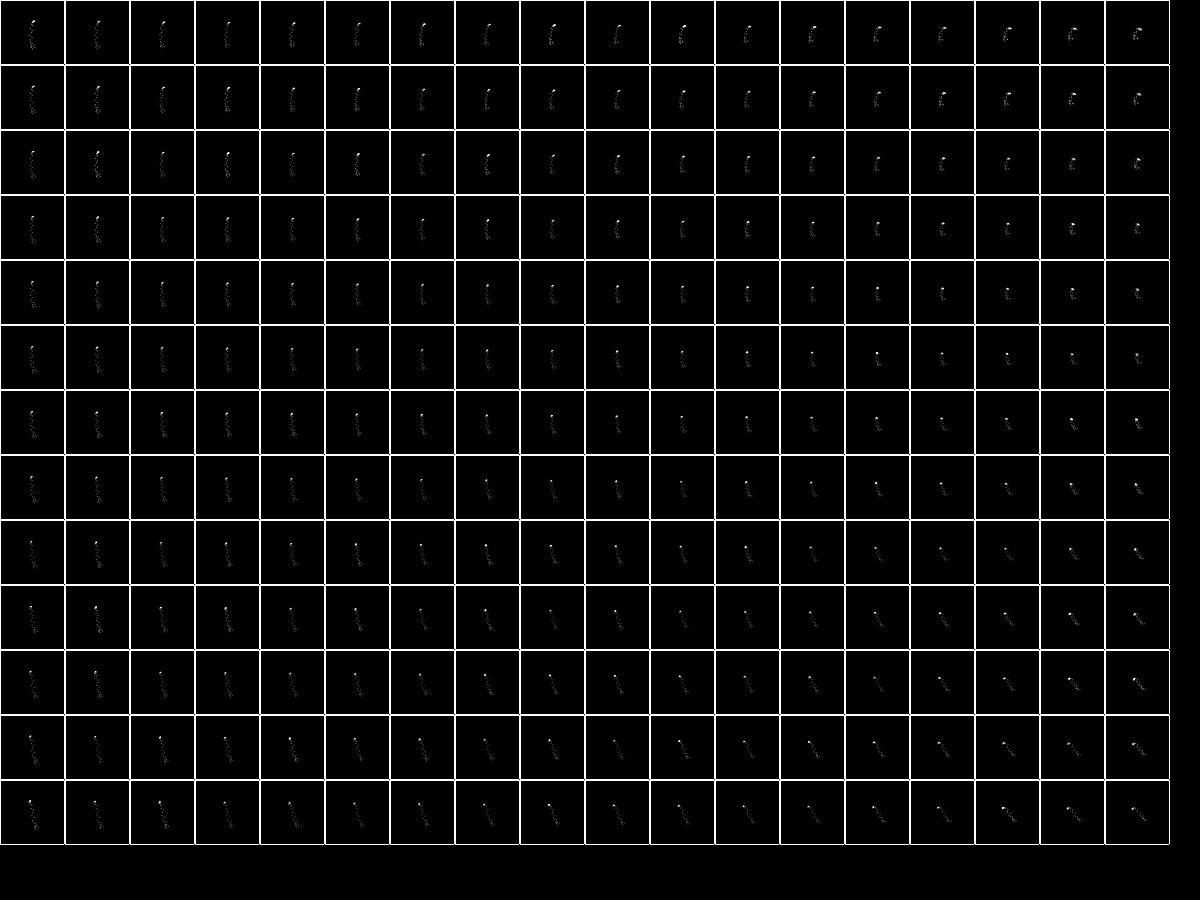}  &
    \includegraphics[width=0.45\linewidth]{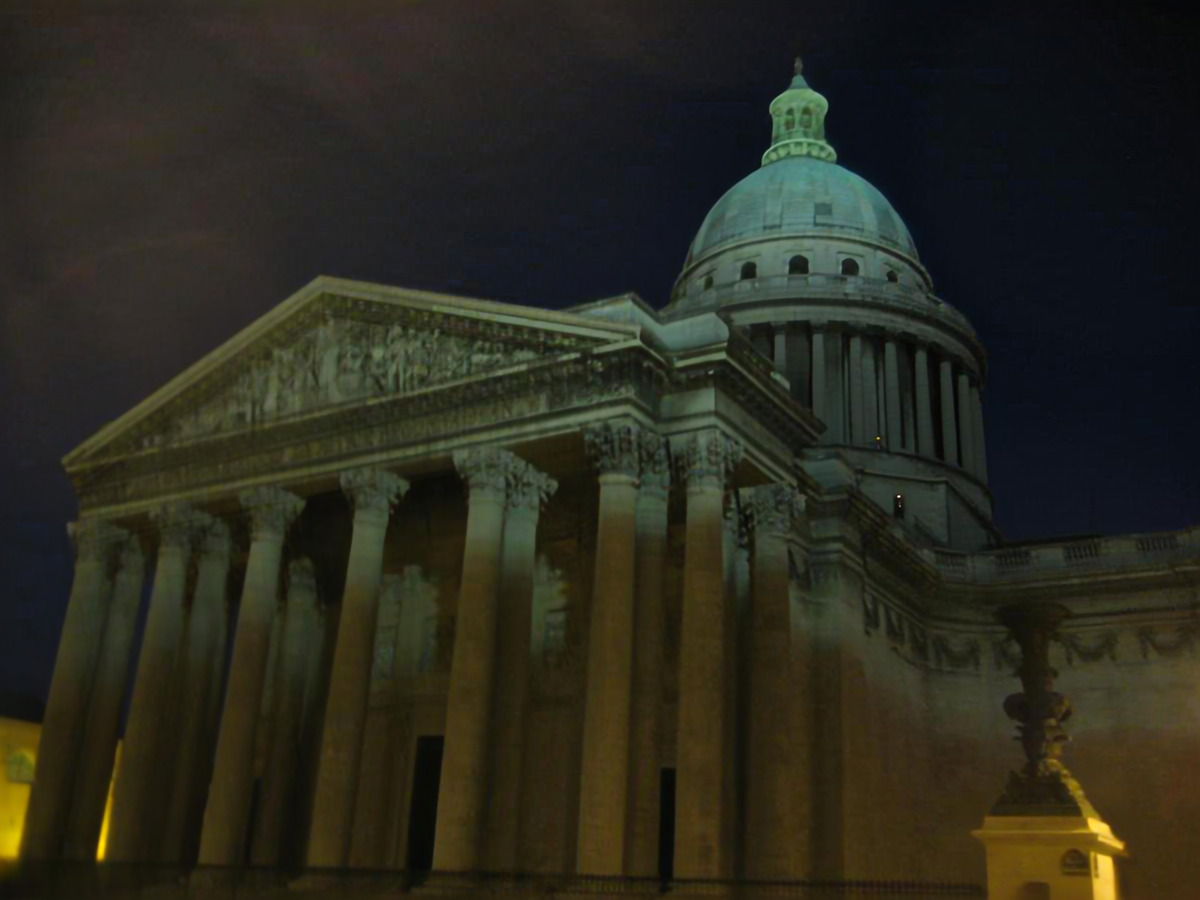} 
    \end{tabular}
    \caption[Restorations on a sample image provided by \cite{whyte2010nonuniform}.]{Restorations on a sample image provided by \cite{whyte2010nonuniform}.  }
    \label{fig:sample_Whyte1}
\end{figure*}

\begin{figure*}[ht]
    \centering
    \begin{tabular}{cc}
    Blurry & \cite{whyte2010nonuniform} \\
    \includegraphics[width=0.45\linewidth]{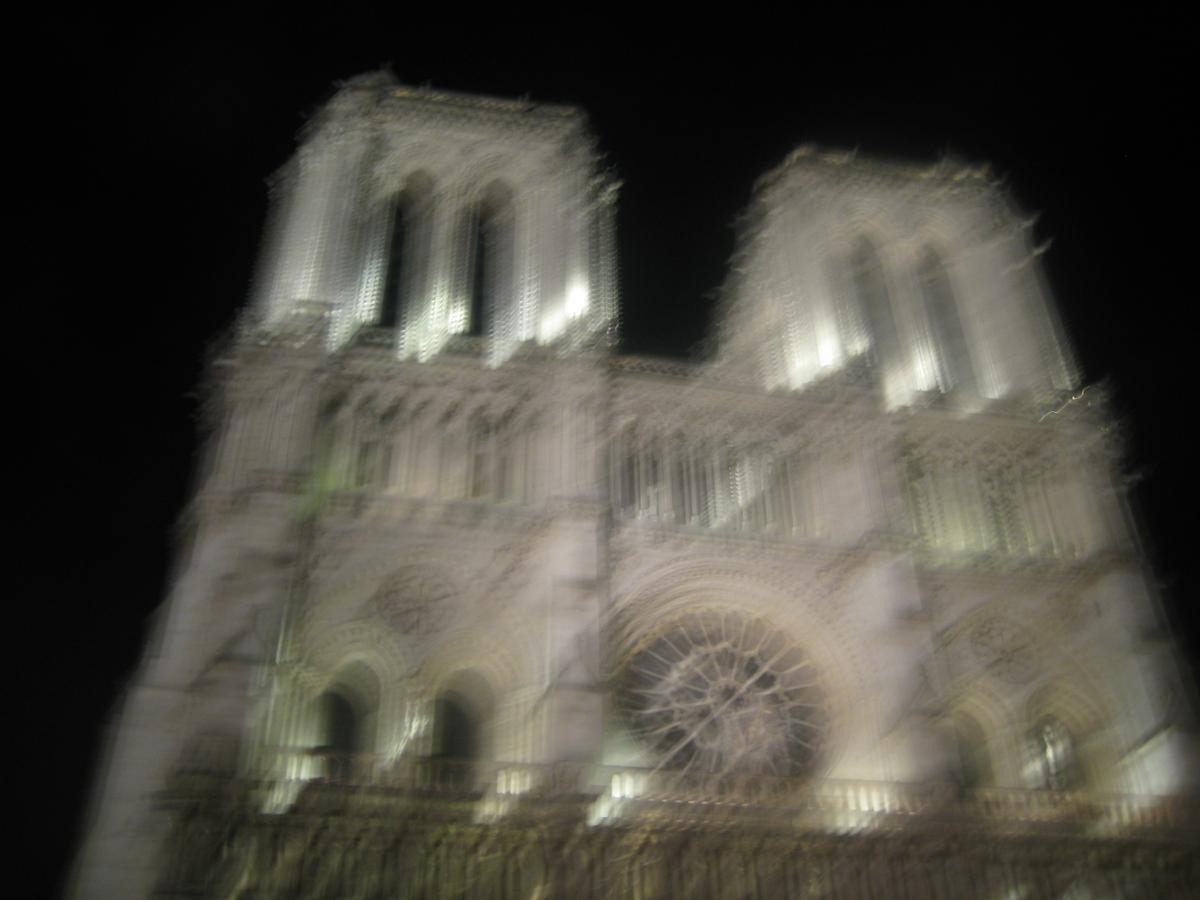}     &  \includegraphics[width=0.45\linewidth]{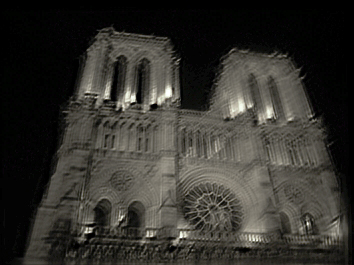}   \\
    Blur2Seq (ours) kernels & Blur2Seq (ours)   \\
    \includegraphics[width=0.45\linewidth]{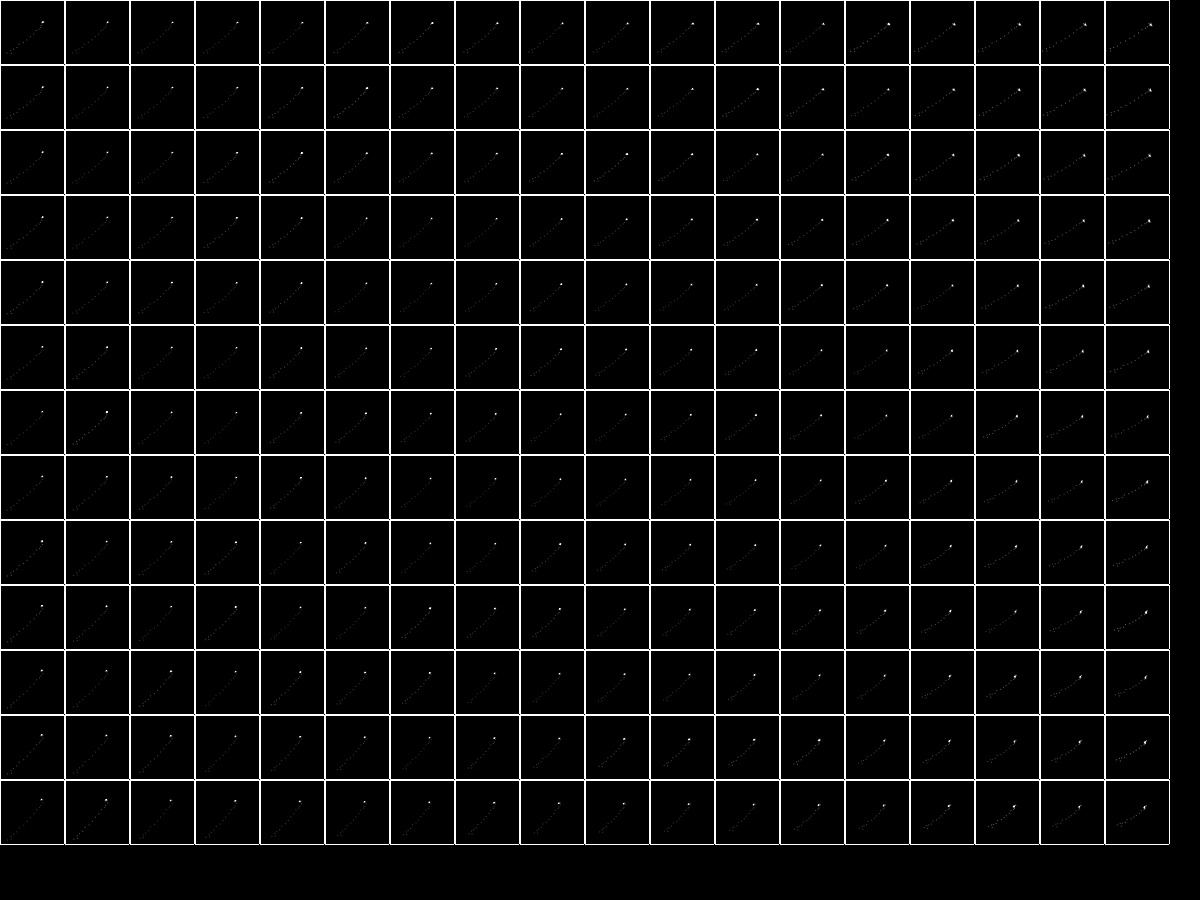}  &
    \includegraphics[width=0.45\linewidth]{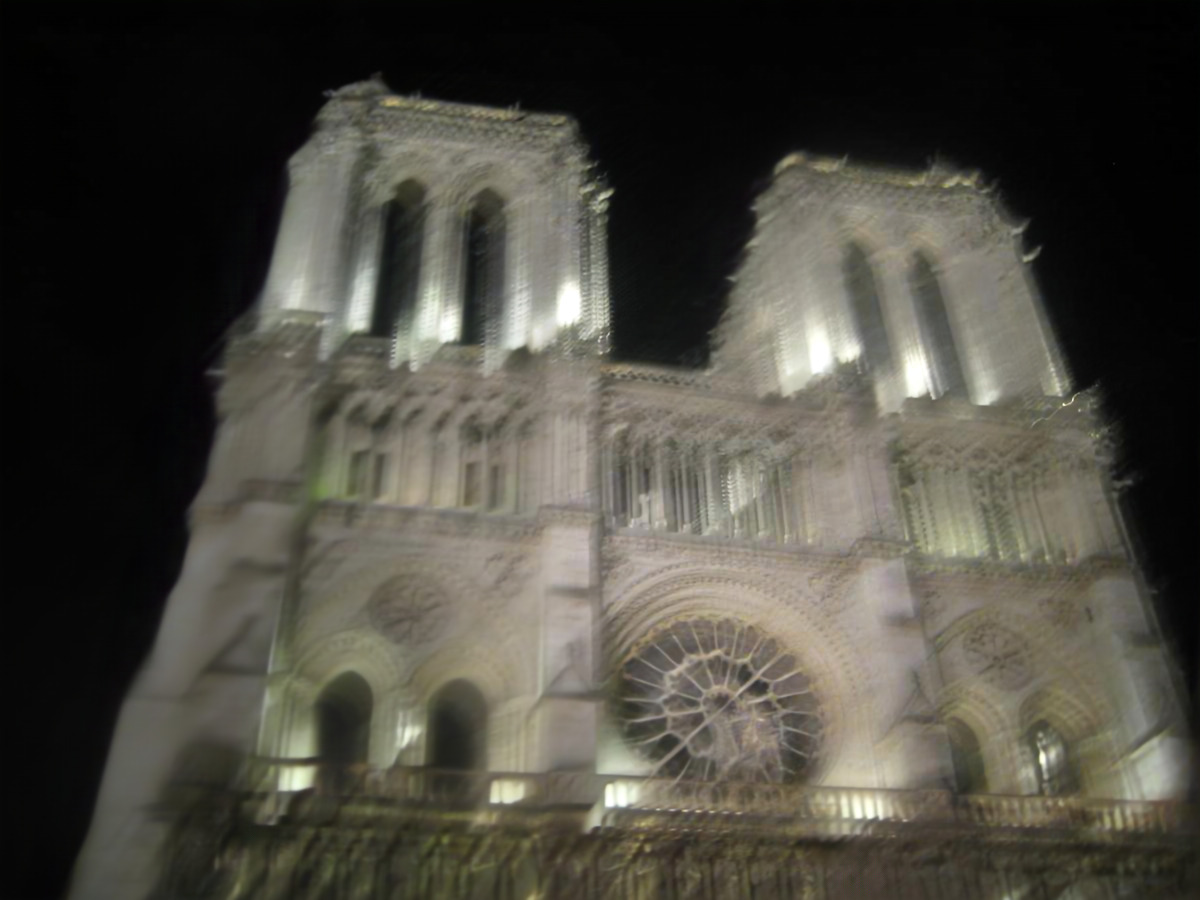} 
    \end{tabular}
    \caption[Restorations on a sample image provided by \cite{whyte2010nonuniform}.]{Restorations on a sample image provided by \cite{whyte2010nonuniform}.  }
    \label{fig:sample_Whyte2}
\end{figure*}

\begin{figure*}[ht]
    \centering
    \begin{tabular}{cc}
    Blurry & \cite{whyte2010nonuniform} \\
    \includegraphics[width=0.45\linewidth]{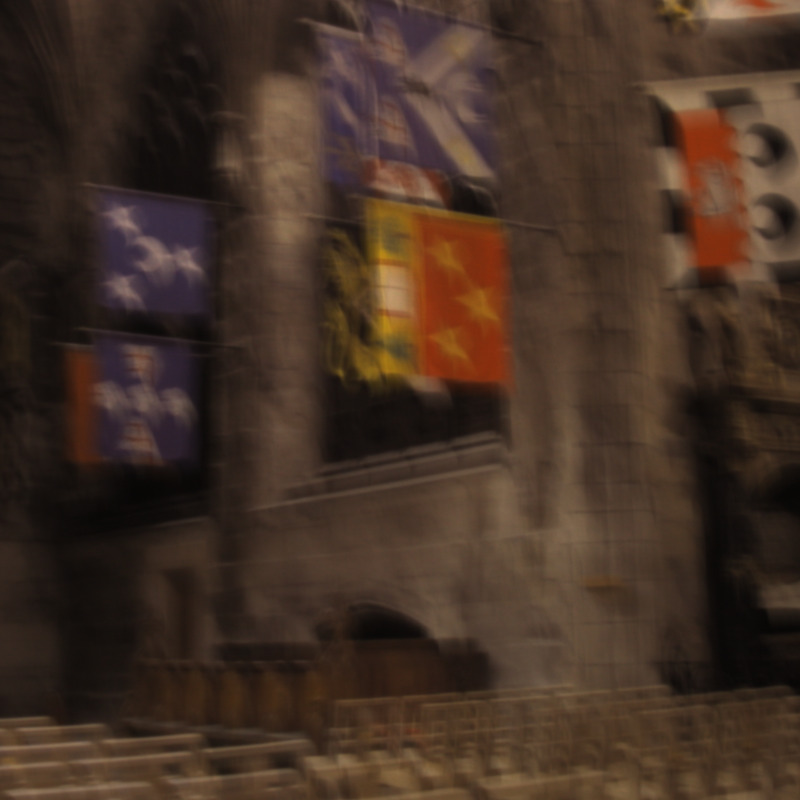}     &  \includegraphics[width=0.45\linewidth]{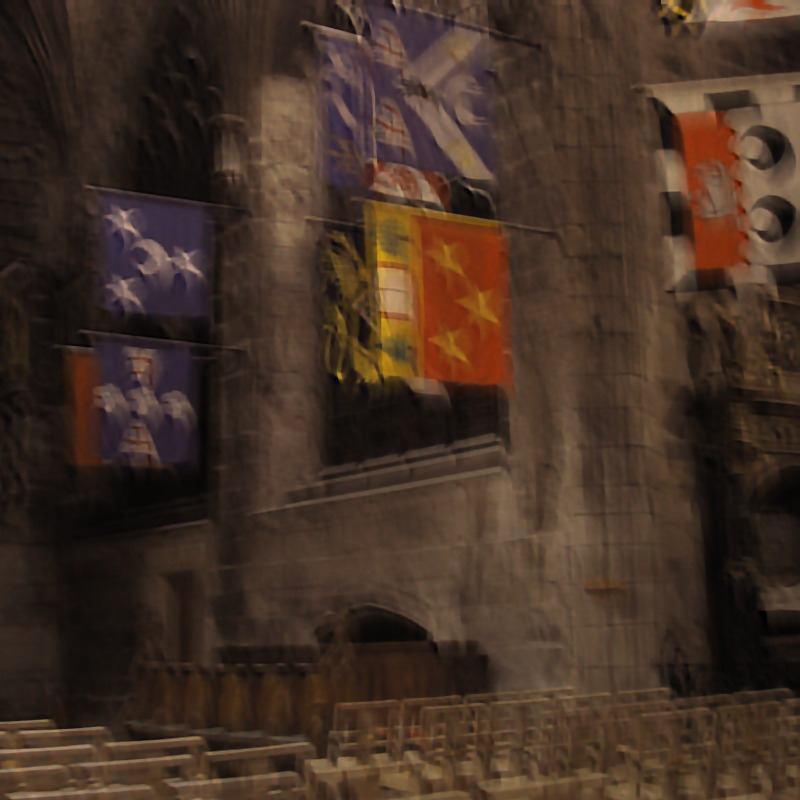}   \\
    Blur2Seq (ours) kernels & Blur2Seq (ours)   \\
    \includegraphics[width=0.45\linewidth]{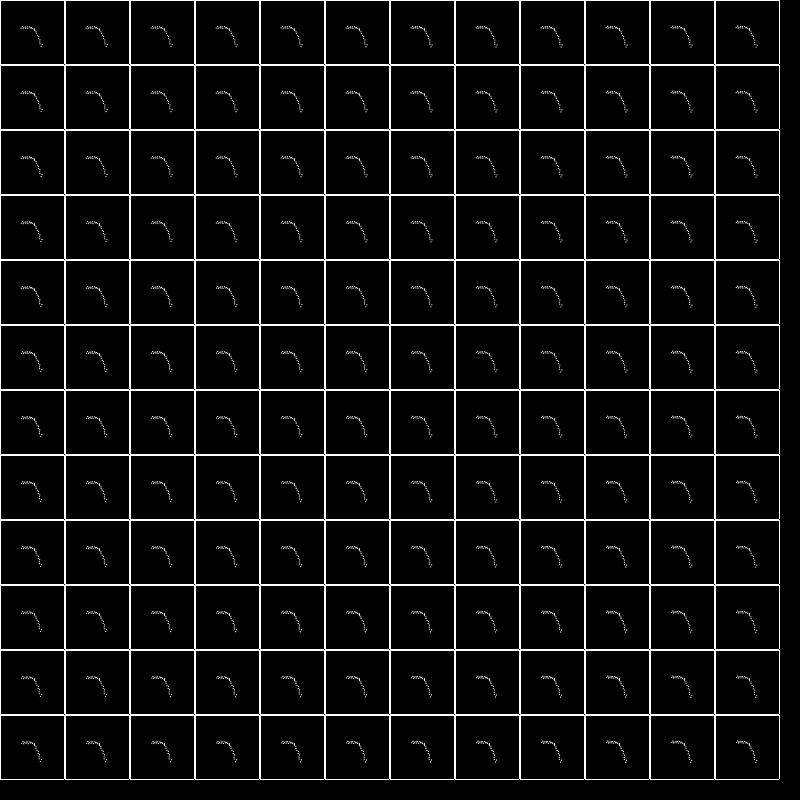}  &
    \includegraphics[width=0.45\linewidth]{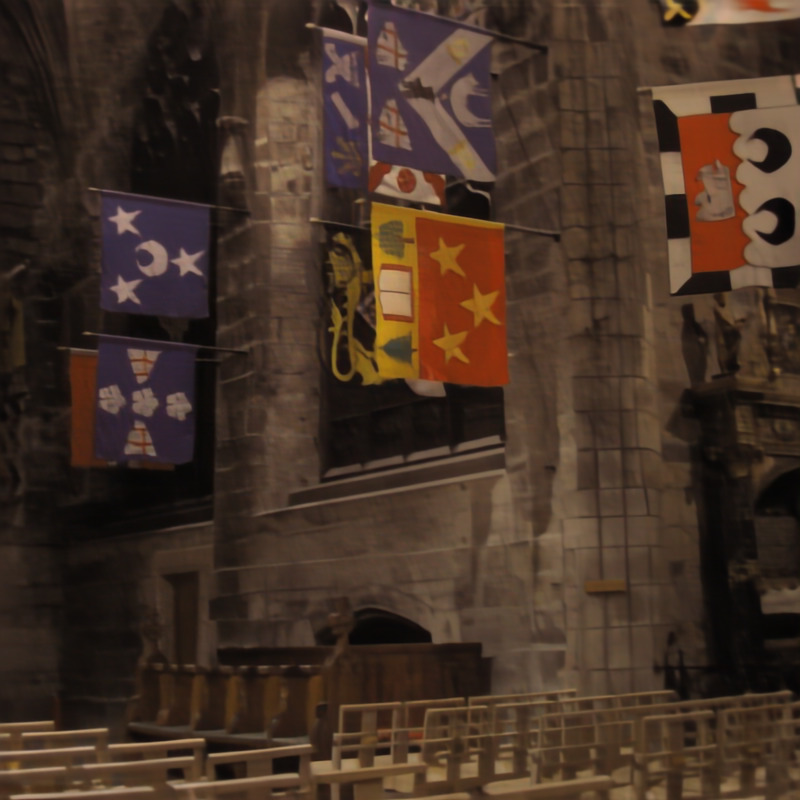} 
    \end{tabular}
    \caption[Restorations on a sample image from the K\"{o}hler dataset \citep{kohler2012recording}.]{Restorations on a sample image from the K\"{o}hler dataset \citep{kohler2012recording}.  }
    \label{fig:sample_Whyte3}
\end{figure*}

\begin{figure*}[ht]
    \centering
    \begin{tabular}{cc}
    Blurry & \cite{whyte2010nonuniform} \\
    \includegraphics[width=0.45\linewidth]{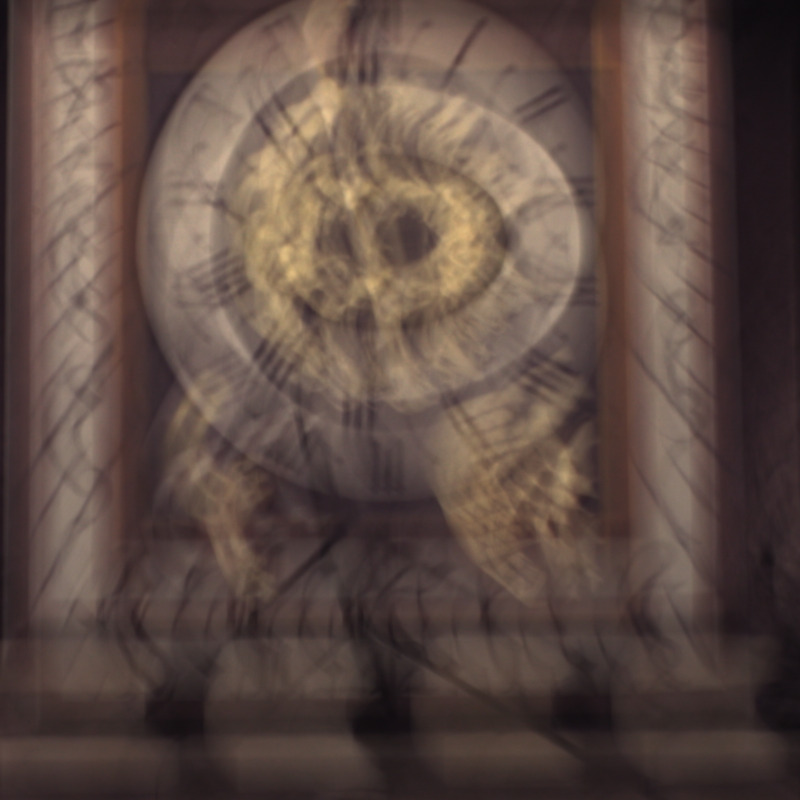}     &  \includegraphics[width=0.45\linewidth]{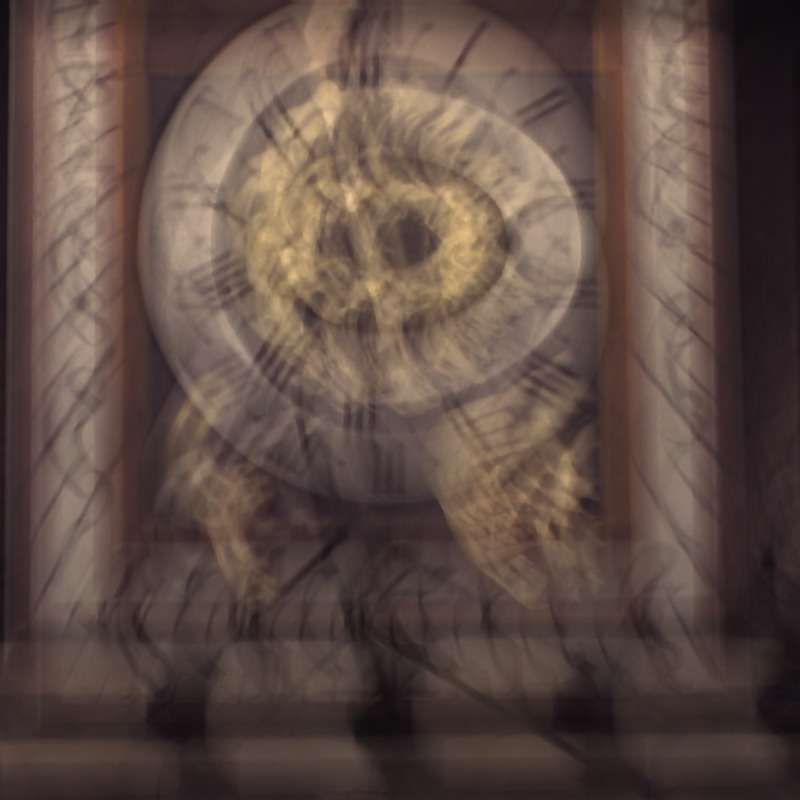}   \\
    Blur2Seq (ours) kernels & Blur2Seq (ours)   \\
    \includegraphics[width=0.45\linewidth]{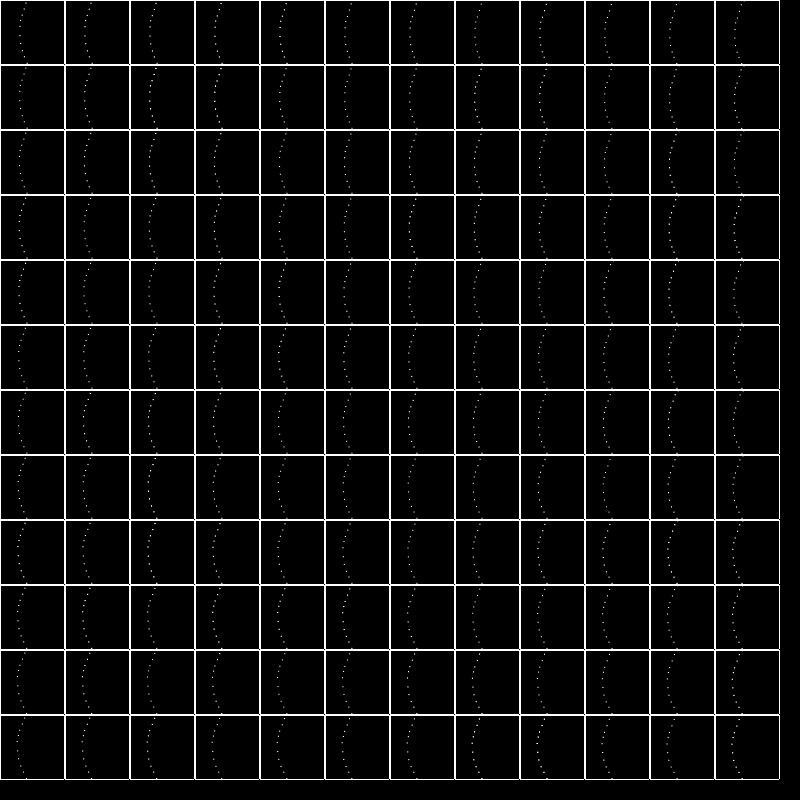}  &
    \includegraphics[width=0.45\linewidth]{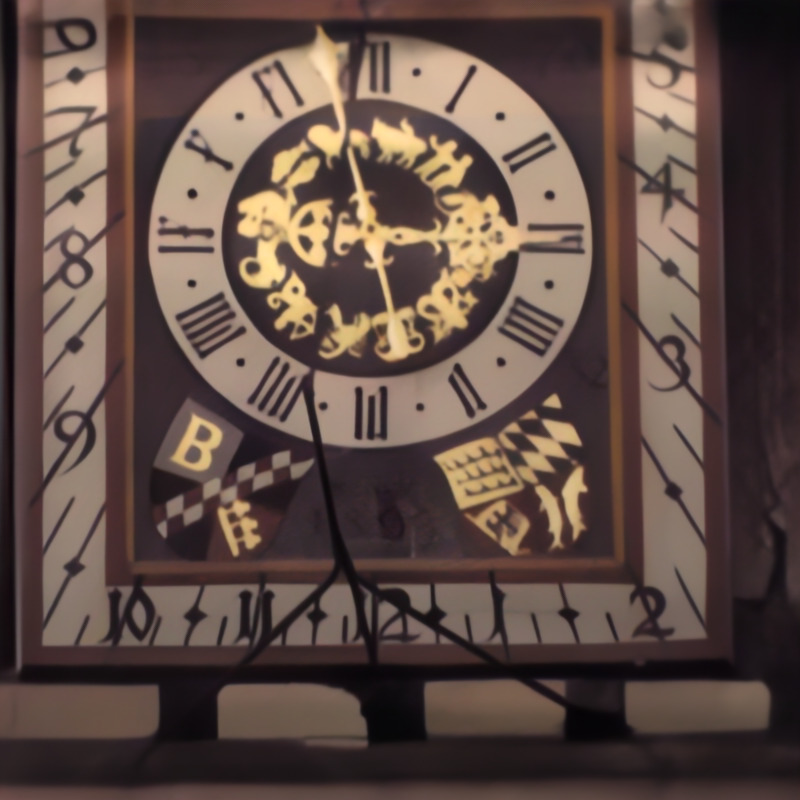} 
    \end{tabular}
    \caption[Restorations on a sample image from the K\"{o}hler dataset \citep{kohler2012recording}.]{Restorations on a sample image from the K\"{o}hler dataset \citep{kohler2012recording}. }
    \label{fig:sample_Whyte4}
\end{figure*}

\begin{figure*}[ht]
    \centering
    \begin{tabular}{ccc}
    Blurry & \cite{gupta_mdf_deblurring} &  \cite{xu2013unnatural} \\
    \includegraphics[width=0.33\linewidth]{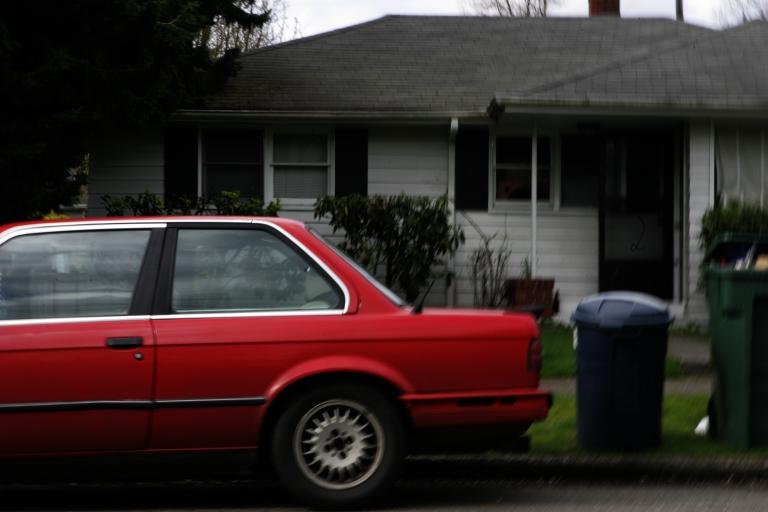}     &  \includegraphics[width=0.33\linewidth]{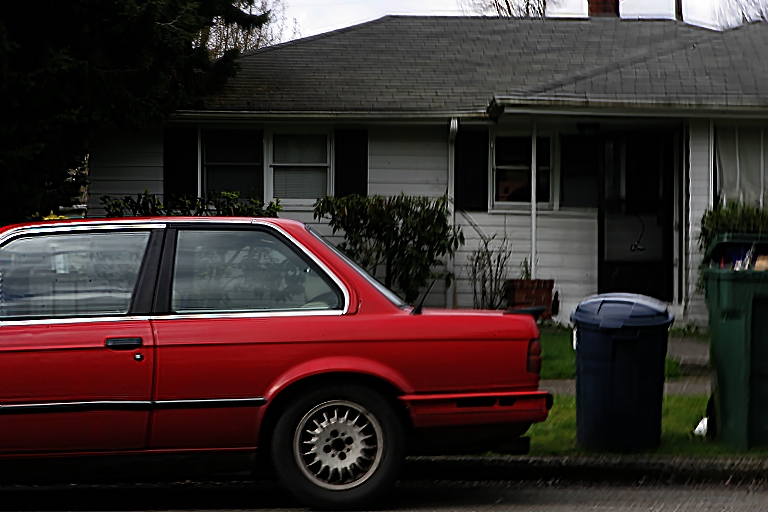} &  \includegraphics[width=0.33\linewidth]{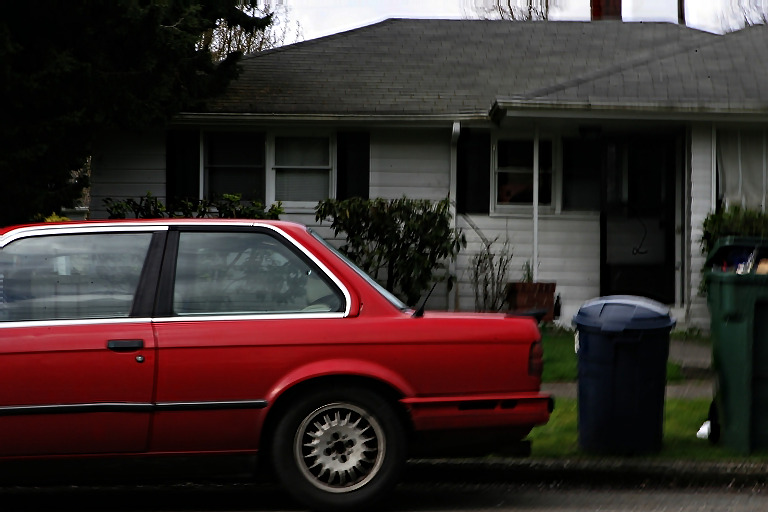}  \\
    \cite{vasu2017local} & Blur2Seq (ours) &  Blur2Seq (ours) kernels \\
    \includegraphics[width=0.33\linewidth]{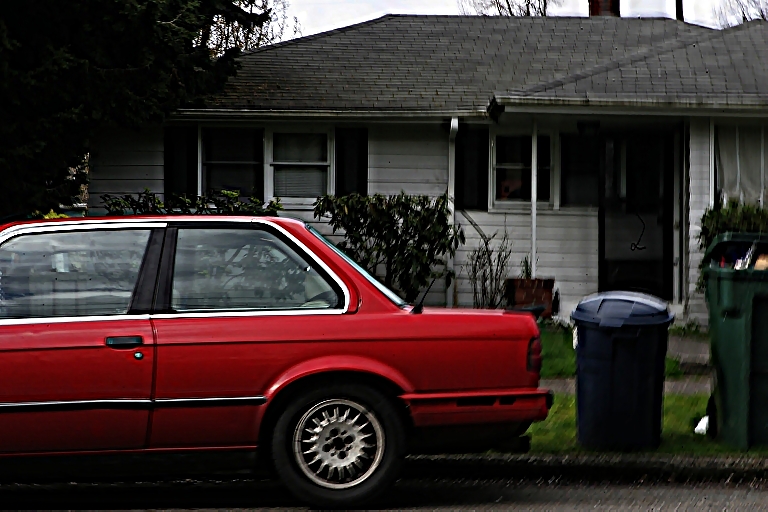}     & \includegraphics[width=0.33\linewidth]{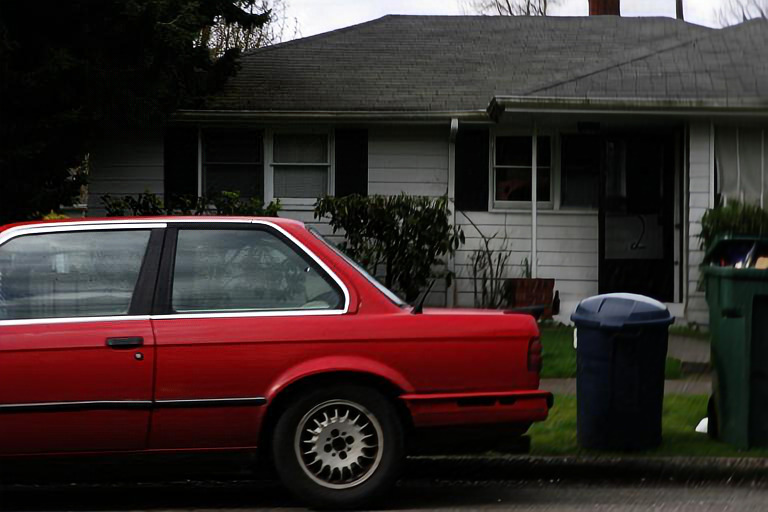} 
    & \includegraphics[width=0.33\linewidth]{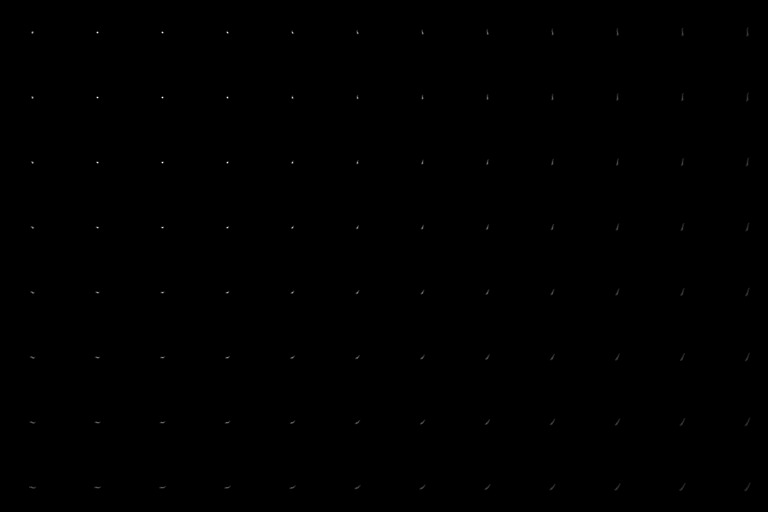} 
    \end{tabular}
    \caption[Restorations on a sample image provided by \cite{gupta_mdf_deblurring}  and \cite{vasu2017local}.]{Restorations on a sample image provided by \cite{gupta_mdf_deblurring}  and \cite{vasu2017local}. The video generated by our method can be seen \href{https://youtu.be/-0HopYPAxDk}{here}. }
    \label{fig:sample_Gupta_car_house2}
\end{figure*}

\begin{figure*}[ht]
    \centering
    \begin{tabular}{ccc}
    Blurry & \cite{gupta_mdf_deblurring} &  \cite{xu2013unnatural} \\
    \includegraphics[width=0.33\linewidth]{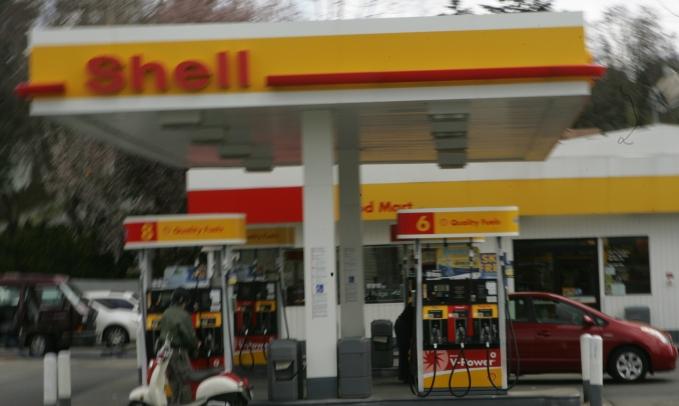}     &  \includegraphics[width=0.33\linewidth]{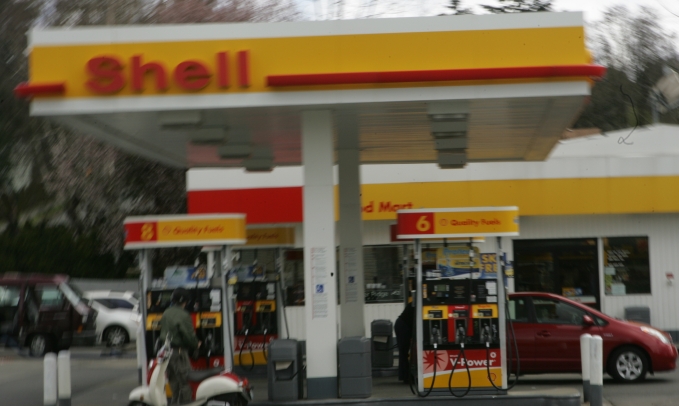} &  \includegraphics[width=0.33\linewidth]{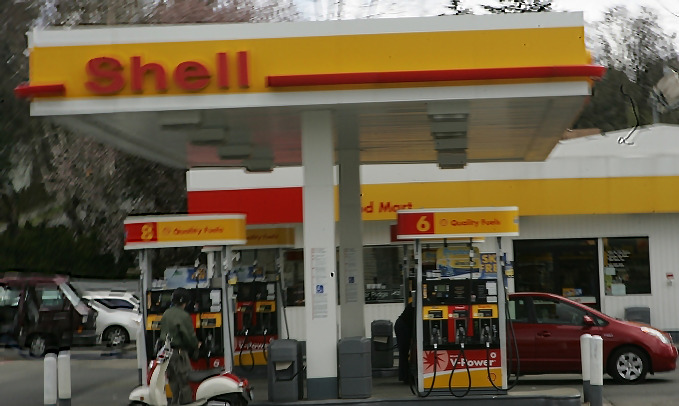}  \\
    \cite{vasu2017local} & Blur2Seq (ours) &  Blur2Seq (ours) kernels \\
    \includegraphics[width=0.33\linewidth]{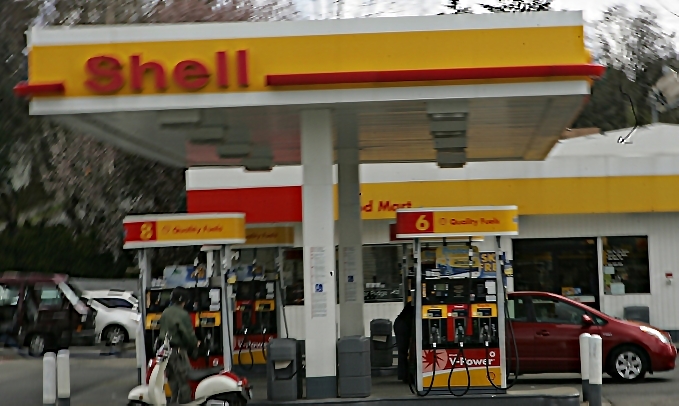}     & \includegraphics[width=0.33\linewidth]{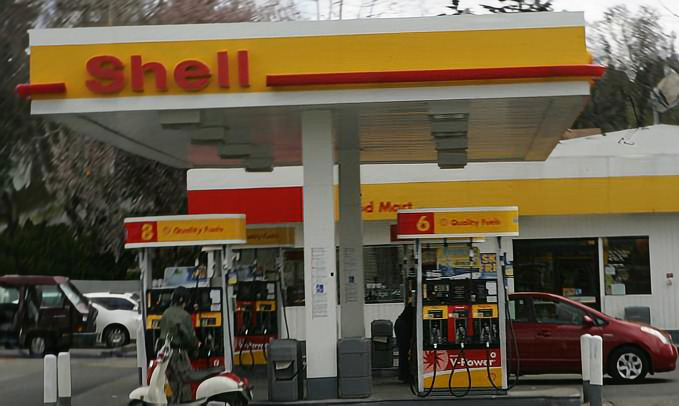} 
    & \includegraphics[width=0.33\linewidth]{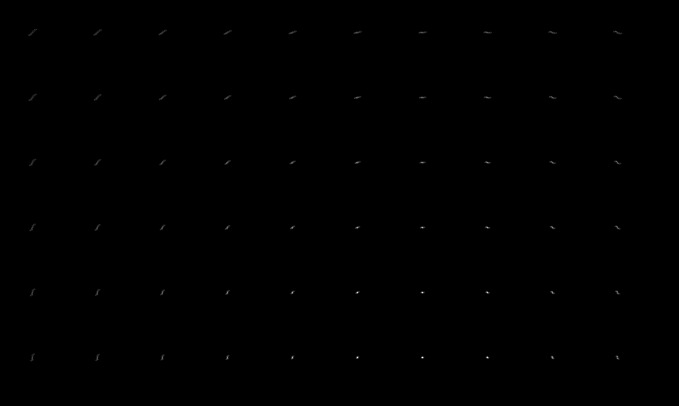} 
    \end{tabular}
    \caption{Restorations on a sample image provided by \cite{gupta_mdf_deblurring} and \cite{vasu2017local}. }
    \label{fig:sample_Gupta_petrol_station}
\end{figure*}

\cref{fig:sample_Gupta_book1,fig:sample_Gupta_book2} show the results on sample images provided by \cite{gupta_mdf_deblurring}, and \cref{fig:sample_Whyte1,fig:sample_Whyte2} illustrate results on images from \cite{whyte2010nonuniform}.
In the case of \cite{whyte2010nonuniform}, we can also compare results on an image from the K\"{o}hler dataset (\cref{fig:sample_Whyte3,fig:sample_Whyte4}), since the corresponding outputs are publicly available on the website associated with \cite{kohler2012recording}. \cref{fig:sample_Gupta_car_house2,fig:sample_Gupta_car_house2,fig:sample_Gupta_petrol_station} show results on sample images provided by \cite{gupta_mdf_deblurring} and \cite{vasu2017local}.

\begin{figure*}[ht]
    \centering
    \begin{tabular}{ccc}
    Blurry & \cite{gupta_mdf_deblurring} &  \cite{xu2013unnatural} \\
    \includegraphics[width=0.33\linewidth]{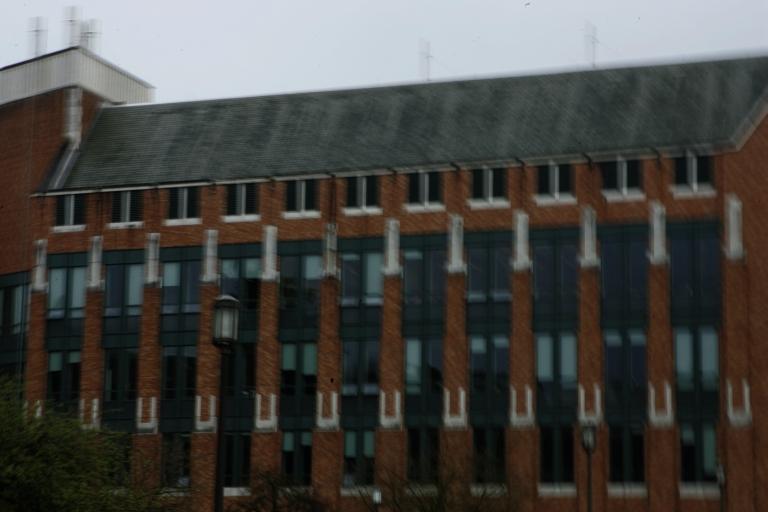}     &  \includegraphics[width=0.33\linewidth]{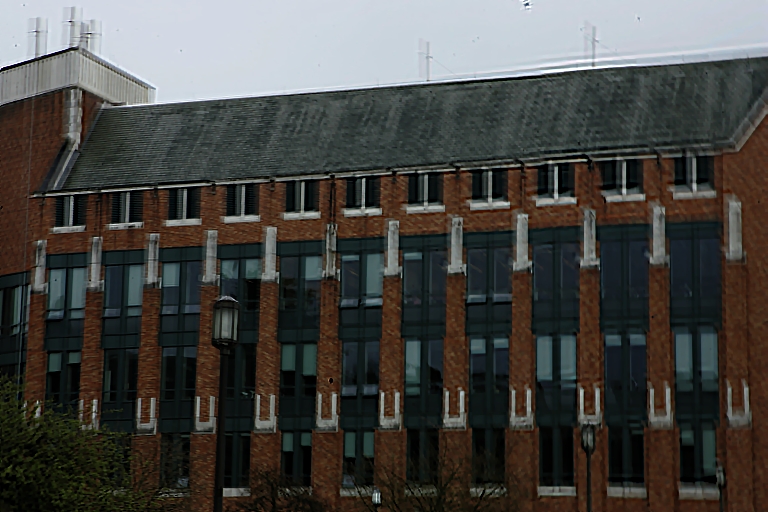} &  \includegraphics[width=0.33\linewidth]{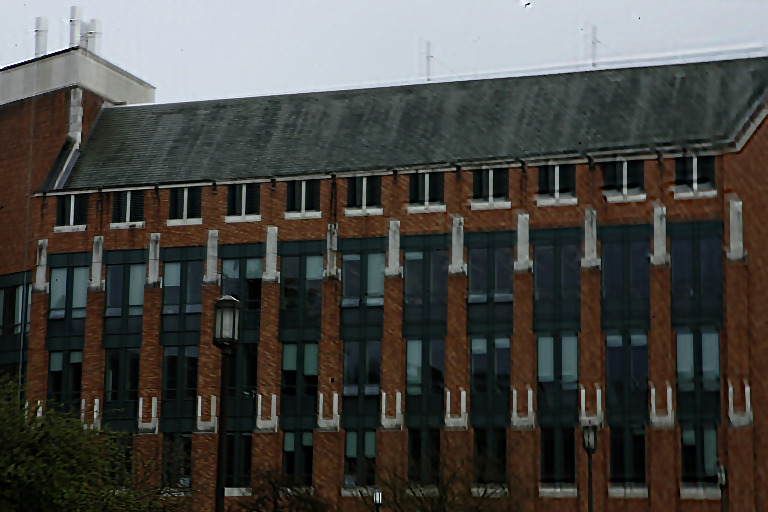}  \\
    \cite{vasu2017local} & Blur2Seq (ours) &  Blur2Seq (ours) kernels \\
    \includegraphics[width=0.33\linewidth]{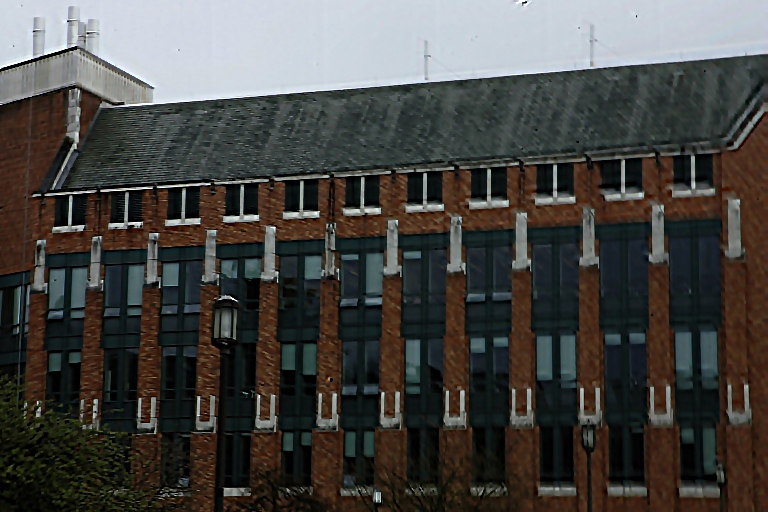}     & \includegraphics[width=0.33\linewidth]{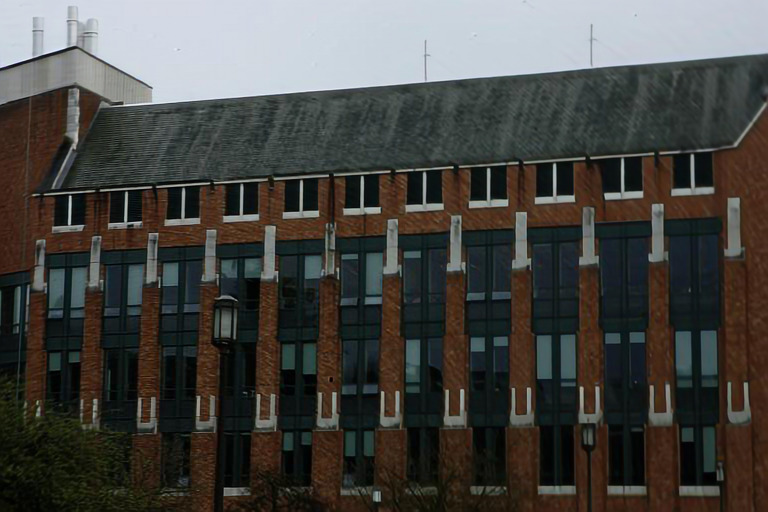} 
    & \includegraphics[width=0.33\linewidth]{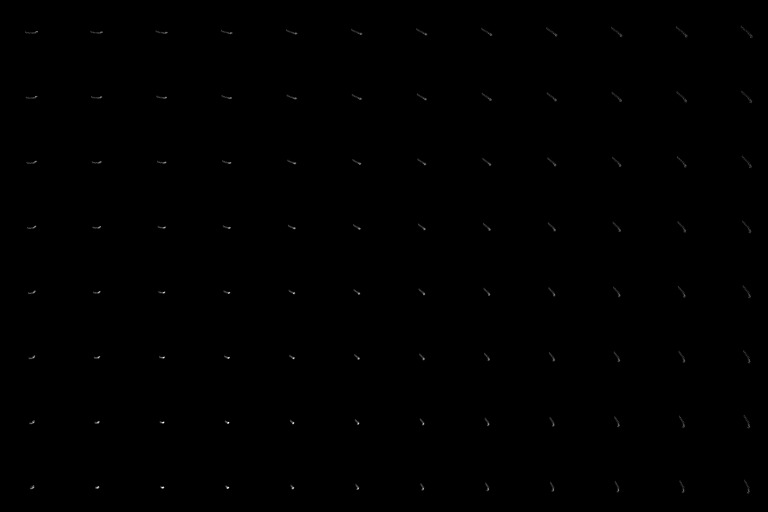} 
    \end{tabular}
    \caption{Restorations on a sample image provided by \cite{gupta_mdf_deblurring}  and \cite{vasu2017local}. }
    \label{fig:sample_Gupta_house2}
\end{figure*}

\FloatBarrier

\bibliography{referencias}
\bibliographystyle{sn-basic}